\documentclass[10pt,twocolumn,letterpaper]{article}

\usepackage{3dv}
\usepackage{times}
\usepackage{epsfig}
\usepackage{graphicx}
\usepackage{amsmath}
\usepackage{amssymb}

\usepackage{subcaption}
\usepackage{graphicx}
\usepackage{comment}
\usepackage{placeins}
\usepackage{booktabs}
\usepackage{multirow}
\usepackage{wrapfig}
\usepackage{amsbsy}
\usepackage{graphics}
\usepackage{grffile}
\usepackage{color}
\usepackage{bm}
\usepackage{url}
\usepackage{comment}

\usepackage[noend]{algpseudocode}
\usepackage{algorithm}
\usepackage{makecell}
\usepackage[square, comma, numbers, sort&compress]{natbib}

\usepackage{enumitem}
\usepackage[table,dvipsnames]{xcolor}
\usepackage{amsthm}

\usepackage[pagebackref=true,breaklinks=true,letterpaper=true,colorlinks,bookmarks=false]{hyperref}

\hypersetup{
linkcolor=BrickRed
,citecolor=Green
,filecolor=Mulberry
,urlcolor=NavyBlue
,menucolor=BrickRed
,runcolor=Mulberry
,linkbordercolor=BrickRed
,citebordercolor=Green
,filebordercolor=Mulberry
,urlbordercolor=NavyBlue
,menubordercolor=BrickRed
,runbordercolor=Mulberry
}

\threedvfinalcopy 

\def\threedvPaperID{152} 

\ifthreedvfinal\fi
\begin{document}

\title{A Dataset-Dispersion Perspective on Reconstruction Versus Recognition in Single-View 3D Reconstruction Networks}

\author{Yefan Zhou\\
UC Berkeley\\
{\tt\small yefan0726@berkeley.edu}
\and
Yiru Shen\\
Clemson University\\
{\tt\small yirus@g.clemson.edu}

\and
Yujun Yan\\
University of Michigan\\
{\tt\small yujunyan@umich.edu}

\and
Chen Feng\\
New York University\\
{\tt\small cfeng@nyu.edu}

\and
Yaoqing Yang\\
UC Berkeley\\
{\tt\small yqyang@berkeley.edu}
}

\maketitle
\thispagestyle{empty}

\begin{abstract}
    Neural networks (NN) for single-view 3D reconstruction (SVR) have gained in popularity. 
    Recent work points out that for SVR, most cutting-edge NNs have limited performance on reconstructing unseen objects because they rely primarily on recognition (i.e., classification-based methods) rather than shape reconstruction.
    To understand this issue in depth, we provide a systematic study on when and why NNs prefer recognition to reconstruction and vice versa.
    Our finding shows that a leading factor in determining recognition versus reconstruction is how ``dispersed'' the training data is.
    Thus, we introduce the \emph{dispersion score}, a new data-driven metric, to 
    quantify this leading factor and study its effect on NNs.
    We hypothesize that NNs are biased toward recognition when training images are more dispersed and training shapes are less dispersed.
    Our hypothesis is supported and the dispersion score is proved effective through our experiments on synthetic and benchmark datasets.
    We show that the proposed metric is a principal way to analyze reconstruction quality and provides novel information in addition to the conventional reconstruction score. We have open-sourced our code.\footnote{https://github.com/YefanZhou/dispersion-score}

\end{abstract}

\section{Introduction}\label{sec:intro}

Using deep learning (DL) for single-view 3D reconstructions (SVR) is our main focus.
Numerous recent publications presented innovative neural network (NN) designs to advance the state-of-the-art in SVR \cite{li2018point,park2019deepsdf,fan2017point,tatarchenko2017octree,groueix2018papier,yang2018foldingnet,wang2018pixel2mesh,sun2018pix3d,tulsiani2017multi,wu2017marrnet,yan2016perspective}.
Their primary focus is on improving the quality of reconstruction on benchmark datasets, which is measured by reconstruction metrics, such as Chamfer distance (CD) \cite{fan2017point}, Earth Mover Distance \cite{fan2017point}, mIoU \cite{choy20163d}, and F-score \cite{tatarchenko2019single}.
While several studies have proposed improved methods to reconstruct shapes, only few focus on the underlying mechanisms of NNs in SVR and whether the actual mechanisms meet the designers' expectations.

Several studies \cite{tatarchenko2019single, Shin2018PixelsVA} have shown that cutting-edge DL models in SVR primarily perform \emph{recognition} rather than reconstruction.
In other words, NNs tend to find a shortcut to solve SVR by using the easy-to-learn classification-based approaches, e.g., by implicitly grouping the training shapes into clusters and memorizing only the mean shapes of these clusters. This mean-shape-based approach is fundamentally different from our intuition of 3D reconstruction. 

More importantly, their finding shows that NN has limited performance on reconstructing novel objects when it relies more on recognition, rendering the question of ``do NNs perform recognition or reconstruction''.
Furthermore, findings~\cite{Shin2018PixelsVA,tatarchenko2019single} show that the bias toward recognition can arise from properties of the dataset, e.g., some training dataset uses the \emph{object-centered (OC) coordinate} \cite{choy20163d, fan2017point, groueix2018papier, Mescheder2019OccupancyNL}. 
In the OC coordinate, the 3D shapes of objects are aligned to the same orientation. 
In the other \emph{viewer-centered (VC) coordinate}, the 3D shapes are aligned to randomly sampled input viewpoints. 
In particular, they observe that using OC can make NNs more biased toward recognition than VC.
Although these examinations provide valuable insights into how NNs perform SVR, the answer to the question depends on a multitude of factors. Determining whether NNs should perform in OC or VC is not enough to resolve the question. 

To address the issue of NNs' bias toward recognition in SVR, we provide a systematic study on recognition versus reconstruction. 
Our investigation leads to a comprehensive evaluation metric (\emph{dispersion score} or DS) and applicable experiment procedures to improve SVR.
In particular, we show that DS can diagnose the trained model's bias toward recognition (in Section~\ref{sec:shapenet-moreimg}). We also illustrate that the use of more dispersed training shapes can improve reconstruction as shown by CD and DS (in Section~\ref{sec:shapenet-moreshape}).
Specifically, our main contributions provide answers to the following questions.

\textit{What causes NNs to prefer recognition in SVR?} 
In our experiments, we showed that whether NNs would perform reconstruction or recognition depends on if the training data is ``dispersed'' or ``clustered''.
For example, we hypothesize that OC makes NNs perform recognition because aligning training objects to a common orientation makes the training shapes more clustered. 
The clustered training shapes can bias NNs toward using recognition-based approaches.\footnote{In the Appendix~\ref{sec:definition-coordinate}, we provide further details of OC and VC, and more experiments on their difference.}
Our main claim is that NNs tend toward reconstruction when 3D training shapes are more dispersed. They are prone to perform recognition when the 2D training images are more dispersed.

\textit{How can we measure the extent of reconstruction or recognition?} 
We propose a new metric, DS, to measure how dispersed or ``unclustered'' the data is.
A larger score indicates that the data is more dispersed and less clustered, whereas a lower score indicates the opposite. 
DS is measured from two perspectives: \emph{input DS} on training data and \emph{output DS} on reconstructed shapes (results of test).
The input DS is calculated to diagnose the training data and describe its relationship with the corresponding trained models.
The output DS is calculated to measure whether the trained models tend to perform reconstruction or recognition. 
We notice that measuring the DS of reconstructed shapes (i.e., output DS) can indicate whether NNs are biased toward recognition because the output shapes tend to form clusters when the NNs rely on using mean shapes to reconstruct.



Finally, it is worth noting that the question of recognition versus reconstruction is not equivalent to memorization versus generalization. The latter has a clear definition, e.g., Eqn. (1) of \cite{feldman2020neural}. The question of recognition versus reconstruction currently does not have a rigorous definition. We provide the first metric to quantify recognition versus reconstruction. However, the DS is not necessarily a one-size-fits-all statistic that can distinguish recognition from reconstruction.
We know that existing reconstruction metrics like CD, which measures a single shape's quality, cannot tell whether the reconstruction uses the mean shape because quantifying the mean shape requires measuring more than one shape. Thus, the proposed DS provides novel information in addition to the conventional reconstruction score when assessing SVR models.



\section{Related Work}\label{sec:related}

\noindent
\textbf{Single-view 3D reconstruction}\label{sec:related}  
There have been lots of studies on DL-based SVR using various 3D representations, including voxels~\cite{choy20163d}, point clouds~\cite{fan2017point, yang2018foldingnet, groueix2018papier}, meshes~\cite{wang2018pixel2mesh, Gkioxari2019MeshR}, and signed distance fields (SDF)~\cite{Xu2019DISNDI, Mescheder2019OccupancyNL}.
These techniques have been proven efficient in improving the quality of shape reconstruction, measured by similarity metrics such as CD, Earth Mover Distance \cite{fan2017point}, mIoU \cite{choy20163d}, and F-score \cite{tatarchenko2019single}. 
A critical difference between these work and ours is that they make a single predicted shape closer to the ground truth shape by neglecting if the NNs use recognition-based or reconstruction-based schemes, which requires the knowledge obtained from the whole dataset.

\noindent
\textbf{Reconstruction vs Recognition} 
Recently, a few studies advocate a rethinking of how NNs perform SVR tasks. 
In particular, the mechanism of SVR is hypothesized to be a combination between reconstruction and recognition \cite{Shin2018PixelsVA, tatarchenko2019single}. 
For example, \cite{Shin2018PixelsVA} proposes that the commonly used shape representation and object-centered coordinate make NNs place more importance on recognizing the object category (or cluster) and thus encourage memorizing the object shape.
This hypothesis is supported by the qualitative results that trained NN models sometimes predict a shape in an entirely different object category than the input image, which is conjectured to be caused by a classification error.
Further, \cite{tatarchenko2019single} shows that state-of-art NNs for SVR tasks rely predominantly on recognition instead of reconstruction. 
The claim is supported by observing that NNs have similar reconstruction performance with recognition-based methods measured by the mIoU score. 
Although prior work points out some factors that could bias NNs toward recognition, their findings are limited to special issues like shape coordinate representation.
In our work, we provide a more systematic view of this problem and give operational ways to guide NNs toward reconstruction.

\noindent
\textbf{Choice of The Coordinate Representation}
The conventional setting for SVR tasks is to predict output shapes in OC coordinate~\cite{choy20163d, fan2017point, groueix2018papier, Xu2019DISNDI, Mescheder2019OccupancyNL}. However, VC coordinate is recommended to alleviate the bias toward recognition \cite{tatarchenko2019single} and improve the generalization ability to reconstruct unseen object classes \cite{Shin2018PixelsVA}. In this work, we study the impact of the two coordinate representations on the DS of datasets and trained models.
\section{Definition and Main Claim}\label{sec:definition}

\subsection{Single-view 3D Reconstruction}

We consider the problem of SVR using NNs. The input $I \in \mathbb{R}^{W \times H}$ is a 2D image with width $W$ and height $H$. The output, denoted as $S$, is a 3D point cloud $\in \mathbb{R}^{N \times 3}$. 
We only consider point-cloud-based shape representation in this work.
A NN model $f$ is trained to reconstruct the shape $S$ from the input image $I$, by minimizing the empirical loss defined for a certain loss function $l$:
\begin{equation}
    \min_f \sum_{i=0}^{n-1} {l(f(I_i), S_i)}.
\end{equation}

\subsection{Recognition vs. Reconstruction}\label{sec:rec_vs_rec}

Recognition and reconstruction are two modes that NNs can perform in the SVR tasks. The basic mechanisms of these two are outlined as the following:

\noindent
\textbf{Recognition} \space A recognition-based model reconstructs shapes in two steps. First, during training, the model partitions the training shapes into clusters based on shape similarity and memorizes the mean shape of each cluster. Then, during testing, the model classifies the input test image into one of the clusters and retrieves the corresponding mean shape as the output.
In this case, the reconstructed shapes are highly clustered because different input images could be classified into one single mean shape.

\noindent
\textbf{Reconstruction}\label{def:reconstruction}
A reconstruction-based model directly generates the 3D reconstruction rather than using any cluster or semantic information. In this case, the reconstructed shapes are dispersed because the feature of each shape corresponds to the low-level image cues.

The two mechanisms described above are distinctive but not disjoint. It is known that a trained NN in practice combines these two to perform SVR.
The investigation of recognition versus reconstruction is different from memorization versus generalization of NN in SVR~\cite{bautista2021generalization}. The latter focuses on reducing the generalization gap between training and test, while our work only studies the working mechanism of NNs for SVR.

\subsection{Dispersion Score Metric}\label{sec:metric_on_dataset}

We define the dispersion score (DS) to measure how dispersed the data is. The metric is defined based on the classical notion of \emph{clustering inertia}~\cite{363440}. 

Given a dataset $D = \{x_{i}\}^{N-1}_{i=0}$ and a distance function $d(x,y)$, we first determine \textit{clustering} of the dataset by using the K-medoids algorithm~\cite{10.1007/978-3-540-87993-0_19}. We provide ablation studies on different clustering methodologies in Appendix~\ref{sec:ablation-study-clustering}.
Given the number of clusters $n$, the clustering result is denoted as $C_n(\cdot)$, where for each sample $x_i\in D$, K-medoids gives the cluster label $C_n(x_i)$. 
 $Ctr_i$ denotes the centroid of the cluster that contains the sample $x_i$. 
Then, the inertia $I$ of the dataset $D$ partitioned by $C_n(\cdot)$ is defined as:
\begin{equation}
    I_{C_{n}}(D) = \sum_{i=0}^{N-1} d(x_{i}, Ctr_{i}).
\end{equation}
The DS, defined using the inertia, is given by:
\begin{equation}
    DS(D) = \frac{I_{C_{n}}(D)}{N}.
\end{equation}

DS measures the average distance of each sample to its assigned cluster centroid. 
Thus, with a larger DS, the sample is further away from its cluster centroid and the dataset is more dispersed.
For example, if the NN performs pure recognition, each reconstructed shape is equal to the corresponding cluster's mean shape (cluster centroid). 
In this case, the DS of the reconstructed shapes equals 0, which is the limit of pure recognition.

The K-medoids algorithm requires assigning the number of clusters. We automatically determine this hyperparameter using the ``Kneedle'' method~\cite{5961514}. The detail is provided in Appendix~\ref{sec:hyperp-tuning}.
When evaluating DS in SVR tasks, we need to define the distance function $d(x,y)$ to measure pairwise distance between data samples. For 3D point cloud data, we define $d(x,y)$ as CD which measures the distance between two point sets. 
For the 2D image data, we define $d(x,y)$ as the \emph{feature reconstruction loss} which compares image contents in a high dimensional feature space~\cite{Johnson2016PerceptualLF}.

\subsection{Dispersion Relationship Hypothesis}
\label{sec:dispersion_relationship}

\begin{figure}[!thb]
\centering
    \includegraphics[width=.98\columnwidth]{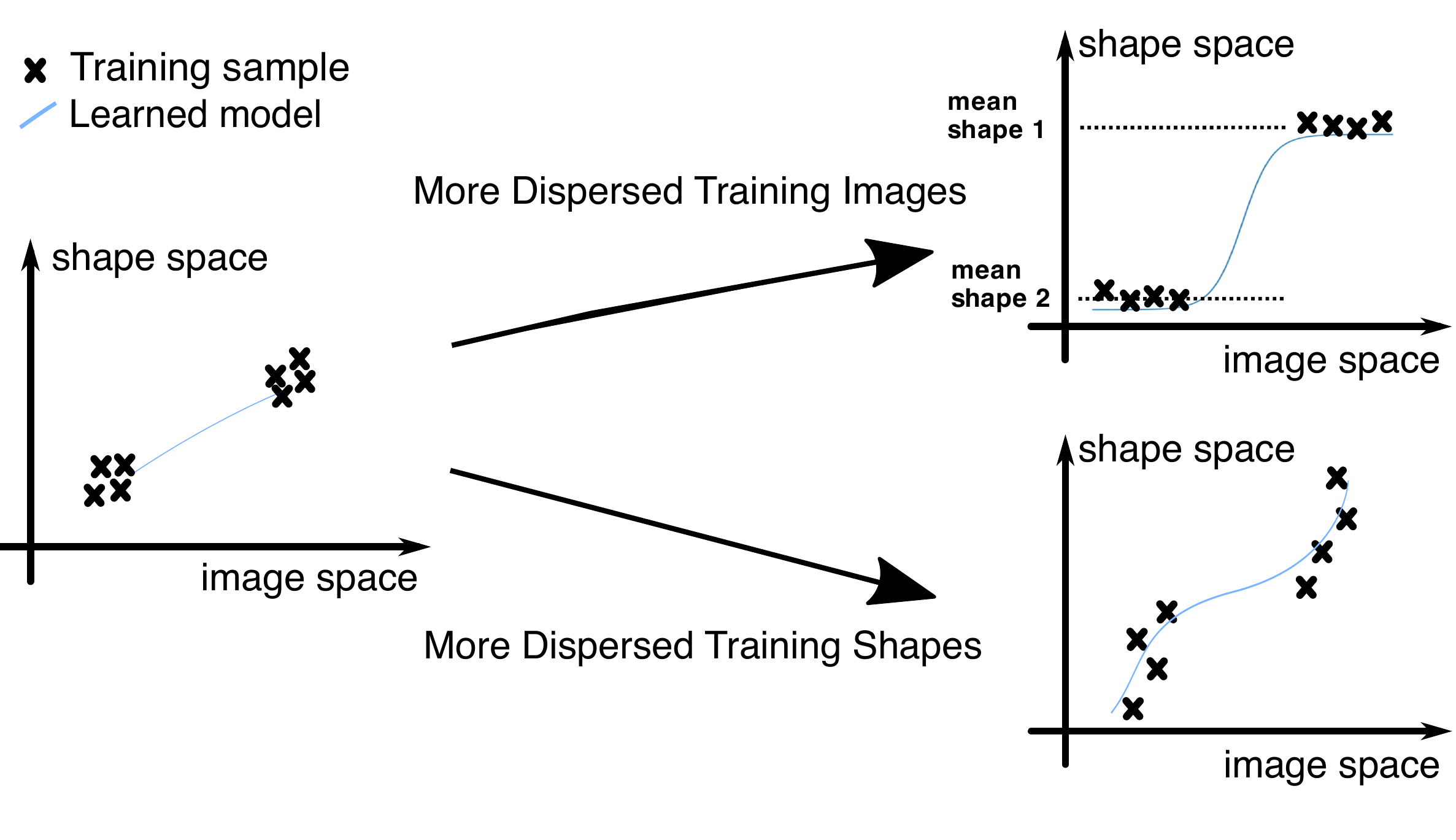}\vspace{-3mm}
    \caption{Caricature of the two approaches to change training datasets’ DS: using more dispersed training images makes the reconstructed shapes more clustered, whereas using more dispersed training shapes makes the reconstructed shapes more dispersed.}
    \label{fig:overview-translation}
\end{figure}

In this paper, we experimentally demonstrate that the bias of trained NNs toward either recognition or reconstruction depends on whether the training dataset is clustered or dispersed. 
Specifically, we consider two ways to change datasets' DS: \emph{more dispersed training images} and \emph{more dispersed training shapes}, as illustrated in Figure \ref{fig:overview-translation}. We will experimentally demonstrate the following claim.

\noindent
\textbf{(Main claim)} \space \textit{In SVR, the more dispersed training images make NNs biased toward recognition, whereas the more dispersed training shapes guide NNs to use reconstruction.}

The claim is motivated by prior SVR work. First, \cite{tatarchenko2019single} proposes to use VC instead of OC. Training shapes in VC are more dispersed than OC, while training images in both cases are the same. Second, it is common to use image augmentations to enhance SVR~\cite{groueix2018papier, Mescheder2019OccupancyNL}, in which training images become more dispersed while training shapes remain unchanged.

We illustrate intuition behind the main claim in Figure~\ref{fig:overview-translation}.
By making training images more dispersed with the clustered training shapes, we make the trained model exhibit a higher tendency toward recognition, i.e., the shape predictions concentrate on the mean shapes and are highly clustered. The clustered shape predictions are illustrated in Figure \ref{fig:overview-translation} as the intersections between the two dashed lines and the $y$-axis.
On the other hand, making training shapes more dispersed in the shape space guides the model to learn more dispersed shape data. Thus, the NNs learn to reconstruct more dispersed shapes and rely less on mean shapes.

\section{Experiments on Synthetic Dataset}\label{sec:exp-toydataset}

In this section, we verify our claim on synthetic datasets.
We first describe the designs of the synthetic datasets, which correspond to the two transitions proposed in Section \ref{sec:dispersion_relationship}, specifically, more dispersed training images and shapes.
We then show the effects of the two transitions on NNs' tendency toward recognition or reconstruction by analyzing both \emph{distance matrices} and input/output DS. The definition of distance matrices is in Section \ref{sec:toy-imple}.



\subsection{Dataset}\label{sec:toy_data_generation}


We create and split a synthetic base dataset into training and test sets. By sampling instances from the training set of the base dataset, we generate two groups of sub-trainsets representing the varying DS for training images and shapes.

\noindent
\textbf{Synthetic Shape Generation} The base dataset is generated by interpolating between a cube and a sphere of a similar size. We use Blender~\cite{Hess:2010:BFE:1893021} to implement the interpolation, and more details are in Appendix~\ref{sec:synthetic-data}. The interpolation generates 1000 intermediate shapes.
Given each intermediate shape in mesh format, we render an image from the isometric view and sample a point cloud consisting of 2500 points. We use the image and the point cloud to comprise an image-shape pair as a data sample.

\begin{figure}[!htb]
  \begin{subfigure}[b]{0.48\columnwidth}
       \centering
       \includegraphics[width=\linewidth]{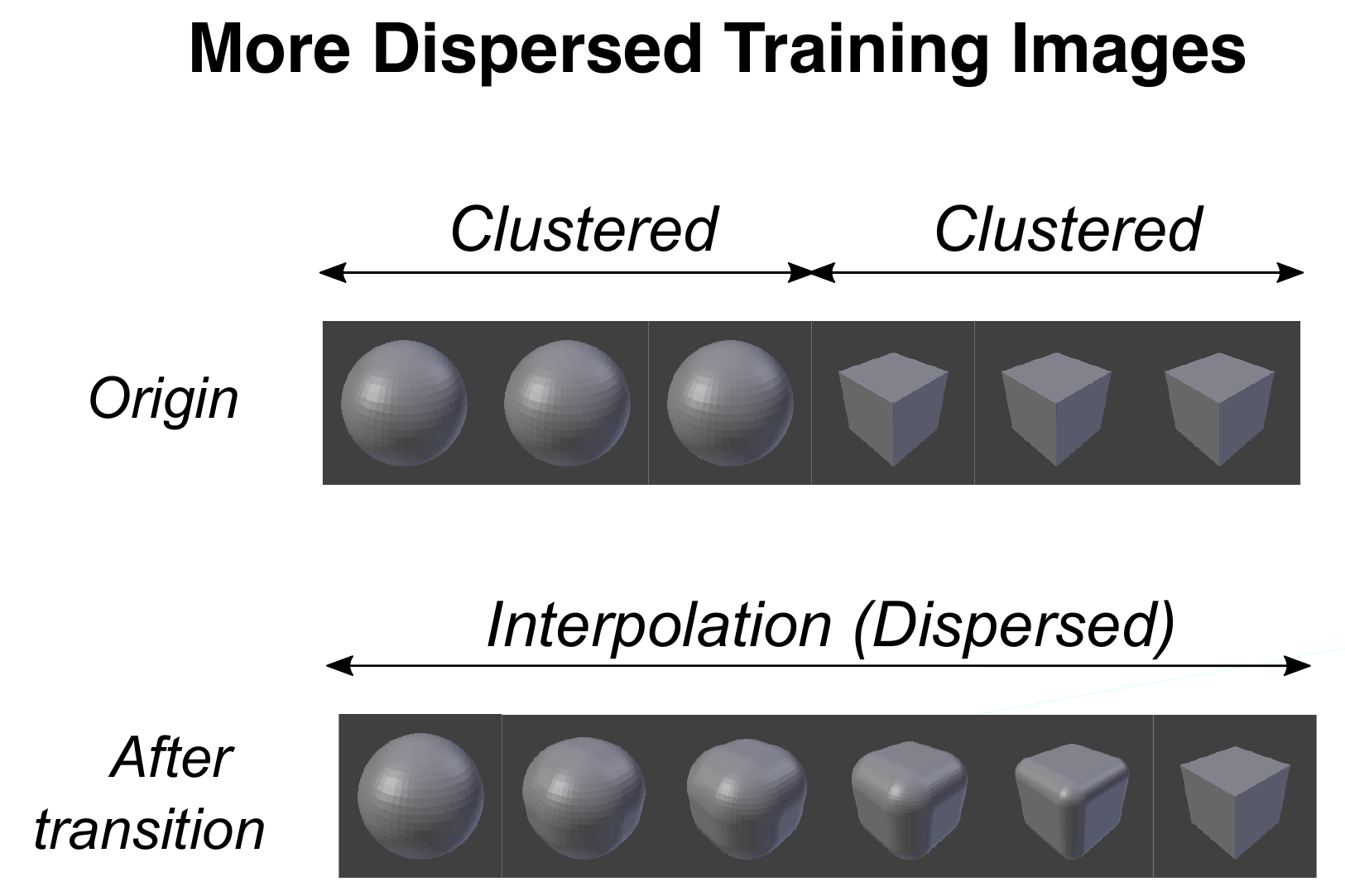}
       \caption{}
       \label{fig:toydata-examples-images}
   \end{subfigure}
  \begin{subfigure}[b]{0.48\columnwidth}
       \centering
       \includegraphics[width=\linewidth]{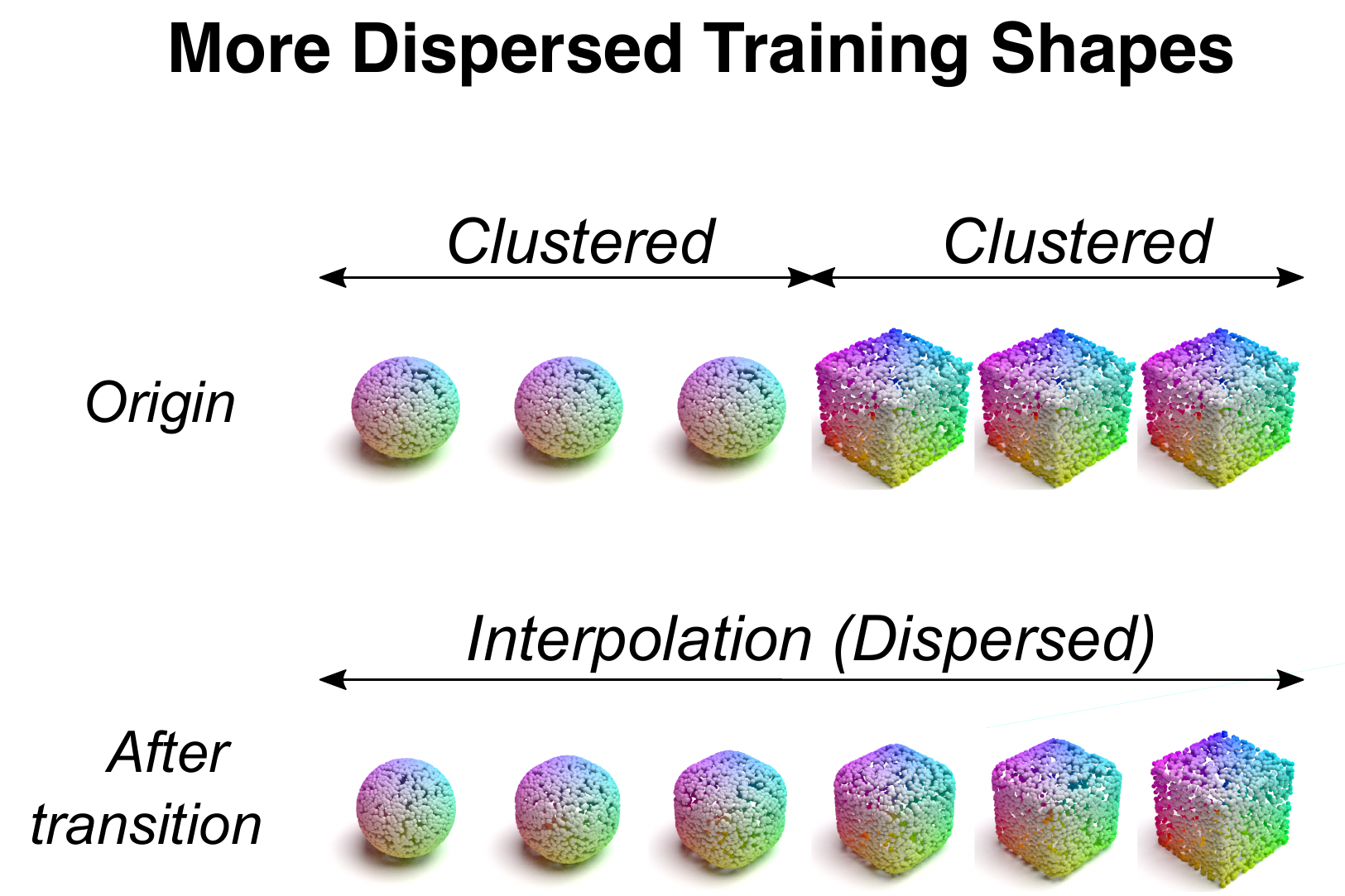}
       \caption{}
       \label{fig:toydata-examples-shapes}
   \end{subfigure} 
    \caption{The illustration of the two transitions implemented by toy examples.
    (a) Interpolating more images between sphere and cube images makes training images more dispersed. 
    (b) Interpolating more shapes between sphere and cube shapes makes training shapes more dispersed. 
    }
    \label{fig:toydata-examples}
\end{figure}

\begin{figure}[ht!]
\begin{centering}
	\begin{tabular}{c@{\hspace{0.1cm}}c@{\hspace{0.07cm}}c@{\hspace{0.05cm}}c|c@{\hspace{0.07cm}}c@{\hspace{0.05cm}}c@{\hspace{0.1cm}}c}
	\toprule
    \multicolumn{4}{c|}{More Dispersed Images} & \multicolumn{4}{c}{More Dispersed Shapes} \\
	\toprule
	 &\small{Images} & \small{Shapes} & \small{Recons.} & \small{Images} & \small{Shapes} & \small{Recons.}& \\
	 \toprule
	\small{$S_{1}I_{1}$} &\includegraphics[width=0.12\columnwidth,keepaspectratio]{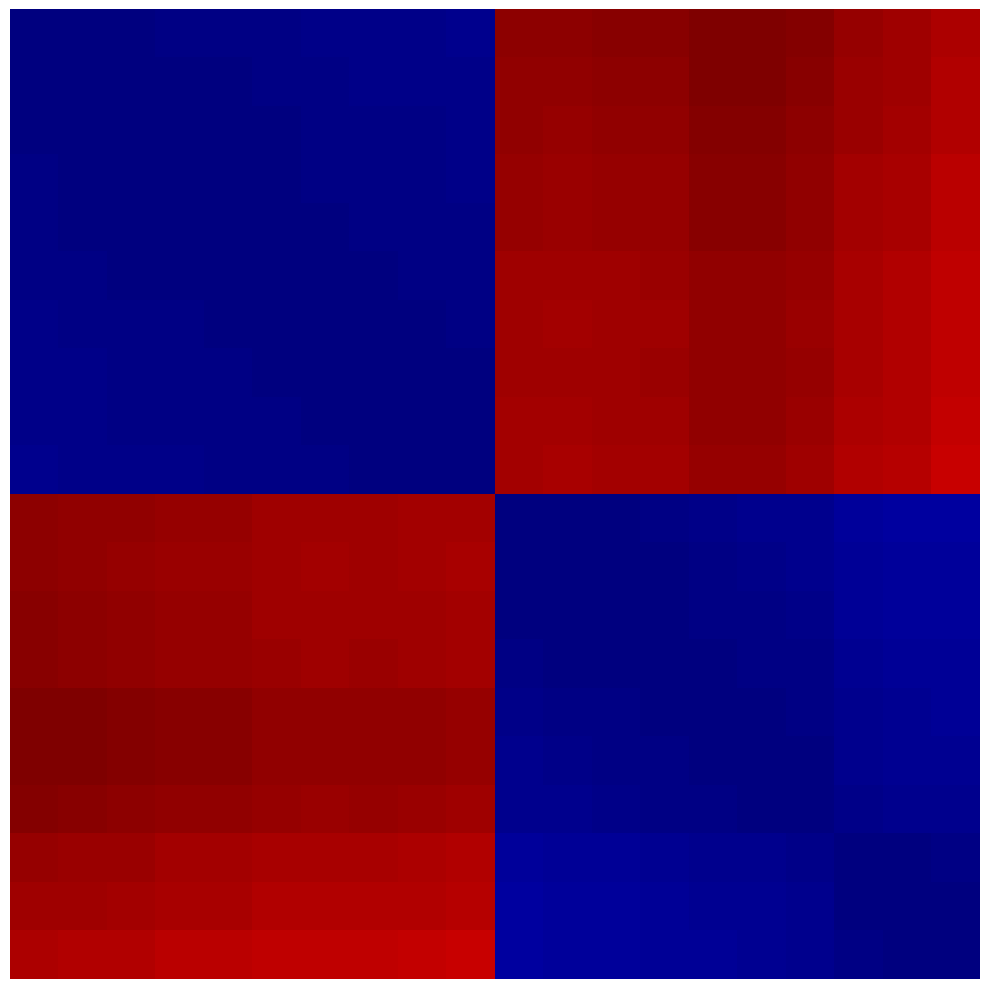} &
	\includegraphics[width=0.12\columnwidth,keepaspectratio]{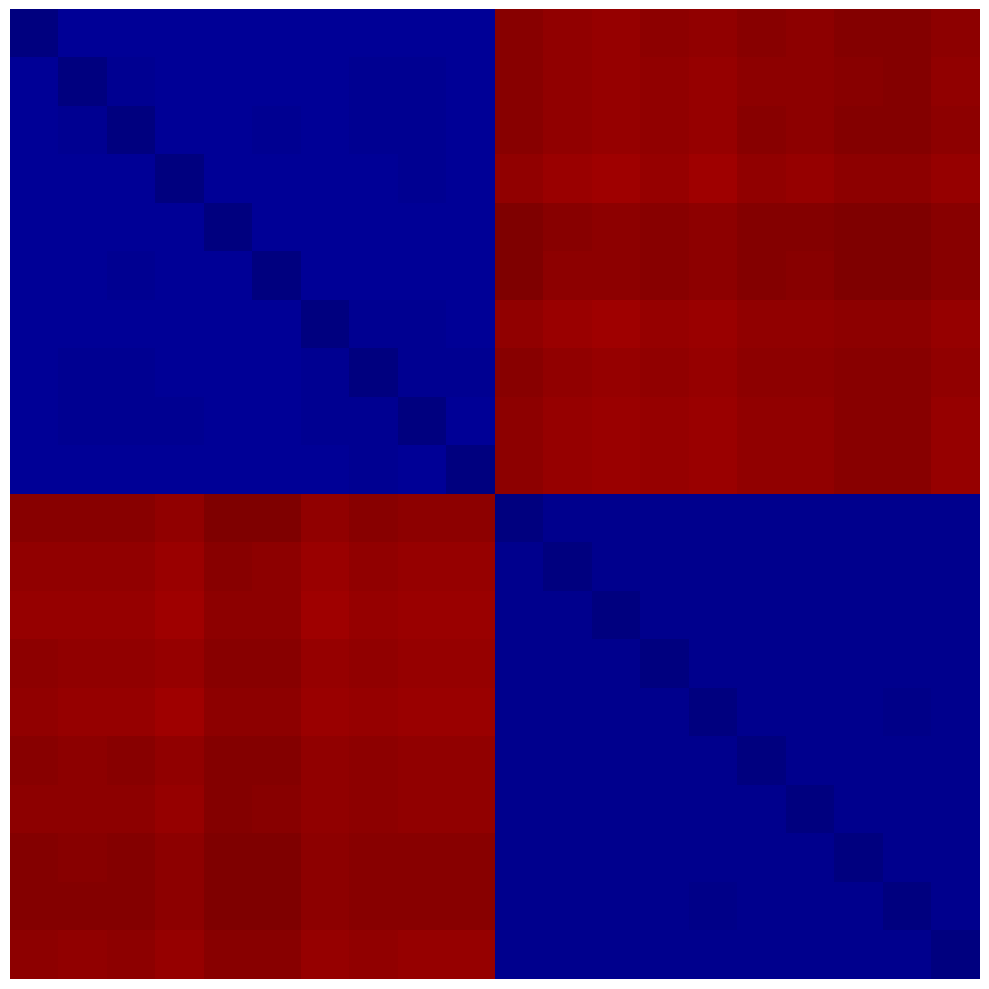} &
	\includegraphics[width=0.12\columnwidth,keepaspectratio]{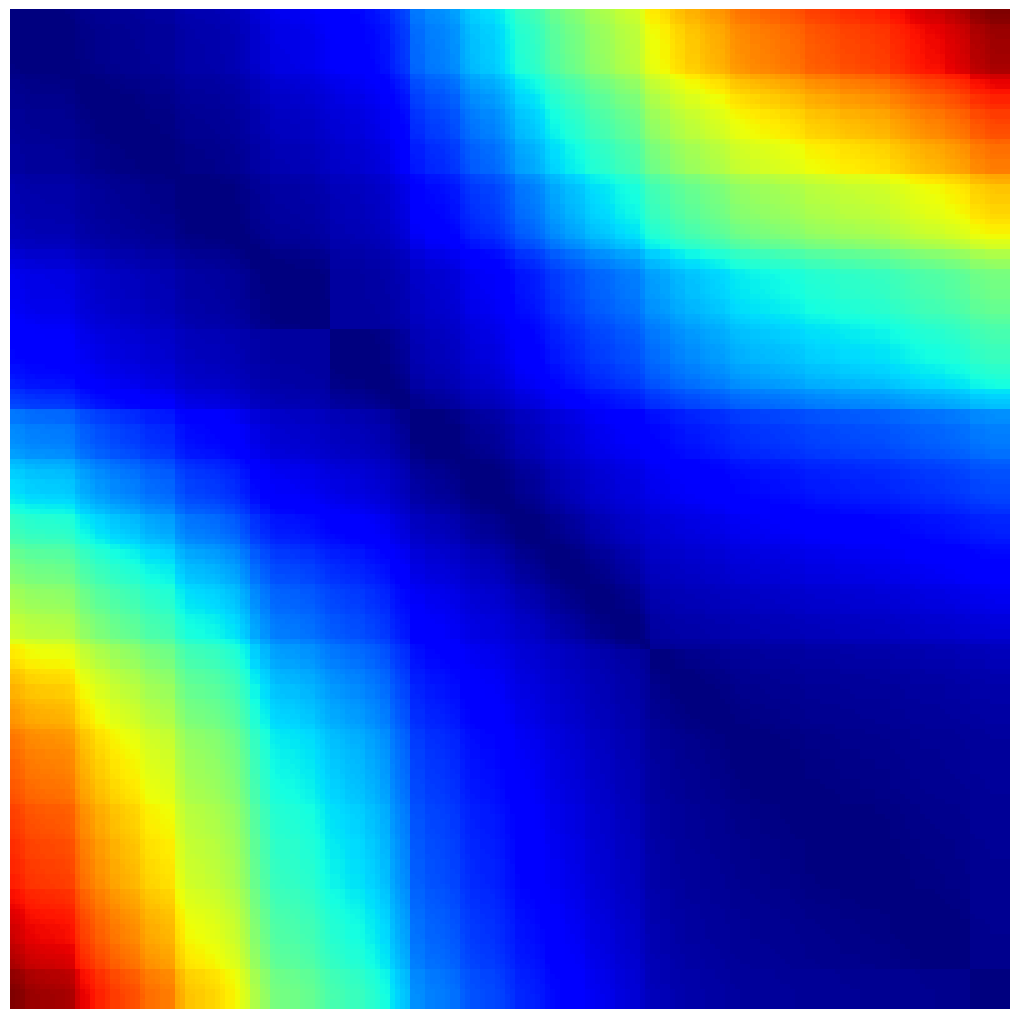} &  
	\includegraphics[width=0.12\columnwidth,keepaspectratio]{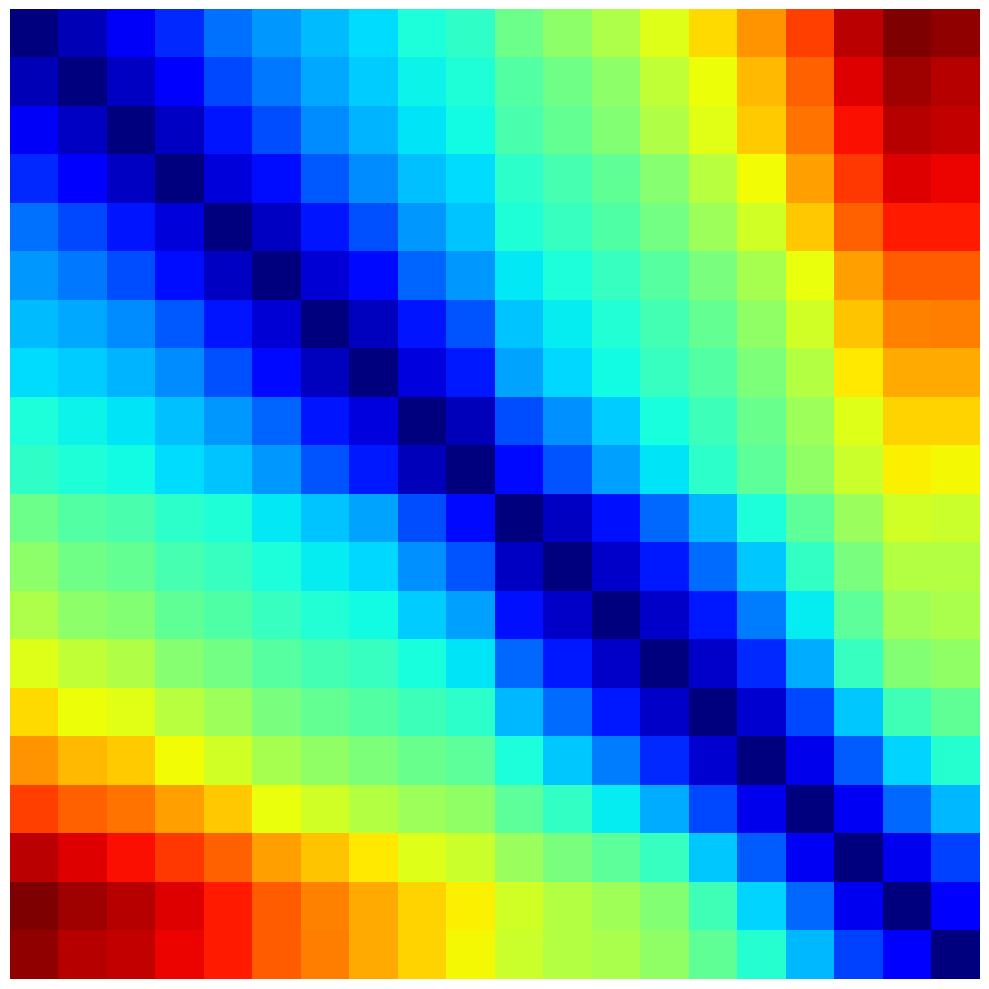} &
	\includegraphics[width=0.12\columnwidth,keepaspectratio]{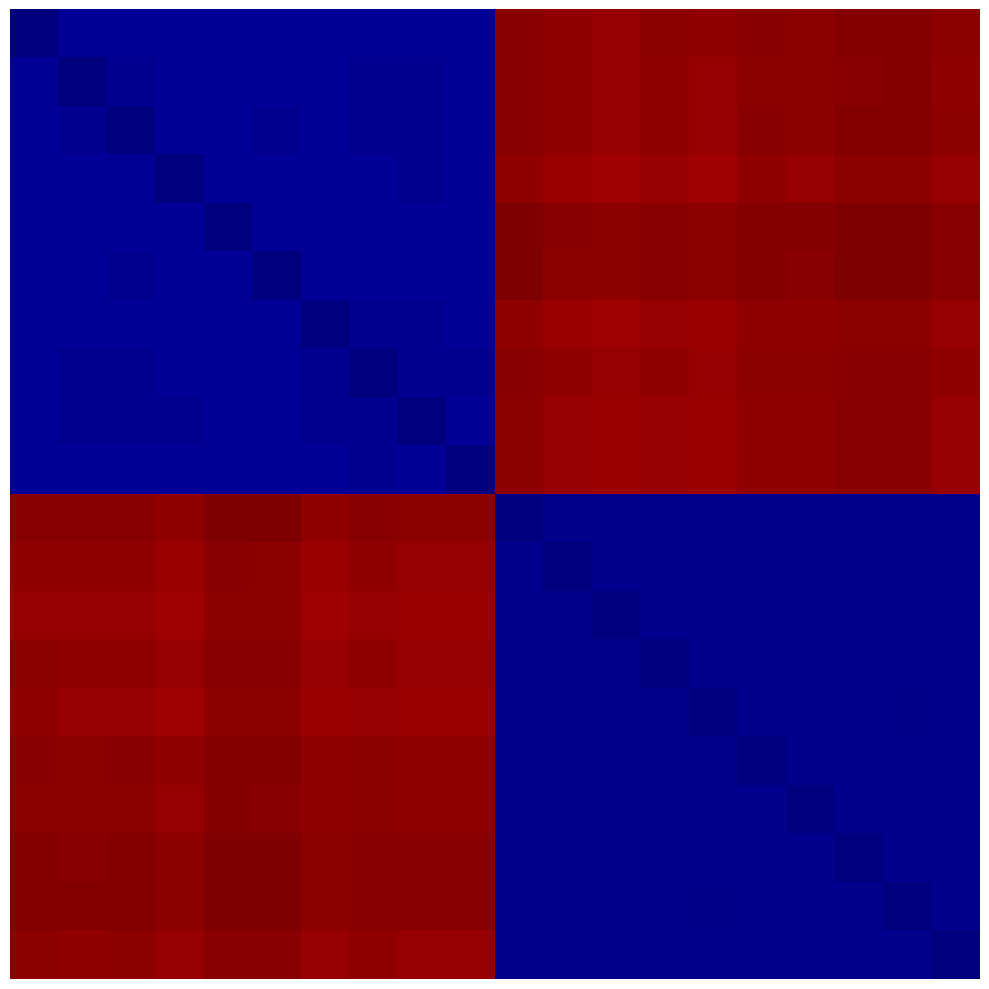} &
	\includegraphics[width=0.12\columnwidth,keepaspectratio]{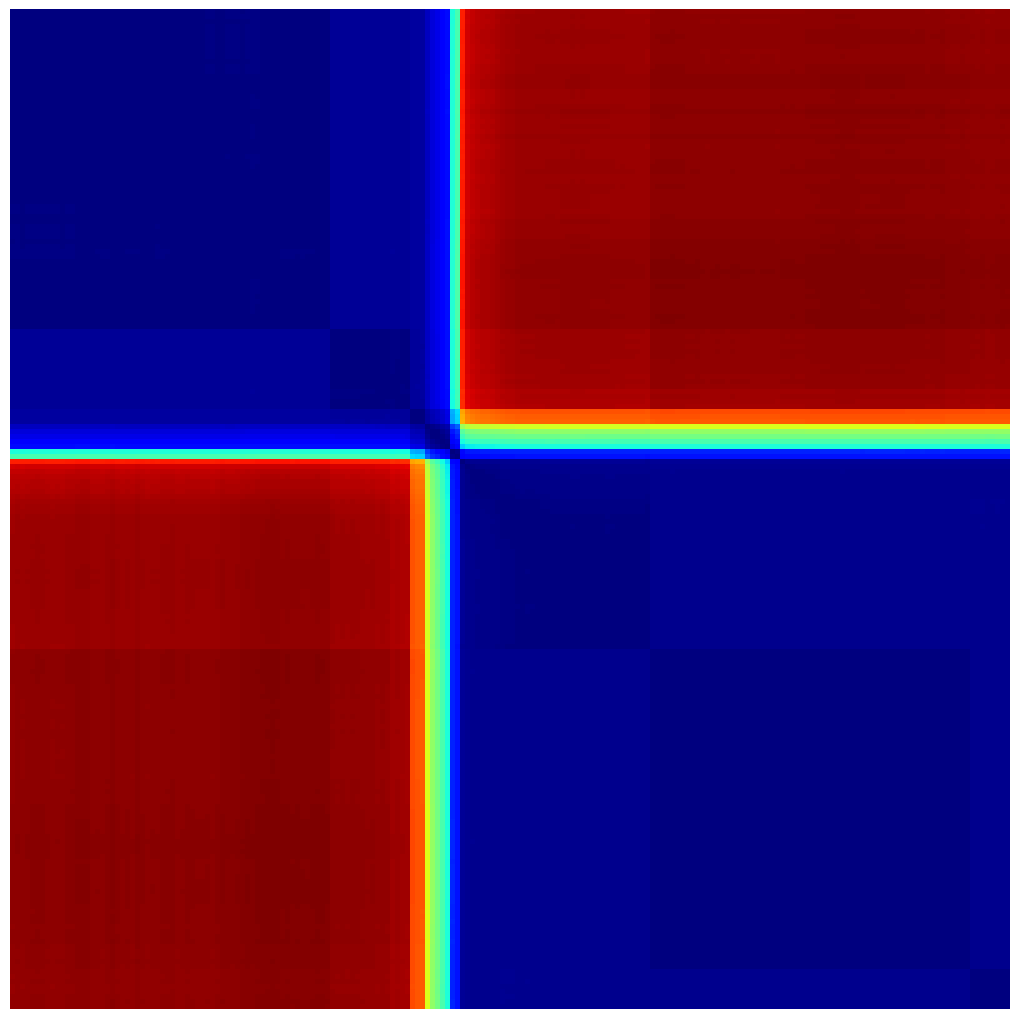}& \small{$S_{1}I_{8}$}\\
	\small{$S_{1}I_{2}$} &\includegraphics[width=0.12\columnwidth,keepaspectratio]{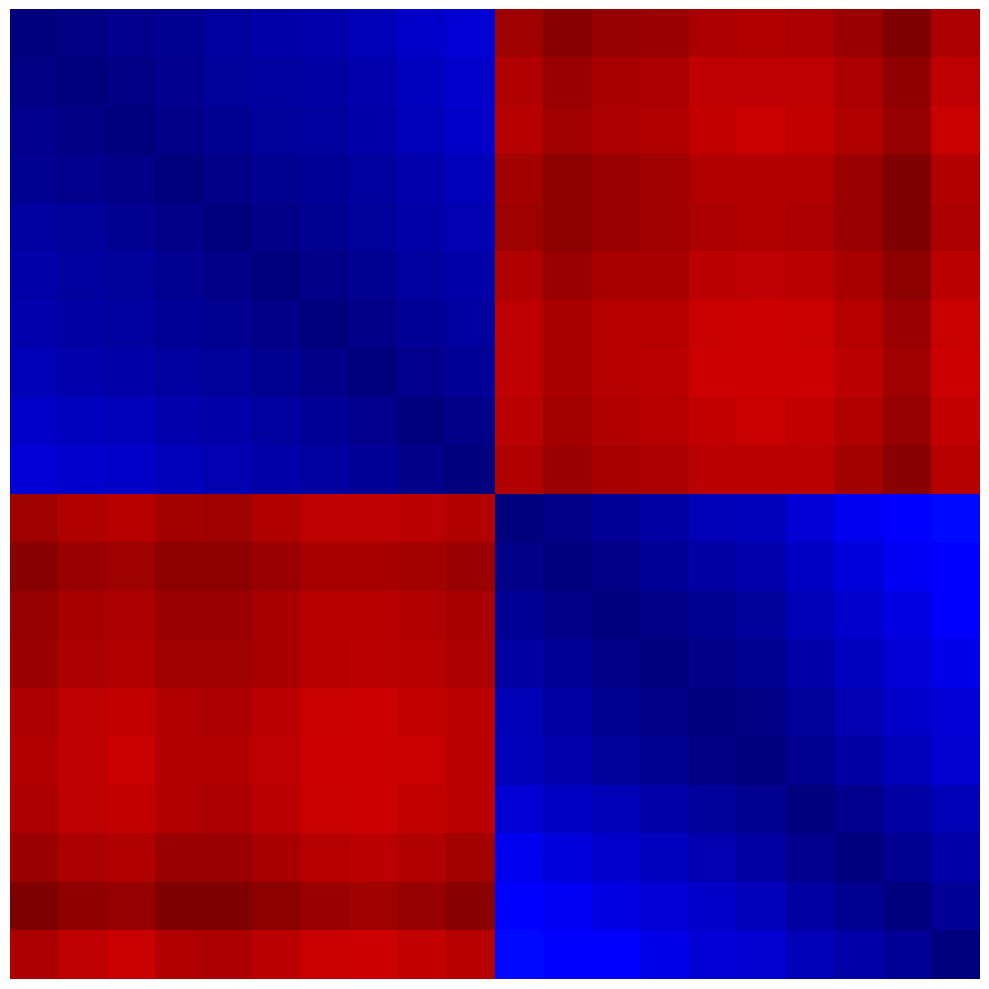} &
	\includegraphics[width=0.12\columnwidth,keepaspectratio]{figs/toy_dataset/TrainPoints/ImageAug/dismat_viewer_train_points_cluster_image_aug_2by10_cltsize01_numsample20.png} &
	\includegraphics[width=0.12\columnwidth,keepaspectratio]{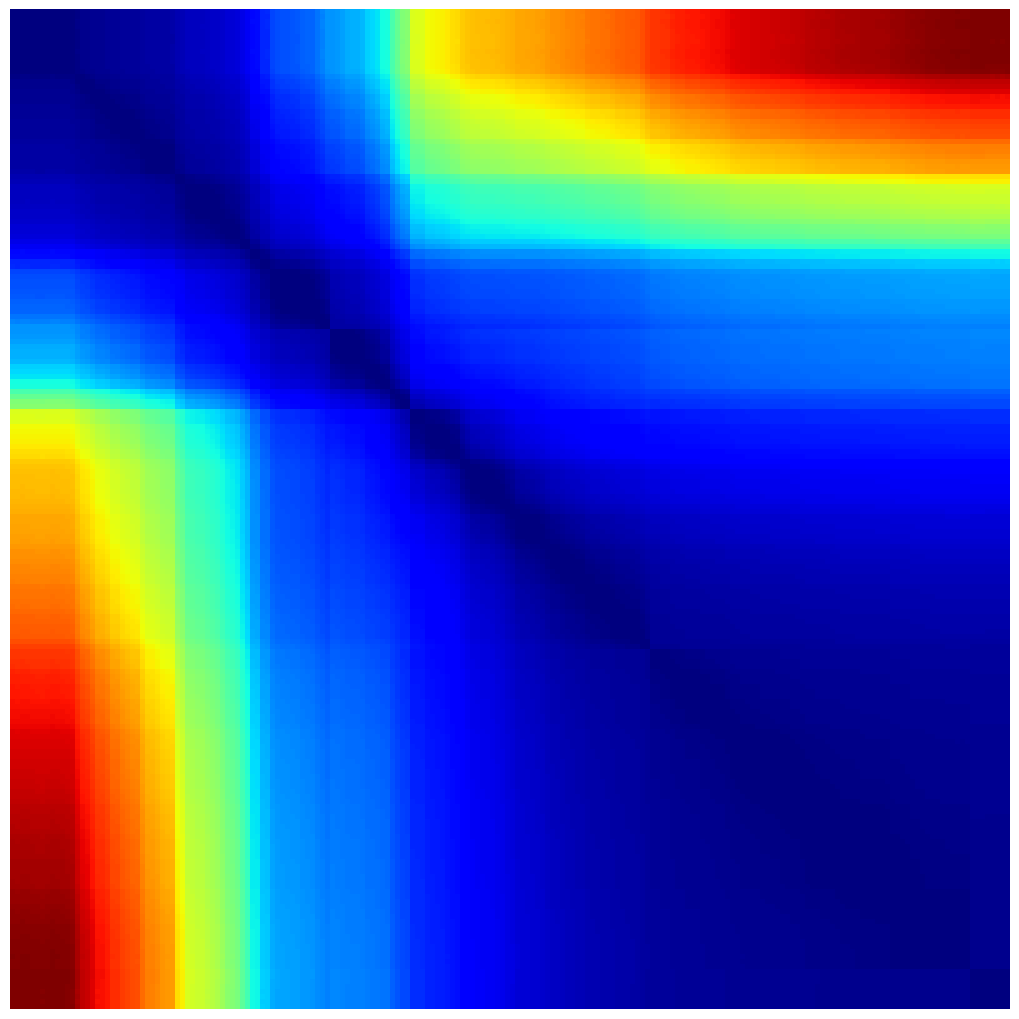} &  
	\includegraphics[width=0.12\columnwidth,keepaspectratio]{figs/toy_dataset/TrainImage/ShapeAug/dismat_viewer_train_image_cluster_shape_aug_2by10_cltsize01_numsample20.png} &
	\includegraphics[width=0.12\columnwidth,keepaspectratio]{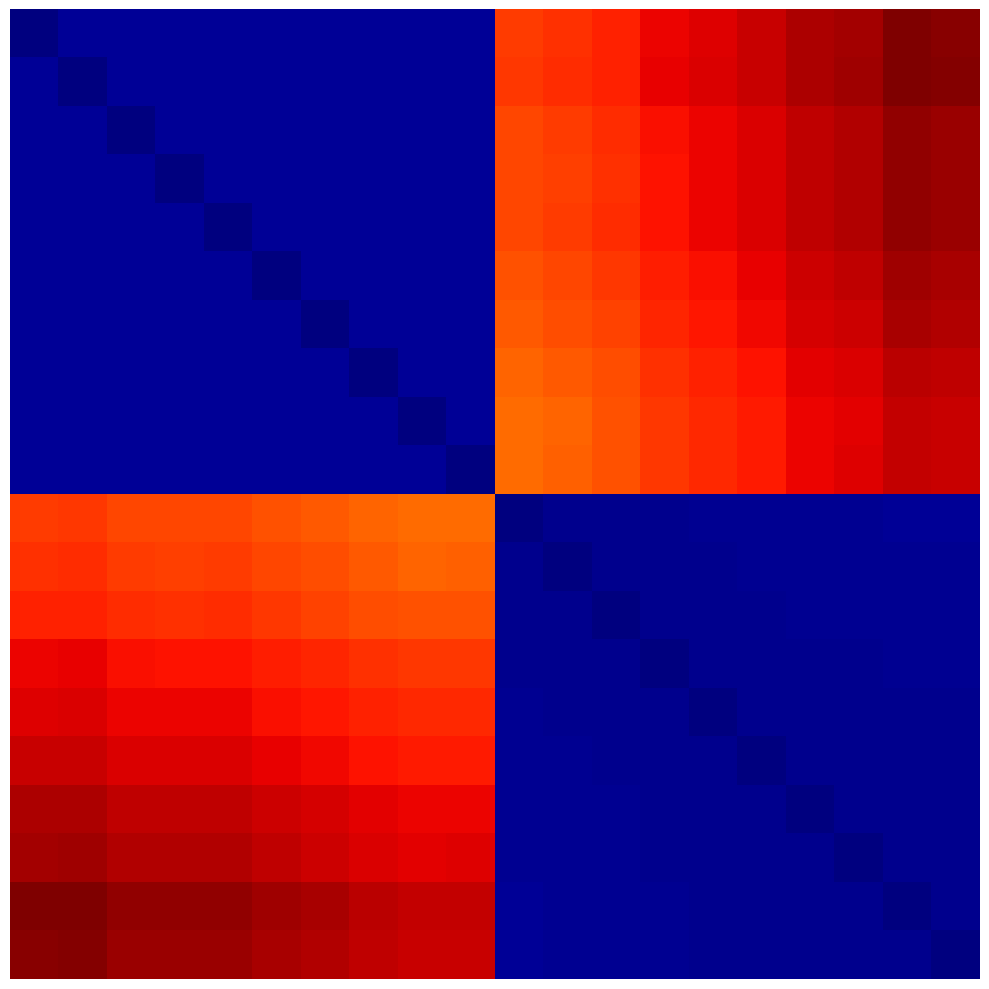} &
	\includegraphics[width=0.12\columnwidth,keepaspectratio]{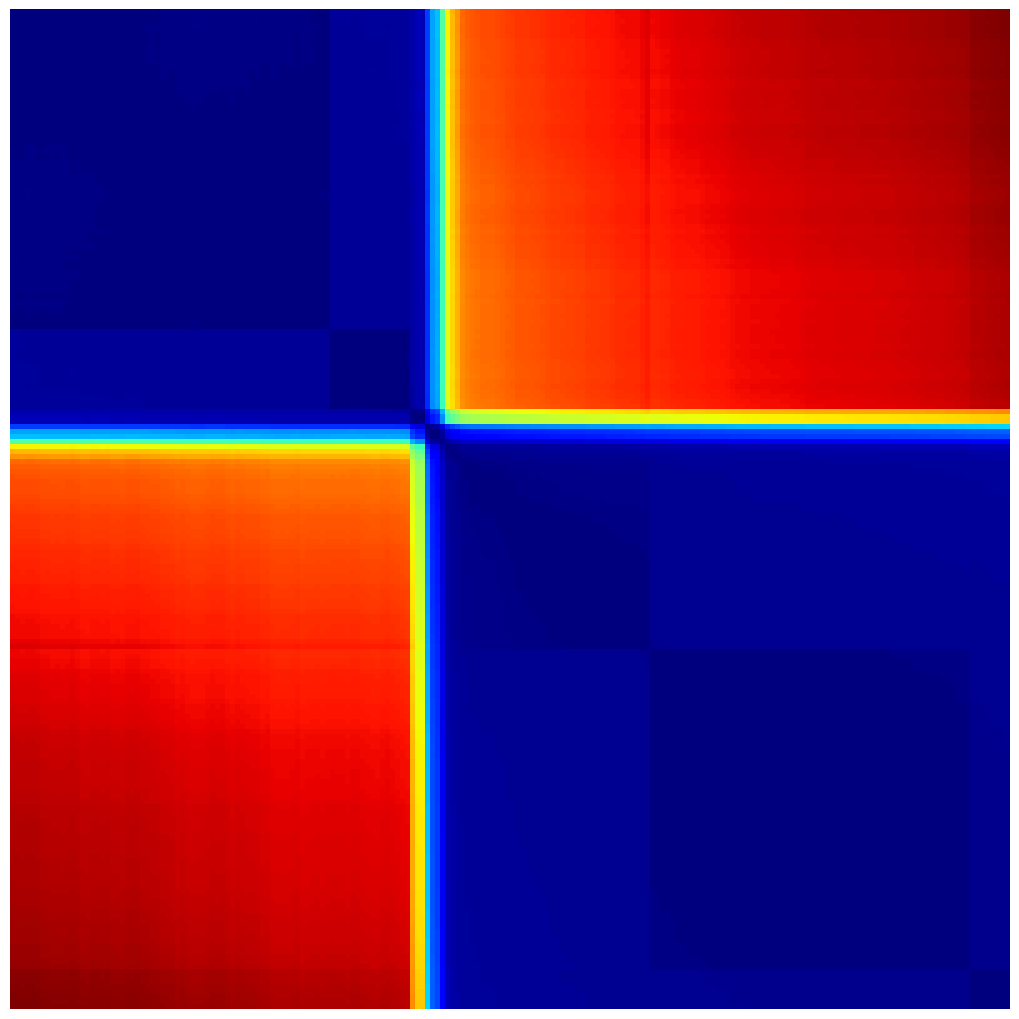}& \small{$S_{2}I_{8}$}\\
	\small{$S_{1}I_{3}$} &\includegraphics[width=0.12\columnwidth,keepaspectratio]{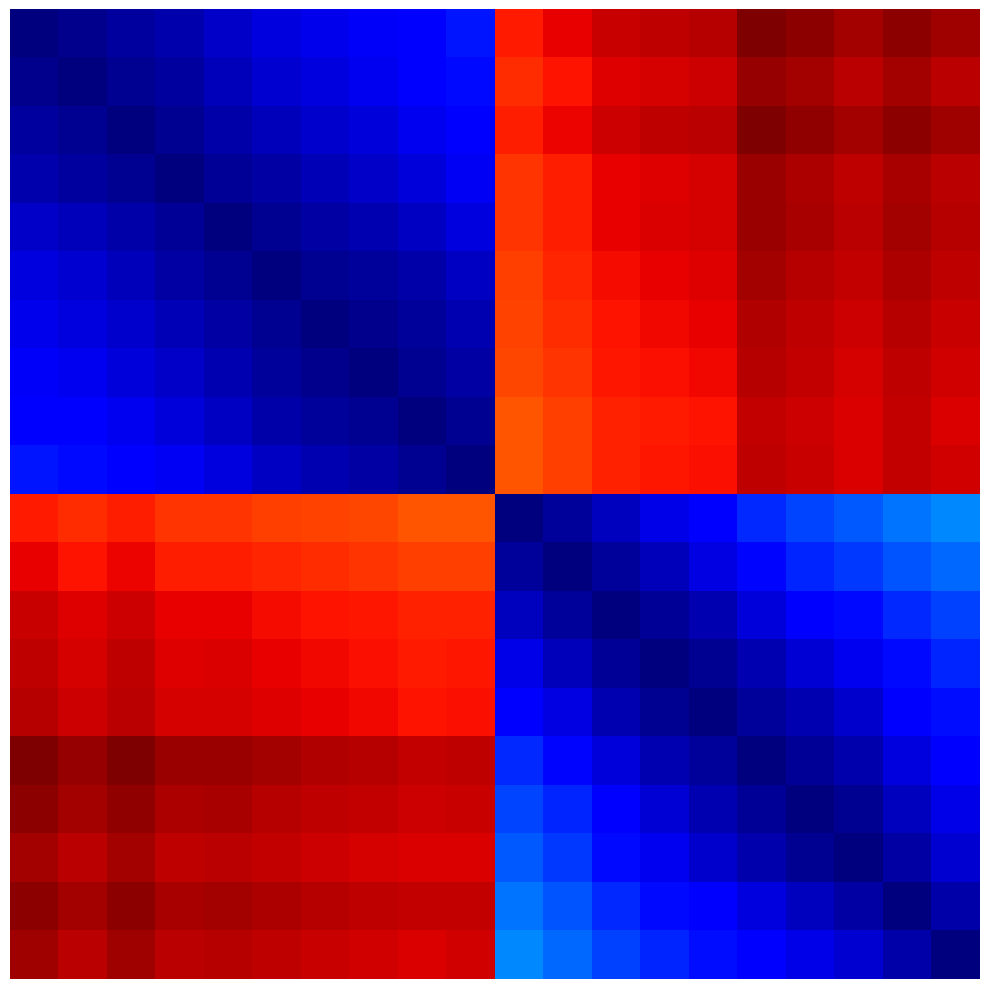} &
	\includegraphics[width=0.12\columnwidth,keepaspectratio]{figs/toy_dataset/TrainPoints/ImageAug/dismat_viewer_train_points_cluster_image_aug_2by10_cltsize01_numsample20.png} &
	\includegraphics[width=0.12\columnwidth,keepaspectratio]{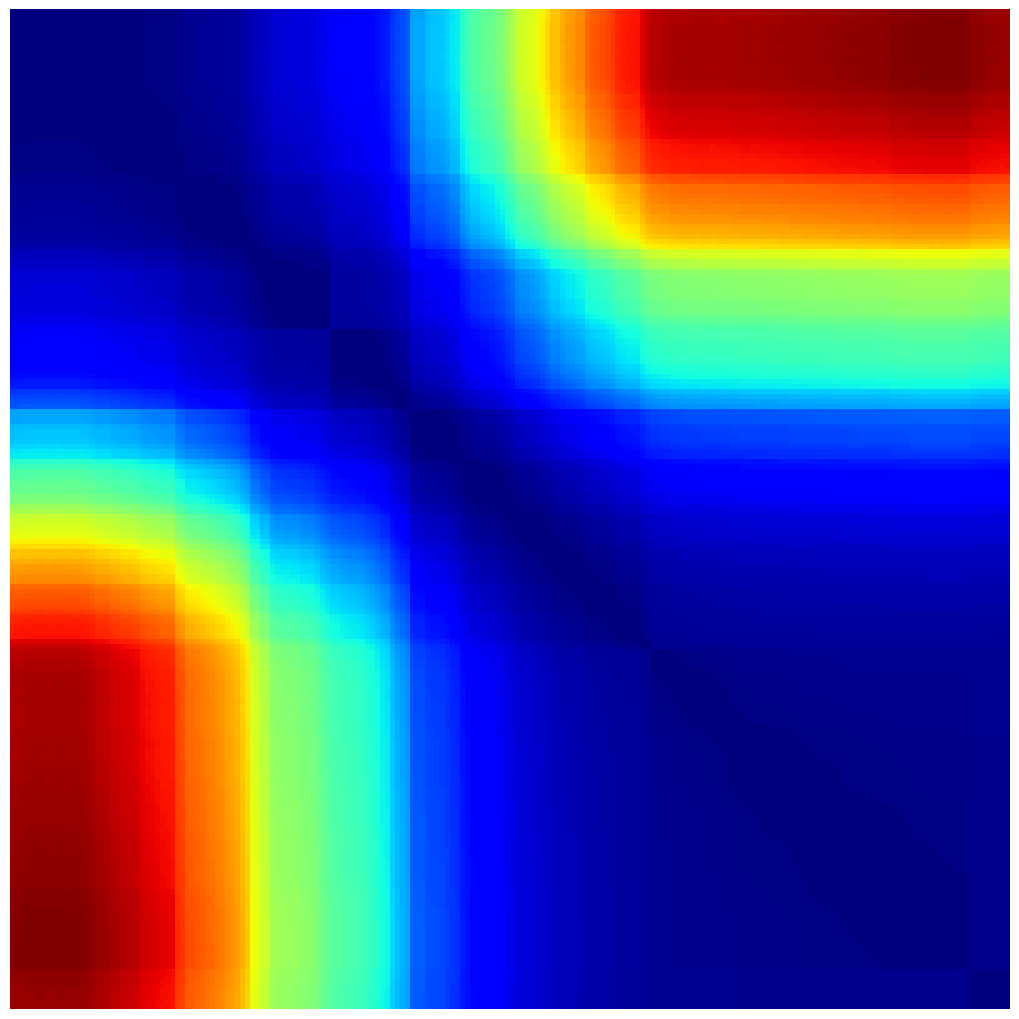} &  
	\includegraphics[width=0.12\columnwidth,keepaspectratio]{figs/toy_dataset/TrainImage/ShapeAug/dismat_viewer_train_image_cluster_shape_aug_2by10_cltsize01_numsample20.png} &
	\includegraphics[width=0.12\columnwidth,keepaspectratio]{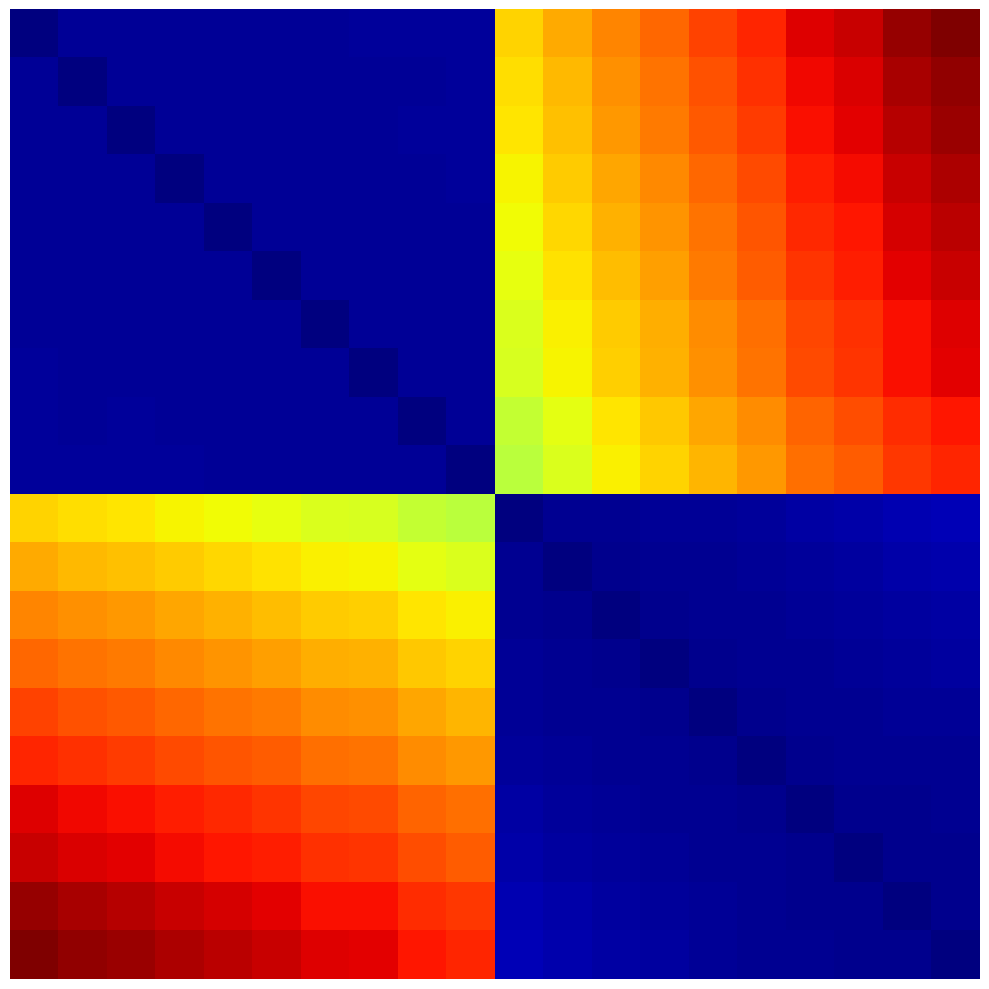} &
	\includegraphics[width=0.12\columnwidth,keepaspectratio]{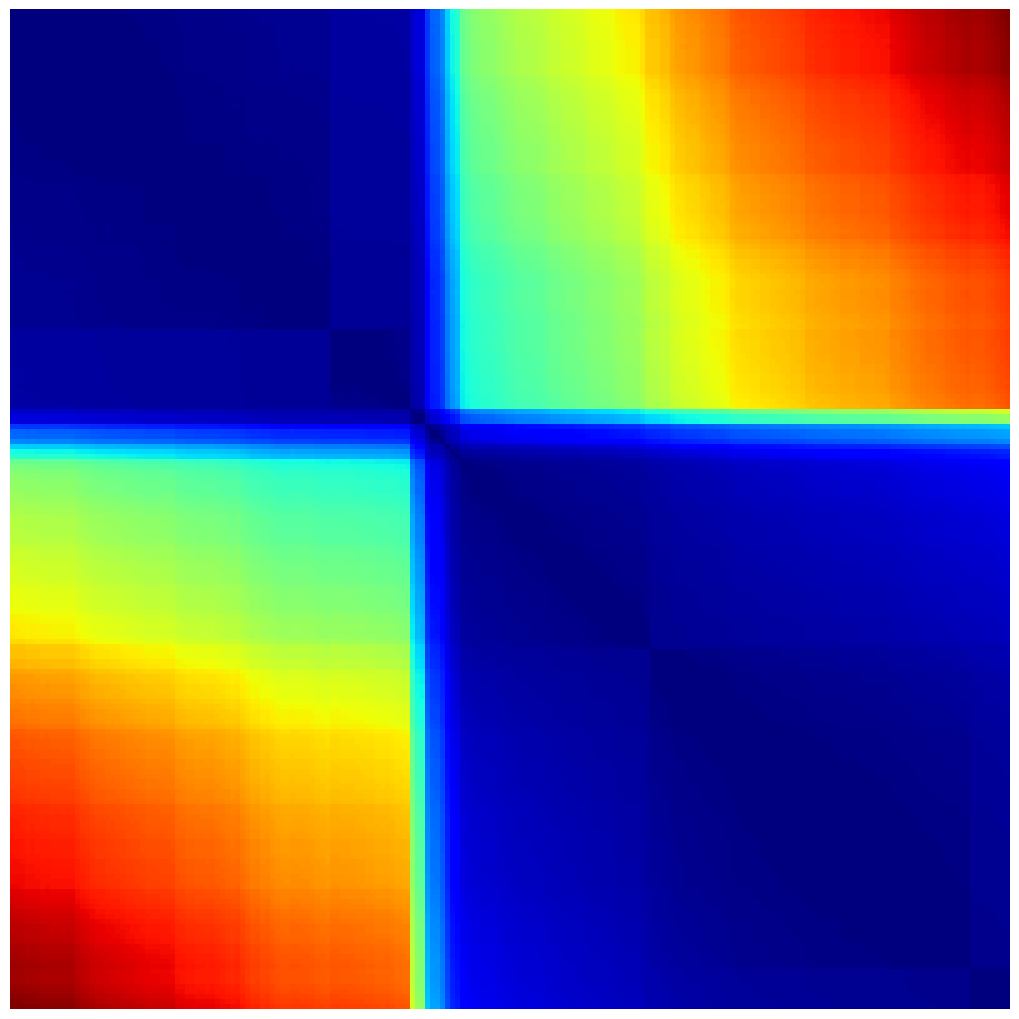} & \small{$S_{3}I_{8}$}\\
	\small{$S_{1}I_{4}$} &\includegraphics[width=0.12\columnwidth,keepaspectratio]{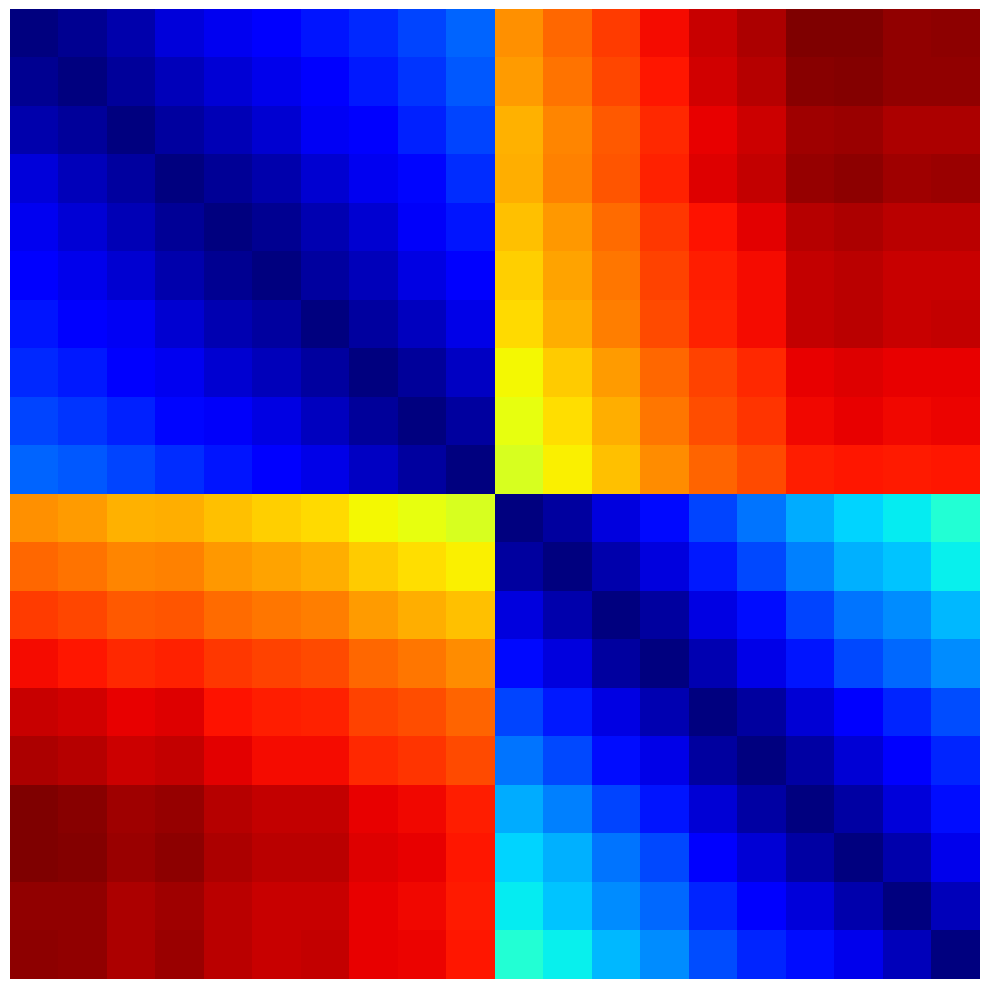} &
	\includegraphics[width=0.12\columnwidth,keepaspectratio]{figs/toy_dataset/TrainPoints/ImageAug/dismat_viewer_train_points_cluster_image_aug_2by10_cltsize01_numsample20.png} &
	\includegraphics[width=0.12\columnwidth,keepaspectratio]{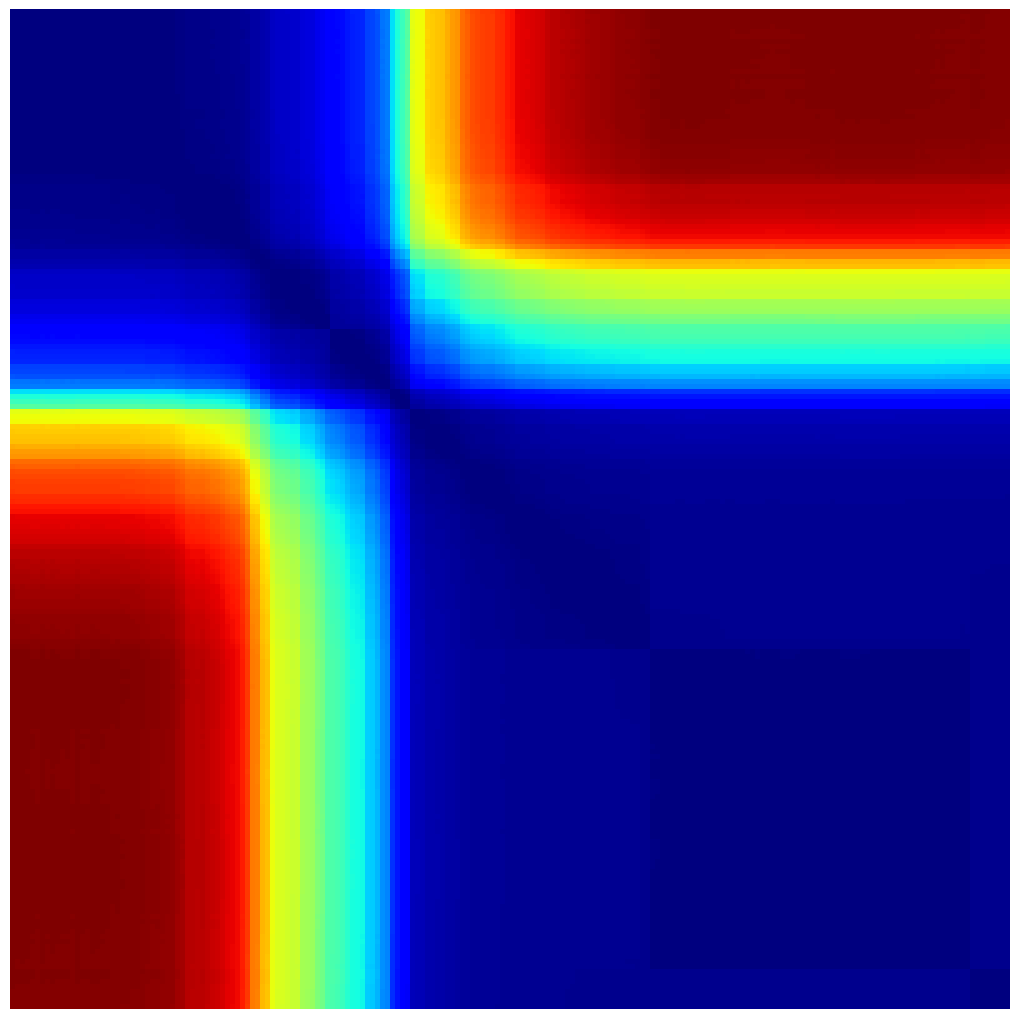} &  
	\includegraphics[width=0.12\columnwidth,keepaspectratio]{figs/toy_dataset/TrainImage/ShapeAug/dismat_viewer_train_image_cluster_shape_aug_2by10_cltsize01_numsample20.png} &
	\includegraphics[width=0.12\columnwidth,keepaspectratio]{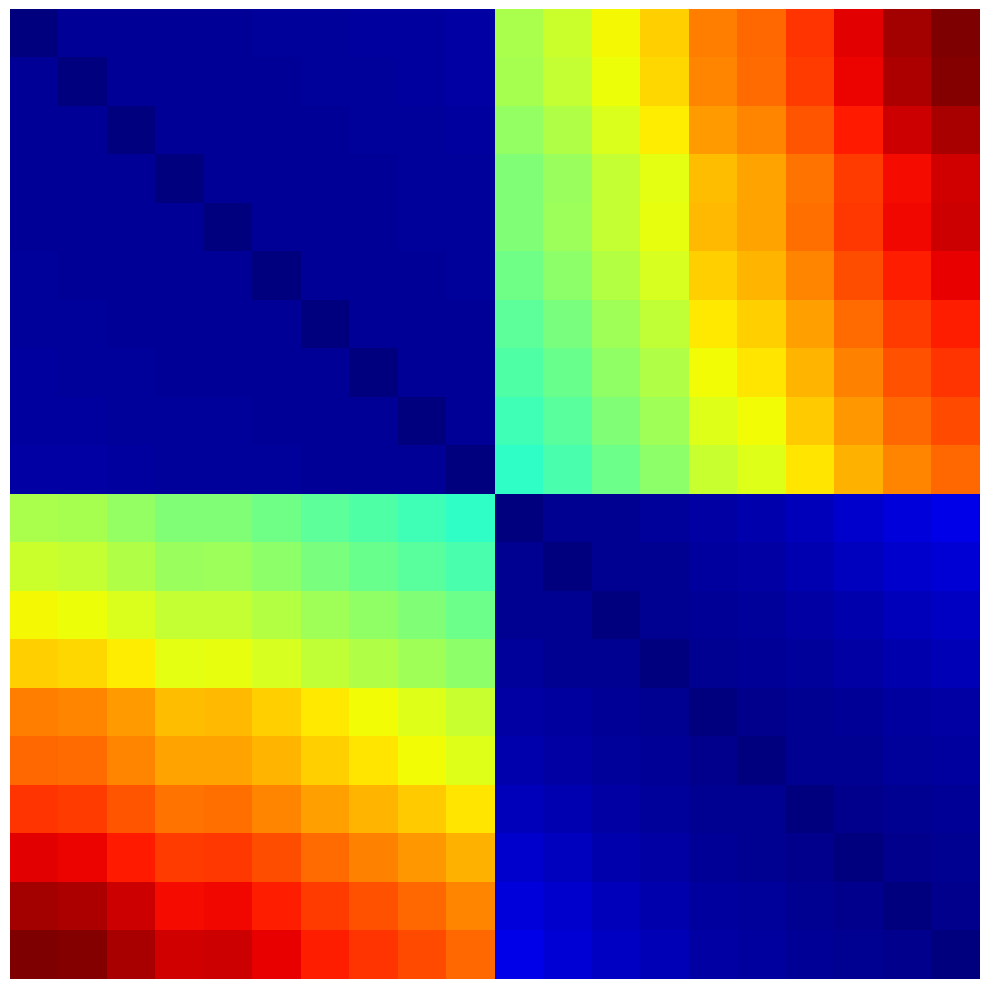} &
	\includegraphics[width=0.12\columnwidth,keepaspectratio]{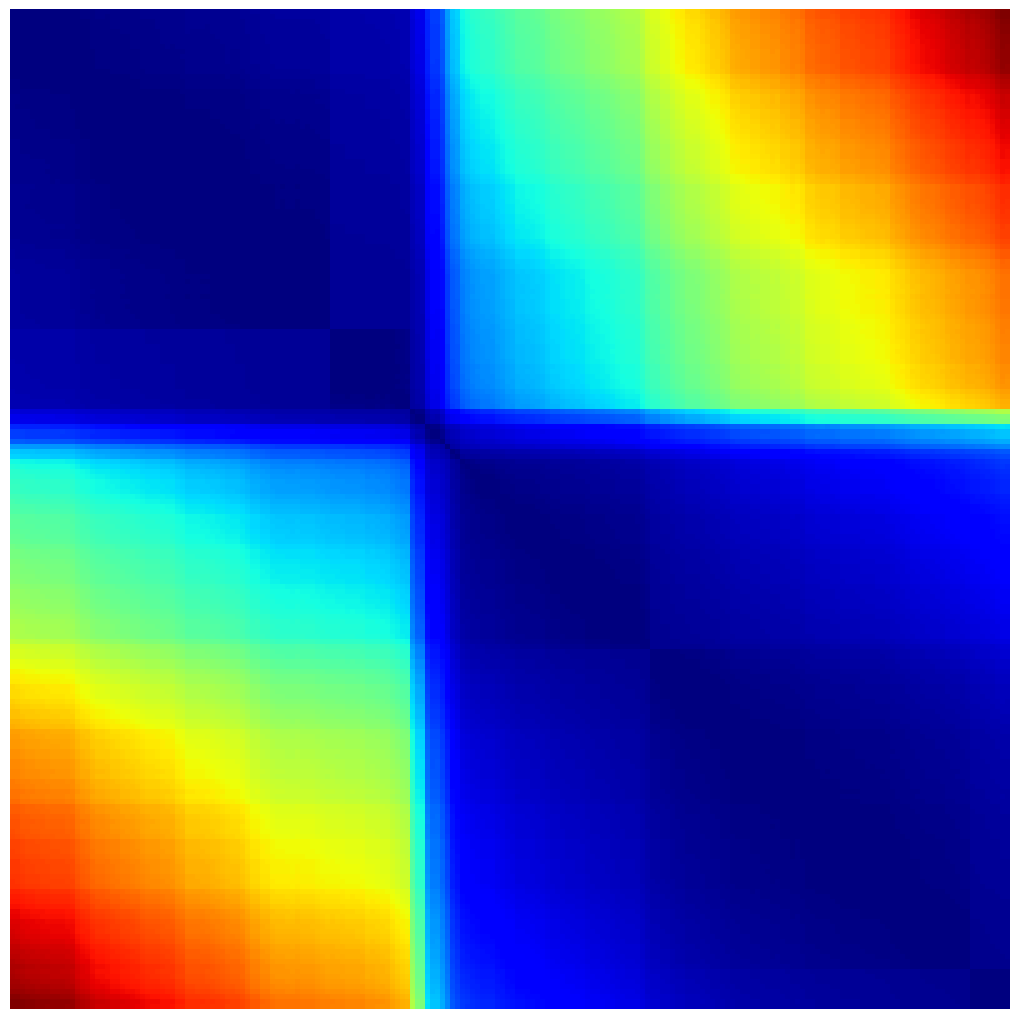} & \small{$S_{4}I_{8}$}\\
	\small{$S_{1}I_{5}$} &\includegraphics[width=0.12\columnwidth,keepaspectratio]{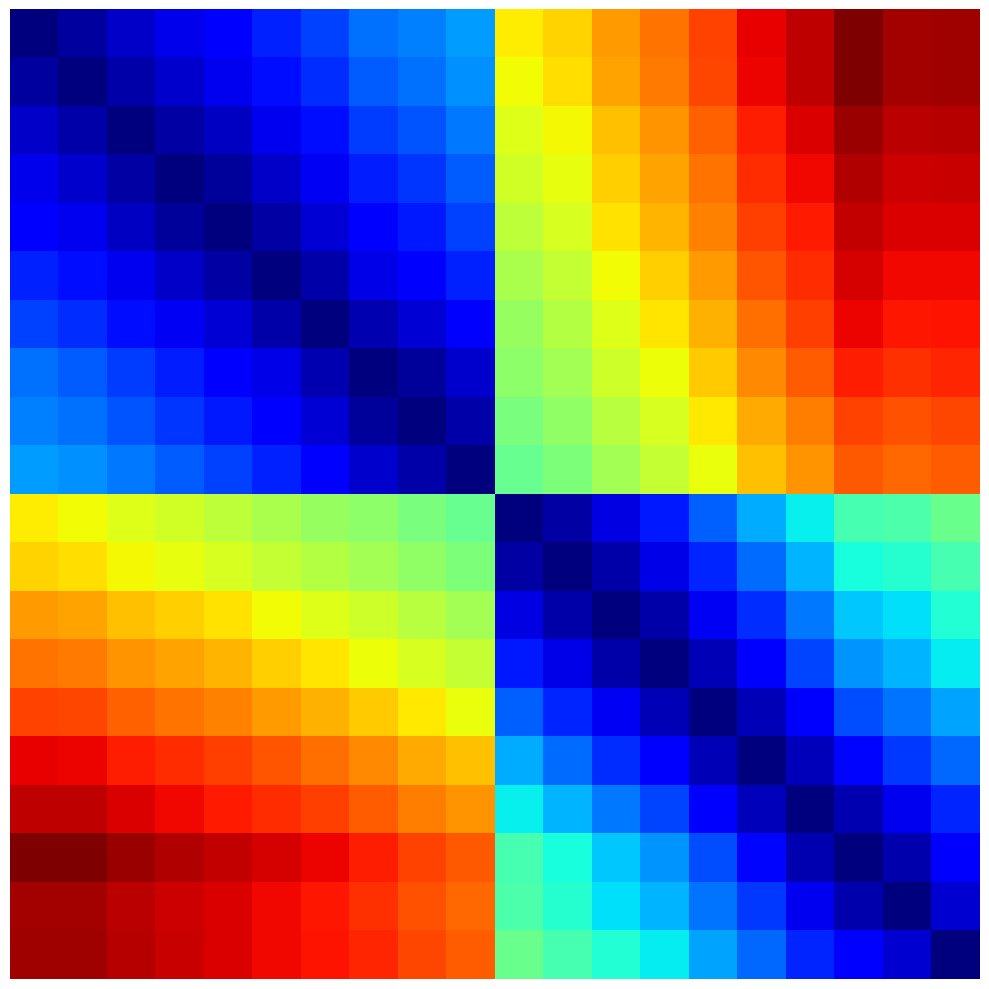} &
	\includegraphics[width=0.12\columnwidth,keepaspectratio]{figs/toy_dataset/TrainPoints/ImageAug/dismat_viewer_train_points_cluster_image_aug_2by10_cltsize01_numsample20.png} &
	\includegraphics[width=0.12\columnwidth,keepaspectratio]{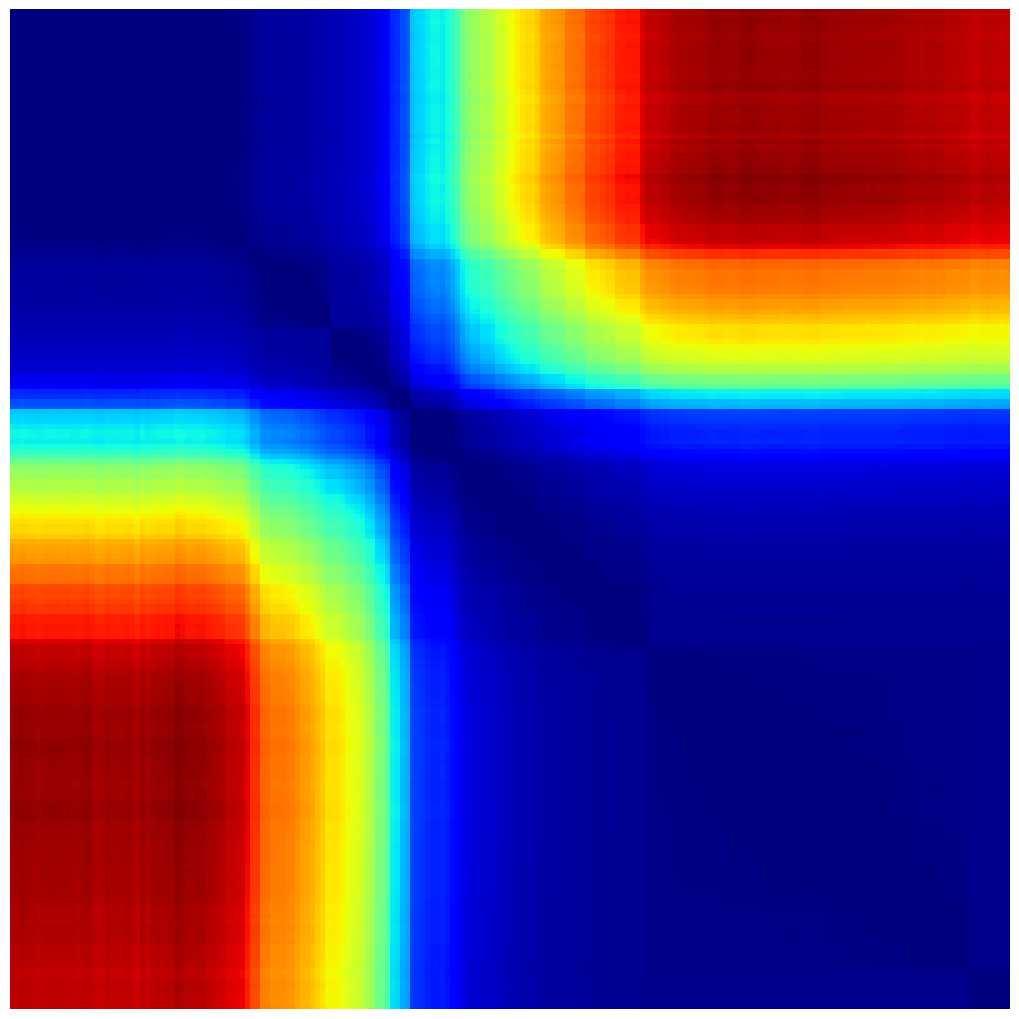} &  
	\includegraphics[width=0.12\columnwidth,keepaspectratio]{figs/toy_dataset/TrainImage/ShapeAug/dismat_viewer_train_image_cluster_shape_aug_2by10_cltsize01_numsample20.png} &
	\includegraphics[width=0.12\columnwidth,keepaspectratio]{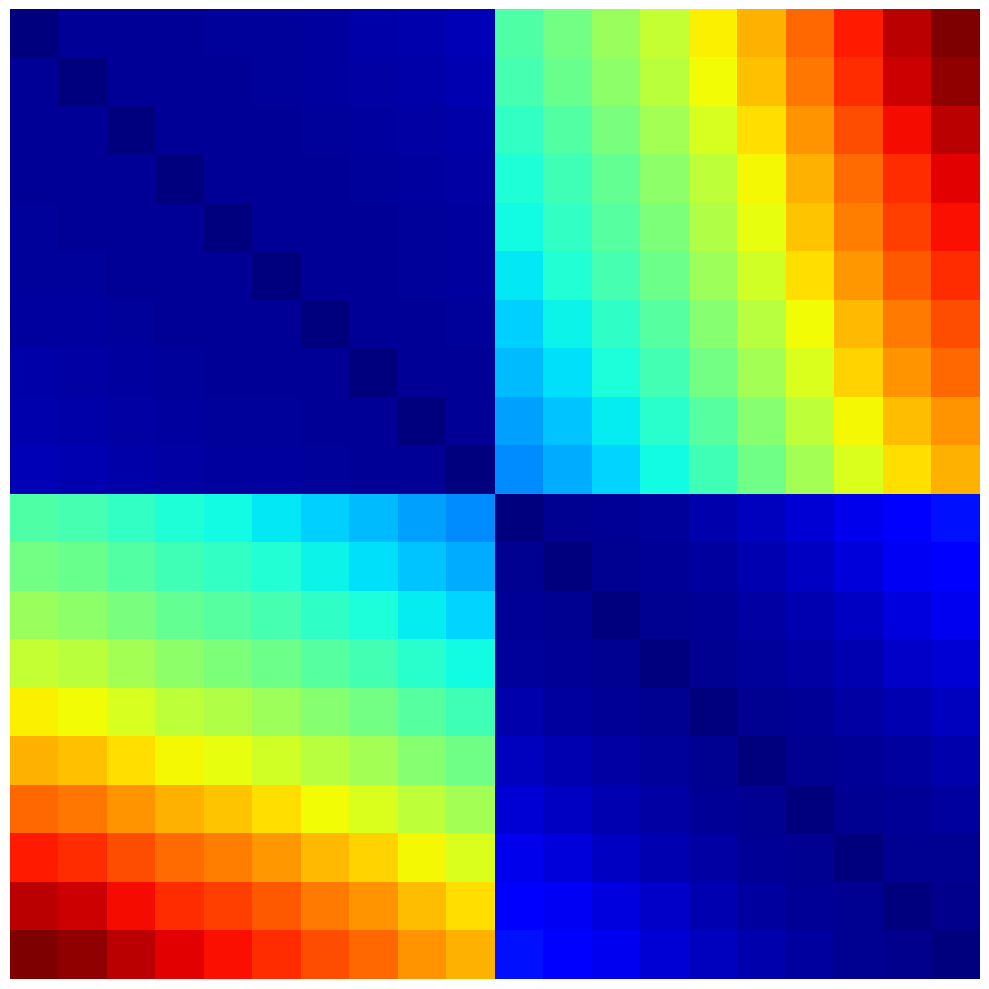} &
	\includegraphics[width=0.12\columnwidth,keepaspectratio]{figs/toy_dataset/Prediction/ShapeAug/toydata_lastepoch_viewer_pred_points_200_20210222_1847_cluster_shape_aug_2by10_cltsize20_seed1_BS8_Epoch1200.png}& \small{$S_{5}I_{8}$}\\
	\small{$S_{1}I_{6}$} &\includegraphics[width=0.12\columnwidth,keepaspectratio]{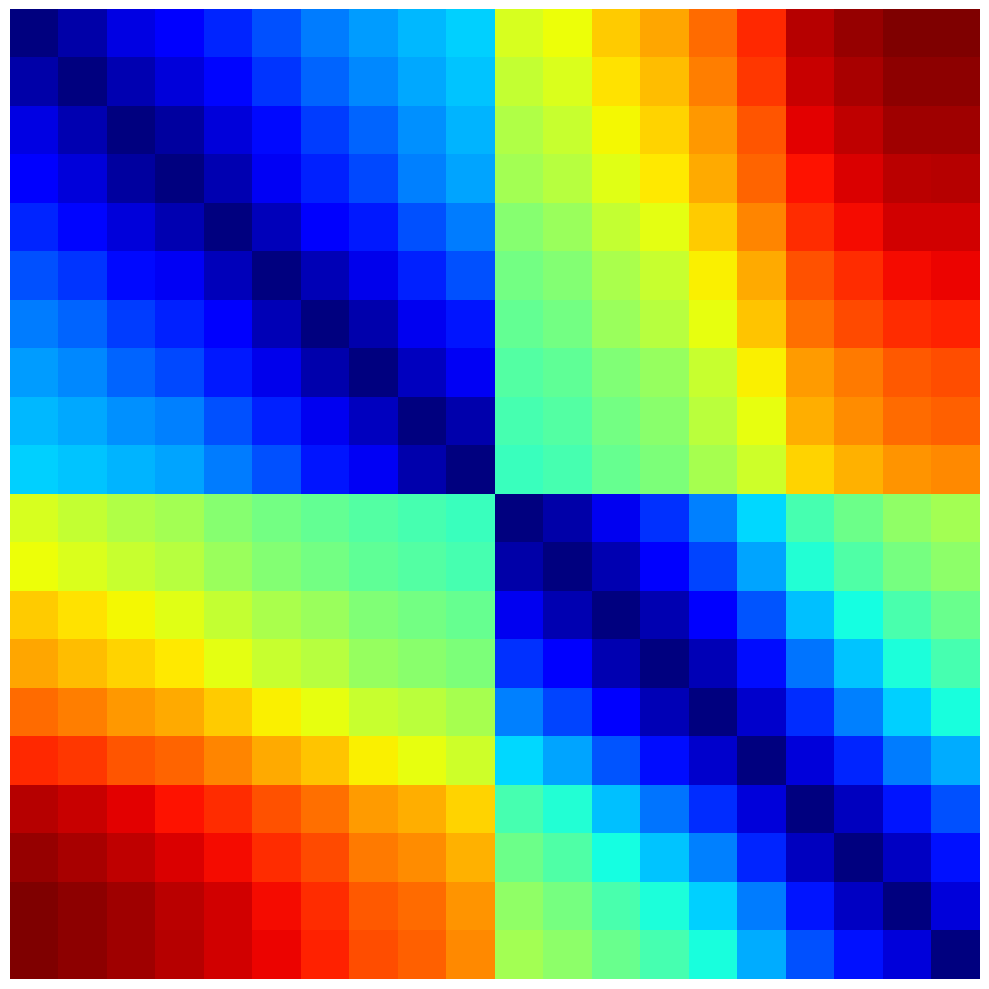} &
	\includegraphics[width=0.12\columnwidth,keepaspectratio]{figs/toy_dataset/TrainPoints/ImageAug/dismat_viewer_train_points_cluster_image_aug_2by10_cltsize01_numsample20.png} &
	\includegraphics[width=0.12\columnwidth,keepaspectratio]{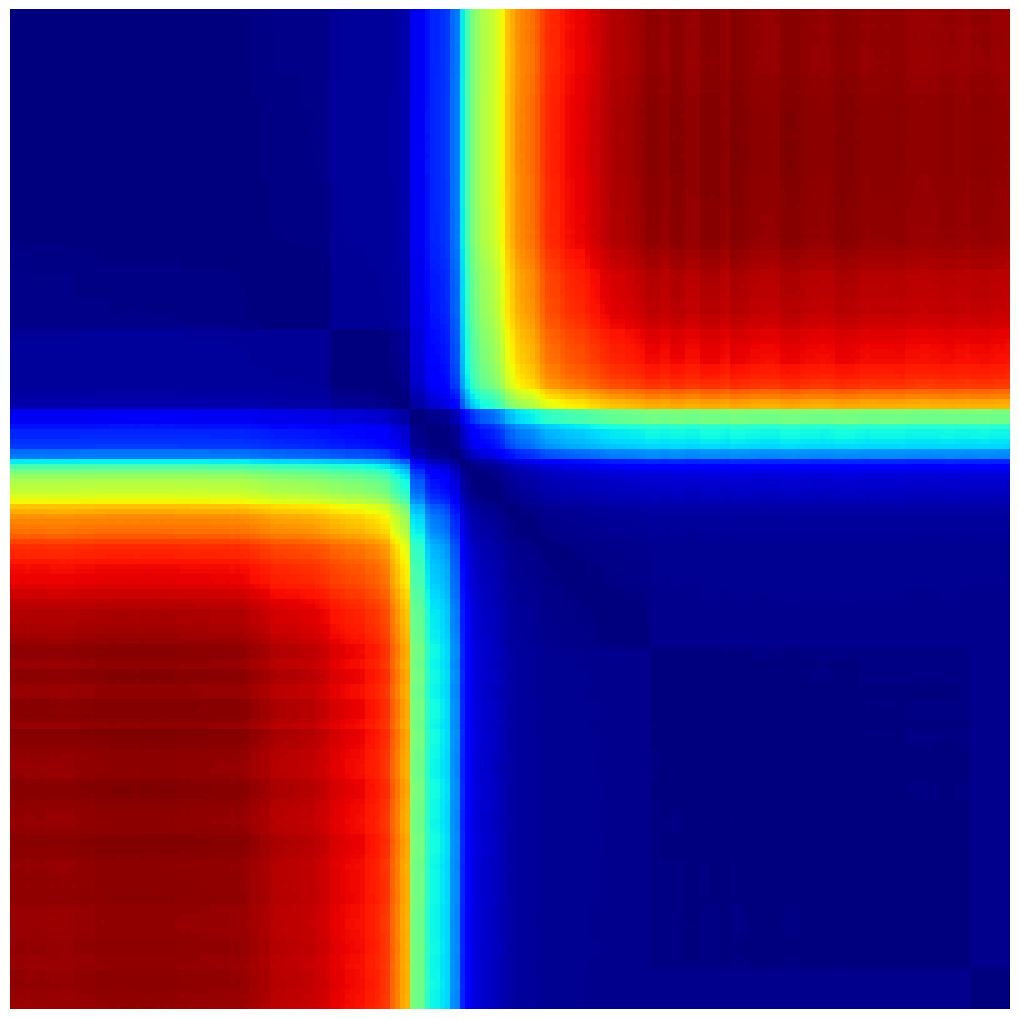} &  
	\includegraphics[width=0.12\columnwidth,keepaspectratio]{figs/toy_dataset/TrainImage/ShapeAug/dismat_viewer_train_image_cluster_shape_aug_2by10_cltsize01_numsample20.png} &
	\includegraphics[width=0.12\columnwidth,keepaspectratio]{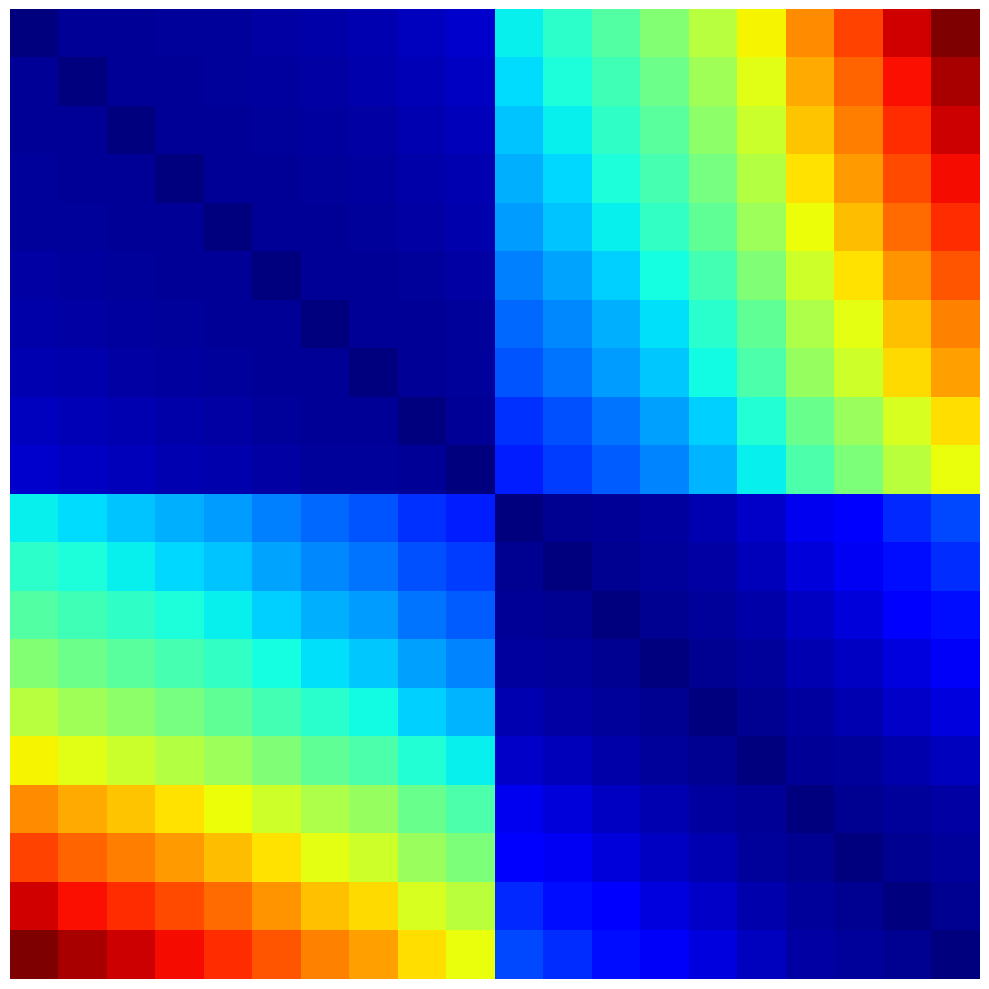} &
	\includegraphics[width=0.12\columnwidth,keepaspectratio]{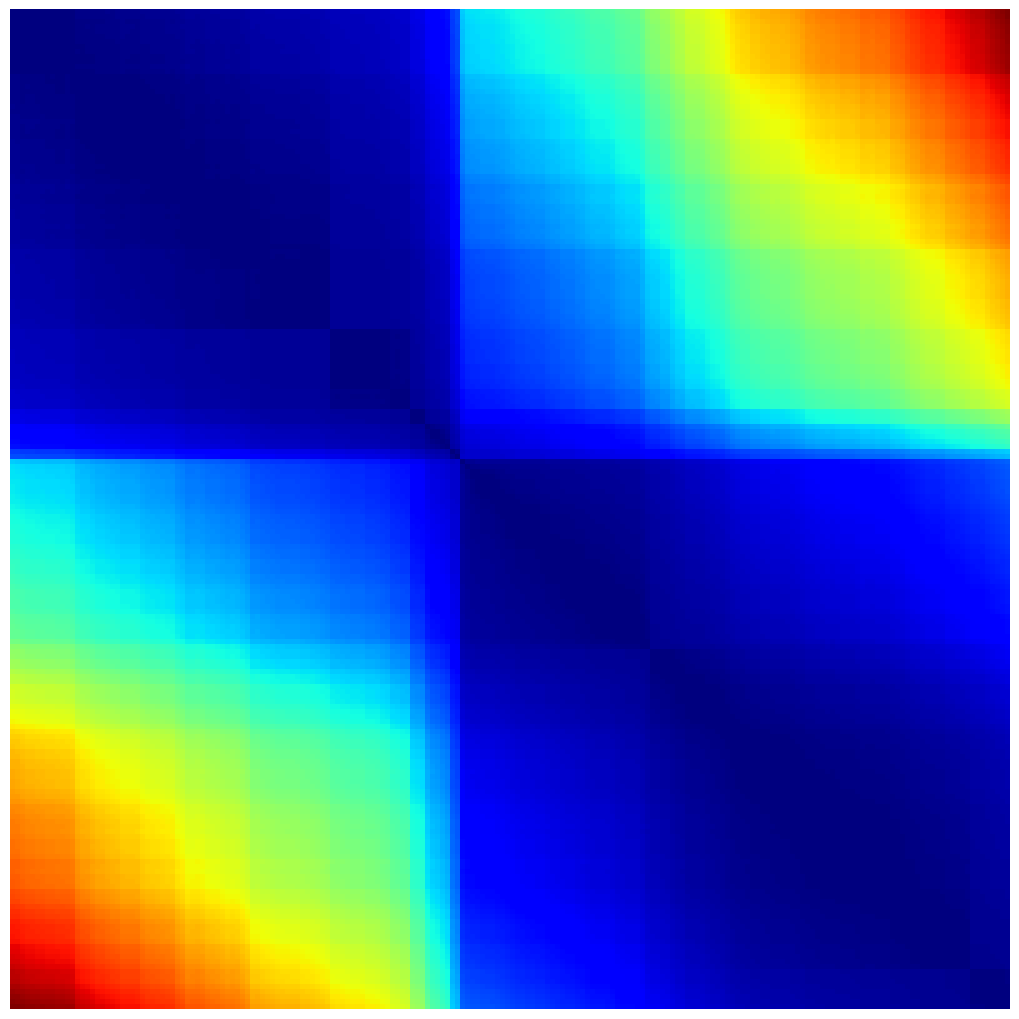}& \small{$S_{6}I_{8}$}\\
	\small{$S_{1}I_{7}$} &\includegraphics[width=0.12\columnwidth,keepaspectratio]{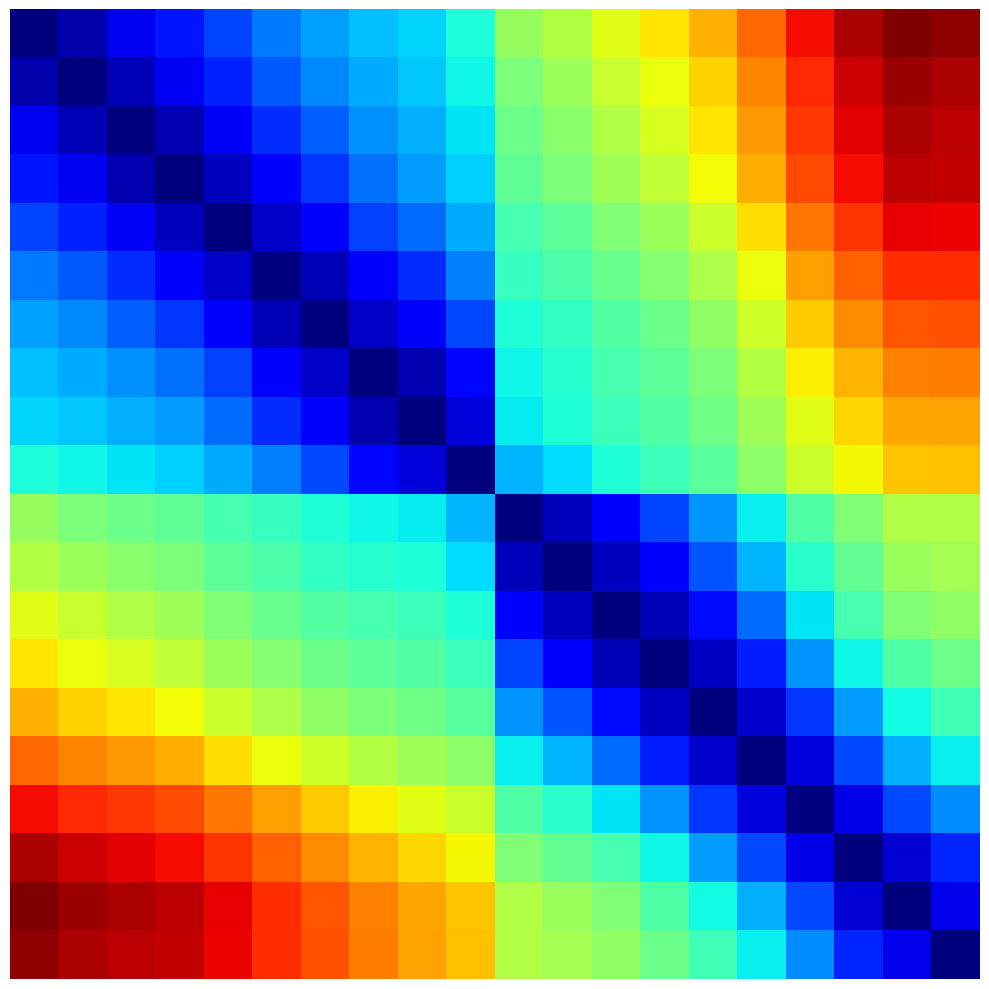} &
	\includegraphics[width=0.12\columnwidth,keepaspectratio]{figs/toy_dataset/TrainPoints/ImageAug/dismat_viewer_train_points_cluster_image_aug_2by10_cltsize01_numsample20.png} &
	\includegraphics[width=0.12\columnwidth,keepaspectratio]{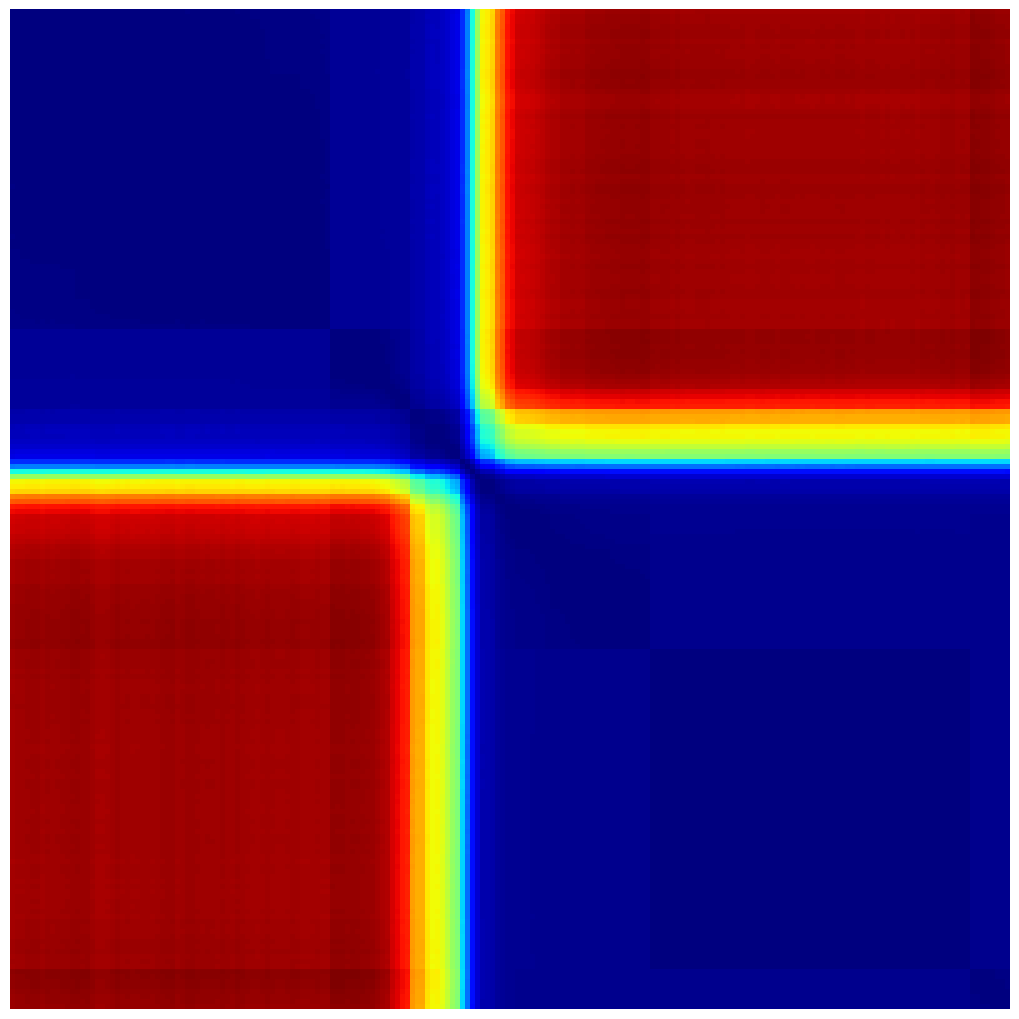} &  
	\includegraphics[width=0.12\columnwidth,keepaspectratio]{figs/toy_dataset/TrainImage/ShapeAug/dismat_viewer_train_image_cluster_shape_aug_2by10_cltsize01_numsample20.png} &
	\includegraphics[width=0.12\columnwidth,keepaspectratio]{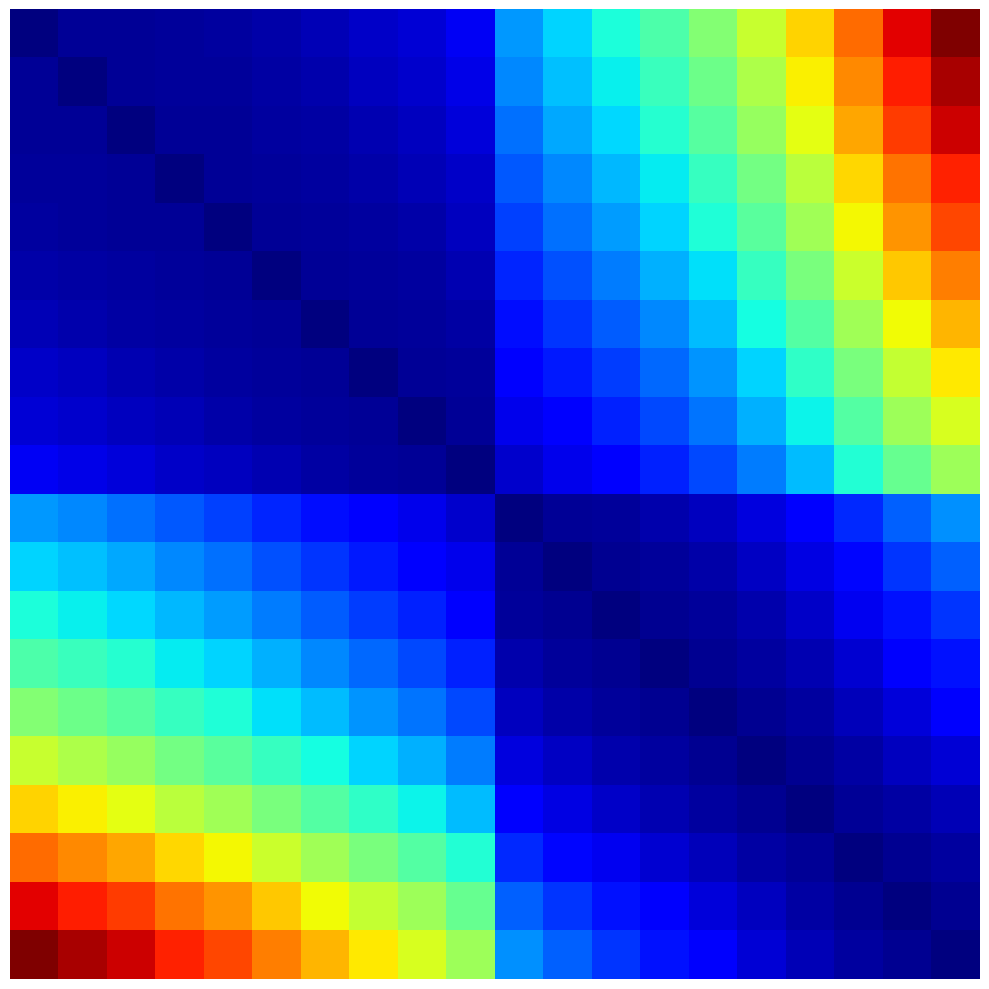} &
	\includegraphics[width=0.12\columnwidth,keepaspectratio]{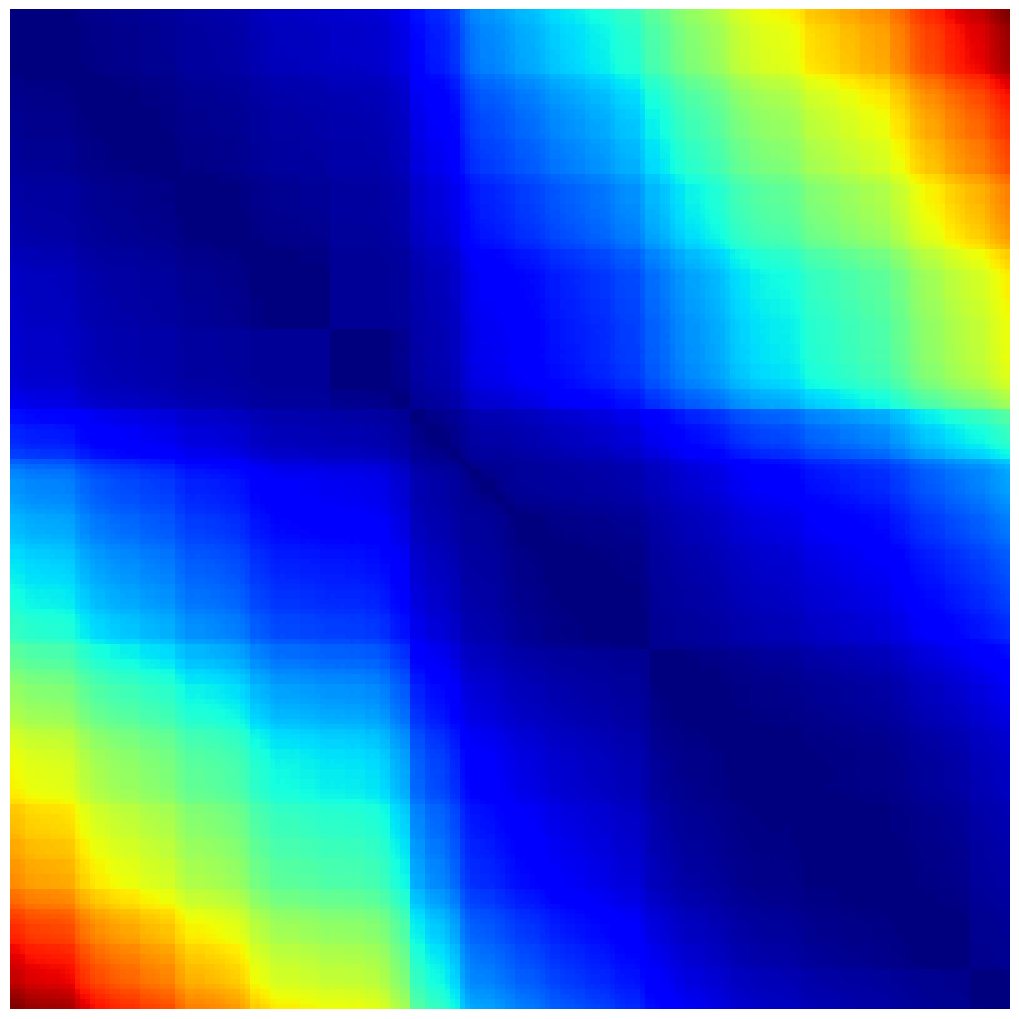} & \small{$S_{7}I_{8}$}\\
	
	\small{$S_{1}I_{8}$} &\includegraphics[width=0.12\columnwidth,keepaspectratio]{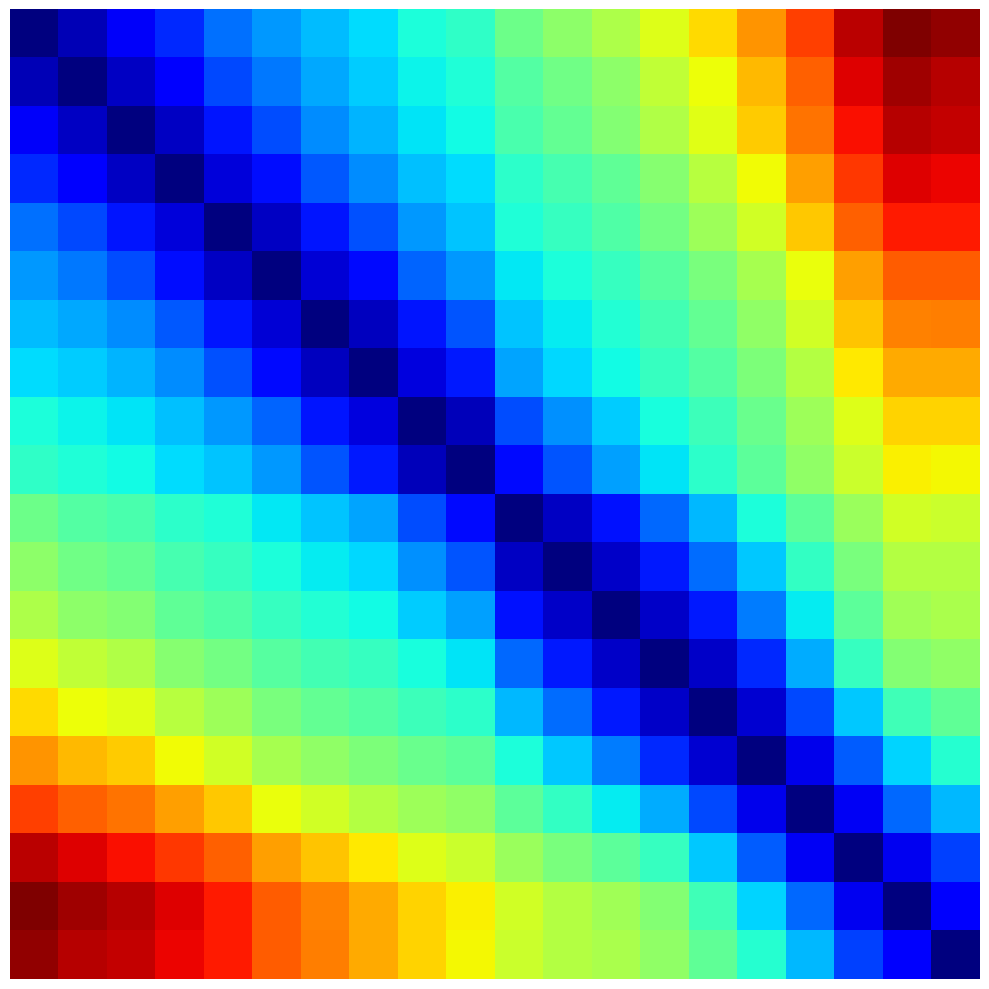} &
	\includegraphics[width=0.12\columnwidth,keepaspectratio]{figs/toy_dataset/TrainPoints/ImageAug/dismat_viewer_train_points_cluster_image_aug_2by10_cltsize01_numsample20.png} &
	\includegraphics[width=0.12\columnwidth,keepaspectratio]{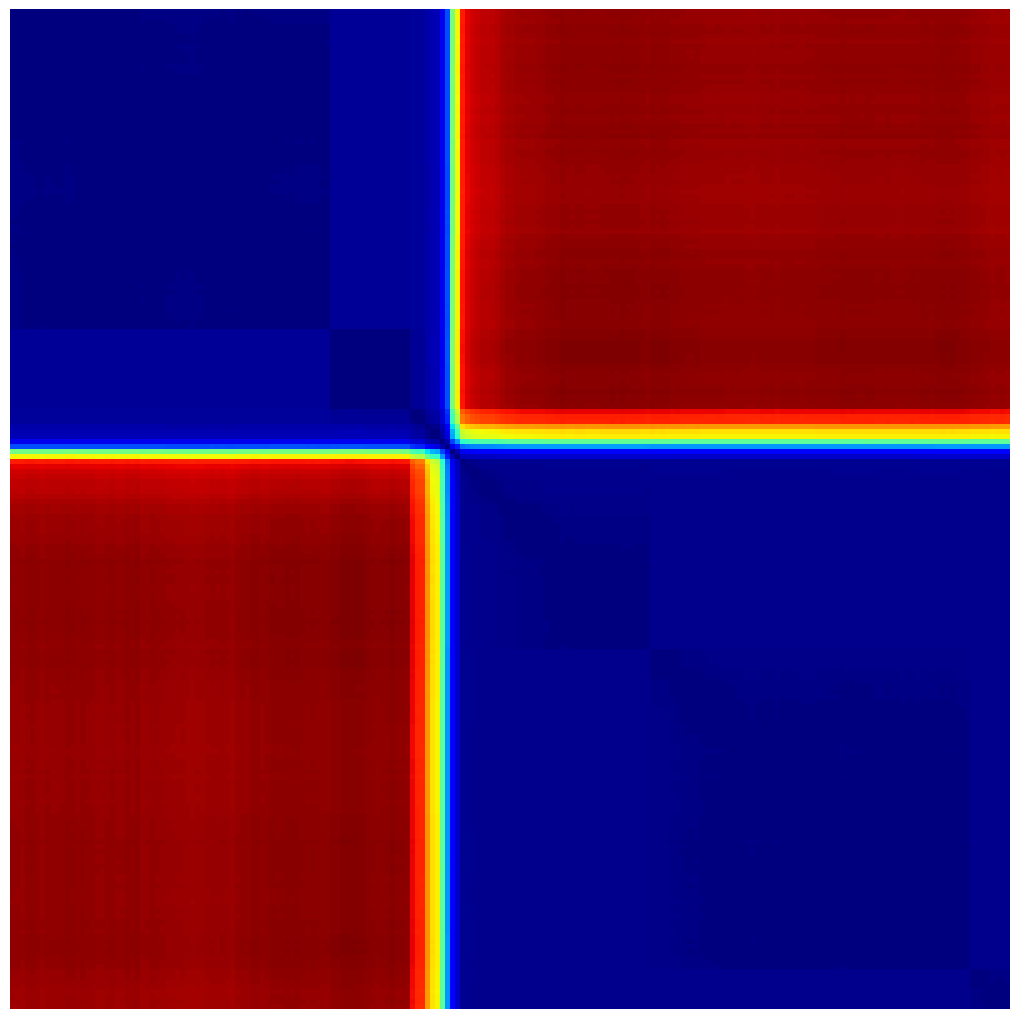} &  
	\includegraphics[width=0.12\columnwidth,keepaspectratio]{figs/toy_dataset/TrainImage/ShapeAug/dismat_viewer_train_image_cluster_shape_aug_2by10_cltsize01_numsample20.png} &
	\includegraphics[width=0.12\columnwidth,keepaspectratio]{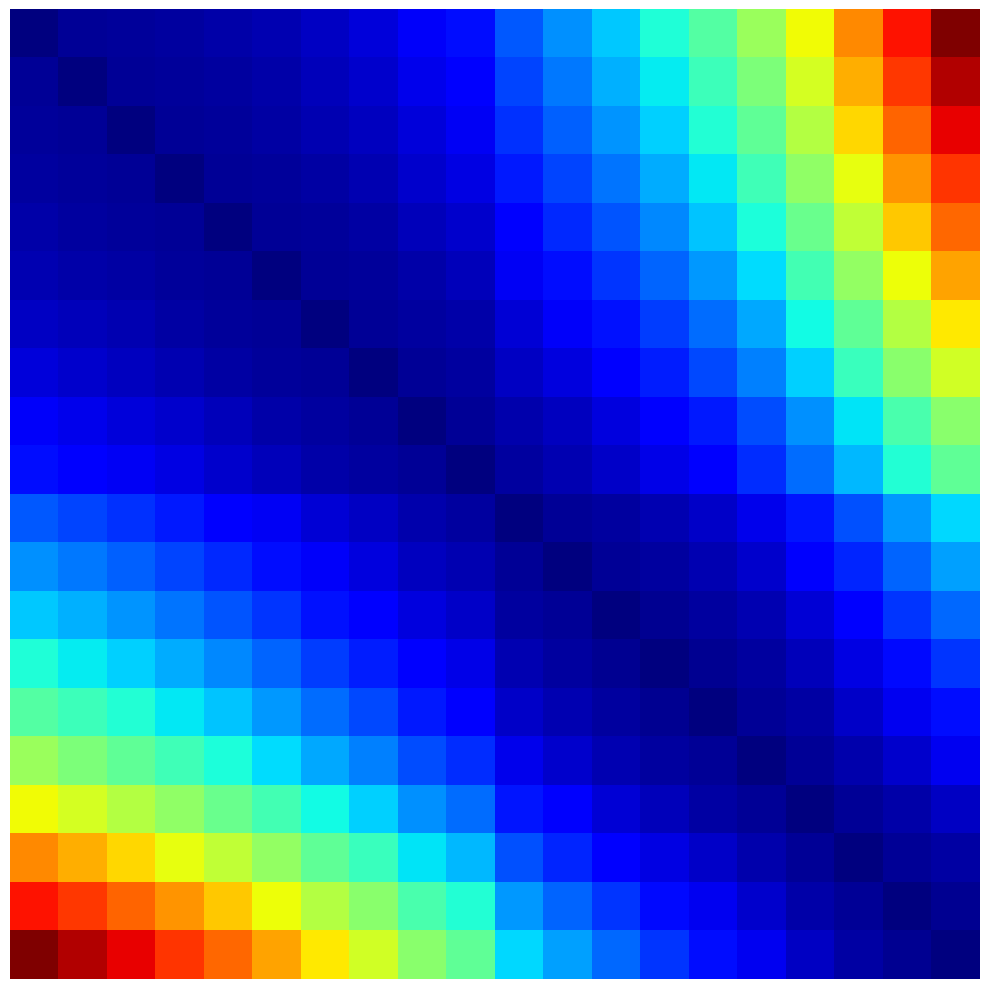} &
	\includegraphics[width=0.12\columnwidth,keepaspectratio]{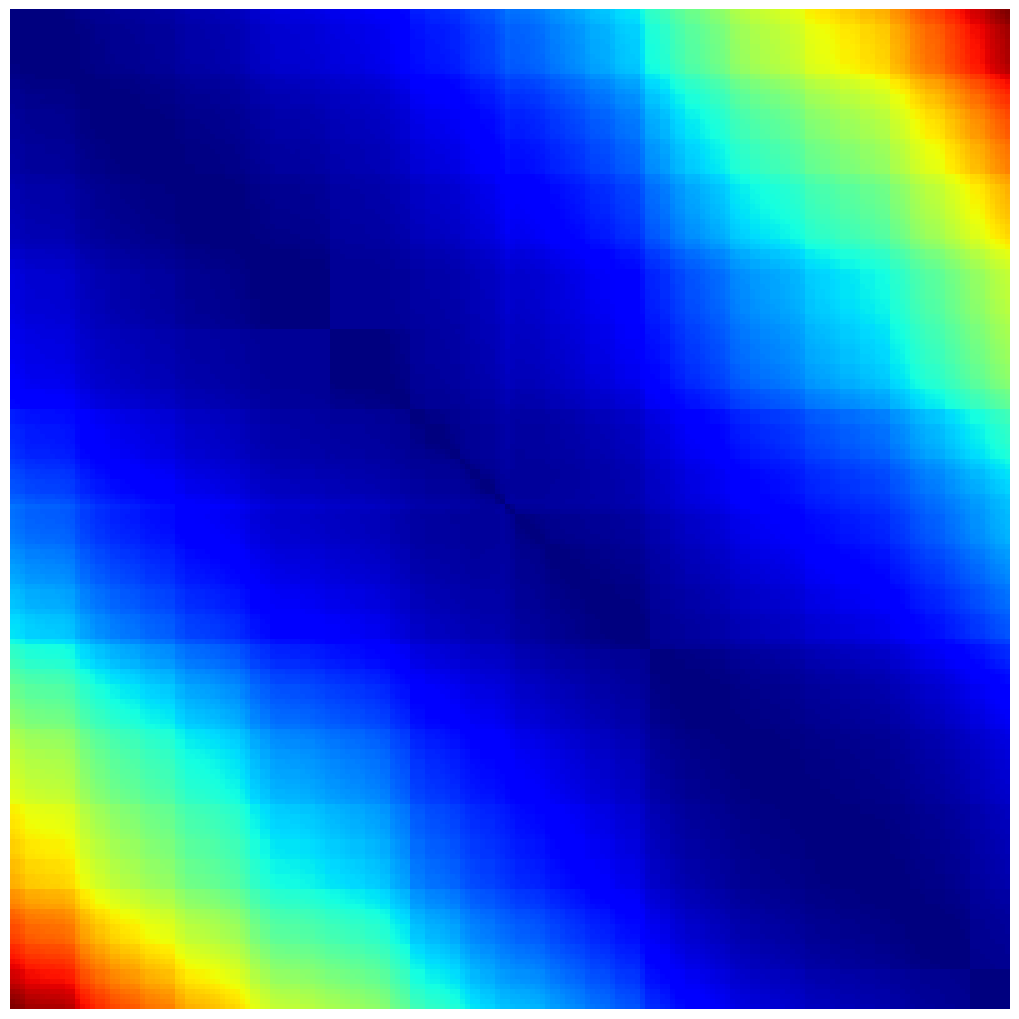} & \small{$S_{8}I_{8}$}\\ \hline
	\end{tabular} 
	\caption{Visualization of the distance matrices from experiments on sub-trainsets $S_1I_{i}$ and $S_{j}I_8$. Blue represents small distance and red represents large distance. 
	\emph{Images}: training images. \emph{Shapes}: training shapes. \emph{Recons}: trained models' reconstructed shapes on an identical test set. 
	\emph{More Dispersed Images (Left Columns)}: Reconstructed shapes become more clustered when training images are more dispersed. \emph{More Dispersed Shapes (Right Columns)}: Reconstructed shapes become more dispersed when training shapes are more dispersed.
	}
	\label{fig:cluster_visualization}
	\vspace{-2mm}
\end{centering}
\end{figure}

\noindent
\textbf{Dataset Composition} 
We use the same test set generated from the base dataset for all the experiments. We produce the training sets in different experiments to create scenarios with varying dispersed training images or shapes.
In particular, we build sub-trainsets with uncorrelated image and shape DS by sampling images and shapes independently from the base training set.


Figure \ref{fig:toydata-examples-images} shows the transition with more dispersed training images, in which the training images are highly clustered initially and more dispersed after the transition. 
We build a group of 8 sub-trainsets to gradually increase image DS by interpolating between the two ends of this transition. 
This procedure gives the first group of the 8 sub-trainsets. 
The corresponding training shape sets are identical across the 8 sub-trainsets and are specifically designed to be highly clustered. 
The design aims to approximate the most common scenario when training shapes are clustered, e.g. when the OC coordinates are used.


Figure \ref{fig:toydata-examples-shapes} shows the other transition with more dispersed training shapes. The training shapes change from being clustered to being dispersed. Similarly, we build a group of 8 sub-trainsets to gradually increase shape DS by interpolating between the two ends of this transition. Also, note that the corresponding training image sets are the same across the 8 sub-trainsets and are kept highly dispersed to approximate the most common scenario in ShapeNet that training images have a large variety of lighting, viewpoint, texture.

We use $S_{1}I_{i}$ and $S_{j}I_{8}$ ($i,j =1, 2, \dots, 8$) to symbolize these two groups of sub-trainsets. $S$ represents training shapes while $I$ represents training images. The subscript $i$ and $j$ represent the DS's order of training images and training shapes, respectively.
Thus, $S_{1}I_{i}$ ($i =1, 2, \dots, 8$) represents the sub-trainsets with the least dispersed training shapes and gradually more dispersed training images, corresponding to the scenario in Figure \ref{fig:toydata-examples-images}.
Similarly, $S_{j}I_{8}$ ($j =1, 2, \dots, 8$) represents the sub-trainsets with the most dispersed training images and gradually more dispersed training shapes, corresponding to the scenario in Figure \ref{fig:toydata-examples-shapes}.

\begin{figure*}
  \centering
  \begin{subfigure}[b]{0.28\textwidth}
       \includegraphics[width=\linewidth]{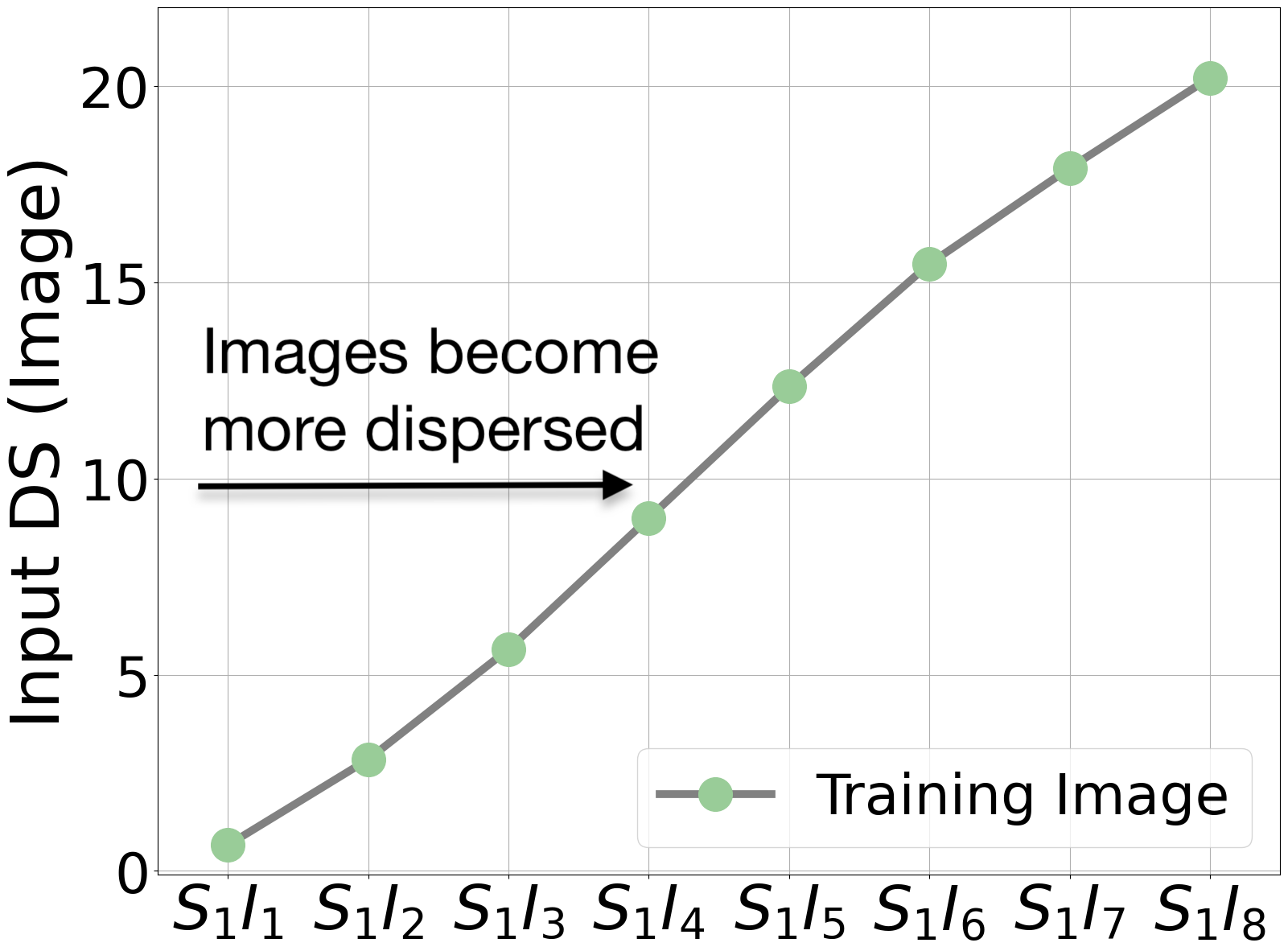}
       \caption{}\centering 
       \label{fig:toydata-imgaug-trainimage-ds}
   \end{subfigure}
   \hspace{5mm}
   \begin{subfigure}[b]{0.28\textwidth}
        \includegraphics[width=\linewidth]{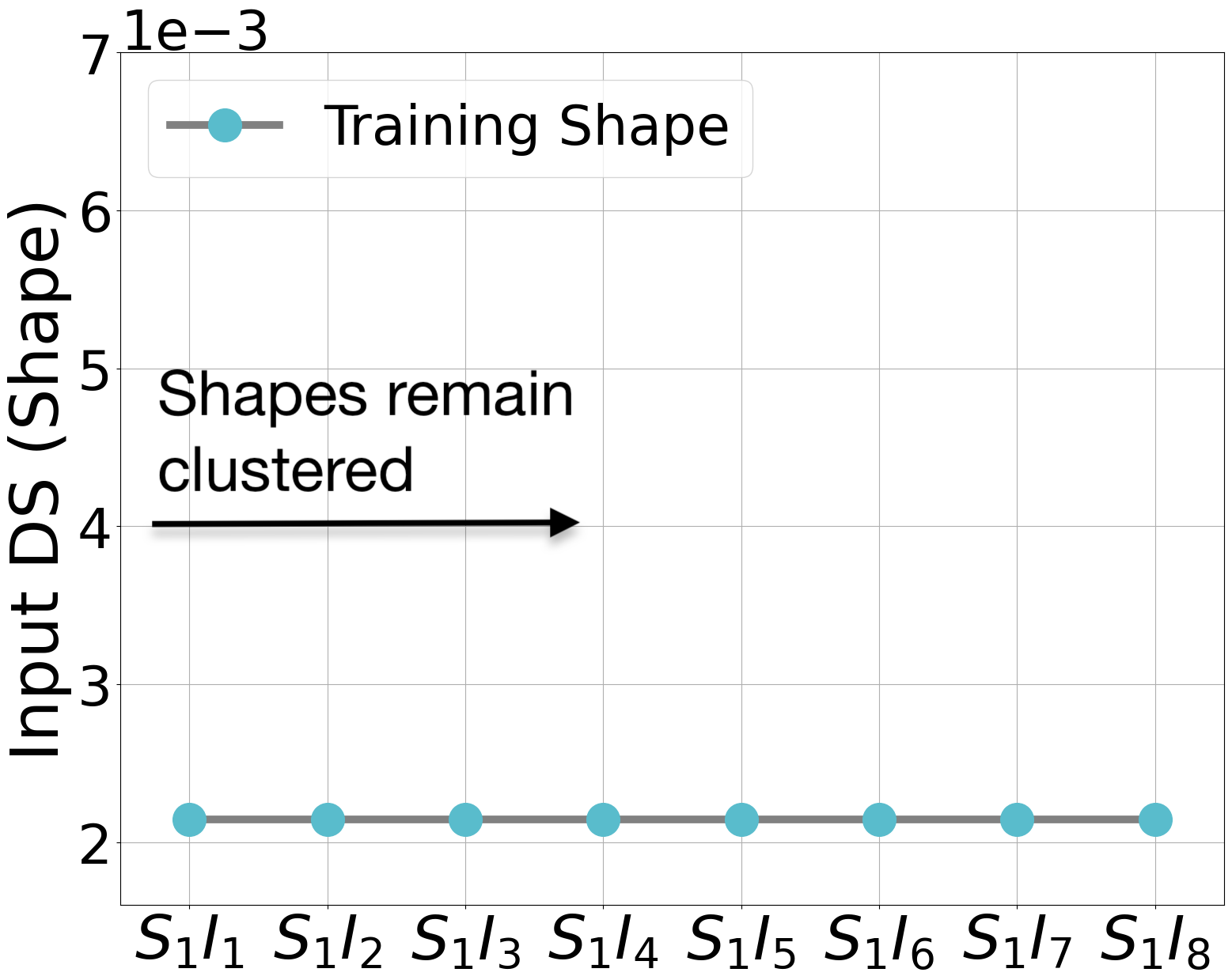}
        \caption{}\centering
        \label{fig:toydata-imgaug-trainshape-ds}
   \end{subfigure} 
   \hspace{4mm}
   \begin{subfigure}[b]{0.29\textwidth}
        \includegraphics[width=\linewidth]{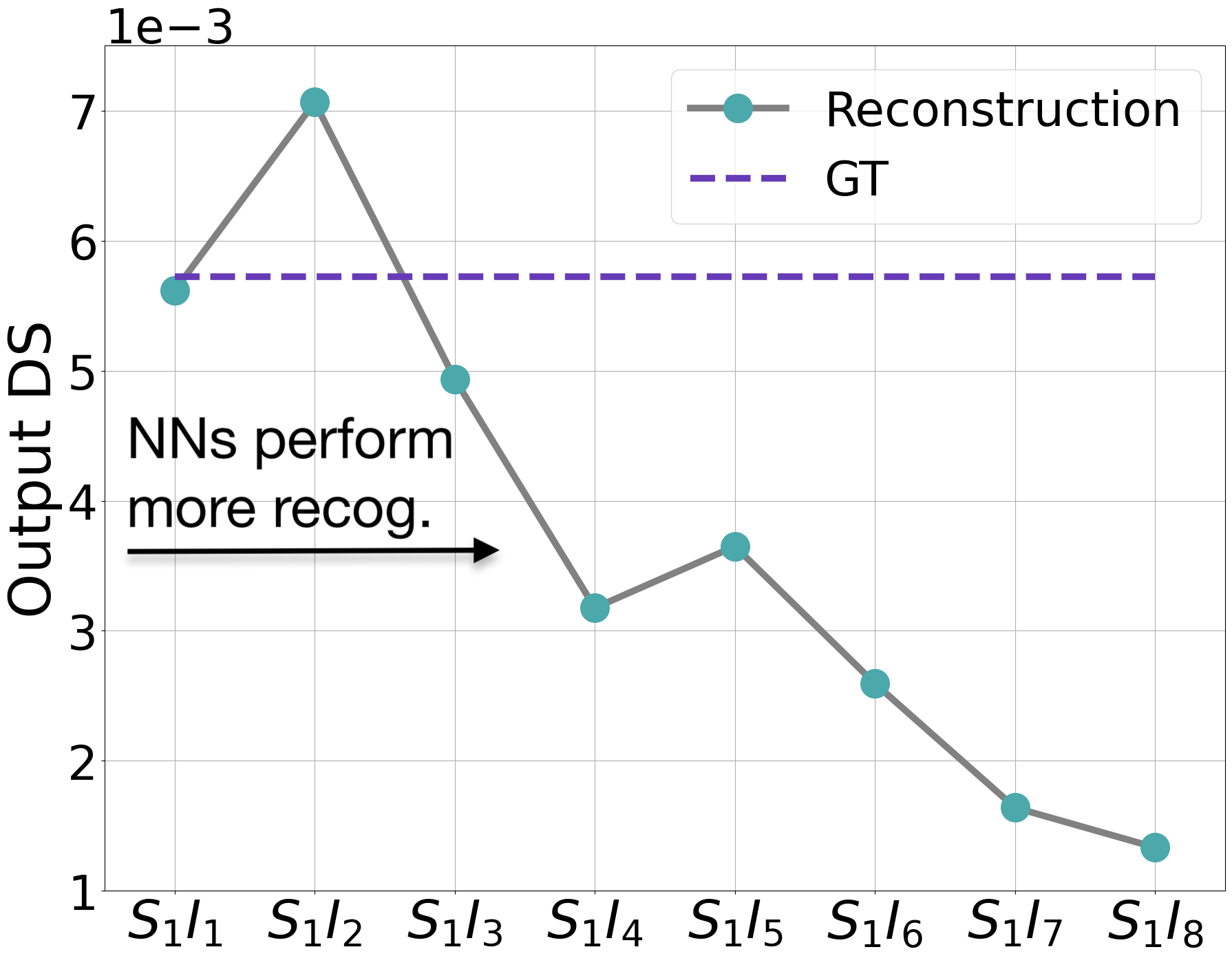}
        \caption{}\centering
        \label{fig:toydata-imgaug-pred-ds}
   \end{subfigure} 
   
  \begin{subfigure}[b]{0.28\textwidth}
       \includegraphics[width=\linewidth]{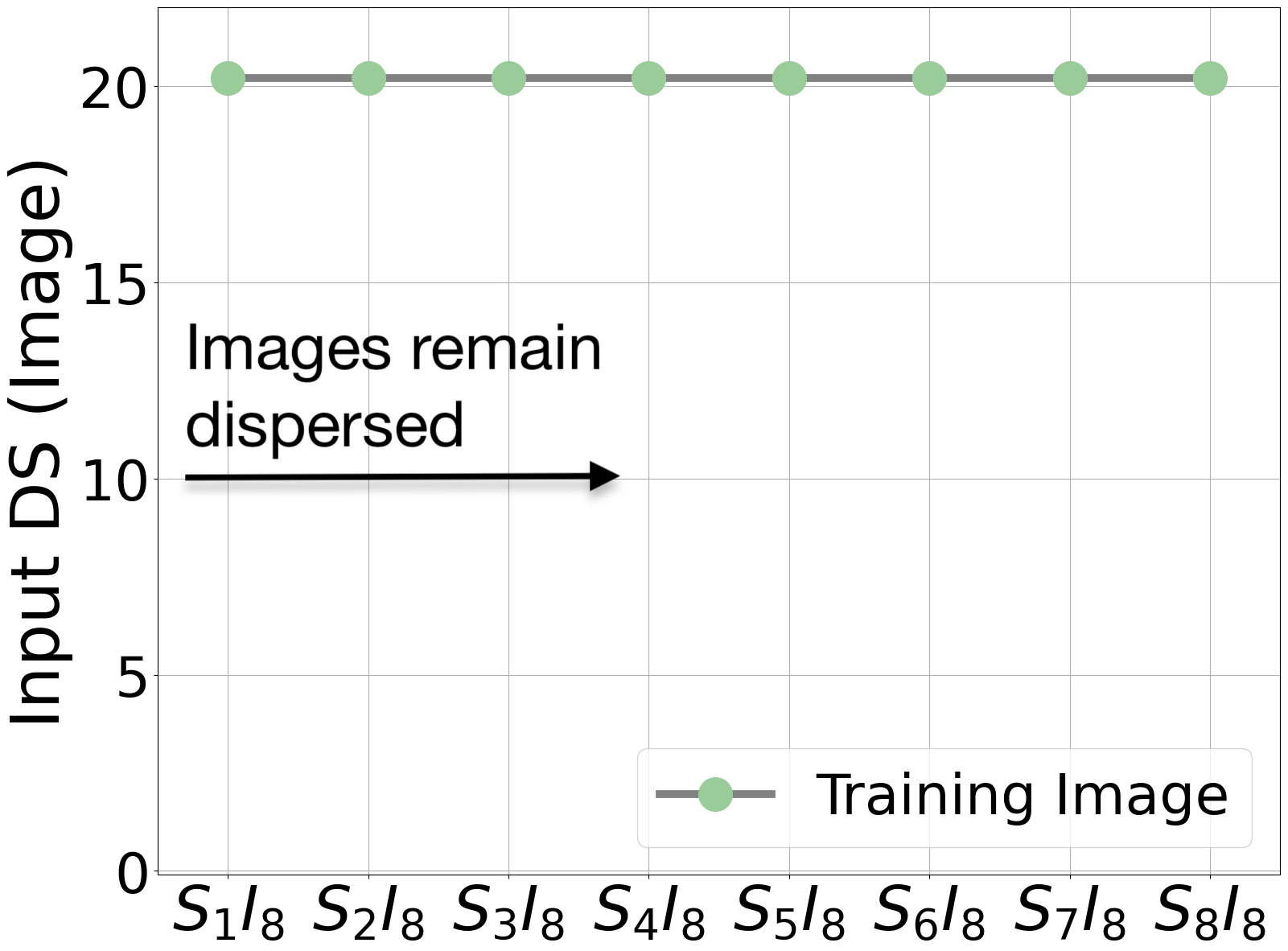}
       \caption{}\centering
       \label{fig:toydata-shapeaug-trainimage-ds}
   \end{subfigure}
   \hspace{5mm}
   \begin{subfigure}[b]{0.28\textwidth}
        \includegraphics[width=\linewidth]{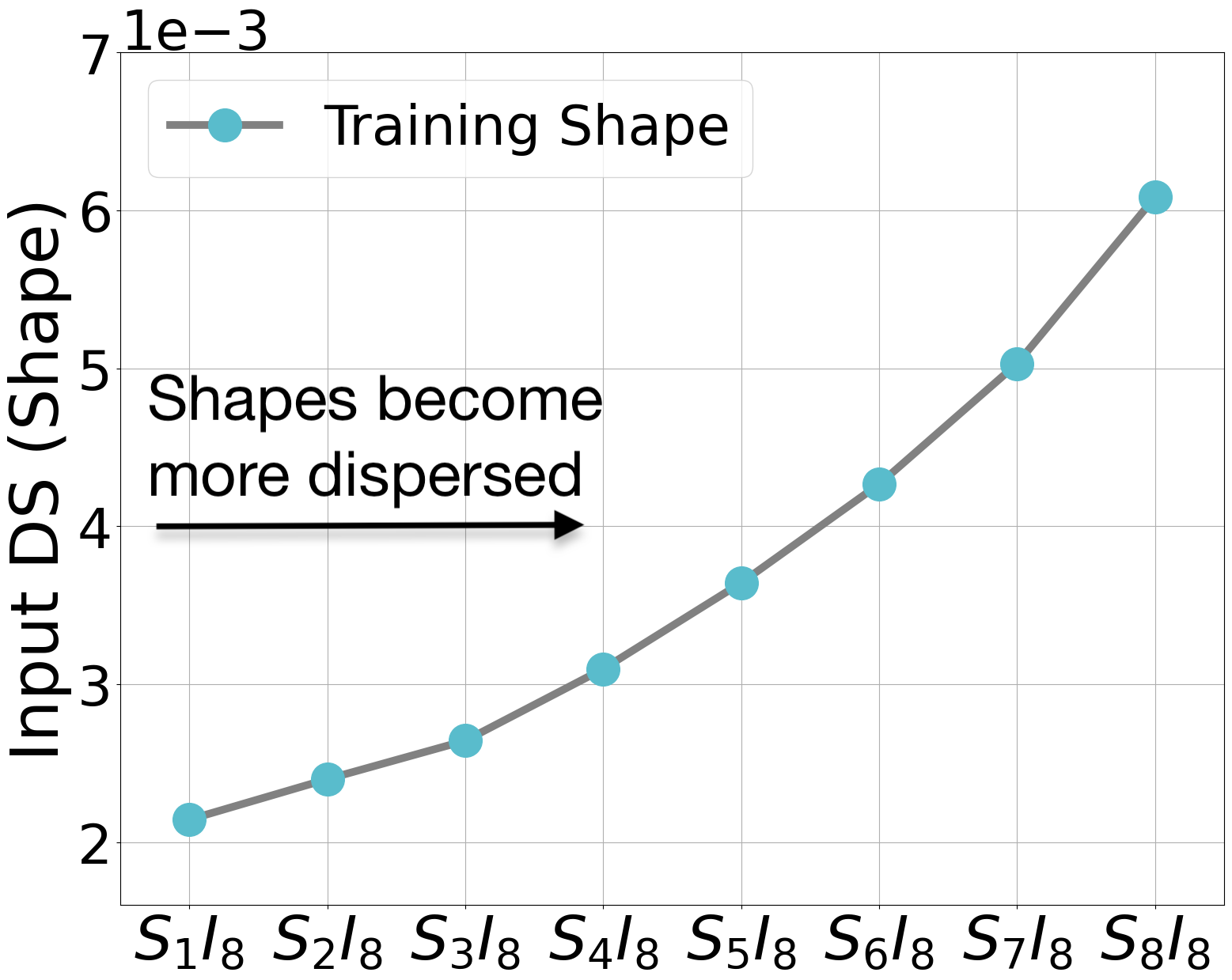}
        \caption{}\centering
        \label{fig:toydata-shapeaug-trainshape-ds}
   \end{subfigure} 
   \hspace{4mm}
   \begin{subfigure}[b]{0.29\textwidth}
        \includegraphics[width=\linewidth]{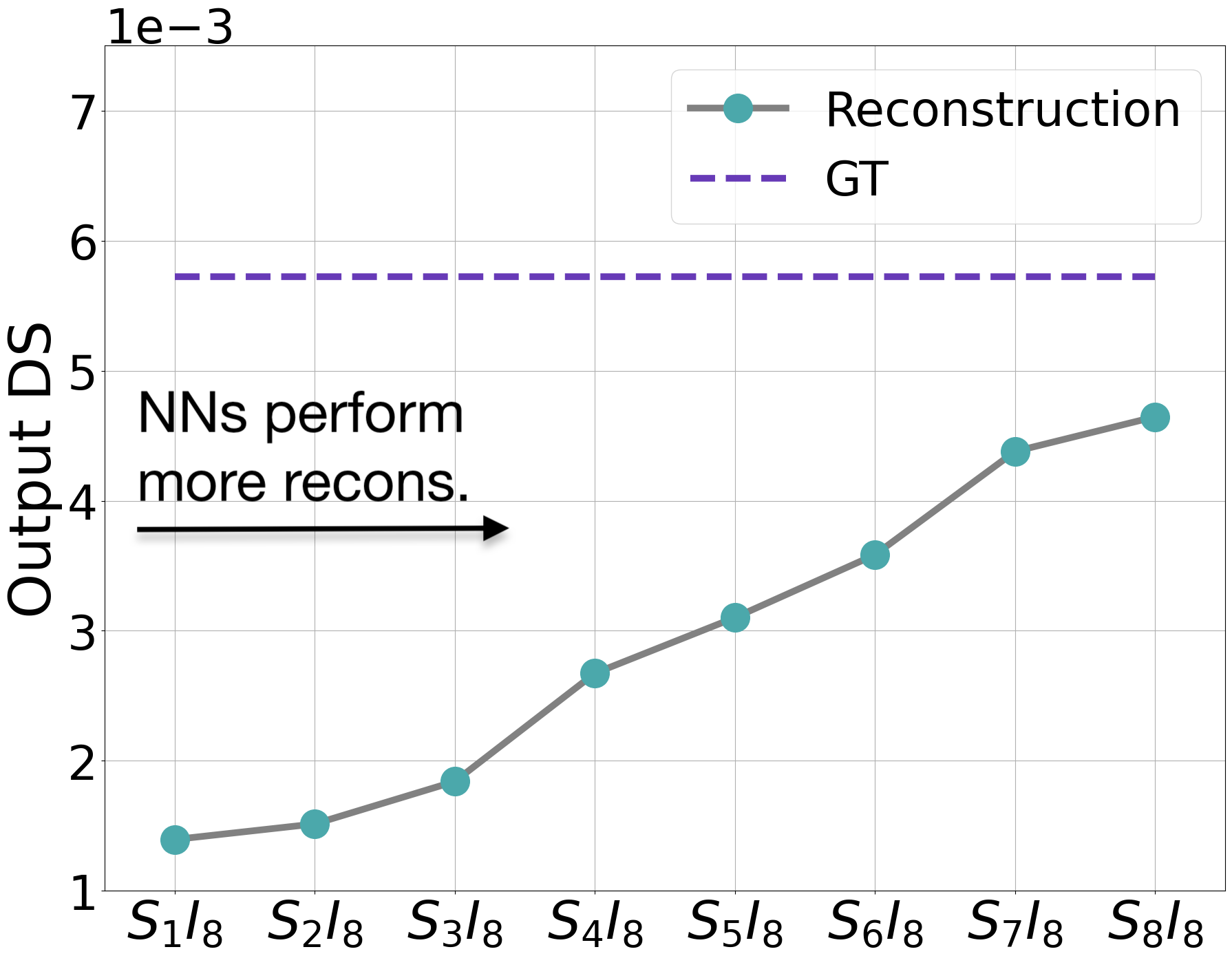}
        \caption{}\centering
        \label{fig:toydata-shapeaug-pred-ds}
   \end{subfigure} 
   \caption {Input/Output DS of synthetic datasets and trained models. (a-c) \textbf{More dispersed images} make NNs tend toward recognition as the reconstructed shapes become less dispersed when training images become more dispersed. 
   (d-f) \textbf{More dispersed shapes} make NNs tend toward reconstruction as the reconstructed shapes become more dispersed when training shapes become more dispersed. GT: ground-truth shapes of test set.}
   \label{fig:toydata-ds}
\end{figure*}

\subsection{Implementation Details}\label{sec:toy-imple}

We adopt AtlasNet-Sphere~\cite{groueix2018papier} as the baseline model.
For two groups of 8 sub-trainsets, we train 16 models separately using training protocol detailed in Appendix~\ref{sec:supp-imple} and evaluate the trained models from the last epoch.
We compute the distance matrices of the 16 reconstructed shape sets and corresponding training data, visualizing them in Figure \ref{fig:cluster_visualization}. The distance functions for different types of data are defined in Section \ref{sec:metric_on_dataset}.
We further evaluate the input/output DS of the 16 sub-trainsets and the trained models. The results are reported in Figure \ref{fig:toydata-ds}.
The cluster-number hyperparameter of input/output DS is set to be 2.


\subsection{More Dispersed Training Images}
We now show results to support the main claim in Section \ref{sec:dispersion_relationship}. Note that one part of the claim is that NNs tend toward recognition when training images are more dispersed. We use two evaluation methods to analyze our experiments: distance matrices (shown in Figure \ref{fig:cluster_visualization}) and DS (shown in Figure \ref{fig:toydata-ds}).

We first analyze Figure \ref{fig:cluster_visualization} to show how the distance matrices support the claim. 
Each subfigure in Figure \ref{fig:cluster_visualization} represents a distance matrix measured on one of the datasets or reconstructed shape sets. The reconstructed shape sets are generated at test time. In this subsection, we parse the left three columns of Figure \ref{fig:cluster_visualization}.

We measure the distance matrices of training images (first column), training shapes (second column), and reconstructed shapes on the test set (third column). 
Now, look at the first row marked by $S_{1}I_{1}$. 
The two distance matrices in the first and second columns, titled ``Images'' and ``Shapes'' respectively, show sudden color mutation from blue to red, indicating training images and shapes are highly clustered. 
The distance matrix in the column titled ``Recons'' shows a continuous color change from blue (low value) to red (high value), meaning the reconstructed shapes are dispersed.
Then, we can analyze the following rows marked by $S_{1}I_{i}$ ($i=2,\dots, 8$) similarly. 
The distance matrices on the ``Recons'' column show increasingly clustered patterns, while the distance matrices under the "Images" column show increasingly dispersed patterns.
It indicates that the reconstructed shapes of NNs become more clustered, which supports our claim that NNs tend toward recognition when training images are more dispersed.

Next, we look at the values of DS for training images, training shapes, and reconstructed shapes, shown in Figure \ref{fig:toydata-imgaug-trainimage-ds}, \ref{fig:toydata-imgaug-trainshape-ds}, and \ref{fig:toydata-imgaug-pred-ds}, respectively.
Figure \ref{fig:toydata-imgaug-trainimage-ds} shows that the input DS of the training image sets gradually increases, indicating that the training image sets become more dispersed. Note that this increasingly dispersed pattern matches the first column in Figure \ref{fig:cluster_visualization}. 
Then, Figure~\ref{fig:toydata-imgaug-trainshape-ds} shows that the training shapes remain unchanged, matching the second column in Figure~\ref{fig:cluster_visualization}. 
Finally, Figure~\ref{fig:toydata-imgaug-pred-ds} shows that the output DS of reconstructed shapes gradually decreases, matching the increasingly clustered color patterns shown in the third column of Figure \ref{fig:cluster_visualization}.

Thus, both the distance matrices (shown in Figure \ref{fig:cluster_visualization}) and the DS trends (shown in Figure \ref{fig:toydata-ds}) show that NNs prone to perform recognition and predict more clustered shapes when training images become more dispersed.

\subsection{More Dispersed Training Shapes}

The other part of our main claim is that NNs tend toward reconstruction when training shapes become more dispersed. 
We conduct the same analysis as the previous subsection using both distance matrices and DS. 

For distance matrices, see the right three columns in Figure~\ref{fig:cluster_visualization}. From top to bottom, the column titled ``Shapes'' shows increasingly dispersed patterns while the column titled ``Images'' remains dispersed.
It indicates a list of training datasets of more dispersed shapes and consistently dispersed images. 
The column titled ``Recons'' shows more dispersed color patterns, meaning that the reconstructed shapes become dispersed. 
Thus, more dispersed training shapes make the reconstructed shapes more dispersed, indicating that NNs tend toward reconstruction.

The results are again verified by DS evaluation, as shown in Figure \ref{fig:toydata-shapeaug-trainimage-ds}, \ref{fig:toydata-shapeaug-trainshape-ds}, and \ref{fig:toydata-shapeaug-pred-ds}. First, Figure \ref{fig:toydata-shapeaug-trainimage-ds} shows that the input DS of training images remains unchanged. Second, Figure \ref{fig:toydata-shapeaug-trainshape-ds} shows that the input DS of training shapes gradually increases. 
Finally, Figure \ref{fig:toydata-shapeaug-pred-ds} shows that the output DS of the reconstructed shapes gradually increases. These trends all match the results shown in Figure~\ref{fig:cluster_visualization}.

Therefore, both of the two evaluation methods support the claim that NNs lean toward reconstruction and predict more dispersed shapes when training shapes also become more dispersed.

\section{Experiments on ShapeNet}\label{sec:exp-shapenet}

In this section, We further verify our main claim by conducting experiments on the commonly used benchmark dataset ShapeNet~\cite{chang2015shapenet}. We show how varying levels of dispersed images and shapes affect the tendency of NNs to perform reconstruction or recognition.
Due to space limitations, we focus on encoder-decoder-based NNs, including PSGN \cite{fan2017point}, FoldingNet \cite{yang2018foldingnet}, and AtlasNet \cite{groueix2018papier}. We conduct additional experiments on SDF-based NNs in Appendix~\ref{sec:sdf-method}.

\subsection{Dataset}

\noindent
\textbf{ShapeNet} We conduct experiments on ShapeNetCore consisting of 3D models in 13 object categories~\cite{chang2015shapenet}. 
We use the train/test split in \cite{choy20163d} and use the point cloud data provided by AtlasNet~\cite{groueix2018papier}. 
In experiment \ref{sec:shapenet-moreimg}, we render new image datasets using the method of \cite{Xu2019DISNDI} to control the rendering viewpoints of training images.
Both OC and VC coordinates are investigated.
In experiments \ref{sec:shapenet-moreshape}, we use images rendered by ~\cite{choy20163d}. In this dataset, each 3D model has been rendered 24 images of random views. We use one fixed view among 24 views in the training/test set in \ref{sec:shapenet-moreshape} and investigate the impact of using more views per shape in \ref{sec:more-training-sample}.

\begin{figure*}
  \centering
  \begin{subfigure}[b]{0.24\textwidth}
       \includegraphics[width=\linewidth]{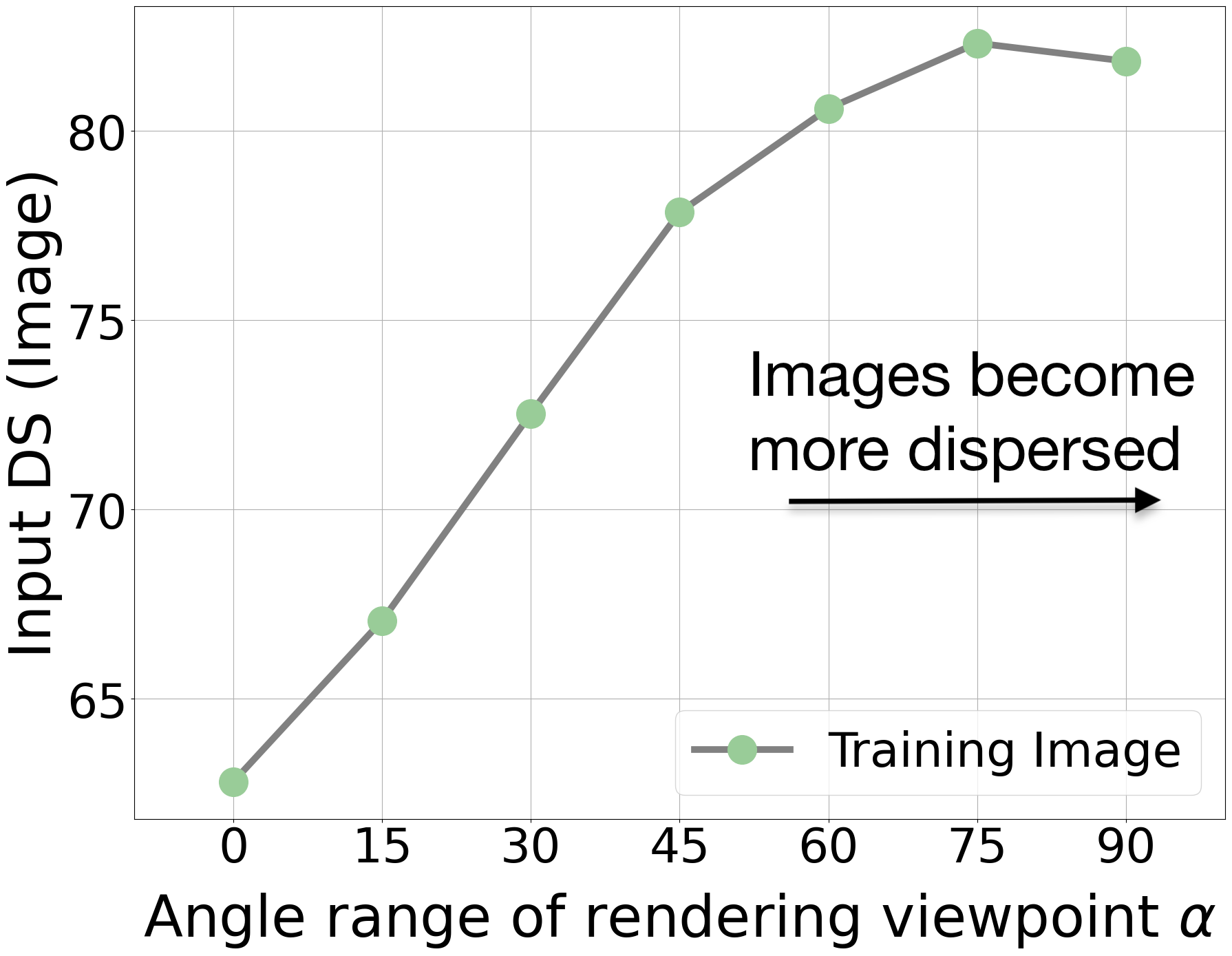}
       \caption{}
       \label{fig:shape13-imgaug-inertia-trainimage-k100}
   \end{subfigure}
   \begin{subfigure}[b]{0.24\textwidth}
        \includegraphics[width=\linewidth]{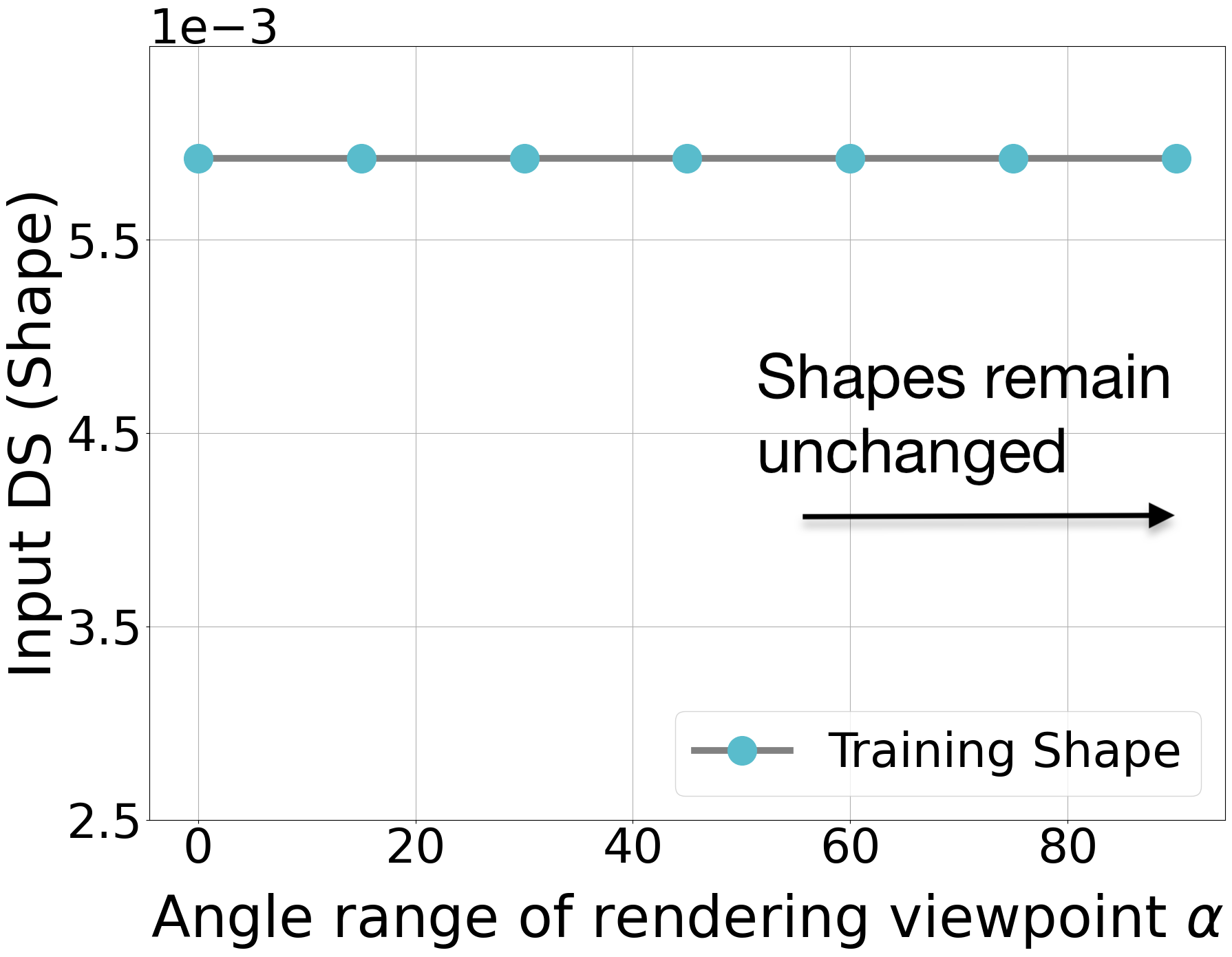}
        \caption{}
        \label{fig:shape13-imgaug-inertia-trainpoints-k500}
   \end{subfigure}
   \begin{subfigure}[b]{0.24\textwidth}
        \includegraphics[width=\linewidth]{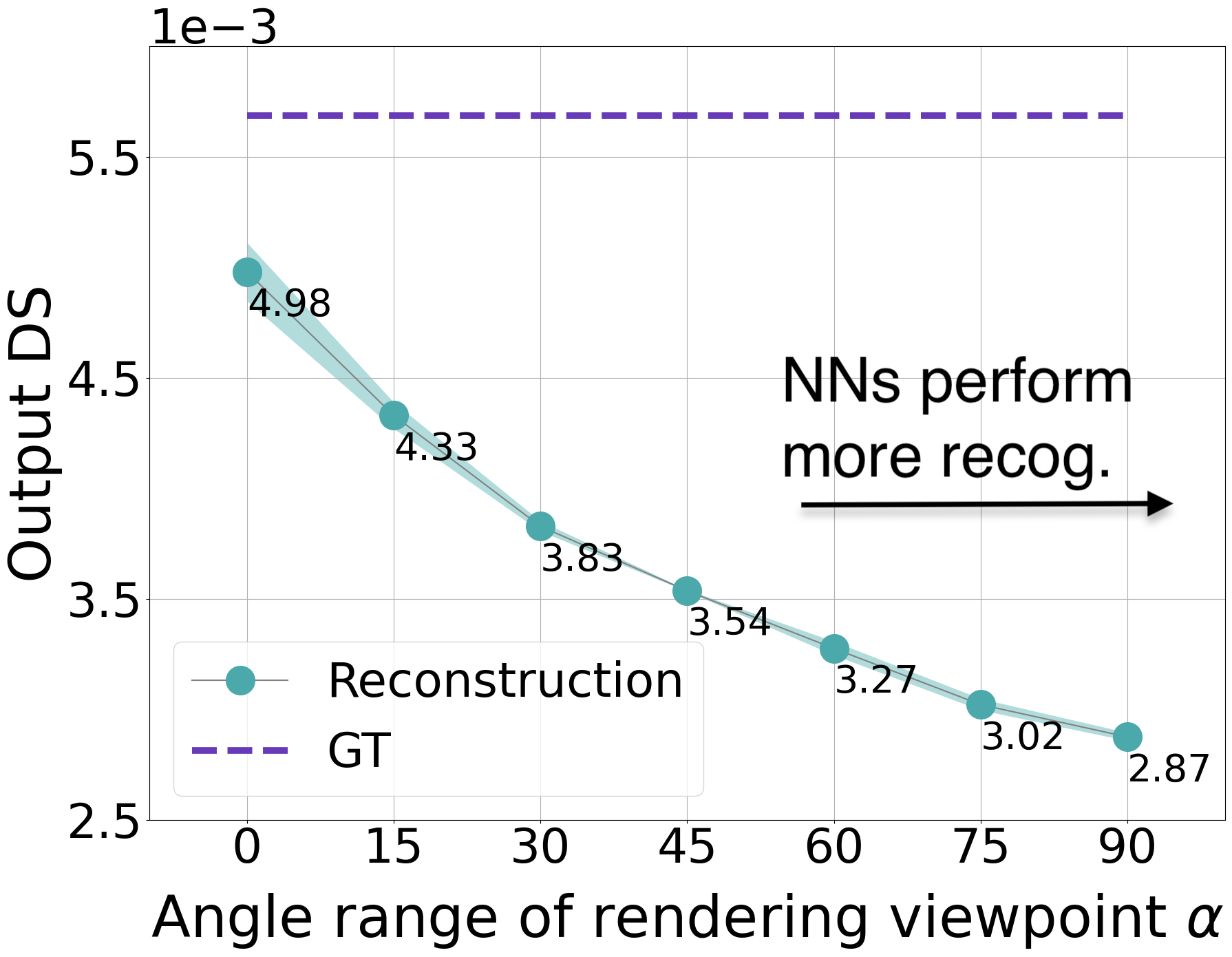}
        \caption{}
        \label{fig:shape13-imgaug-inertia-pred-k500}
   \end{subfigure}
    \begin{subfigure}[b]{0.24\textwidth}
        \includegraphics[width=\linewidth]{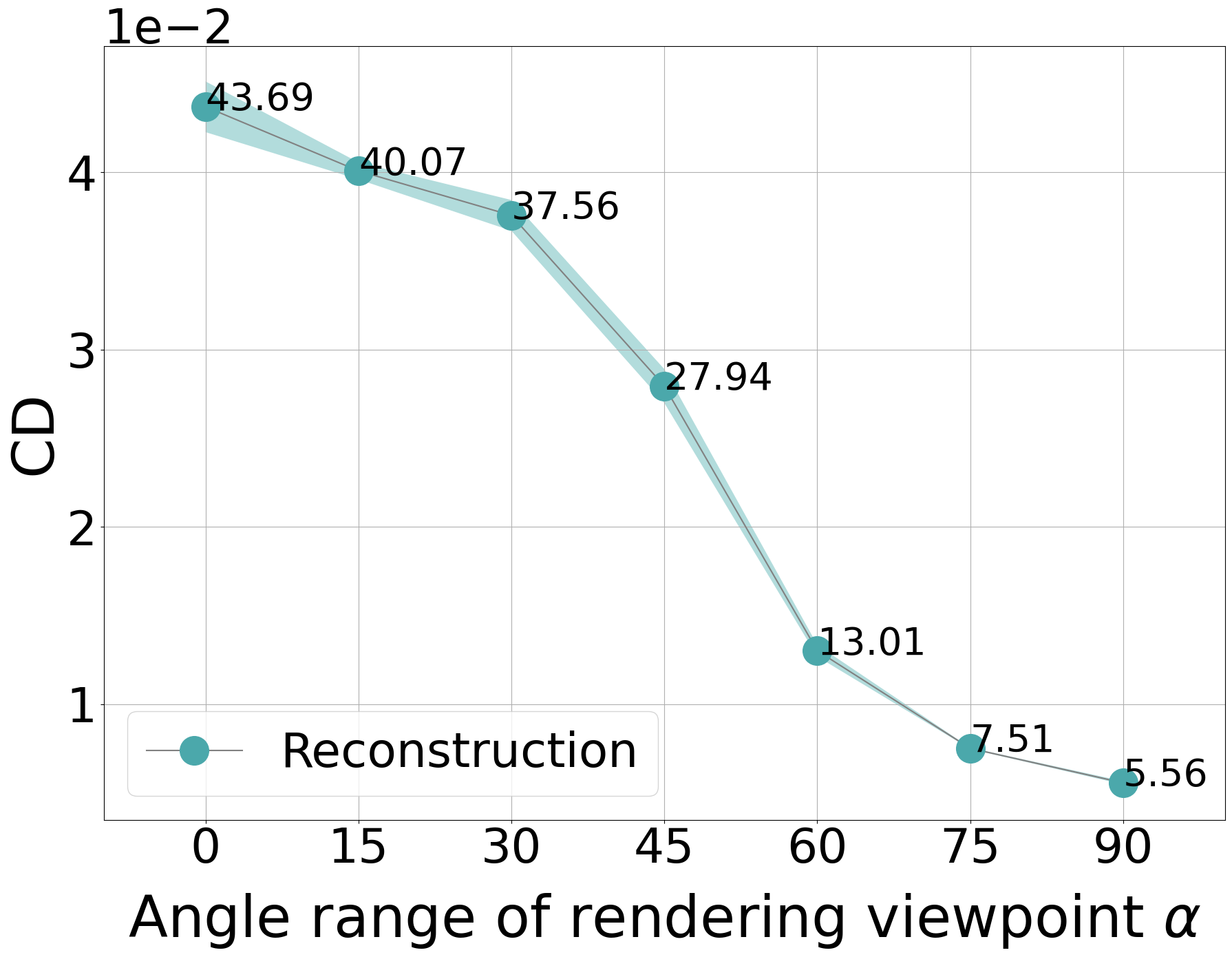}
        \caption{}
        \label{fig:shape13-imgaug-chamfer-pred}
   \end{subfigure}
   
   \caption{DS measure in OC when the training images are rendered by increasing viewpoint angle range $\alpha$. The unit of $\alpha$ is degree. (a) (b) From left to right, training images become more dispersed and training shapes remain unchanged. (c) Reconstructed shapes become less dispersed. (d) CD scores of reconstructed shapes become smaller. The smaller, the better.}
   \vspace{-2mm}
   \label{fig:shape13-imageaug}
\end{figure*}

\subsection{Implementation Details}\label{sec:shapenet-imple}

For experiments in \ref{sec:shapenet-moreimg} and \ref{sec:more-training-sample}, we adopt AtlasNet-Sphere~\cite{groueix2018papier} as the baseline. In \ref{sec:shapenet-moreshape}, we use multiple SVR models. The details of model implementation and training are provided in Appendix~\ref{sec:supp-imple}. We evaluate the trained model from the last epoch.
We run the model on each dataset using three random seeds and report the mean and standard deviation for evaluation. 
Point clouds are used as shape representation. Each point cloud includes 2500 points. The cluster-number hyperparameter of input/output DS is set to be 500. 

\subsection{More Dispersed Training Images}\label{sec:shapenet-moreimg}
We study the transition more dispersed training images. The image set is generated to be more dispersed while the shape set remains clustered. 

\noindent
\textbf{Experiment Design} 
We render a list of new image datasets from the training shapes of ShapeNet and increase the angle range of the rendering viewpoint. 
The angle range is denoted by $\alpha$, and the unit is degree. 
We build seven rendering datasets in this way. For each of them, we only render a single image for each shape. 
During rendering, the $\theta_{\mathrm{az}}$ of viewpoint is randomly sampled from $-\alpha$ to $\alpha$ and $\theta_{\mathrm{el}}$ is sampled from 20 to 30 degree. 
The $\theta_{\mathrm{az}}$ and $\theta_{\mathrm{el}}$ are azimuth angle and elevation angle of viewpoint, respectively. $\alpha$ is selected as 0, 15, 30, 45, 60, 75, 90 for the seven different datasets. 
As $\alpha$ increases, training images become more dispersed. We only use shapes in OC coordinate so that the training shapes remain clustered. 
During testing, the input images of the test set are rendered using an $\alpha$ value equal to 90.

\noindent
\textbf{Results} We show how NNs perform when the training images become more dispersed.
First, Figure \ref{fig:shape13-imgaug-inertia-trainimage-k100} and \ref{fig:shape13-imgaug-inertia-trainpoints-k500} show that our approach makes training images more dispersed while maintaining the DS of training shapes.
Then, Figure \ref{fig:shape13-imgaug-inertia-pred-k500} shows that the trained NNs tend more towards recognition as the value of output DS decreases. This trend matches our main claim that more dispersed training images make the output shapes tend towards recognition.

However, Figure \ref{fig:shape13-imgaug-chamfer-pred} shows that NNs trained on more dispersed images have improved reconstruction score measured by the CD.
This improvement could result from augmenting the training dataset or the case where the distribution of images in the training set becomes closer to that of the test set under this transition.
Thus, if only looking at the CD measure, one may believe that adding more training images can lead to improved 3D reconstruction quality but may neglect the confounding factor that the output shapes become more clustered.
This finding shows that while more dispersed images can potentially improve the CD score, they also incline NN to reconstruct more clustered shapes.
In other words, the improved CD score is not sufficient to capture whether the output shapes become more clustered or not. 
Therefore, we illustrate that the proposed DS provides novel information in addition to the conventional reconstruction score.

\begin{figure}[!htb]
    \centering
    \begin{subfigure}[b]{0.49\columnwidth}
        \includegraphics[width=\linewidth]{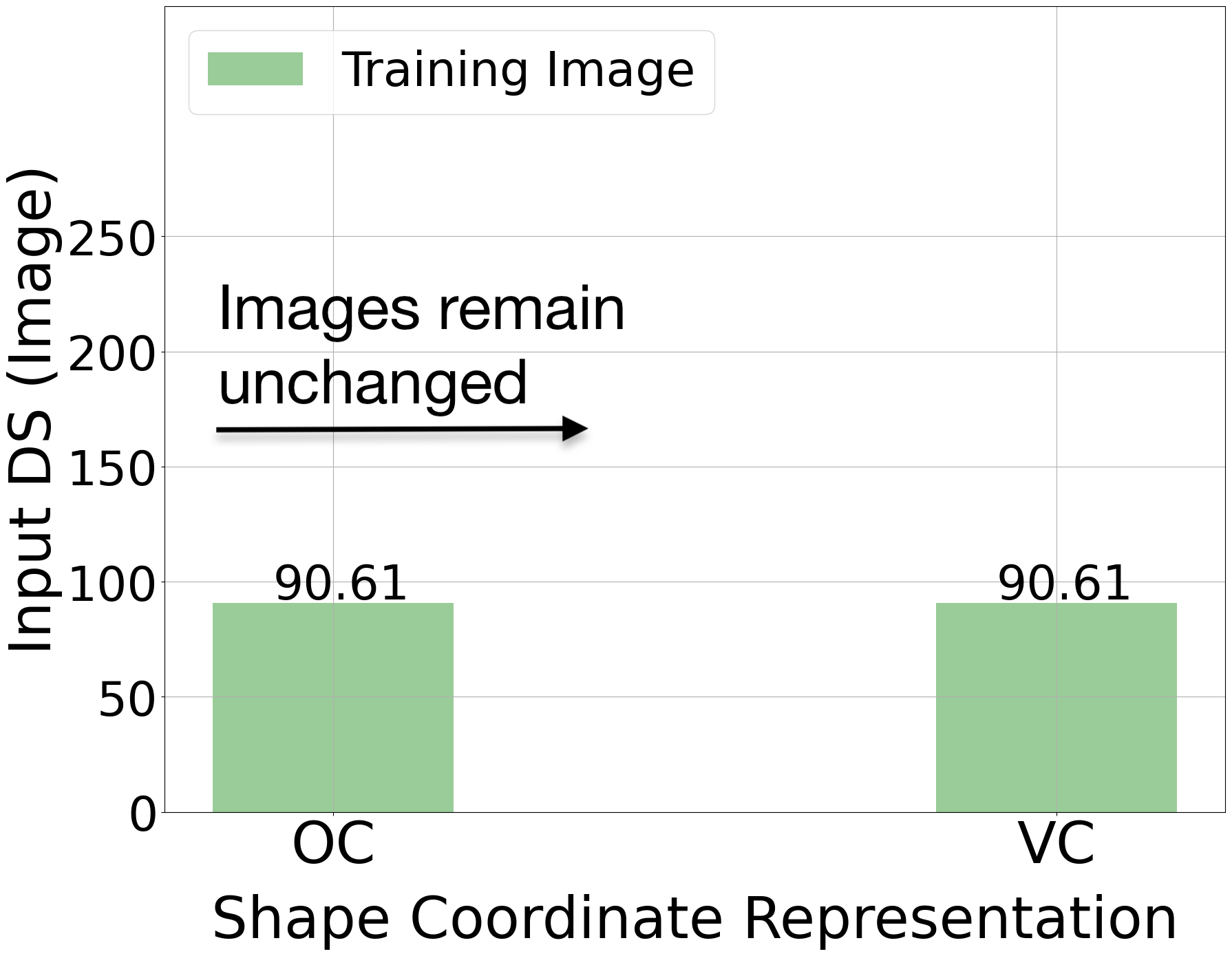}
    \caption{}\centering
    \label{fig:shape13-coordinate-trainimages}
    \end{subfigure}
    \begin{subfigure}[b]{0.49\columnwidth}
        \includegraphics[width=\linewidth]{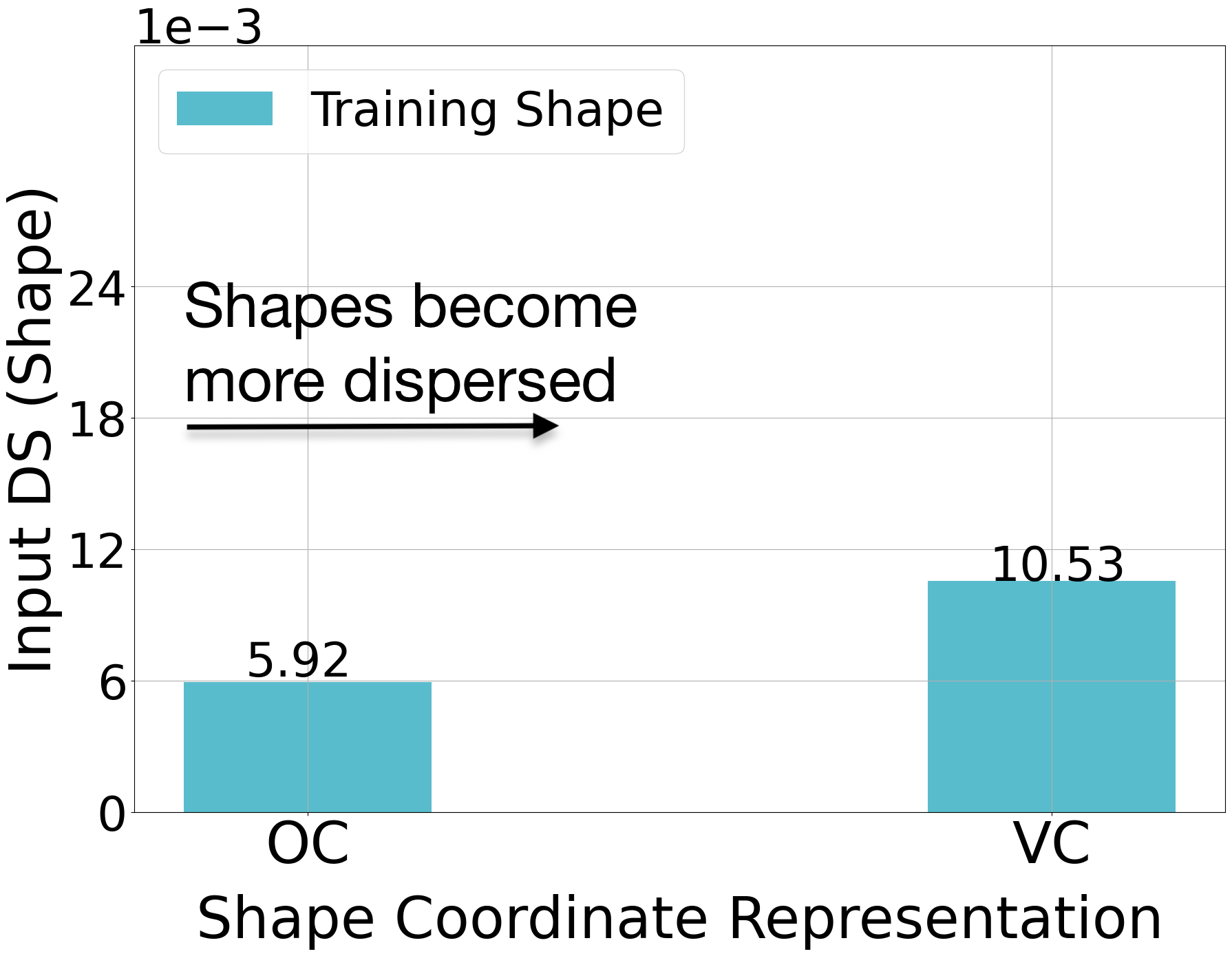}
    \caption{}\centering
    \label{fig:shape13-coordinate-trainpoints}
    \end{subfigure}
   \caption{Comparing Input DS of OC and VC. (a) Training images remain unchanged. (b) Training shapes become more dispersed.}
   \vspace{-2mm}
   \label{fig:shape13-coordinate}
\end{figure}

\subsection{More Dispersed Training Shapes}\label{sec:shapenet-moreshape}

We study what happens when we use gradually more dispersed training shapes while maintaining the training images dispersed.

\noindent
\textbf{Experiment Design}
Two coordinate representations OC and VC are used here to change the input DS of the ShapeNet dataset. 
As mentioned in Section \ref{sec:intro}, shape sets in VC are more dispersed than those in OC. We use the same training images for both two datasets to ensure that the DS of training images remains unchanged.
We train four different SVR models: PSGN~\cite{fan2017point}, AtlasNet-Sphere~\cite{groueix2018papier}, AtlasNet-25~\cite{groueix2018papier}, and FoldingNet~\cite{yang2018foldingnet}.
We measure these models in both OC and VC and use the output DS to see if the predicted shapes are more or less dispersed.
However, note that the output shapes are not in the same coordinates and are not directly comparable if we change the coordinates.
Thus, during the calculation of output DS, we transform the predicted shapes back to OC coordinates to keep the same output configuration for a fair comparison on output DS.

\noindent
\textbf{Results} We now present our results to demonstrate that the transition with more dispersed shapes guides NNs towards more reconstruction. See Figure \ref{fig:shape13-coordinate} for the input DS and Table \ref{tb:more-shapes} for the output DS.
From Figure \ref{fig:shape13-coordinate}, we see that the transition makes training shapes more dispersed while letting training images remain unchanged. 
Then, from Table \ref{tb:more-shapes}, we see that models trained in the VC coordinate have more dispersed shape predictions than OC, as the output DS values of all the models in VC are larger than that in OC. 
Besides, in Table \ref{tb:more-shapes}, we also measure the CD, and we see that all the VC models outperform OC models.
Based on the two observations, we can conclude that more dispersed training shapes encourage NNs to use more reconstruction than recognition as the underlying mechanism to perform.
At the same time, more dispersed training shapes also improve reconstruction performance measured by CD.

Finally, from the results of Section~\ref{sec:shapenet-moreimg} and \ref{sec:shapenet-moreshape}, we see that both more dispersed training images and more dispersed training shapes can lead to improved reconstruction scores. However, more dispersed training images actually let the NNs prefer recognition to reconstruction.
Such results again show that our new way of measuring the dispersion of output shapes provides novel information on assessing the 3D reconstruction quality.

\begin{table} 
\resizebox{\columnwidth}{!}{%
\begin{tabular}{c|cc|cc}
\hline
              & \multicolumn{2}{c|}{Output DS}     & \multicolumn{2}{c}{CD}      \\ \hline
              & OC & VC      & OC & VC      \\ \hline
PSGN          & $1.63\pm0.07$      & $\boldsymbol{2.47\pm0.00}$   & $6.60\pm0.12$       & $\boldsymbol{5.90\pm0.15}$     \\
FoldingNet    & $2.39\pm0.04$      &  $\boldsymbol{3.34\pm0.05}$    & $7.26\pm0.08$      & $\boldsymbol{5.84\pm0.10}$     \\
AtlasNet-Sph. & $2.83\pm0.00$      &  $\boldsymbol{3.60\pm0.02}$    & $7.12\pm0.08$      & $\boldsymbol{5.40\pm0.02}$     \\
AtlasNet-25   & $2.84\pm0.01$      &   $\boldsymbol{3.55\pm0.01}$   &  $6.59\pm0.07$     & $\boldsymbol{5.06\pm0.02}$     \\
\hline
GT            & $5.68$      & $5.68$            &   -   & -      \\ \hline
\end{tabular}
}
\caption{Evaluation (mean$\pm$stdev) of models trained in OC (\emph{Left}) and VC (\emph{Right}) coordinates. Metrics are output DS ($\times$ 0.001, $\uparrow$), CD ($\times$ 0.001, $\downarrow$). NNs trained in VC predict more dispersed and better reconstructed shapes than NNs in OC.}
\label{tb:more-shapes}
\vspace{-2mm}
\end{table}

\subsection{More Training Samples}\label{sec:more-training-sample}

We investigate whether more training samples can guide NNs to perform more reconstruction in SVR. 
Note that in Section~\ref{sec:shapenet-moreimg} and \ref{sec:shapenet-moreshape}, we only make training shapes more dispersed or only make training images more dispersed, to show how the output DS changes with each covariate.
In this subsection, we conduct this additional experiment to change training shapes and images simultaneously because it is often practically convenient to do so, e.g., by adding more samples.

\noindent
\textbf{Experiment Design}
We obtain more training samples using more rendered images that have been given in \cite{choy20163d}. 
The conventional training protocol in \cite{choy20163d, groueix2018papier} is to use one view of the image among 24 views per shape for each epoch. 
However, in this experiment, we use more views per shape for each epoch, which ranges from 1 to 18. 
We use the VC coordinate, and hence shapes are rotated based on the input viewpoint of the rendered images.
Thus, using more views per shape is equivalent to using more training images and more training shapes obtained by performing rotations in the 3D space.
Also, the amount of training data is linear in the number of views per shape. We adopt AtlasNet-Sphere~\cite{groueix2018papier} as the baseline model and use the same protocol in Section \ref{sec:shapenet-imple}.

\begin{figure}[h!]
    \centering
    \begin{subfigure}[b]{0.49\columnwidth}
        \includegraphics[width=\linewidth]{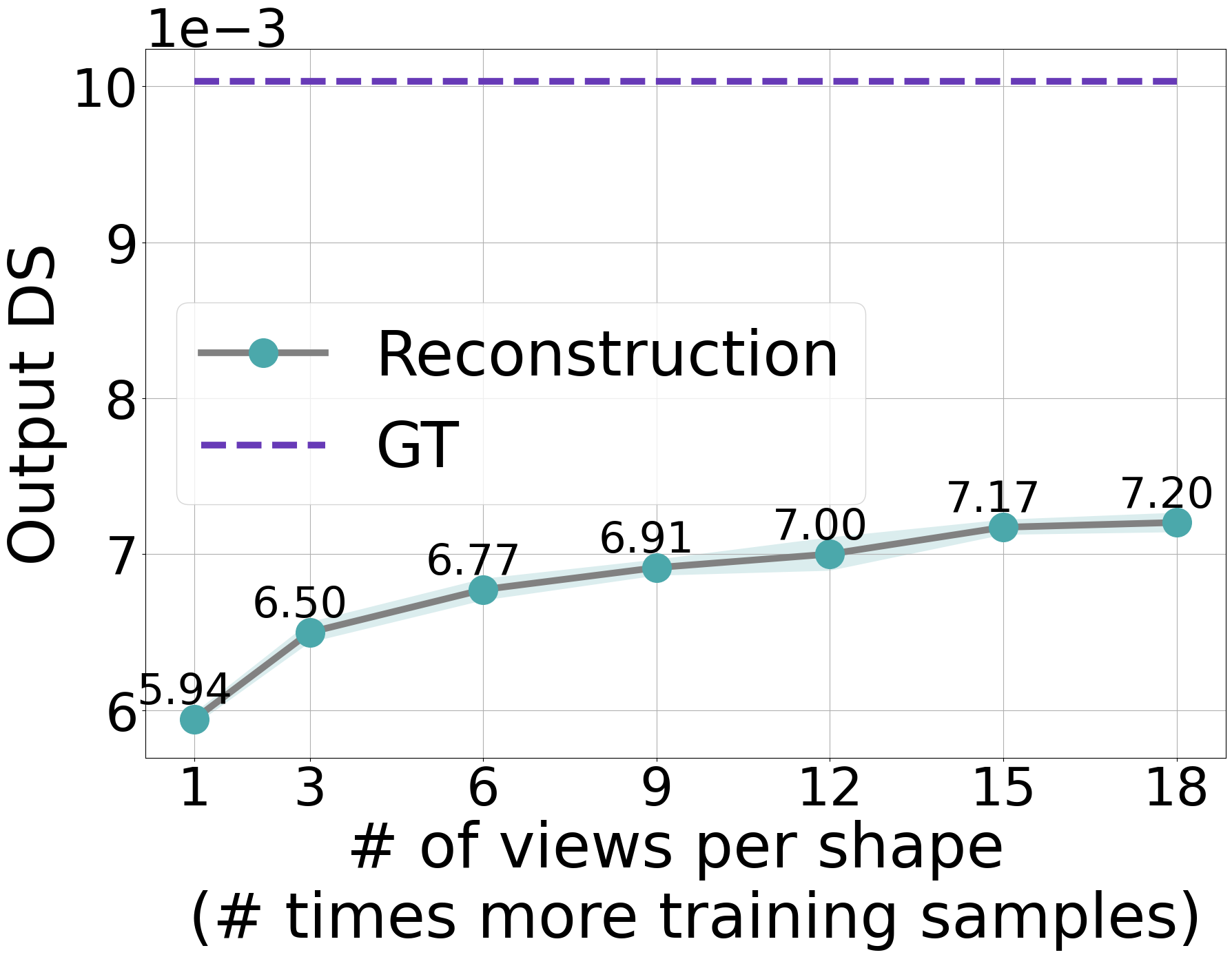}
    \caption{}\centering
    \label{fig:nviews-inertia}
    \end{subfigure}
    \begin{subfigure}[b]{0.49\columnwidth}
        \includegraphics[width=\linewidth]{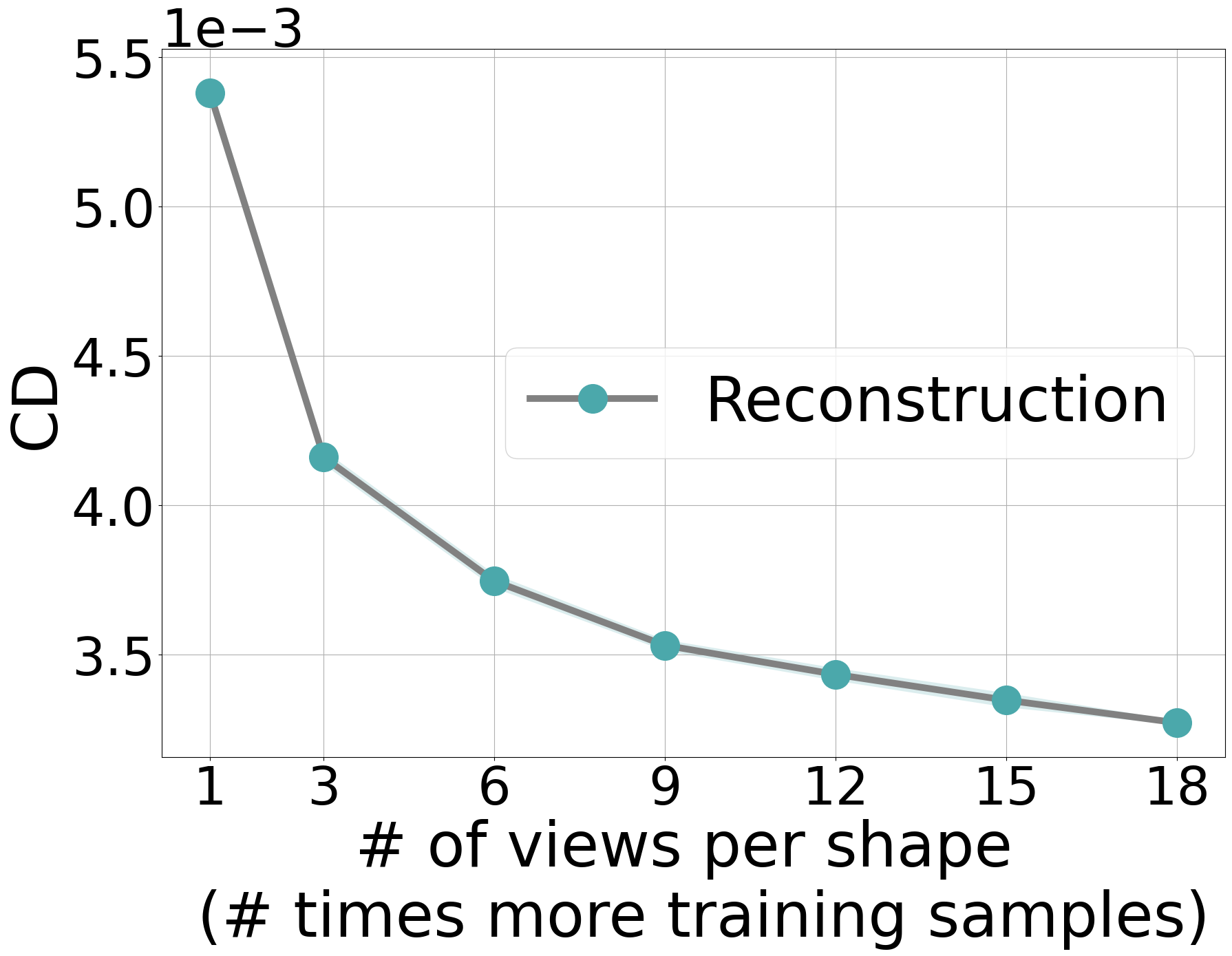}
    \caption{}\centering
    \label{fig:nviews-chamfer}
    \end{subfigure}
   \caption{Evaluation of models trained on more data samples in VC. The amount of training data samples is linear in the number of views per shape. Although more training shapes make NNs prefer reconstruction, the capability of NNs to perform reconstruction is still limited, even using 18$\times$ training data. 
   (a) Output DS ($\uparrow$). (b) CD ($\downarrow$).} \vspace{-2mm}
   \label{fig:nviews-search}
\end{figure}

\noindent
\textbf{Results}
The results are reported in Figure \ref{fig:nviews-search}.
First, both output DS and CD are shown to improve with more training samples. 
More specifically, the increased output DS indicates that more training samples guide NNs to perform more reconstruction. And the decreased CD score indicates that more training samples improve the reconstruction quality. 
Second, Figure \ref{fig:nviews-inertia} shows a noticeable gap between GT (0.01) and the limit of the improved output DS (0.0072) of NNs trained on 18$\times$ more data samples. This indicates that it is challenging to further improve NNs to perform reconstruction based on simply augmenting the public dataset.

\section{Conclusion}\label{sec:conclusion}
In this paper, we study the underlying mechanisms of NNs in SVR tasks. 
First, we show that NNs can be disposed towards recognition or reconstruction depending on how dispersed the training data is. 
We propose a metric called DS to quantify this relationship.
We show that both of the two experiment procedures, i.e., using more dispersed training images and shapes, can improve conventional reconstruction scores such as CD.
However, the DS measure shows that the former (training images) leads NNs to prefer recognition rather than reconstruction while the latter (training shapes) leads NNs to perform more reconstruction.
Thus, the proposed DS provides novel information on how NNs perform SVR tasks.
We suggest measuring the DS in conjunction with conventional reconstruction scores when assessing trained NNs in SVR tasks. 
More studies on other DL techniques, including data augmentation and network architectures, would be necessary to make NNs perform reconstruction instead of recognition.

\section*{Acknowledgments}
This research is partially supported by NSF Future Manufacturing program under EEC-2036870.
\newpage
{\small
\bibliographystyle{ieee_fullname}
\bibliography{egbib}

\begin{thebibliography}{10}\itemsep=-1pt

\bibitem{10.5555/1283383.1283494}
David Arthur and Sergei Vassilvitskii.
\newblock K-means++: The advantages of careful seeding.
\newblock In {\em Proceedings of the Eighteenth Annual ACM-SIAM Symposium on
  Discrete Algorithms}, SODA '07, page 1027–1035, USA, 2007. Society for
  Industrial and Applied Mathematics.

\bibitem{bautista2021generalization}
Miguel~Angel Bautista, Walter Talbott, Shuangfei Zhai, Nitish Srivastava, and
  Joshua~M Susskind.
\newblock On the generalization of learning-based 3d reconstruction.
\newblock In {\em Proceedings of the IEEE/CVF Winter Conference on Applications
  of Computer Vision}, pages 2180--2189, 2021.

\bibitem{chang2015shapenet}
Angel~X Chang, Thomas Funkhouser, Leonidas Guibas, Pat Hanrahan, Qixing Huang,
  Zimo Li, Silvio Savarese, Manolis Savva, Shuran Song, Hao Su, et~al.
\newblock Shapenet: An information-rich 3d model repository.
\newblock {\em arXiv preprint arXiv:1512.03012}, 2015.

\bibitem{363440}
C. {Chinrungrueng} and C.~H. {Sequin}.
\newblock Optimal adaptive k-means algorithm with dynamic adjustment of
  learning rate.
\newblock {\em IEEE Transactions on Neural Networks}, 6(1):157--169, 1995.

\bibitem{choy20163d}
Christopher~B Choy, Danfei Xu, JunYoung Gwak, Kevin Chen, and Silvio Savarese.
\newblock 3d-r2n2: A unified approach for single and multi-view 3d object
  reconstruction.
\newblock In {\em European conference on computer vision}, pages 628--644.
  Springer, 2016.

\bibitem{fan2017point}
Haoqiang Fan, Hao Su, and Leonidas~J Guibas.
\newblock A point set generation network for 3d object reconstruction from a
  single image.
\newblock In {\em Proceedings of the IEEE conference on computer vision and
  pattern recognition}, pages 605--613, 2017.

\bibitem{feldman2020neural}
Vitaly Feldman and Chiyuan Zhang.
\newblock What neural networks memorize and why: Discovering the long tail via
  influence estimation.
\newblock {\em arXiv preprint arXiv:2008.03703}, 2020.

\bibitem{Gkioxari2019MeshR}
Georgia Gkioxari, Jitendra Malik, and Justin~J Johnson.
\newblock Mesh r-cnn.
\newblock {\em 2019 IEEE/CVF International Conference on Computer Vision
  (ICCV)}, pages 9784--9794, 2019.

\bibitem{groueix2018papier}
Thibault Groueix, Matthew Fisher, Vladimir~G Kim, Bryan~C Russell, and Mathieu
  Aubry.
\newblock A papier-m{\^a}ch{\'e} approach to learning 3d surface generation.
\newblock In {\em Proceedings of the IEEE conference on computer vision and
  pattern recognition}, pages 216--224, 2018.

\bibitem{10.1007/978-3-540-87993-0_19}
Francesco Gullo, Giovanni Ponti, and Andrea Tagarelli.
\newblock Clustering uncertain data via k-medoids.
\newblock In Sergio Greco and Thomas Lukasiewicz, editors, {\em Scalable
  Uncertainty Management}, pages 229--242, Berlin, Heidelberg, 2008. Springer
  Berlin Heidelberg.

\bibitem{Hess:2010:BFE:1893021}
Roland Hess.
\newblock {\em Blender Foundations: The Essential Guide to Learning Blender
  2.6}.
\newblock Focal Press, 2010.

\bibitem{10.5555/2968618.2968725}
Geoffrey Hinton and Sam Roweis.
\newblock Stochastic neighbor embedding.
\newblock In {\em Proceedings of the 15th International Conference on Neural
  Information Processing Systems}, NIPS'02, page 857–864, Cambridge, MA, USA,
  2002. MIT Press.

\bibitem{Johnson2016PerceptualLF}
J. Johnson, Alexandre Alahi, and Li Fei-Fei.
\newblock Perceptual losses for real-time style transfer and super-resolution.
\newblock In {\em ECCV}, 2016.

\bibitem{kingma2014adam}
Diederik~P Kingma and Jimmy Ba.
\newblock Adam: A method for stochastic optimization.
\newblock {\em arXiv preprint arXiv:1412.6980}, 2014.

\bibitem{li2018point}
Chun-Liang Li, Manzil Zaheer, Yang Zhang, Barnabas Poczos, and Ruslan
  Salakhutdinov.
\newblock Point cloud gan.
\newblock {\em arXiv preprint arXiv:1810.05795}, 2018.

\bibitem{Mescheder2019OccupancyNL}
Lars~M. Mescheder, Michael Oechsle, M. Niemeyer, Sebastian Nowozin, and Andreas
  Geiger.
\newblock Occupancy networks: Learning 3d reconstruction in function space.
\newblock {\em 2019 IEEE/CVF Conference on Computer Vision and Pattern
  Recognition (CVPR)}, pages 4455--4465, 2019.

\bibitem{park2019deepsdf}
Jeong~Joon Park, Peter Florence, Julian Straub, Richard Newcombe, and Steven
  Lovegrove.
\newblock Deepsdf: Learning continuous signed distance functions for shape
  representation.
\newblock In {\em Proceedings of the IEEE Conference on Computer Vision and
  Pattern Recognition}, pages 165--174, 2019.

\bibitem{5961514}
Ville Satopaa, Jeannie Albrecht, David Irwin, and Barath Raghavan.
\newblock Finding a "kneedle" in a haystack: Detecting knee points in system
  behavior.
\newblock In {\em 2011 31st International Conference on Distributed Computing
  Systems Workshops}, pages 166--171, 2011.

\bibitem{Shin2018PixelsVA}
Daeyun Shin, Charless~C. Fowlkes, and Derek Hoiem.
\newblock Pixels, voxels, and views: A study of shape representations for
  single view 3d object shape prediction.
\newblock {\em 2018 IEEE/CVF Conference on Computer Vision and Pattern
  Recognition}, pages 3061--3069, 2018.

\bibitem{sun2018pix3d}
Xingyuan Sun, Jiajun Wu, Xiuming Zhang, Zhoutong Zhang, Chengkai Zhang, Tianfan
  Xue, Joshua~B Tenenbaum, and William~T Freeman.
\newblock Pix3d: Dataset and methods for single-image 3d shape modeling.
\newblock In {\em Proceedings of the IEEE Conference on Computer Vision and
  Pattern Recognition}, pages 2974--2983, 2018.

\bibitem{tatarchenko2017octree}
Maxim Tatarchenko, Alexey Dosovitskiy, and Thomas Brox.
\newblock Octree generating networks: Efficient convolutional architectures for
  high-resolution 3d outputs.
\newblock In {\em Proceedings of the IEEE International Conference on Computer
  Vision}, pages 2088--2096, 2017.

\bibitem{tatarchenko2019single}
Maxim Tatarchenko, Stephan~R Richter, Ren{\'e} Ranftl, Zhuwen Li, Vladlen
  Koltun, and Thomas Brox.
\newblock What do single-view 3d reconstruction networks learn?
\newblock In {\em Proceedings of the IEEE Conference on Computer Vision and
  Pattern Recognition}, pages 3405--3414, 2019.

\bibitem{tulsiani2017multi}
Shubham Tulsiani, Tinghui Zhou, Alexei~A Efros, and Jitendra Malik.
\newblock Multi-view supervision for single-view reconstruction via
  differentiable ray consistency.
\newblock In {\em Proceedings of the IEEE conference on computer vision and
  pattern recognition}, pages 2626--2634, 2017.

\bibitem{wang2018pixel2mesh}
Nanyang Wang, Yinda Zhang, Zhuwen Li, Yanwei Fu, Wei Liu, and Yu-Gang Jiang.
\newblock Pixel2mesh: Generating 3d mesh models from single rgb images.
\newblock In {\em Proceedings of the European Conference on Computer Vision
  (ECCV)}, pages 52--67, 2018.

\bibitem{wu2017marrnet}
Jiajun Wu, Yifan Wang, Tianfan Xue, Xingyuan Sun, Bill Freeman, and Josh
  Tenenbaum.
\newblock Marrnet: 3d shape reconstruction via 2.5 d sketches.
\newblock In {\em Advances in neural information processing systems}, pages
  540--550, 2017.

\bibitem{Xu2019DISNDI}
Qiangeng Xu, Weiyue Wang, D. Ceylan, R. Mech, and U. Neumann.
\newblock Disn: Deep implicit surface network for high-quality single-view 3d
  reconstruction.
\newblock In {\em NeurIPS}, 2019.

\bibitem{yan2016perspective}
Xinchen Yan, Jimei Yang, Ersin Yumer, Yijie Guo, and Honglak Lee.
\newblock Perspective transformer nets: Learning single-view 3d object
  reconstruction without 3d supervision.
\newblock In {\em Advances in neural information processing systems}, pages
  1696--1704, 2016.

\bibitem{yang2018foldingnet}
Yaoqing Yang, Chen Feng, Yiru Shen, and Dong Tian.
\newblock Foldingnet: Point cloud auto-encoder via deep grid deformation.
\newblock In {\em Proceedings of the IEEE Conference on Computer Vision and
  Pattern Recognition}, pages 206--215, 2018.

\end{thebibliography}
}

\clearpage

\appendix
\section*{Appendices}
\addcontentsline{toc}{section}{Appendices}
\renewcommand{\thesubsection}{\Alph{subsection}}

\section{Choice of The Coordinate Representation}\label{sec:definition-coordinate}
First, we provide the definition of object-centered (OC) and viewer-centered (VC) coordinates. Then, we provide quantitative and qualitative results to show the difference between shapes in the two coordinate representations.

\begin{figure}[!htb]
  \centering
  \begin{subfigure}[b]{0.49\columnwidth}
       \centering
       \includegraphics[width=\linewidth]{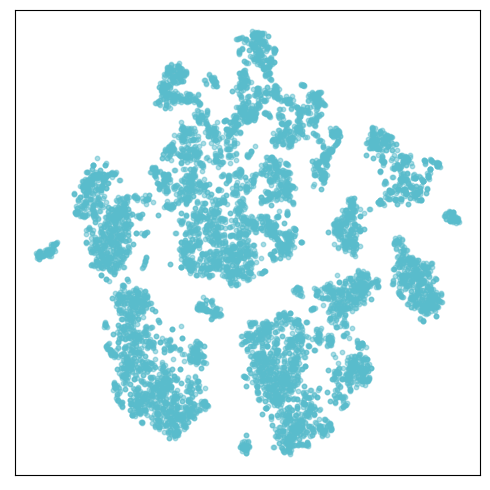}
       \caption{}
       \label{fig:tsne-oc}
   \end{subfigure}
   \begin{subfigure}[b]{0.49\columnwidth}
        \centering
        \includegraphics[width=\linewidth]{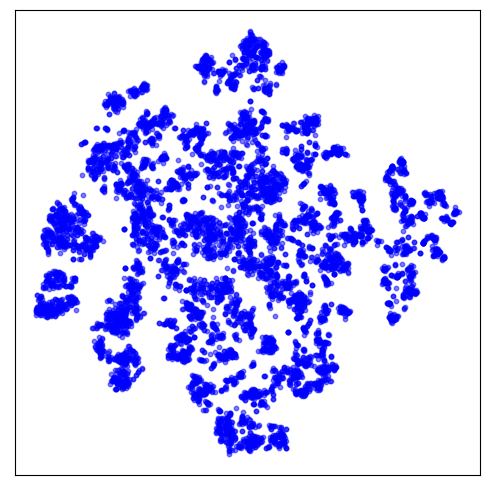}
        \caption{}
        \label{fig:tsne-vc}
   \end{subfigure}
   \begin{subfigure}[b]{0.84\columnwidth}
        \centering
        \includegraphics[width=\linewidth]{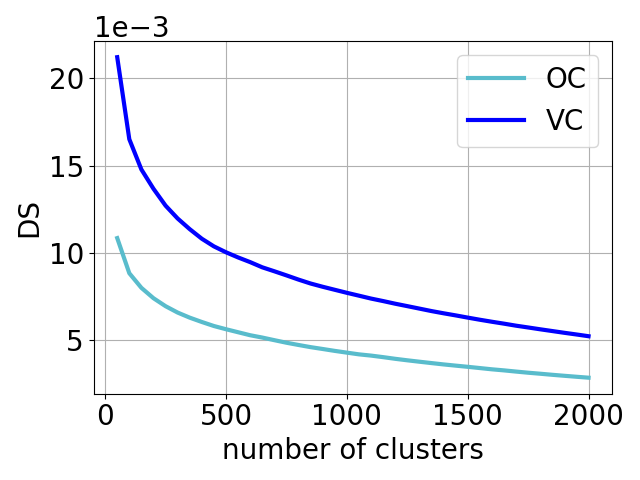}
        \caption{}
        \label{fig:ds-ocvc}
   \end{subfigure} \vspace{-3mm}
   \caption{(a) t-SNE of distance matrix of shapes in OC coordinate. (b) t-SNE of distance matrix of shapes in VC coordinate. (c) The DS of shapes in OC and VC with varying number of clusters.}
   \label{fig:oc-vc-dispersion-demo}
\end{figure}

\begin{figure*}[!h]
\begin{centering}
	\begin{tabular}{c@{} c@{} c@{}|c@{} c@{} c@{}}
	\toprule
    Image & \begin{tabular}[c]{c@{}}OC\\ \small{(z axis aligned)} \end{tabular} & \begin{tabular}[c]{c@{}}VC\\ \small{(viewpoint aligned)} \end{tabular} & Image & \begin{tabular}[c]{c@{}}OC\\ \small{(z axis aligned)} \end{tabular} & \begin{tabular}[c]{c@{}}VC\\ \small{(viewpoint aligned)} \end{tabular}\\
	\toprule
    \includegraphics[width=0.34\columnwidth,keepaspectratio]{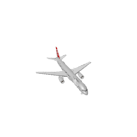} &
    \includegraphics[width=0.32\columnwidth,keepaspectratio]{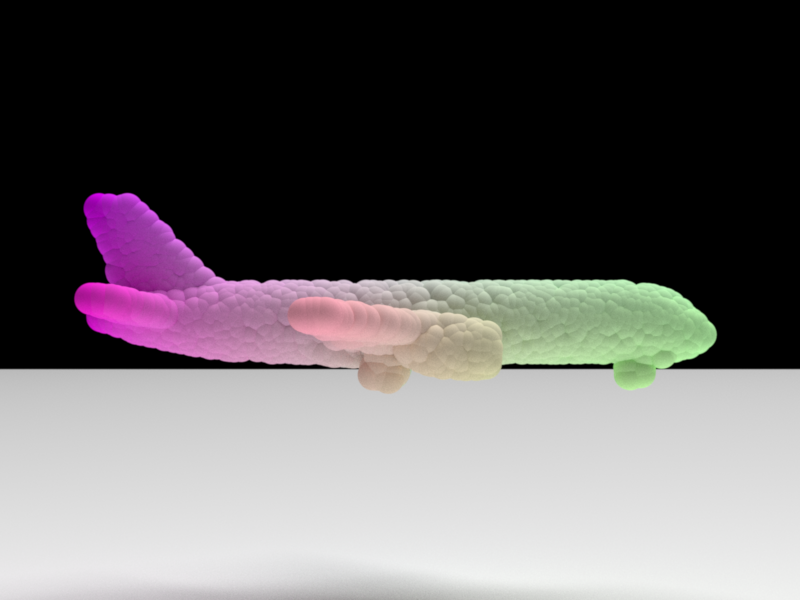} \hspace{1mm} & 
    \includegraphics[width=0.32\columnwidth,keepaspectratio]{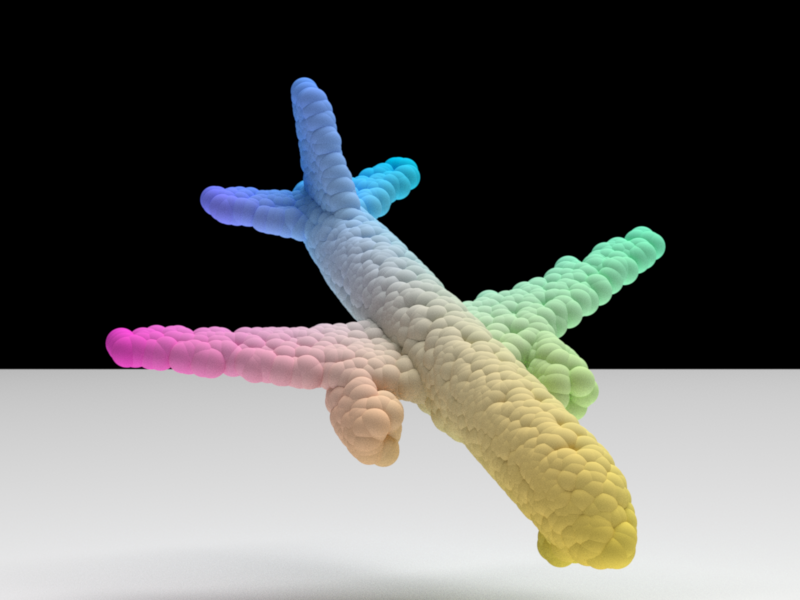} \hspace{1mm} &
    \includegraphics[width=0.34\columnwidth,keepaspectratio]{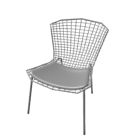} &
    \includegraphics[width=0.32\columnwidth,keepaspectratio]{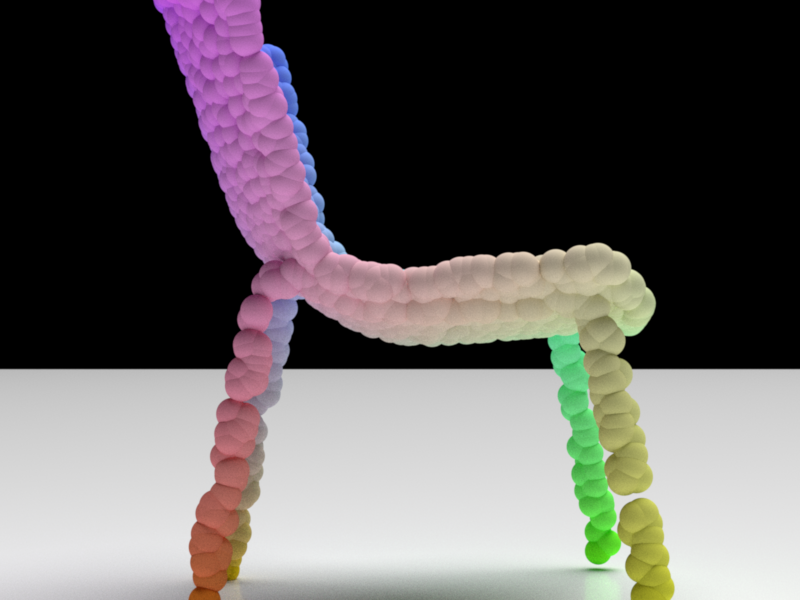} \hspace{1mm} & 
    \includegraphics[width=0.32\columnwidth,keepaspectratio]{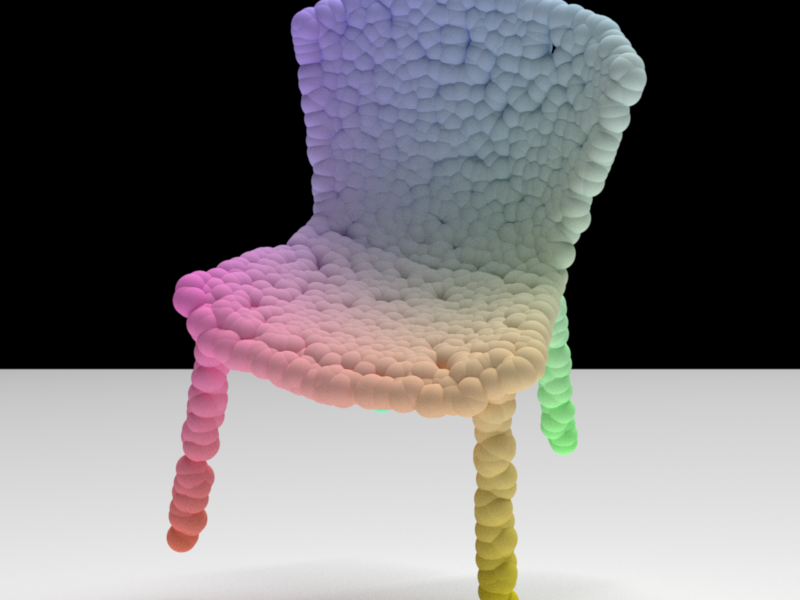} \\
    \includegraphics[width=0.34\columnwidth,keepaspectratio]{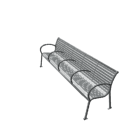} &
    \includegraphics[width=0.32\columnwidth,keepaspectratio]{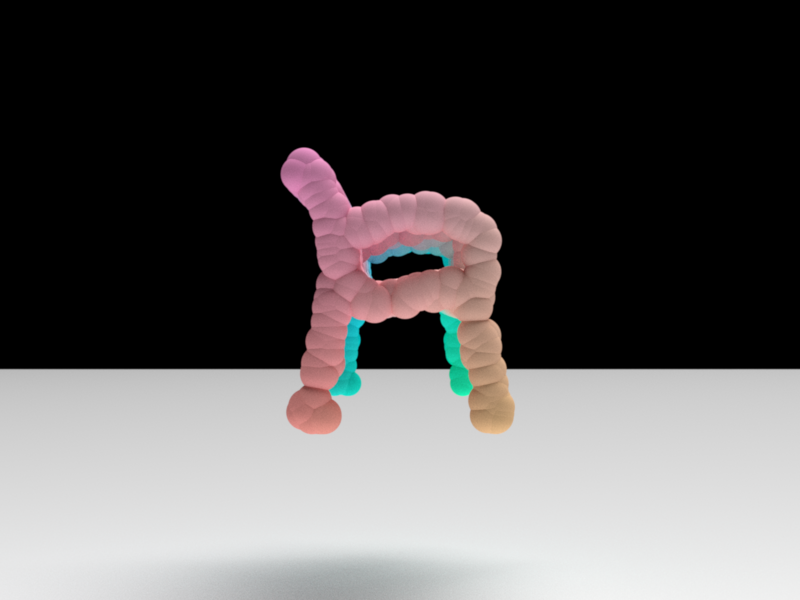} \hspace{1mm} & 
    \includegraphics[width=0.32\columnwidth,keepaspectratio]{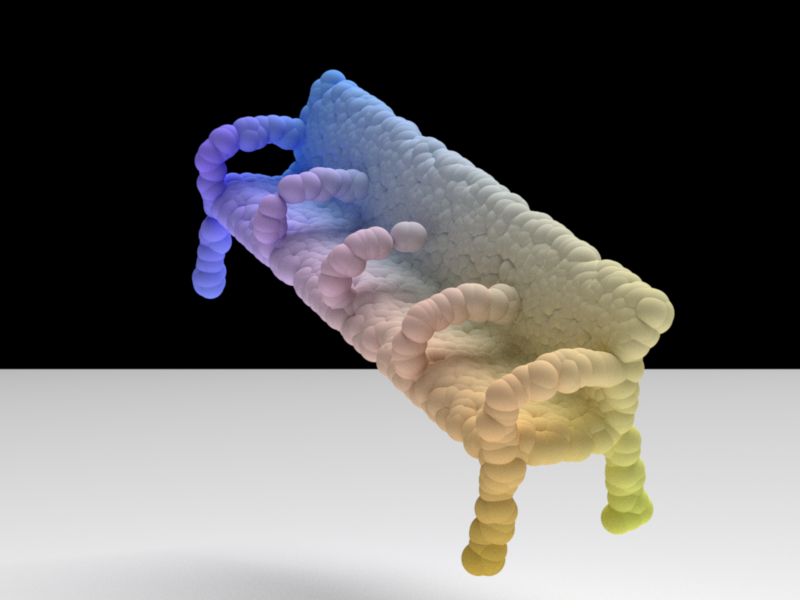} \hspace{1mm} &
    \includegraphics[width=0.34\columnwidth,keepaspectratio]{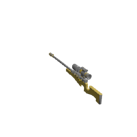} &
    \includegraphics[width=0.32\columnwidth,keepaspectratio]{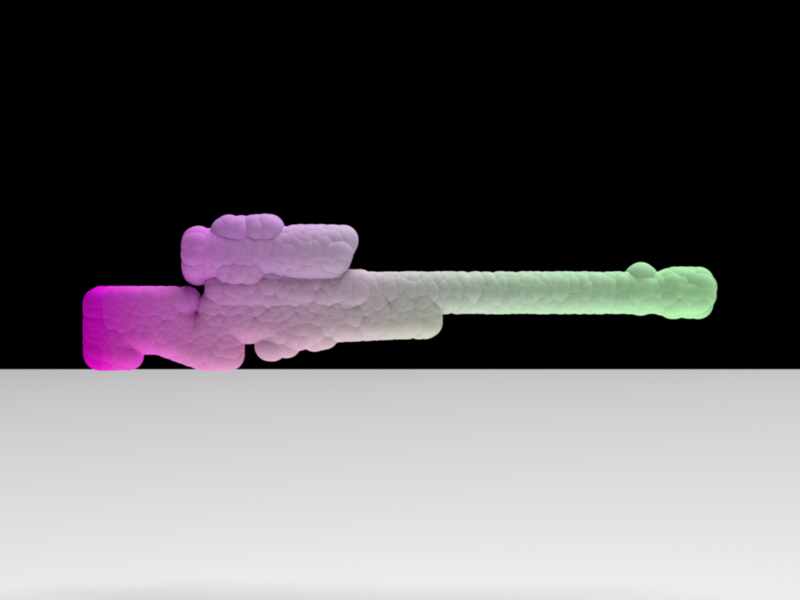} \hspace{1mm} & 
    \includegraphics[width=0.32\columnwidth,keepaspectratio]{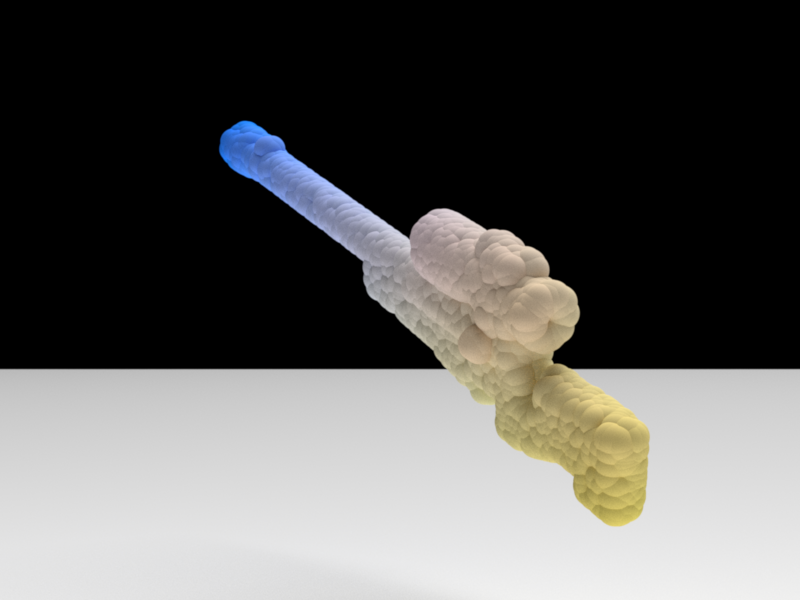}
    \\
    \includegraphics[width=0.34\columnwidth,keepaspectratio]{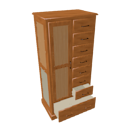} &
    \includegraphics[width=0.32\columnwidth,keepaspectratio]{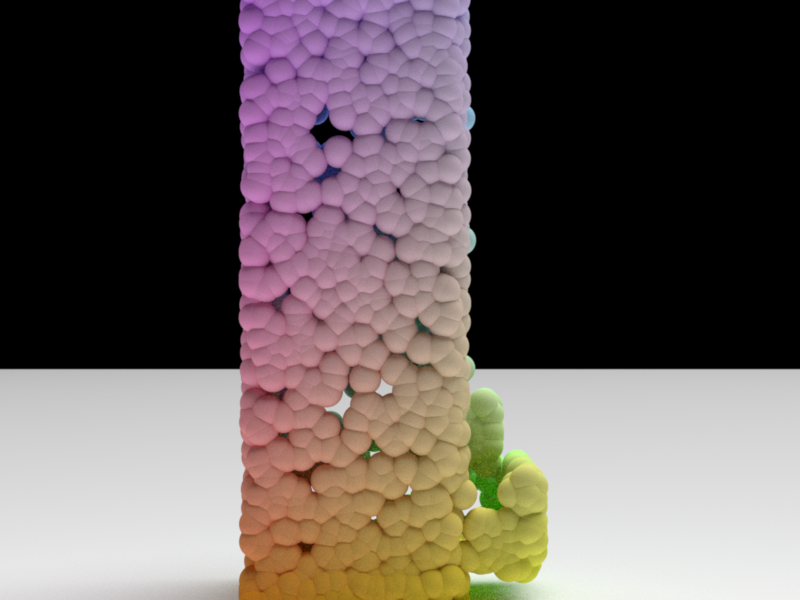} \hspace{1mm} & 
    \includegraphics[width=0.32\columnwidth,keepaspectratio]{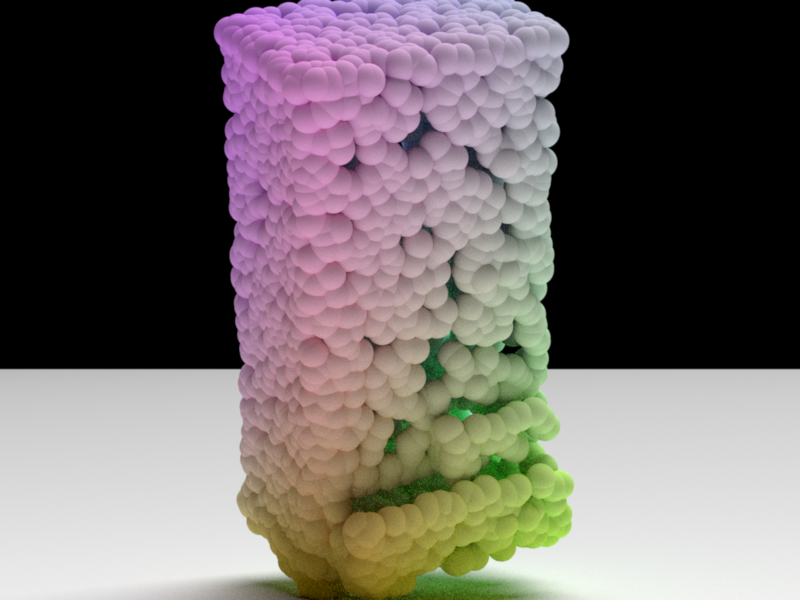} \hspace{1mm} & 
    \includegraphics[width=0.34\columnwidth,keepaspectratio]{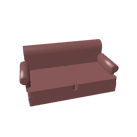} &
    \includegraphics[width=0.32\columnwidth,keepaspectratio]{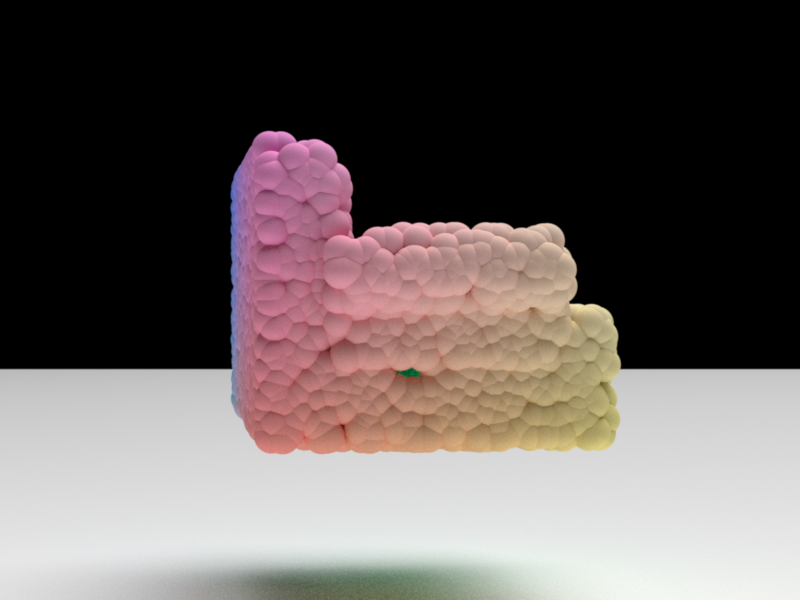} \hspace{1mm} & 
    \includegraphics[width=0.32\columnwidth,keepaspectratio]{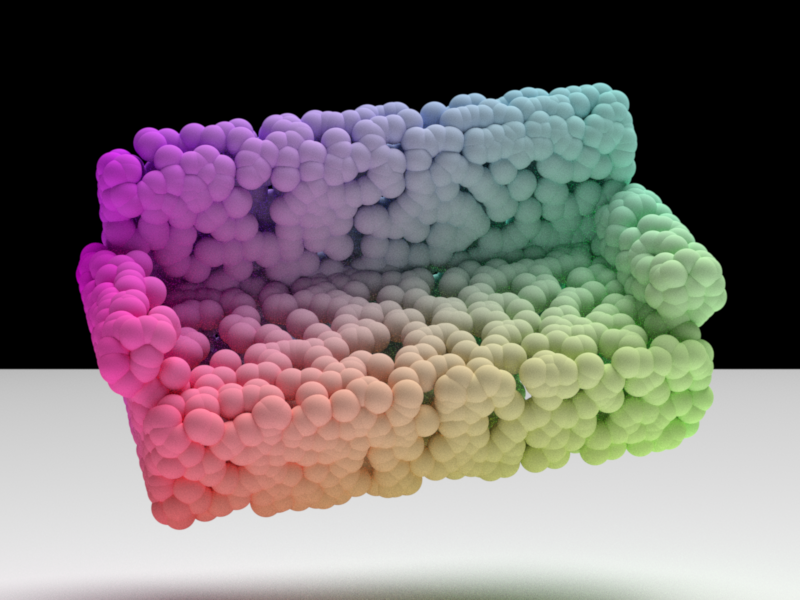}
    \\
    \includegraphics[width=0.34\columnwidth,keepaspectratio]{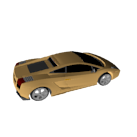} &
    \includegraphics[width=0.32\columnwidth,keepaspectratio]{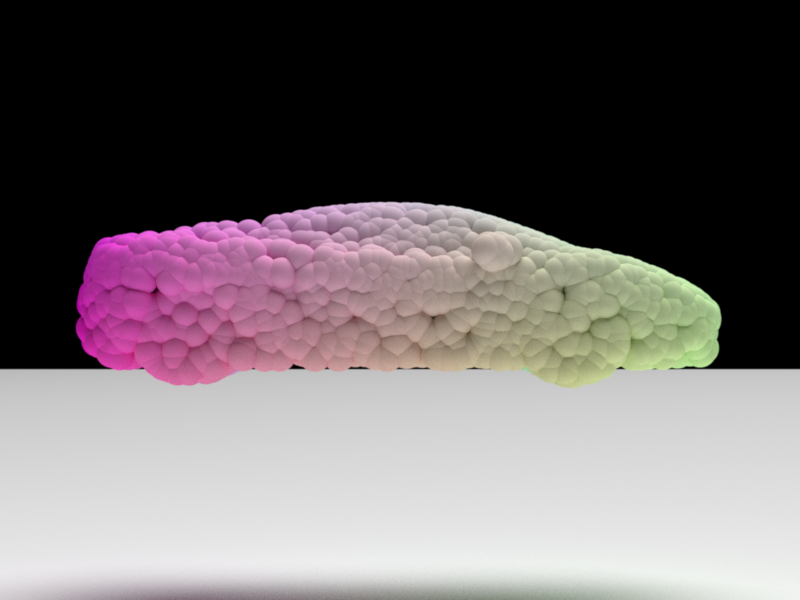} \hspace{1mm} & 
    \includegraphics[width=0.32\columnwidth,keepaspectratio]{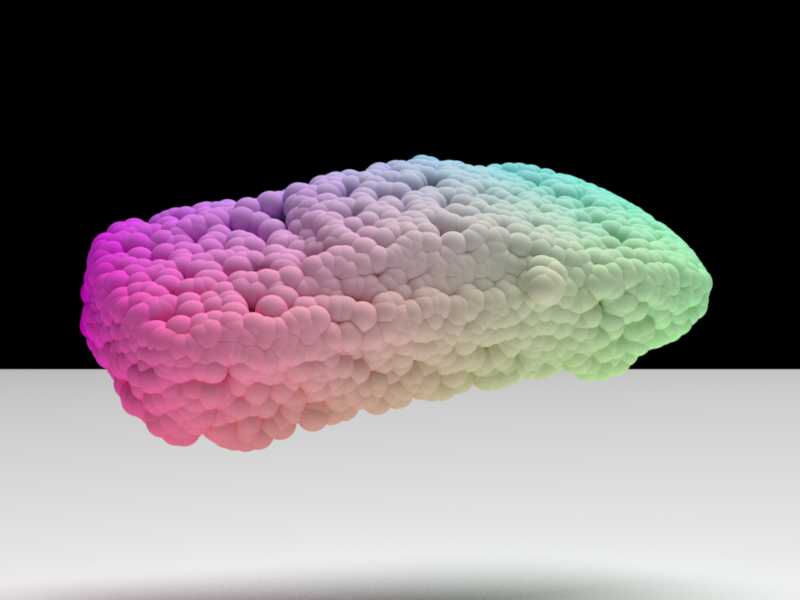}
    \hspace{1mm} &
    \includegraphics[width=0.34\columnwidth,keepaspectratio]{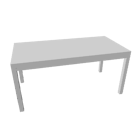} &
    \includegraphics[width=0.32\columnwidth,keepaspectratio]{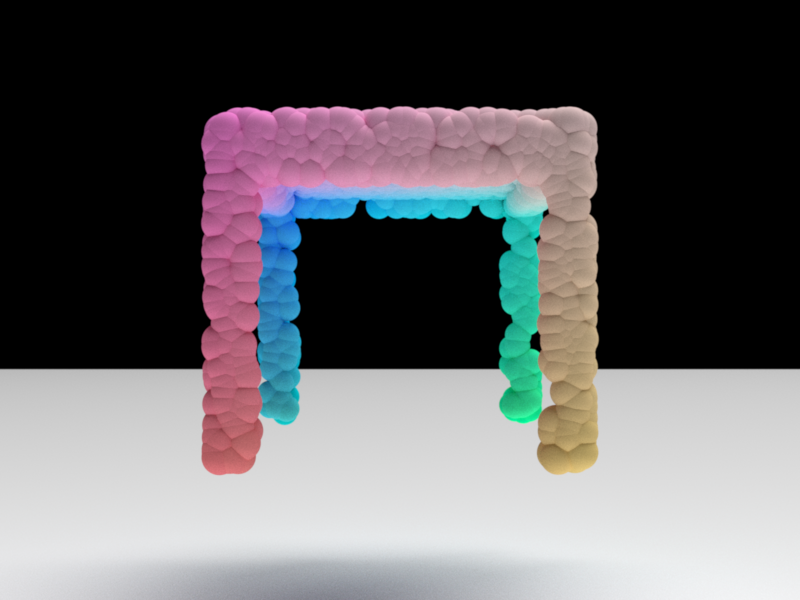} \hspace{1mm} & 
    \includegraphics[width=0.32\columnwidth,keepaspectratio]{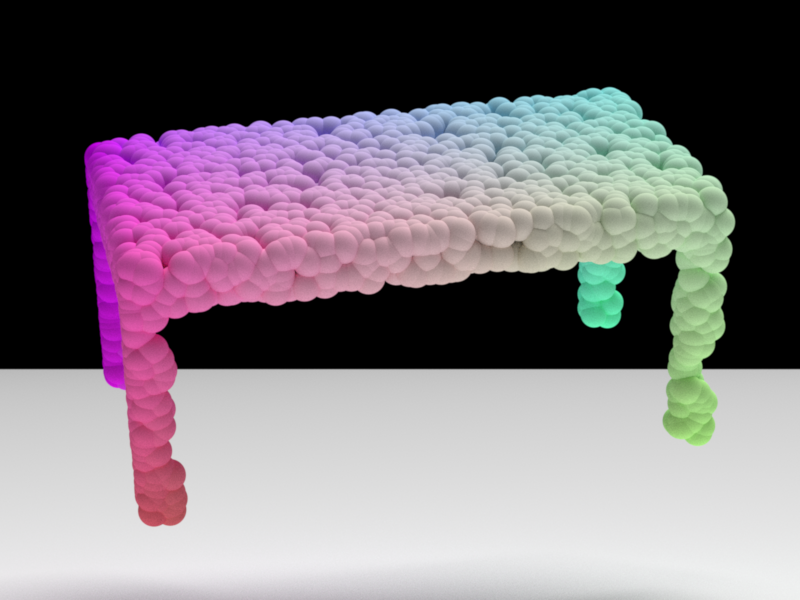}
	\end{tabular}  
	\caption{Visualization of input images and shapes in OC and VC coordinates of ShapeNet~\cite{chang2015shapenet}.}
	\vspace{-3mm}
	\label{fig:oc-vc-vis}
\end{centering}
\end{figure*}

As shown in Figure \ref{fig:oc-vc-vis}, we visualize some shapes in ShapeNet both in OC and VC coordinates. Given a single RGB image as input, we want to predict the 3D shape of the object from which the image is taken. In the OC coordinate, the shapes are predicted in canonical coordinates specified in the training set. For example, in the ShapeNetCore~\cite{chang2015shapenet} dataset, the $\left(\theta_{\mathrm{az}}=0^{\circ}, \theta_{\mathrm{el}}=0^{\circ}\right)$ direction corresponds to the commonly agreed front of the object, where $\theta_{\mathrm{az}}$ and $\theta_{\mathrm{el}}$ are the azimuth and elevation angle of viewpoint. 
In the VC coordinate, the NN is supervised to predict a pre-aligned 3D shape in the input image’s reference frame. The image-shape pair ensures that $\left(\theta_{\mathrm{az}}=0^{\circ}, \theta_{\mathrm{el}}=0^{\circ}\right)$ in the output coordinate system always corresponds to the input viewpoint. 

We further show the different impacts of the two coordinates on training shapes. The main difference is that shapes in OC are more clustered, while shapes in VC are much more dispersed. We use all the shapes of the ShapeNetCore~\cite{chang2015shapenet} test set split by \cite{choy20163d} and represent them both in OC and VC coordinates. There are 8762 shapes in total. These shapes cover 13 semantic classes, and each of them is represented as a point cloud with 2500 3D points.
First, we compute distance matrices of shapes using Chamfer distance as the distance function. 
Then, we visualize the matrices by t-SNE~\cite{10.5555/2968618.2968725} in Figure \ref{fig:tsne-oc} and \ref{fig:tsne-vc}. 
Comparing Figure \ref{fig:tsne-oc} with \ref{fig:tsne-vc}, we see that Figure \ref{fig:tsne-oc} shows a more clustered pattern, while Figure \ref{fig:tsne-vc} shows a more dispersed pattern. 
It indicates that shapes in OC are more clustered than those in VC. Besides, we also measure the DS of shapes. We sweep the number of clusters (NC) from 50 to 2000 with step size 50. The results are shown in Figure \ref{fig:ds-ocvc}. The DS of shapes in VC is clearly larger, indicating that the VC coordinate makes shapes more dispersed.

\section{Ablation Study of Clustering Methodology}\label{sec:ablation-study-clustering}
\begin{figure*}[htb!]
  \centering
  \begin{subfigure}[b]{0.24\linewidth}
       \includegraphics[width=\linewidth]{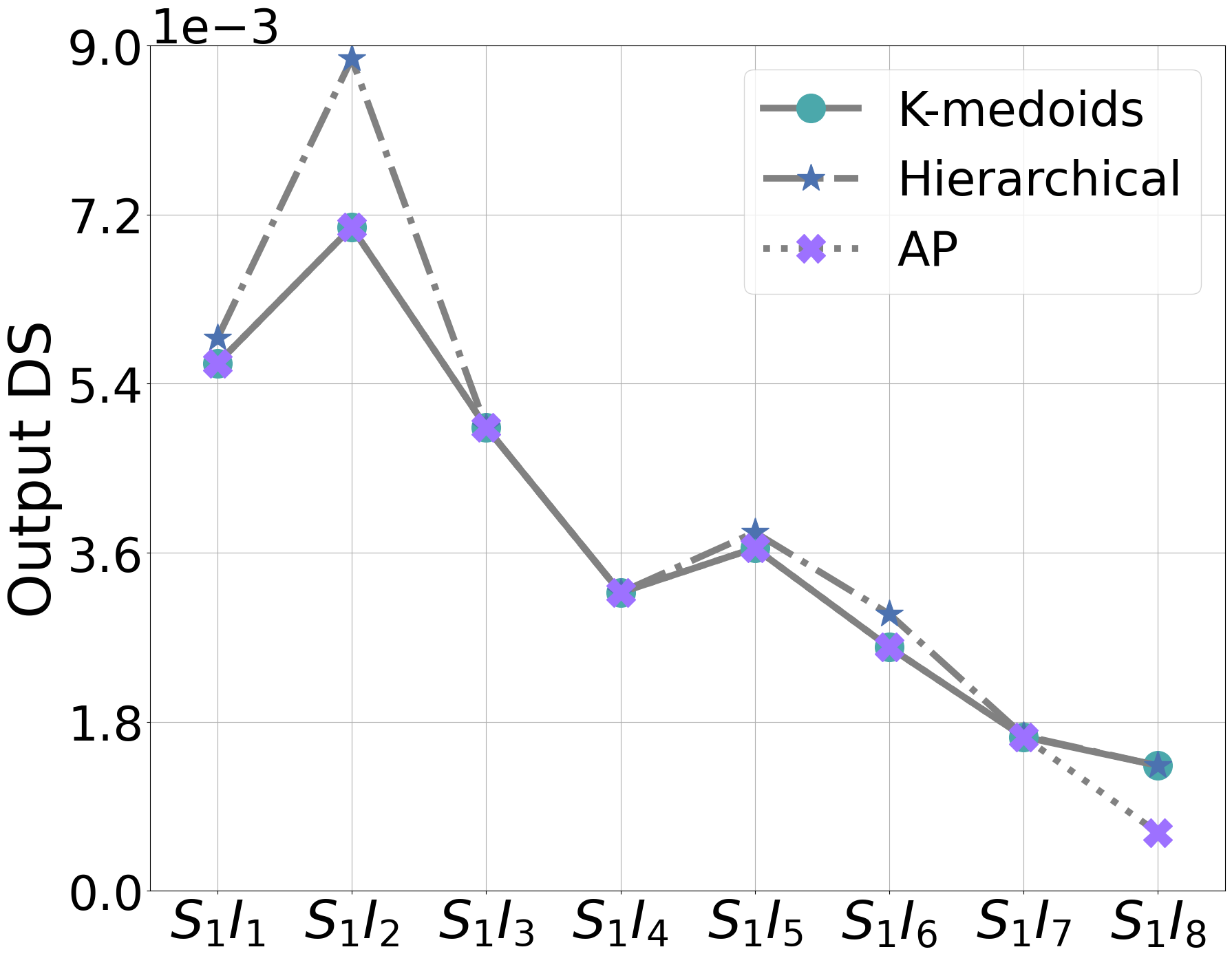}
       \caption{}
       \label{fig:toy-pred-moreimg}
   \end{subfigure}
   \begin{subfigure}[b]{0.24\linewidth}
        \includegraphics[width=\linewidth]{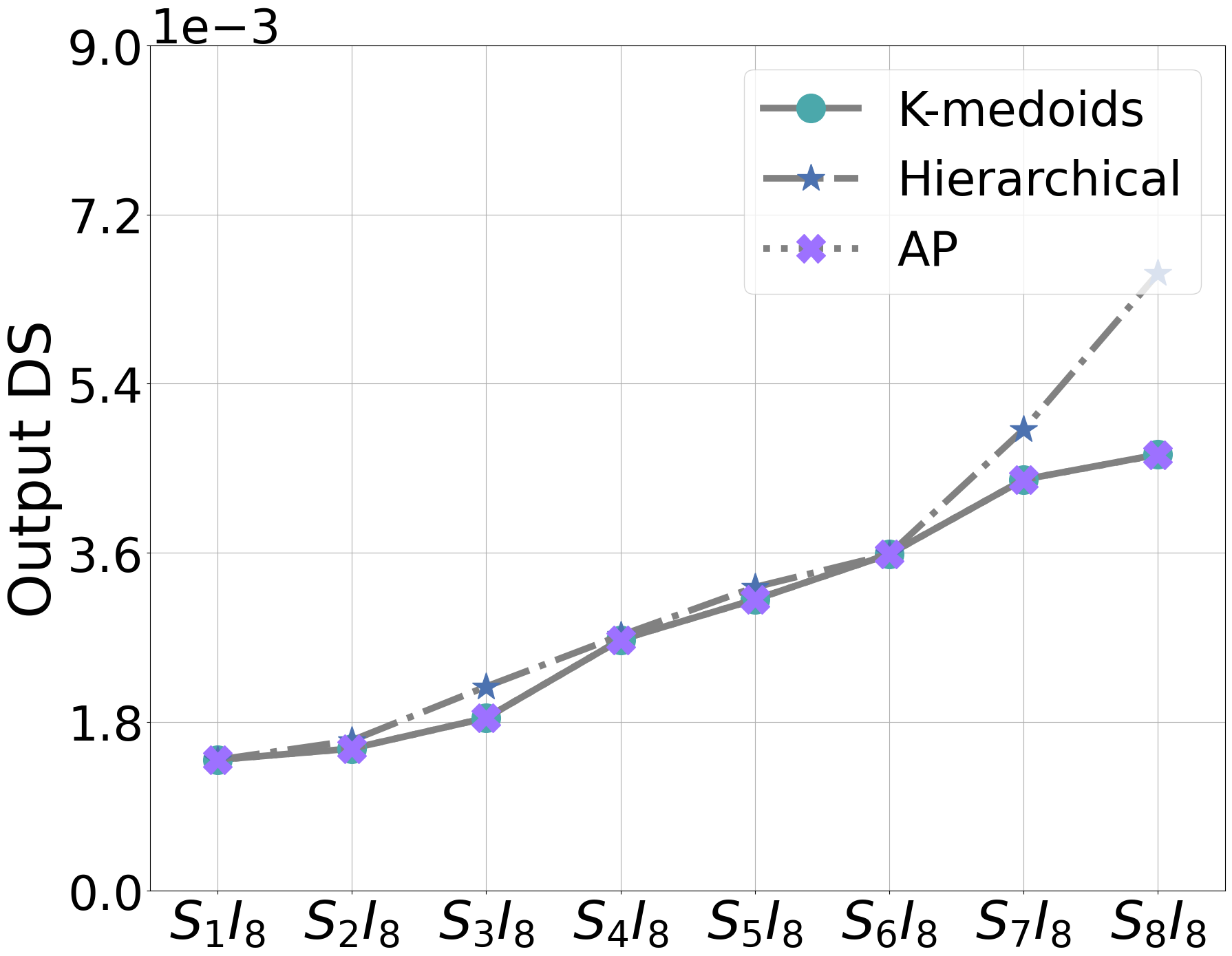}
        \caption{}
        \label{fig:toy-pred-moreshape}
   \end{subfigure}
   \begin{subfigure}[b]{0.24\linewidth}
        \includegraphics[width=\linewidth]{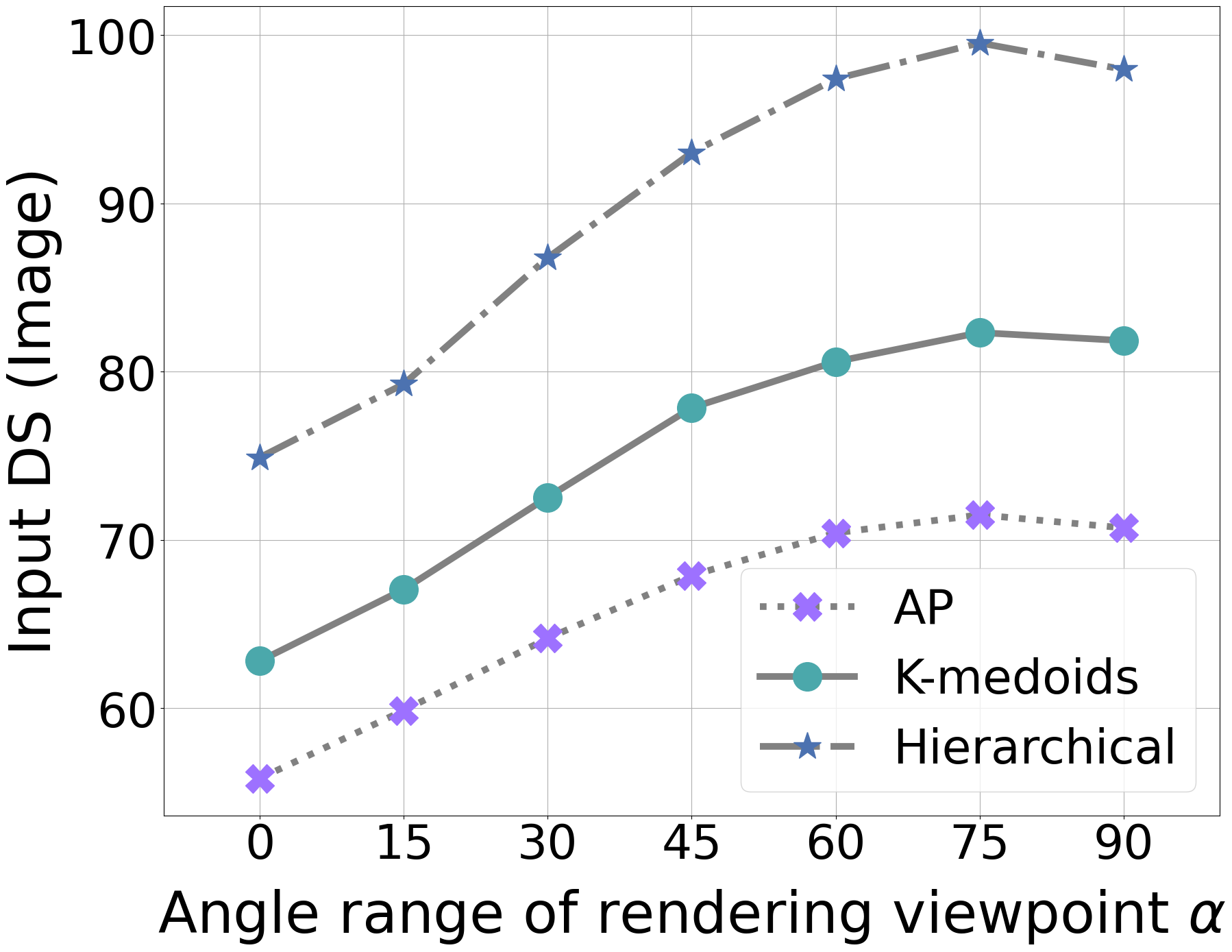}
        \caption{}
        \label{fig:shape-input-moreimg}
   \end{subfigure}
    \begin{subfigure}[b]{0.24\linewidth}
        \includegraphics[width=\linewidth]{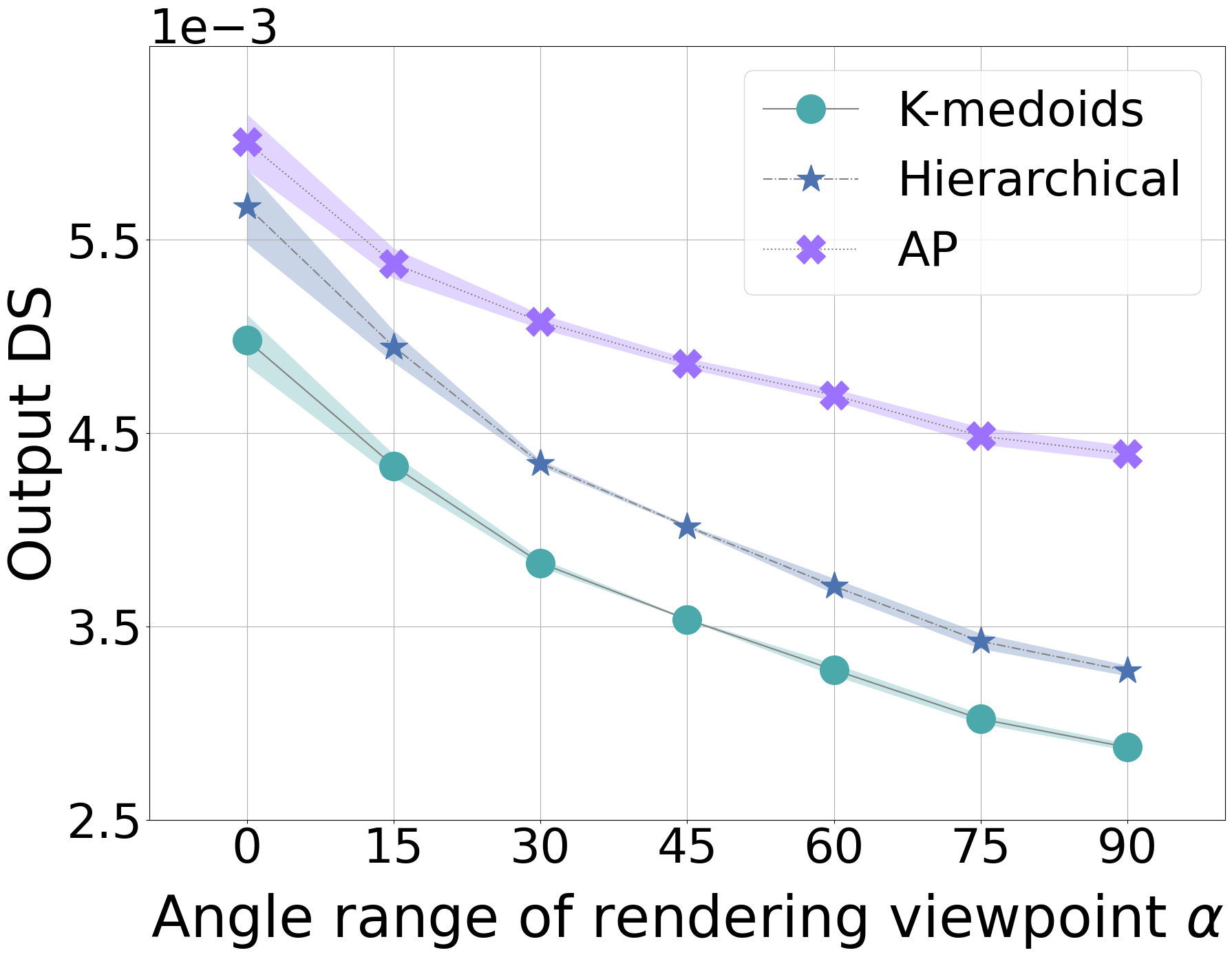}
        \caption{}
        \label{fig:shape-pred-moreimg}
   \end{subfigure}
   \caption{DS from different clustering methods. (a)(b) Output DS of models trained on synthetic dataset with more dispersed training images (a) and shapes (b). (c)(d) Input/Ouput DS of models trained on ShapeNet with more dispersed training images.}
   \label{fig:two-more-clustering}
\end{figure*}

We study two more common clustering methods, namely hierarchical clustering and affinity propagation (AP), besides the K-medoids method.
Figure \ref{fig:two-more-clustering} and Table \ref{tb:two-more-clustering} show the DS obtained with the three clustering methods evaluated on all the main experiments on both synthetic and ShapeNet datasets.
The results show consistent trends for different clustering algorithms. 

We provide the implementation detail of clustering methods~\footnote{All the methods are implemented based on scikit-learn package.}. K-medoids and hierarchical clustering require assigning the NC. The value of NC for the synthetic dataset is 2 and for Shapenet is 500. It is automatically determined by the ``Kneedle'' method detailed in Appendix~\ref{sec:hyperp-tuning}. K-medoids is initialized by the ``k-means++''~\cite{10.5555/1283383.1283494}. The linkage criterion of hierarchical clustering is the maximum distances between all observations of the two sets. For affinity propagation, we construct the affinity matrix by normalizing the distance matrix by standard deviation and negative exponential. Then we set the hyperparameter ``perference'', which controls how many exemplars are used, to be the top $q$-th percentile of entries in the affinity matrix. For synthetic dataset, $q = 4$. For ShapeNet, $q = 60$.

\begin{table}[htb!] 
\resizebox{\columnwidth}{!}{%
\begin{tabular}{c|cc|cc|cc}
\hline
              & \multicolumn{2}{c|}{K-medoids}     & \multicolumn{2}{c}{Hierarchical} \vline  &  \multicolumn{2}{c}{Affinity Propagation}   \\ \hline
              & OC & VC      & OC & VC     & OC & VC  \\ \hline
PSGN          & $1.63\pm0.07$   & $\boldsymbol{2.47\pm0.00}$   & $1.84\pm0.09$  & $\boldsymbol{2.72\pm0.00}$      & $2.51\pm0.13$       & $\boldsymbol{3.31\pm0.00}$   \\
FoldingNet    & $2.39\pm0.04$   &  $\boldsymbol{3.34\pm0.05}$    & $2.66\pm0.03$  & $\boldsymbol{3.78\pm0.07}$    & $3.55\pm0.02$       & $\boldsymbol{4.53\pm0.04}$   \\
AtlasNet-Sph. & $2.83\pm0.00$   &  $\boldsymbol{3.60\pm0.02}$    & $3.18\pm0.00$  & $\boldsymbol{4.04\pm0.04}$    & $4.21\pm0.01$       & $\boldsymbol{5.04\pm0.05}$   \\
AtlasNet-25   & $2.84\pm0.01$   &   $\boldsymbol{3.55\pm0.01}$   &  $3.17\pm0.01$   & $\boldsymbol{3.95\pm0.03}$  & $4.20\pm0.04$       & $\boldsymbol{4.96\pm0.01}$   \\
\hline
\end{tabular}
}
\caption{Output DS (mean$\pm$stdev, $\times$ 0.001, $\uparrow$) of models trained on more dispersed training shapes in ShapeNet.}
\label{tb:two-more-clustering}
\end{table}

\section{Hyperparameter of Dispersion Score}\label{sec:hyperp-tuning}

\begin{figure} 
  \centering
  \begin{subfigure}[b]{0.48\columnwidth}
       \includegraphics[width=\linewidth]{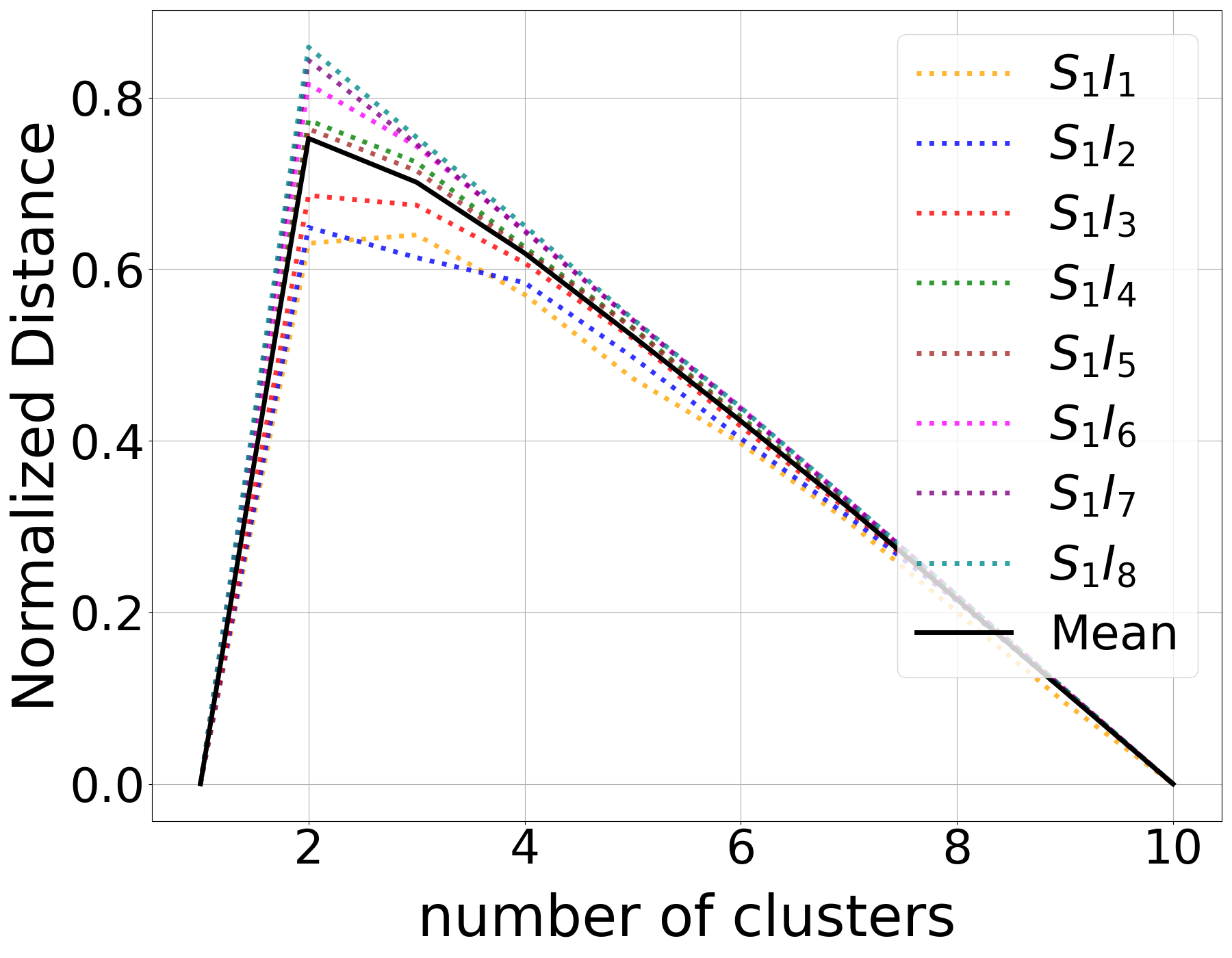}
       \caption{}\centering 
       \label{fig:knee-toy-image}
   \end{subfigure}
     \begin{subfigure}[b]{0.48\columnwidth}
       \includegraphics[width=\linewidth]{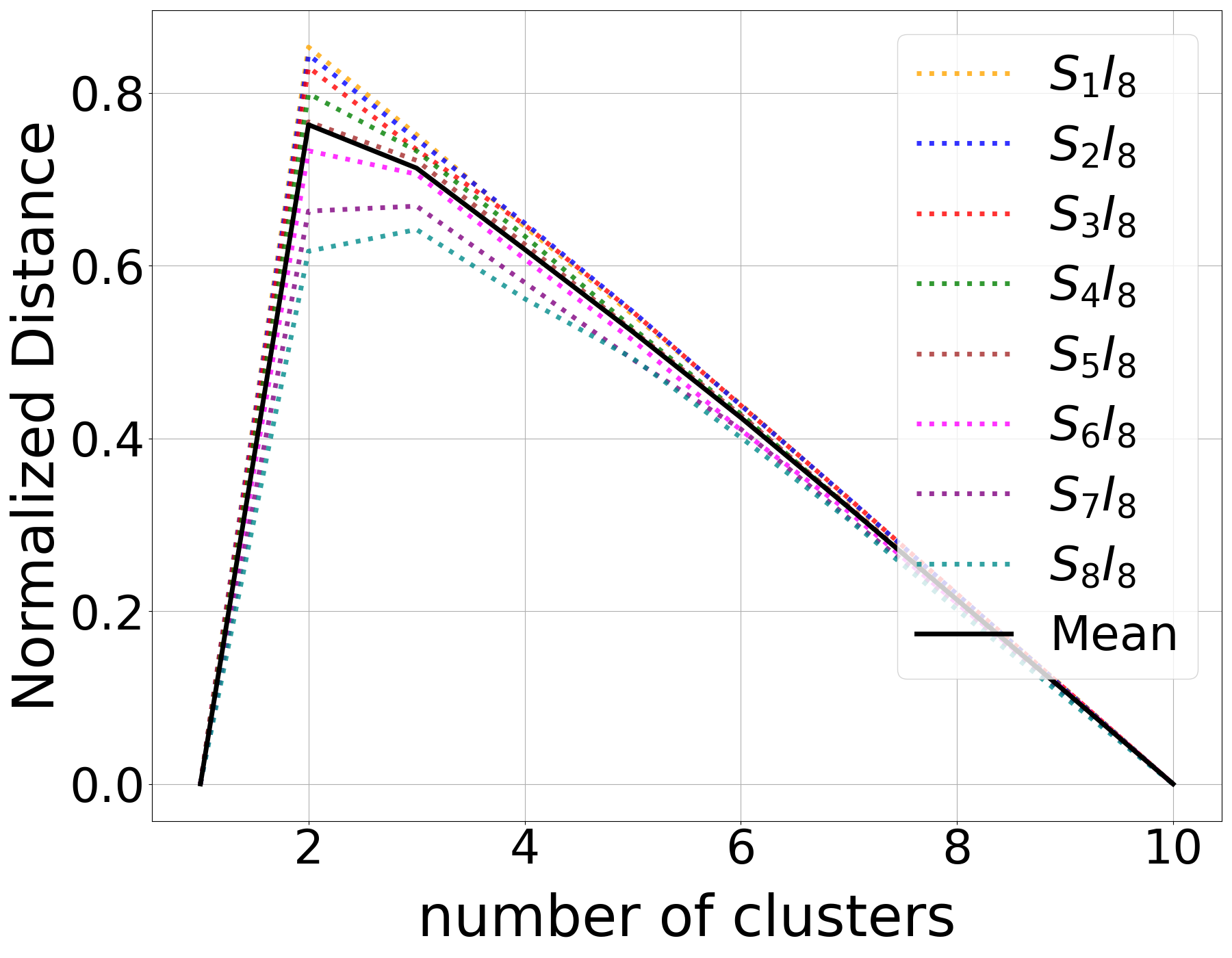}
       \caption{}\centering 
       \label{fig:knee-toy-shape}
   \end{subfigure}
   
   \begin{subfigure}[b]{0.48\columnwidth}
        \includegraphics[width=\linewidth]{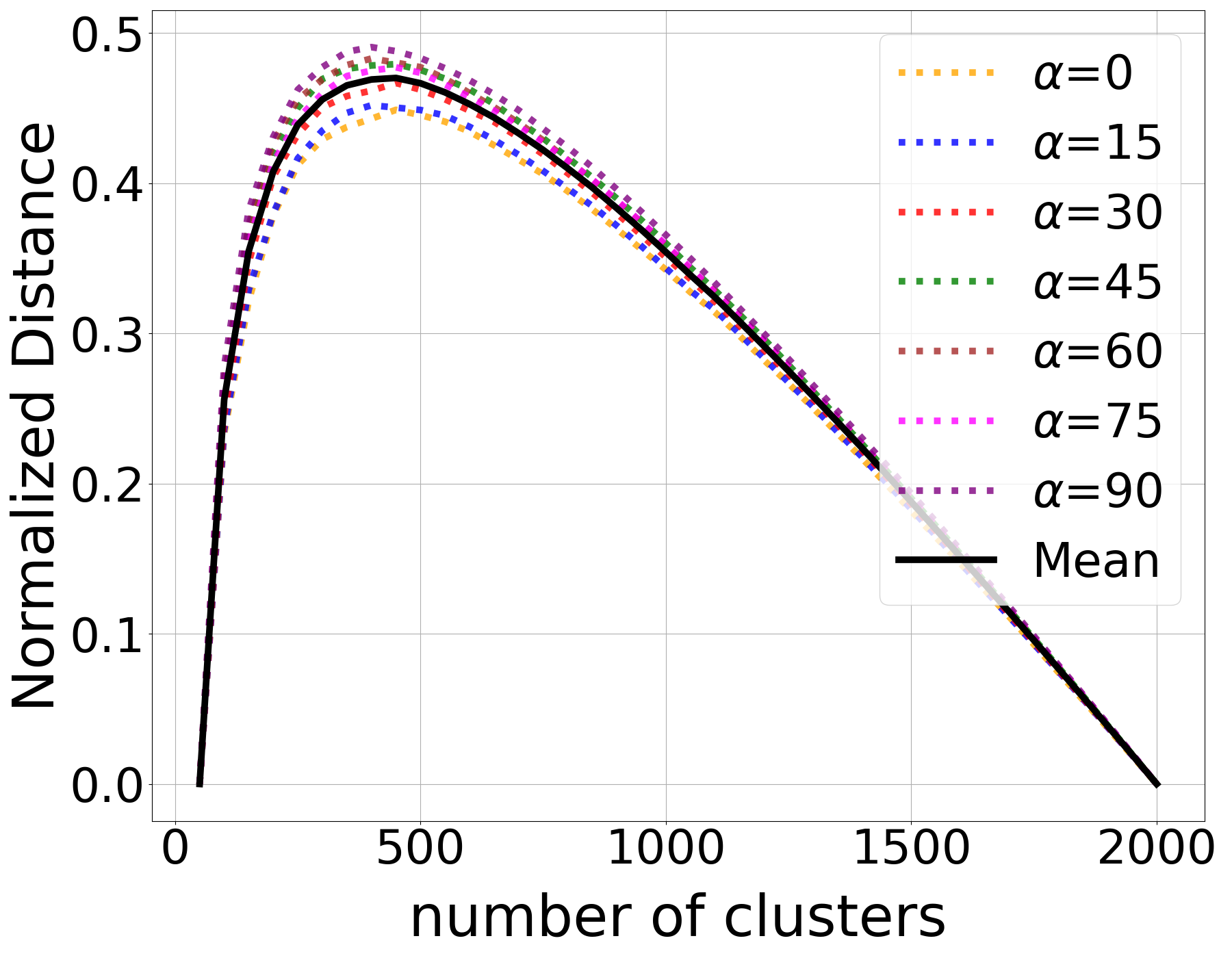}
        \caption{}\centering
        \label{fig:knee-image}
   \end{subfigure} 
   \begin{subfigure}[b]{0.48\columnwidth}
        \includegraphics[width=\linewidth]{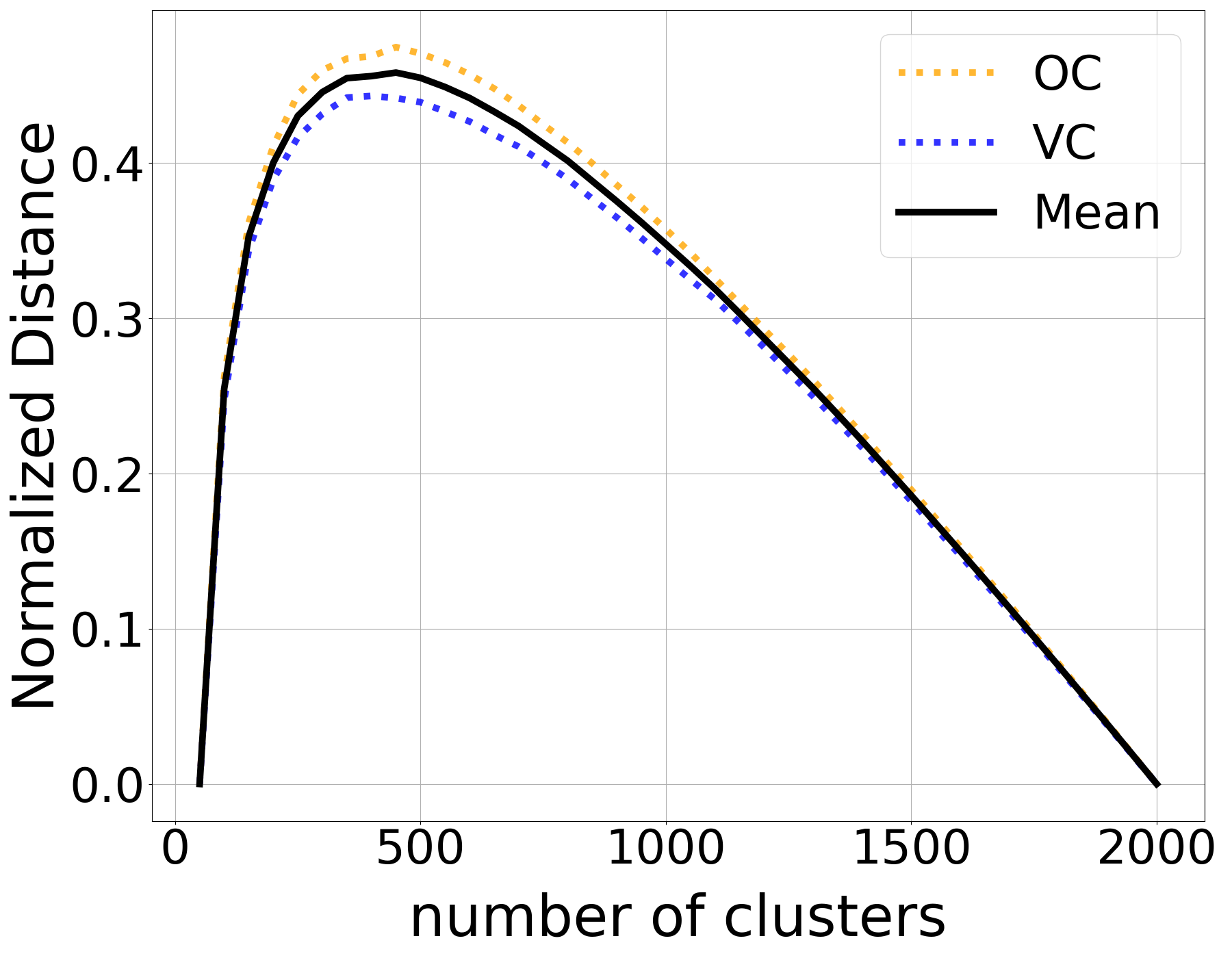}
        \caption{}\centering
        \label{fig:knee-shape}
   \end{subfigure} 
   \caption {The normalized distance computed in Kneedle. (a)(b) Synthetic dataset with more dispersed training images and shapes. (c)(d) ShapeNet with more dispersed training images and shapes.} 
   \label{fig:knee}
\end{figure}

In this section, we explain our method to tune the number of clusters (NC) in the proposed DS metric.
To choose the best NC, we do a parameter sweeping and choose the value equal to the elbow of the DS curves. We automatically determine the elbow using the ``Kneedle'' method in \cite{5961514}. This method approximately finds the point with the maximum curvature using a score called ``normalized distance'' and selects that as the elbow. See Figure \ref{fig:knee}. For each experiment, we compute the mean of the normalized distance across different trained models to determine the NC. For the synthetic dataset, the NC is 2. For ShapeNet, the NC is 500.

\section{Synthetic Data Generation}\label{sec:synthetic-data}

\begin{figure}[!ht]
	\begin{tabular}{c@{}c@{}c@{}c@{}c@{}c@{}c@{}c@{}c@{}c@{}c@{}}
	    sphere & & & & & $\Longleftrightarrow$ & & & &  cube \\
		\includegraphics[width=.10\columnwidth,keepaspectratio]{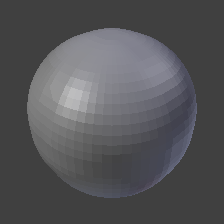} & \hspace{-2mm}
		\includegraphics[width=.10\columnwidth,keepaspectratio]{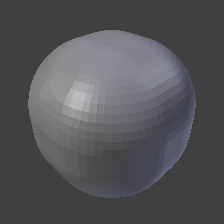} &
		\includegraphics[width=.10\columnwidth,keepaspectratio]{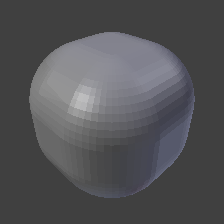} &
		\includegraphics[width=.10\columnwidth,keepaspectratio]{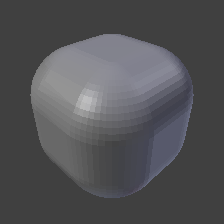} &
		\includegraphics[width=.10\columnwidth,keepaspectratio]{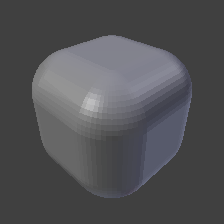} &
		\includegraphics[width=.10\columnwidth,keepaspectratio]{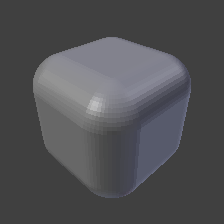} &
		\includegraphics[width=.10\columnwidth,keepaspectratio]{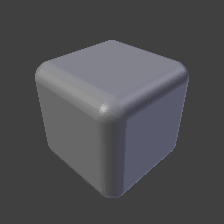} &
		\includegraphics[width=.10\columnwidth,keepaspectratio]{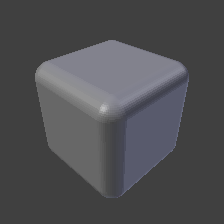} &
		\includegraphics[width=.10\columnwidth,keepaspectratio]{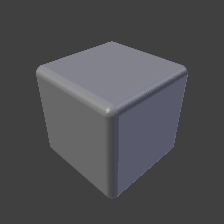} &
		\includegraphics[width=.10\columnwidth,keepaspectratio]{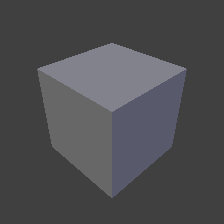} \\  
	    \vspace{2mm}
	    sphere & & & & & $\Longleftrightarrow$ & & & &  cube \\
		\includegraphics[width=.10\columnwidth,keepaspectratio]{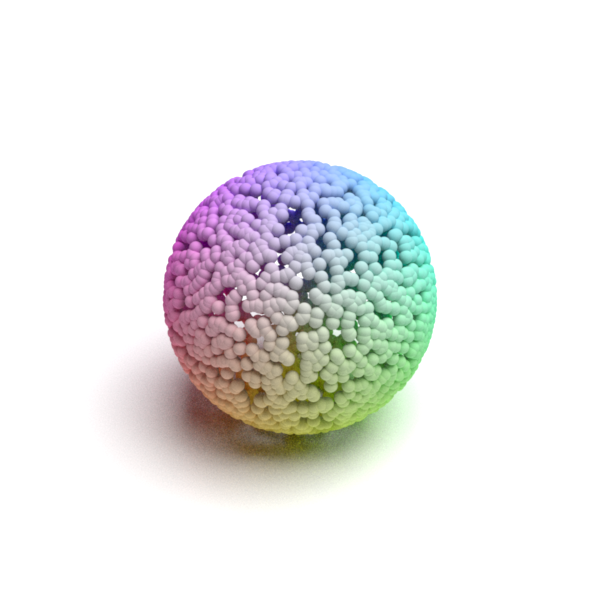} & \hspace{-2mm}
		\includegraphics[width=.10\columnwidth,keepaspectratio]{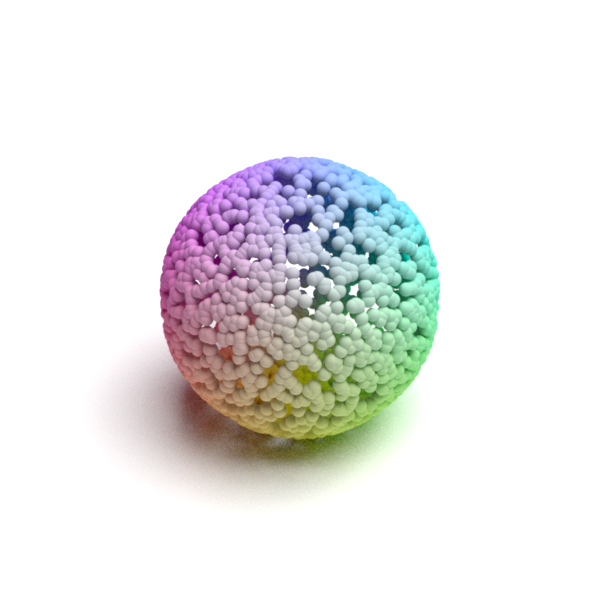} &
		\includegraphics[width=.10\columnwidth,keepaspectratio]{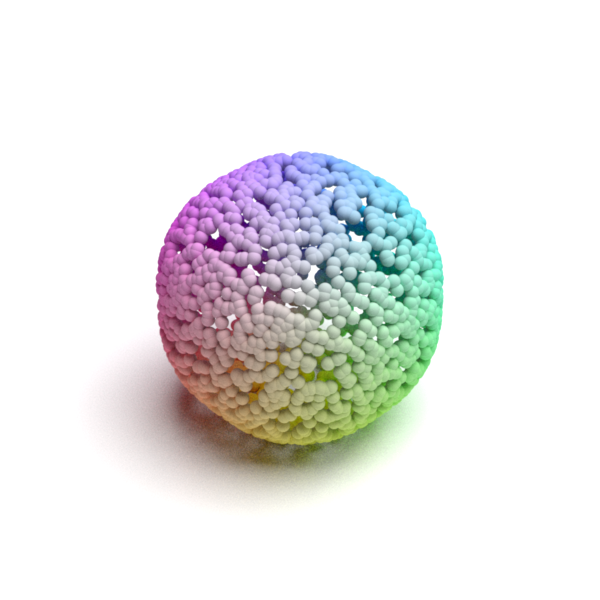} &
		\includegraphics[width=.10\columnwidth,keepaspectratio]{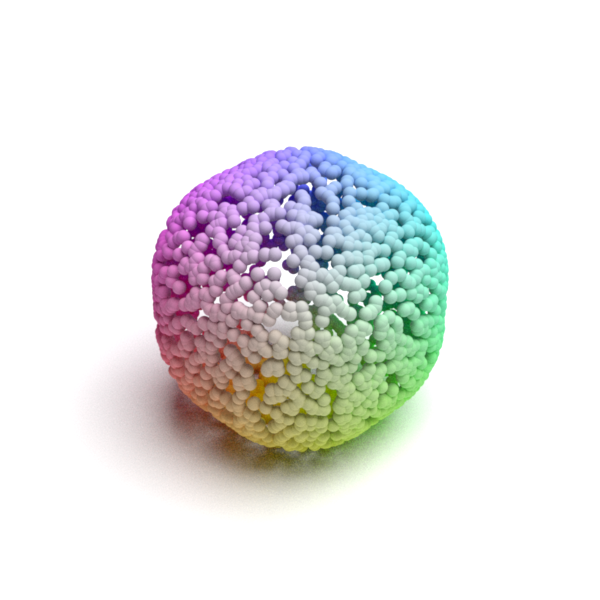} &
		\includegraphics[width=.10\columnwidth,keepaspectratio]{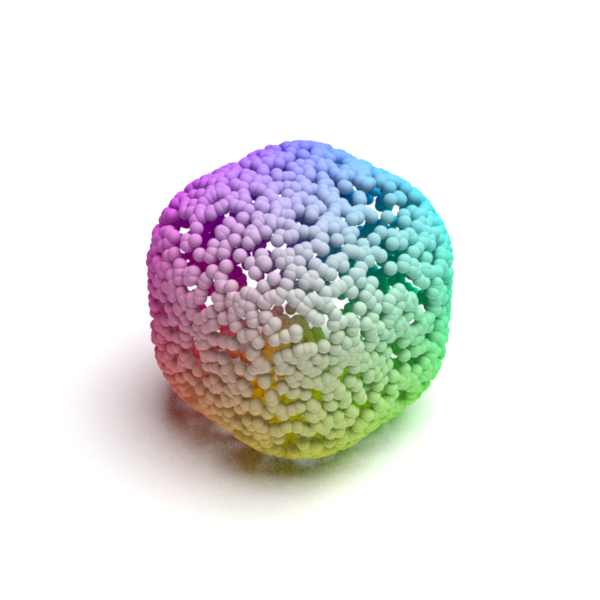} &
		\includegraphics[width=.10\columnwidth,keepaspectratio]{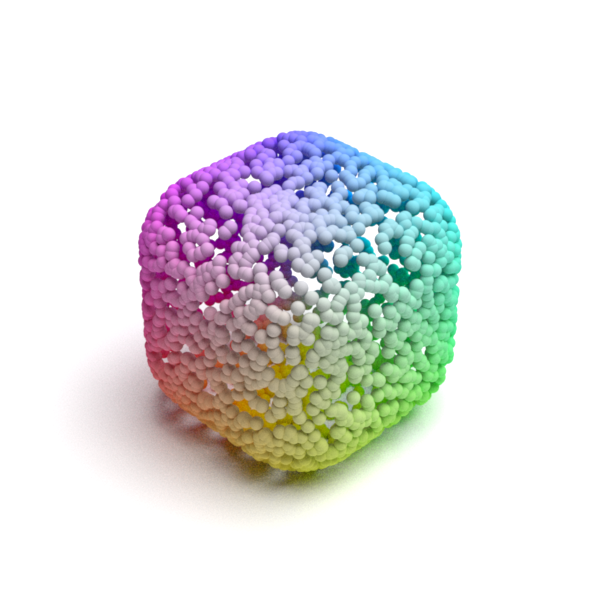} &
		\includegraphics[width=.10\columnwidth,keepaspectratio]{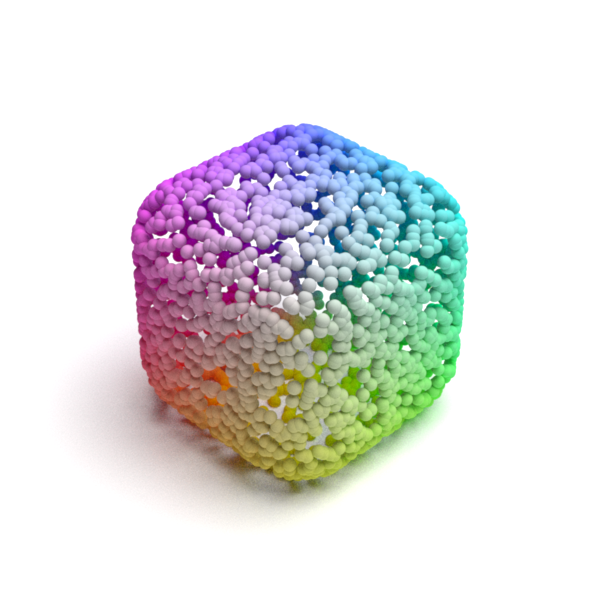} &
		\includegraphics[width=.10\columnwidth,keepaspectratio]{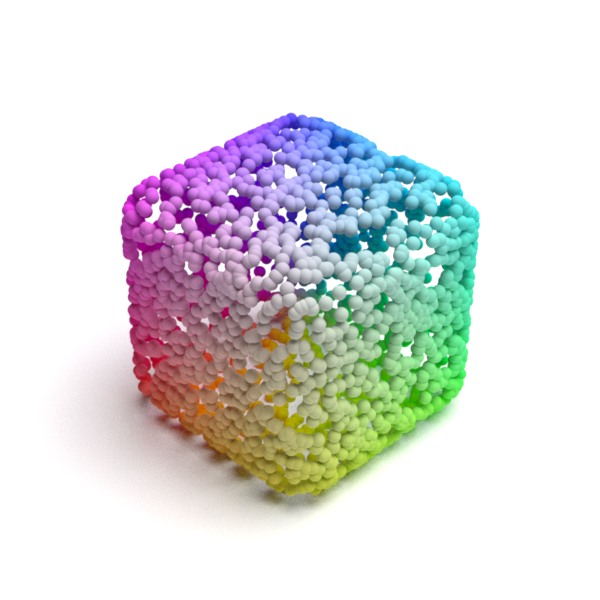} &
		\includegraphics[width=.10\columnwidth,keepaspectratio]{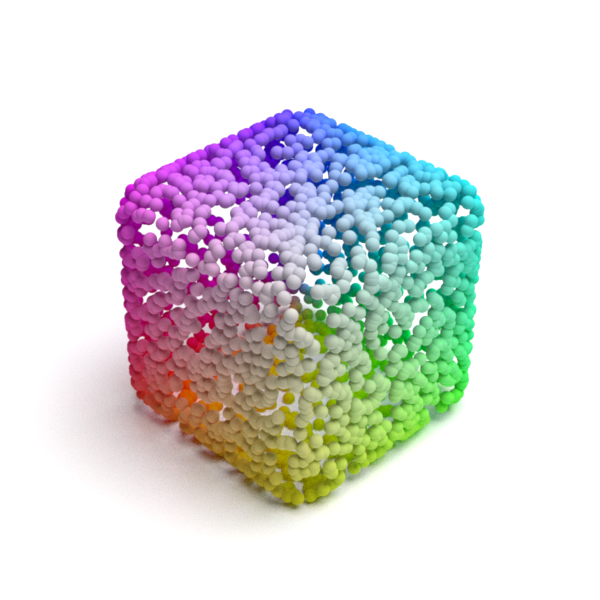} &
		\includegraphics[width=.10\columnwidth,keepaspectratio]{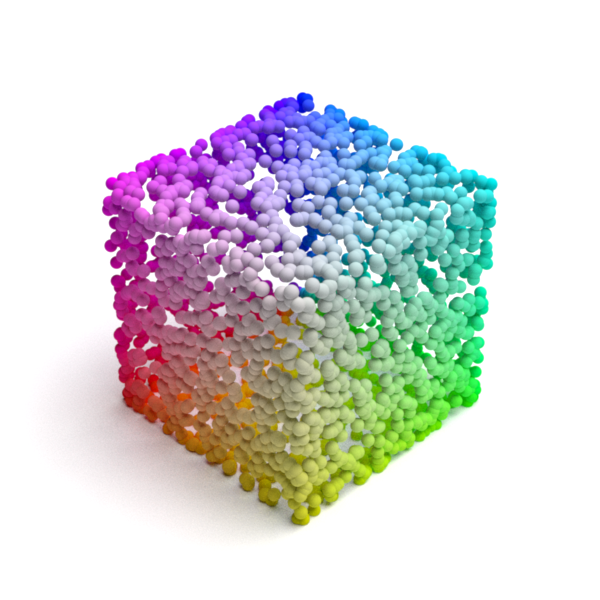} \\  
	\end{tabular}\vspace{-3mm}
	\caption{Image and shape examples of the synthetic dataset built by interpolating between a sphere and a cube. \emph{Upper row} shows rendered images. \emph{Lower row} shows shapes represented by point clouds.}
    \label{fig:cube-sphere-pointcloud} \vspace{-2mm}
\end{figure}

We provide the details of synthetic data generation in this section. The image and shape examples are shown in Figure \ref{fig:cube-sphere-pointcloud}. We use the software Blender to generate base shapes in a mesh format. Then, we use the Shrinkwrap modifier in the ``Nearest Vertex'' mode to define the shape morphing between the two base shapes and control the interpolation progress by the Blender Shape Keys panel. After creating the mesh dataset, we render images and sample point clouds from meshes.

\section{Implementation Details}\label{sec:supp-imple}

In this section, we provide implementation details, including baseline models and training protocol. 

\noindent
\textbf{Baselines}
For NN-based methods, we include PSGN~\cite{fan2017point}, FoldingNet~\cite{yang2018foldingnet}, AtlasNet-Sphere~\cite{groueix2018papier}, AtlasNet-25~\cite{groueix2018papier}. We use a ResNet-18 image encoder without any pre-training, the encoder outputs a 1024 dimensional latent vector. We use the same image encoder for all the models. We implement the decoder of each model according to architectures in the original publications.

\noindent
\textbf{Training Protocol}
Among all the experiments, the loss function is Chamfer distance~\cite{fan2017point}, and optimizer is Adam~\cite{kingma2014adam}.
For experiments on the synthetic dataset, each model is trained for 3600 iterations, using batch size 8. The initial learning rate is 1e-3, and the learning rate decays at 2400, 3000, 3300 iterations by a ratio of 0.1. 
For experiments on ShapeNet, each model is trained for 120 epochs, the batch size is 64, the initial learning rate is 1e-3, and it decays at 90, 110, 115 epoch by ratio 0.1. The weight decay is set to be 0.

\section{Verifying the Dispersion Relationship on SDF-based NNs}\label{sec:sdf-method}

\begin{figure}[!htb]
  \centering
  \begin{subfigure}[b]{0.48\columnwidth}
       \includegraphics[width=\linewidth]{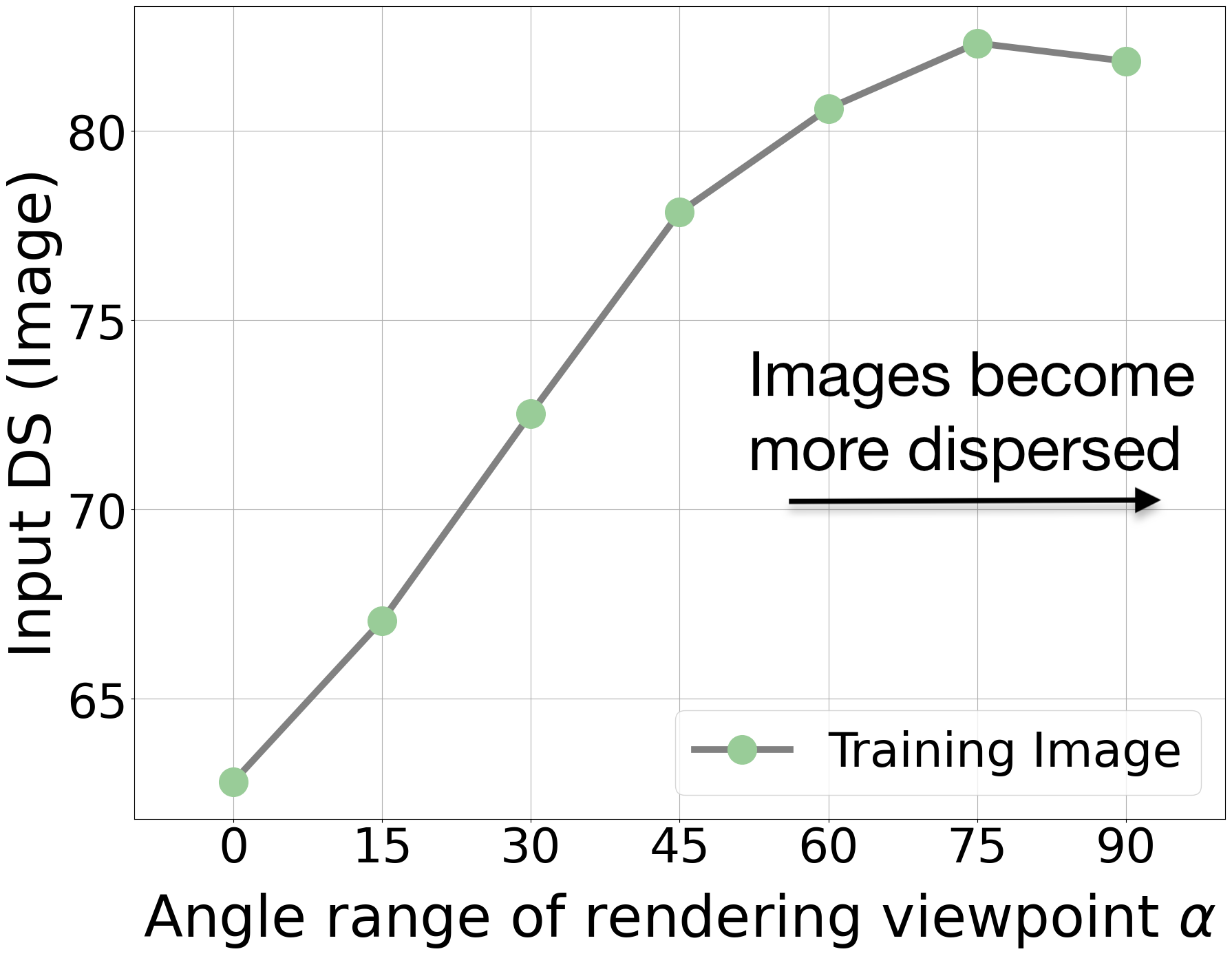}
       \caption{}\centering 
       \label{fig:supp-moreimg-trainimg}
   \end{subfigure}
   \begin{subfigure}[b]{0.48\columnwidth}
        \includegraphics[width=\linewidth]{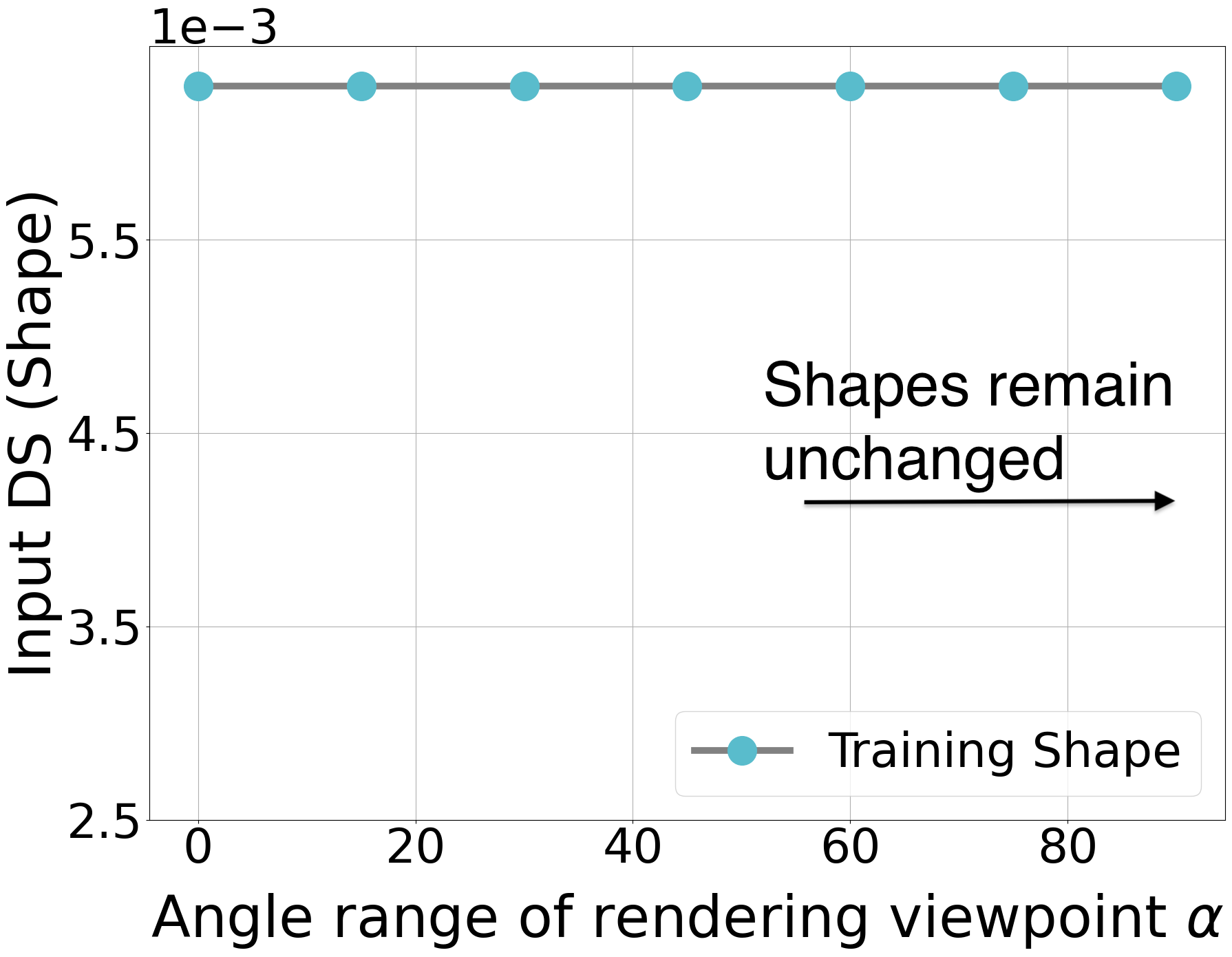}
        \caption{}\centering
        \label{fig:supp-moreimgs-trainshape}
   \end{subfigure} 
  \centering
  \begin{subfigure}[b]{0.48\columnwidth}
       \includegraphics[width=\linewidth]{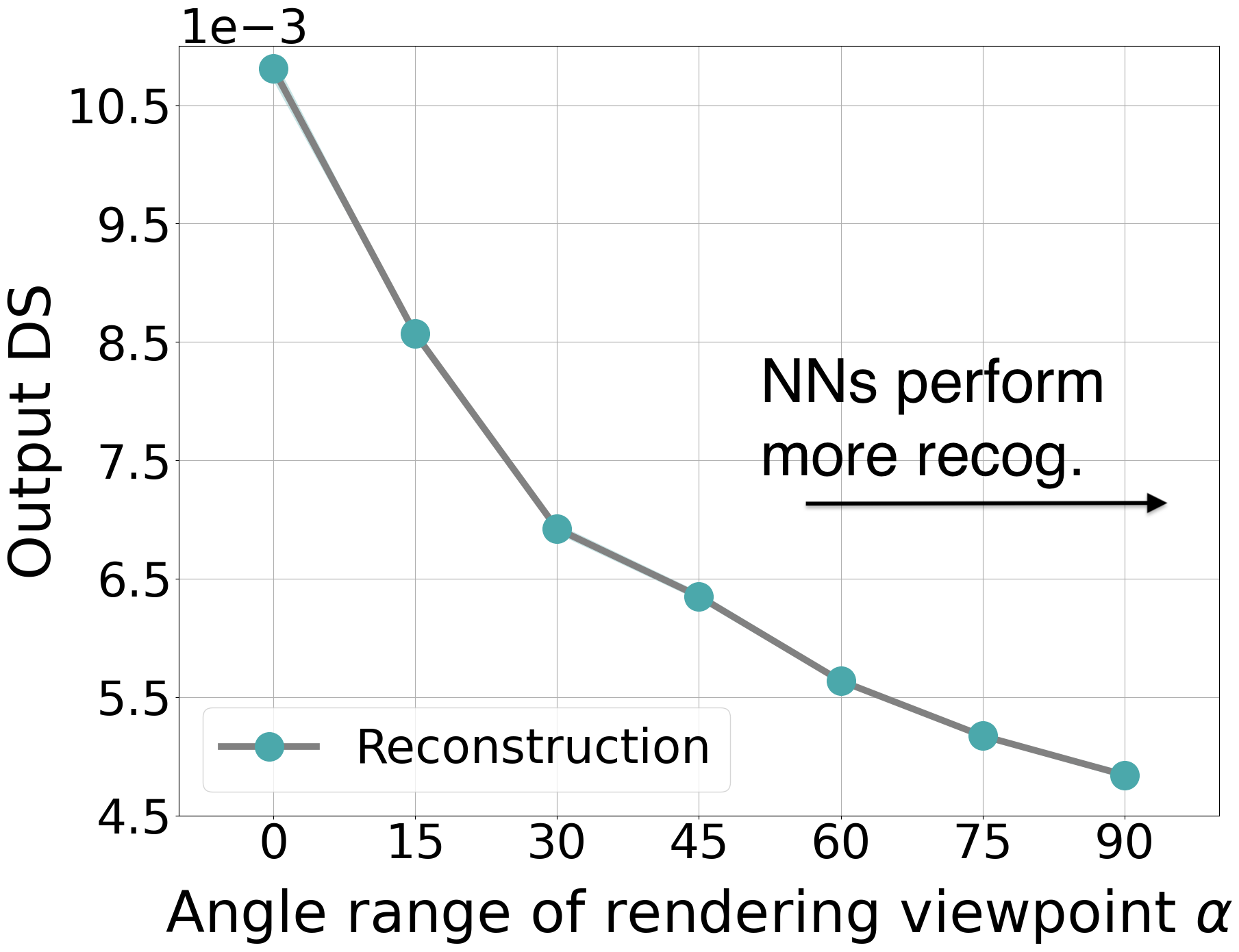}
       \caption{}\centering 
       \label{fig:supp-moreimg-ds}
   \end{subfigure}
   \begin{subfigure}[b]{0.48\columnwidth}
        \includegraphics[width=\linewidth]{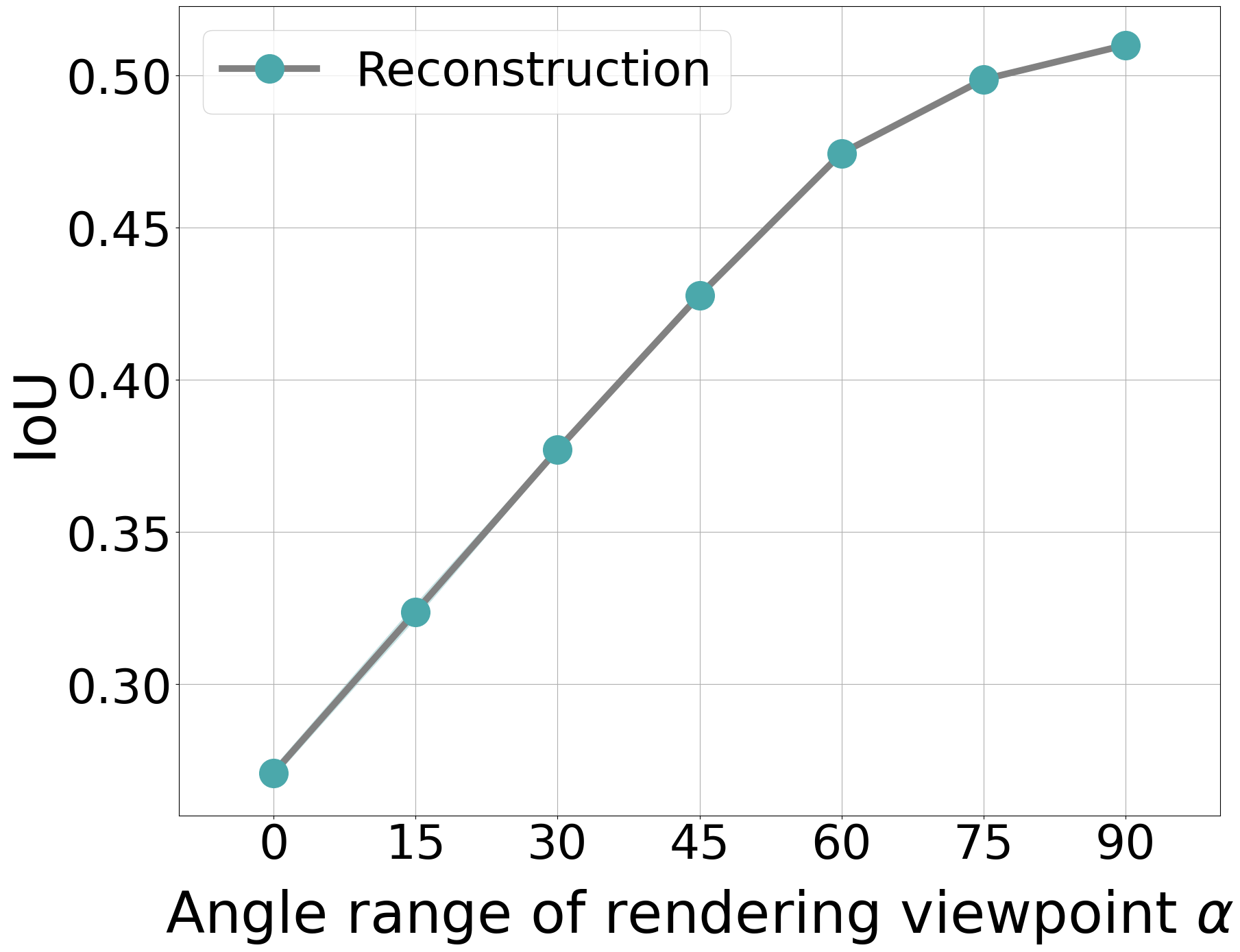}
        \caption{}\centering
        \label{fig:supp-moreimgs-iou}
   \end{subfigure} 

   \caption {Measuring input/output DS and IoU in OC coordinate when the training images are rendered by increasing viewpoint angle range $\alpha$. The unit of $\alpha$ is degree. (a) (b) From left to right, training images become more dispersed and training shapes remain unchanged. (c) Reconstructed shapes become less dispersed. (d) IoU of prediction become larger, the larger the better.}
   \label{fig:supp-moreimgs}
\end{figure}

In the main paper, we have verified our hypothesis of dispersion relationship using point-cloud-based NN methods. 
In this section, we conduct experiments with methods based on signed distance fields (SDF). We adopt Occupancy Network (ONet)~\cite{Mescheder2019OccupancyNL} as the baseline.
For the transition \emph{more dispersed training images}, we reuse the group of increasingly dispersed training image datasets and use the training protocol in Section \ref{sec:shapenet-moreimg}.
While the training images become increasingly dispersed, the DS of training shapes remains unchanged because we use OC coordinate. 
The input DS of training images and shapes are shown in Figure \ref{fig:supp-moreimg-trainimg} and \ref{fig:supp-moreimgs-trainshape}. For evaluation, we use Volumetric IoU~\cite{Mescheder2019OccupancyNL} to measure reconstruction quality. To calculate DS, we extract mesh from predicted SDF following \cite{Mescheder2019OccupancyNL} and uniformly sample 2500 points from each mesh surface. 
The following procedure is the same as Section \ref{sec:metric_on_dataset}, and we omit the details here.

As our main claim predicts, the models tend more towards recognition as the decreasing output DS shows in Figure~\ref{fig:supp-moreimg-ds}.
We also observe that ONet trained on more dispersed images achieves improved Volumetric IoU, as shown in Figure~\ref{fig:supp-moreimgs-iou}.
We notice that the results of SDF-based experiment are consistent with the observations of point-cloud-based experiment shown in Figure~\ref{fig:shape13-imageaug}, which further verifies that more dispersed training images make NNs do more recognition.


\end{document}


\title{Supplementary document for Reconstruction or Recognition: Justifying Single-view 3D Reconstruction Networks}

\author{First Author\\
Institution1\\
Institution1 address\\
{\tt\small firstauthor@i1.org}
\and
Second Author\\
Institution2\\
First line of institution2 address\\
{\tt\small secondauthor@i2.org}
}

\maketitle
\ificcvfinal\thispagestyle{empty}\fi

\section{Emphasizing the Main Contribution}
The main purpose of this work is to study the mechanism behind the neural networks (NN) currently used in single-view 3D reconstruction (SVR). We do not intend to propose new methodologies to improve the reconstruction performance. Previous papers mostly focus on making a single predicted shape closer to the ground truth shape, while we focus on making the set of predicted shapes more dispersed, i.e., making NN tend more towards using reconstruction-based ways for SVR instead of recognition-based ways.

\section{Choice of The Coordinate System}\label{sec:definition-coordinate}

\begin{figure}[!htb]
  \centering
  \begin{subfigure}[b]{0.49\columnwidth}
       \centering
       \includegraphics[width=\linewidth]{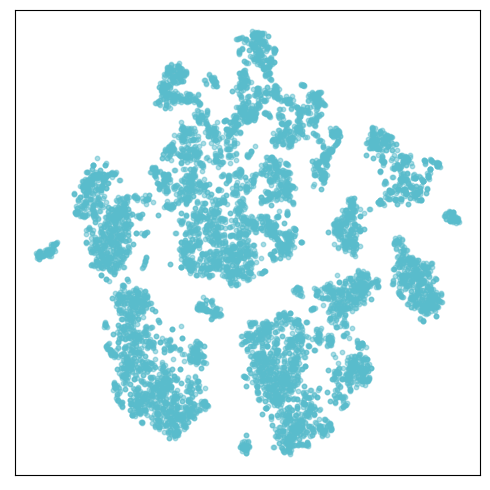}
       \caption{}
       \label{fig:tsne-oc}
   \end{subfigure}
   \begin{subfigure}[b]{0.49\columnwidth}
        \centering
        \includegraphics[width=\linewidth]{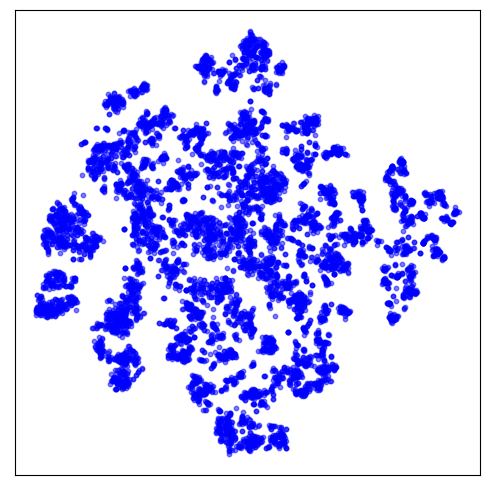}
        \caption{}
        \label{fig:tsne-vc}
   \end{subfigure}
   \begin{subfigure}[b]{0.84\columnwidth}
        \centering
        \includegraphics[width=\linewidth]{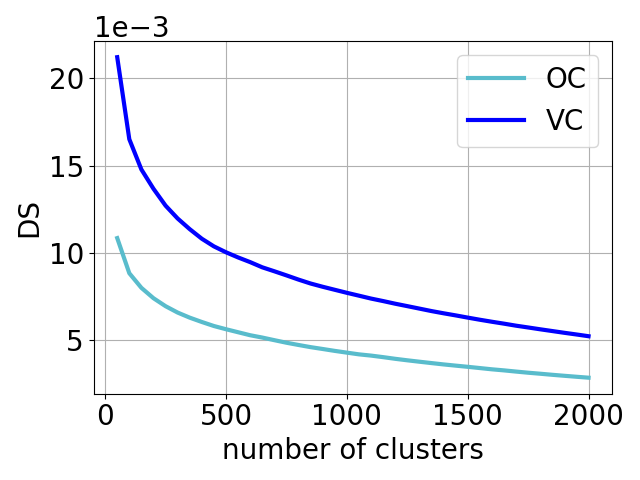}
        \caption{}
        \label{fig:ds-ocvc}
   \end{subfigure} \vspace{-3mm}
   \caption{(a) t-SNE of distance matrix of shapes in OC coordinate. (b) t-SNE of distance matrix of shapes in VC coordinate. (c) The dispersion score (DS) of shapes in OC and VC with varying cluster-number hyperparameter.}
   \label{fig:oc-vc-dispersion-demo}
\end{figure}

First, we provide the definition of object-centered (OC) and viewer-centered (VC) coordinates. Then, we provide quantitative and qualitative results to show the difference between shapes in the two coordinate representations.

\begin{figure*}[!h]
\begin{centering}
	\begin{tabular}{c@{} c@{} c@{}|c@{} c@{} c@{}}
	\toprule
    Image & \begin{tabular}[c]{c@{}}OC\\ \small{(z axis aligned)} \end{tabular} & \begin{tabular}[c]{c@{}}VC\\ \small{(viewpoint aligned)} \end{tabular} & Image & \begin{tabular}[c]{c@{}}OC\\ \small{(z axis aligned)} \end{tabular} & \begin{tabular}[c]{c@{}}VC\\ \small{(viewpoint aligned)} \end{tabular}\\
	\toprule
    \includegraphics[width=0.34\columnwidth,keepaspectratio]{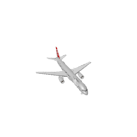} &
    \includegraphics[width=0.32\columnwidth,keepaspectratio]{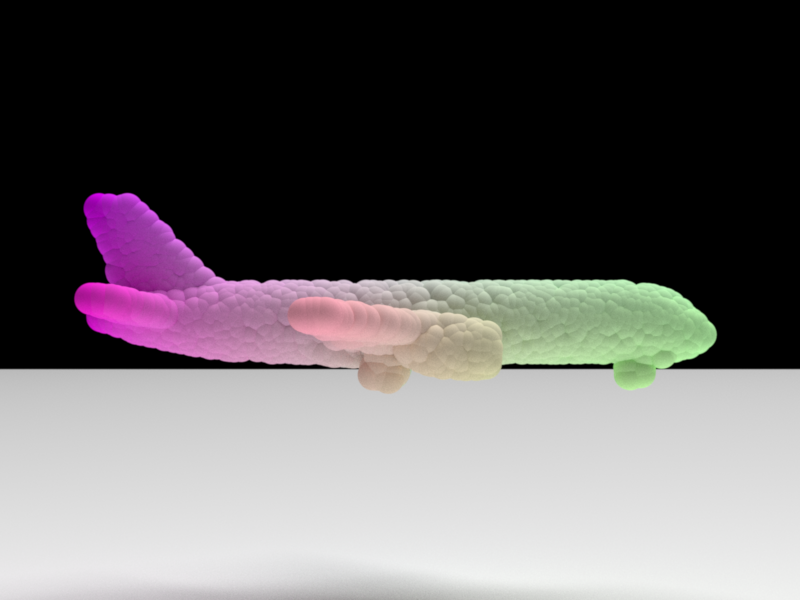} \hspace{1mm} & 
    \includegraphics[width=0.32\columnwidth,keepaspectratio]{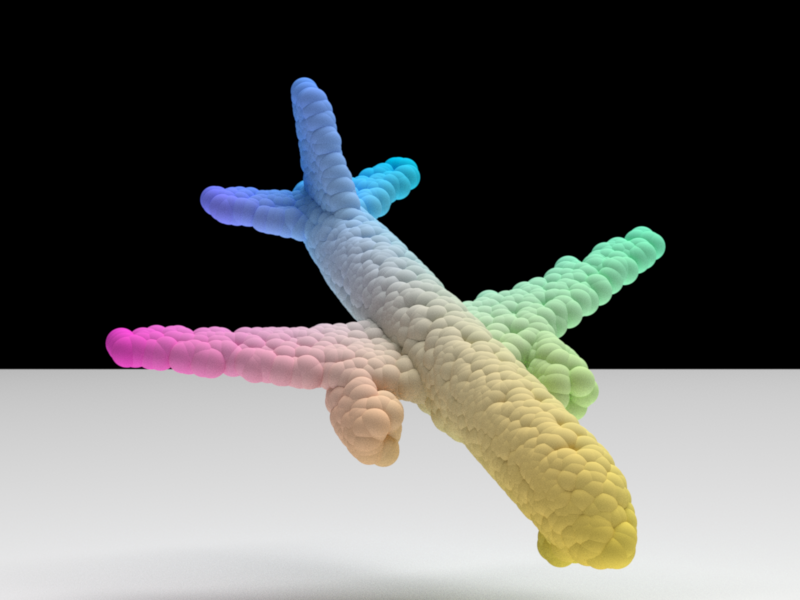} \hspace{1mm} &
    \includegraphics[width=0.34\columnwidth,keepaspectratio]{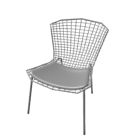} &
    \includegraphics[width=0.32\columnwidth,keepaspectratio]{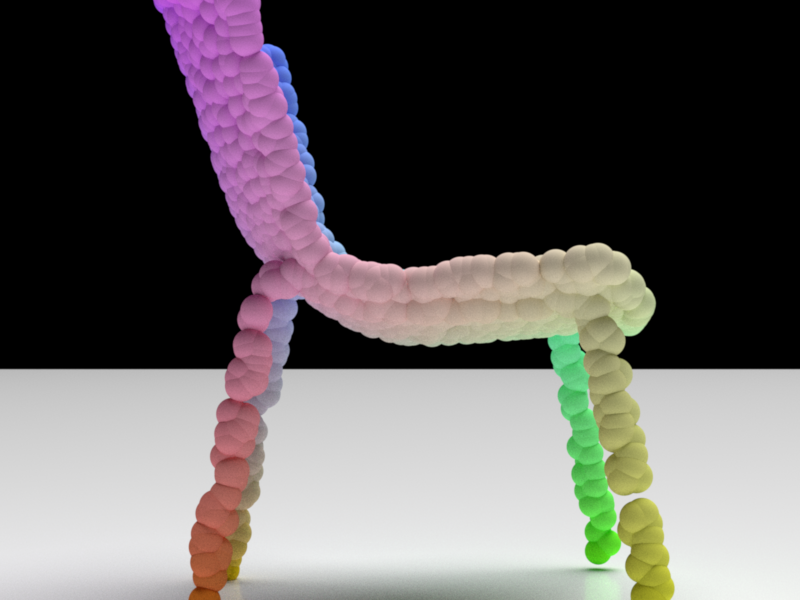} \hspace{1mm} & 
    \includegraphics[width=0.32\columnwidth,keepaspectratio]{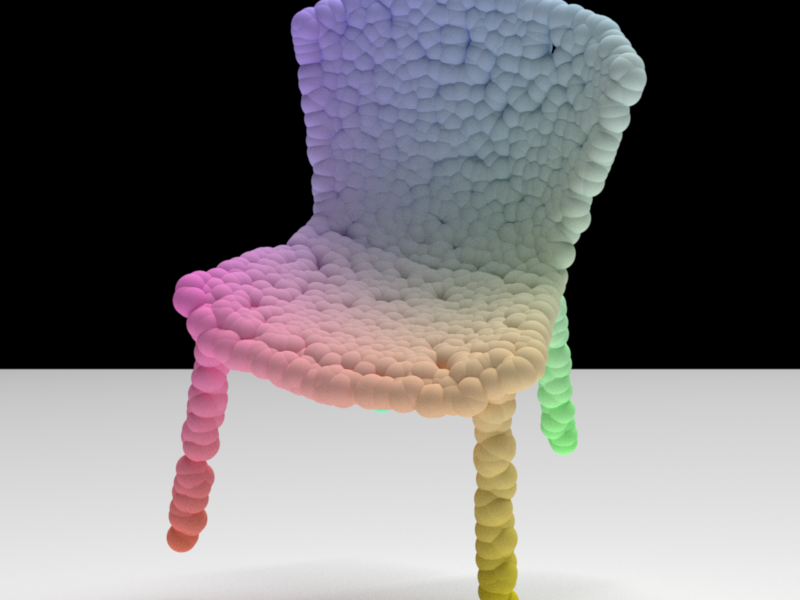} \\
    \includegraphics[width=0.34\columnwidth,keepaspectratio]{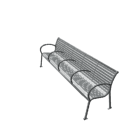} &
    \includegraphics[width=0.32\columnwidth,keepaspectratio]{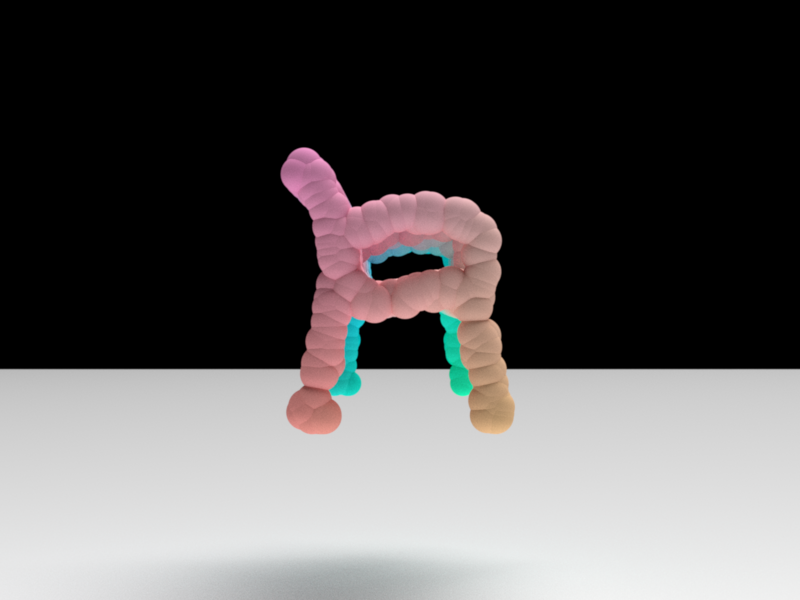} \hspace{1mm} & 
    \includegraphics[width=0.32\columnwidth,keepaspectratio]{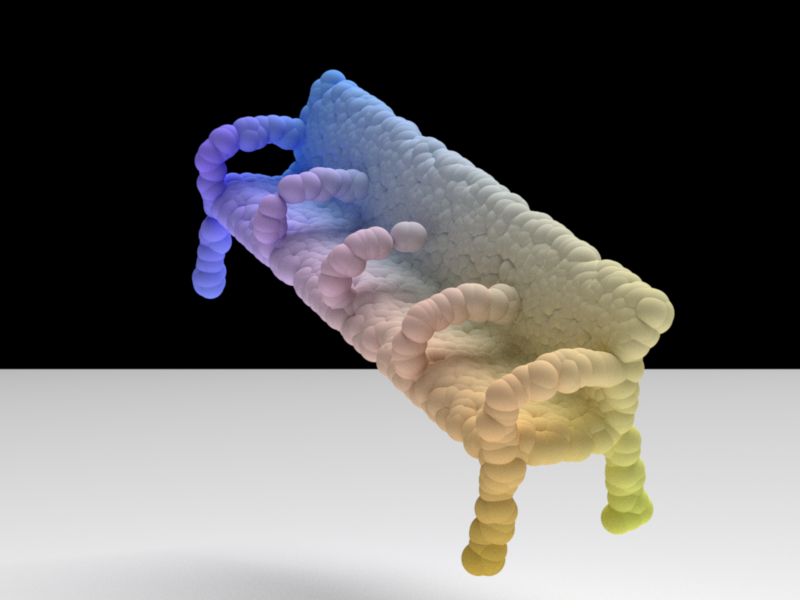} \hspace{1mm} &
    \includegraphics[width=0.34\columnwidth,keepaspectratio]{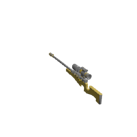} &
    \includegraphics[width=0.32\columnwidth,keepaspectratio]{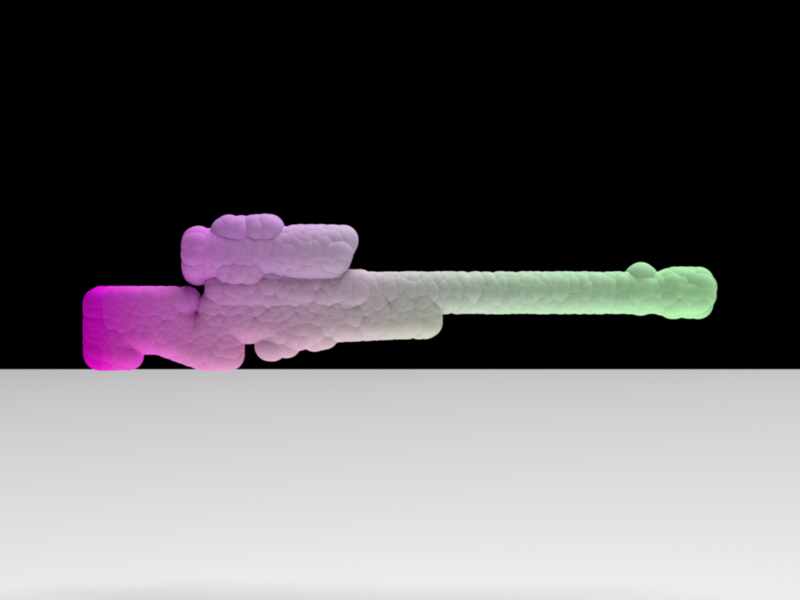} \hspace{1mm} & 
    \includegraphics[width=0.32\columnwidth,keepaspectratio]{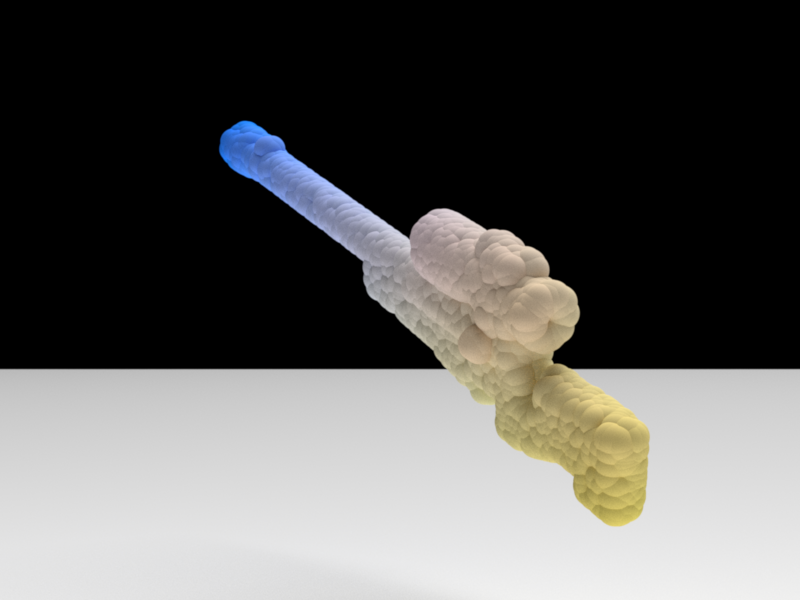}
    \\
    \includegraphics[width=0.34\columnwidth,keepaspectratio]{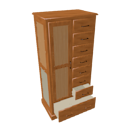} &
    \includegraphics[width=0.32\columnwidth,keepaspectratio]{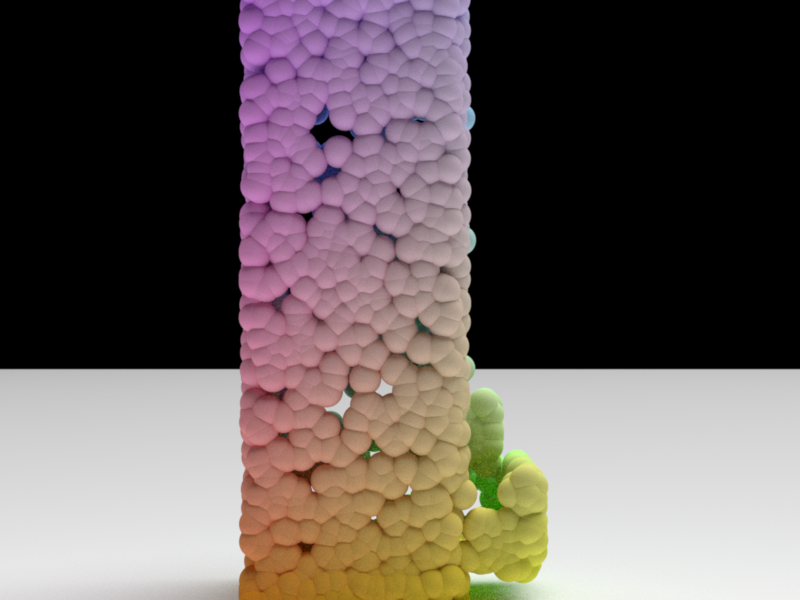} \hspace{1mm} & 
    \includegraphics[width=0.32\columnwidth,keepaspectratio]{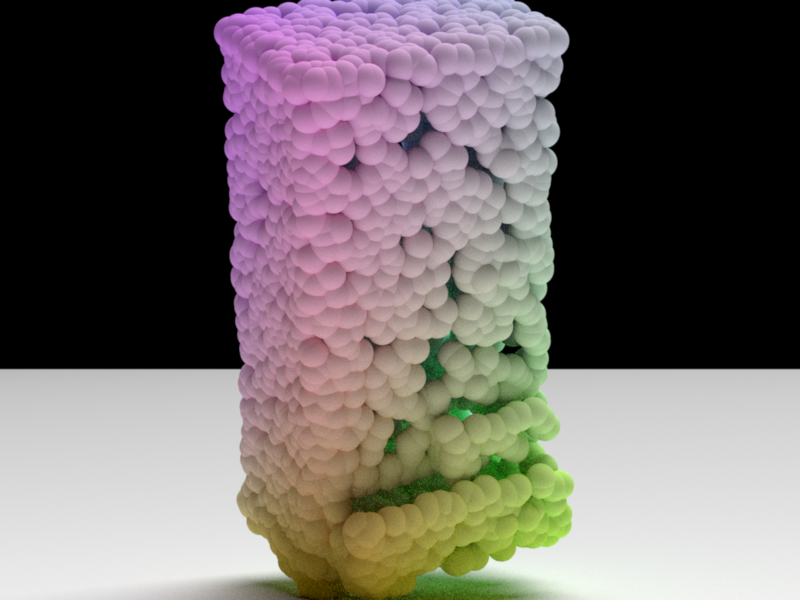} \hspace{1mm} & 
    \includegraphics[width=0.34\columnwidth,keepaspectratio]{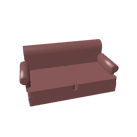} &
    \includegraphics[width=0.32\columnwidth,keepaspectratio]{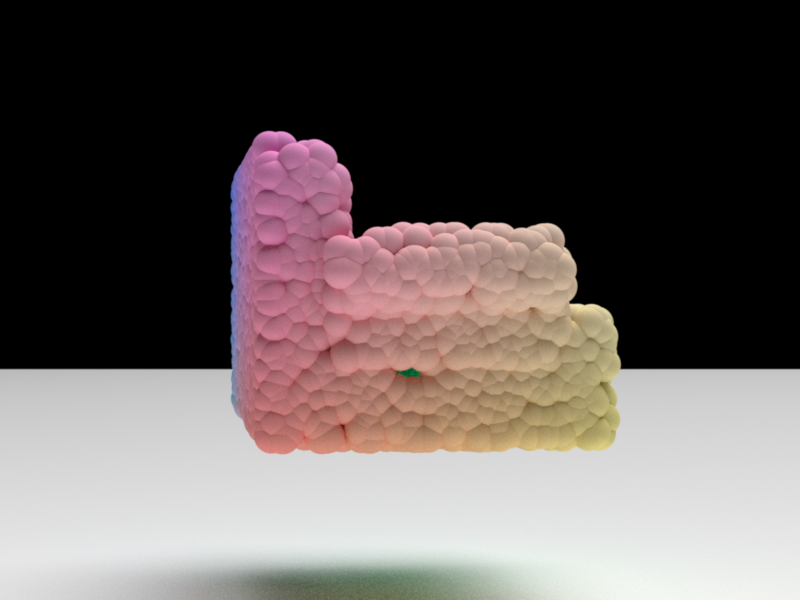} \hspace{1mm} & 
    \includegraphics[width=0.32\columnwidth,keepaspectratio]{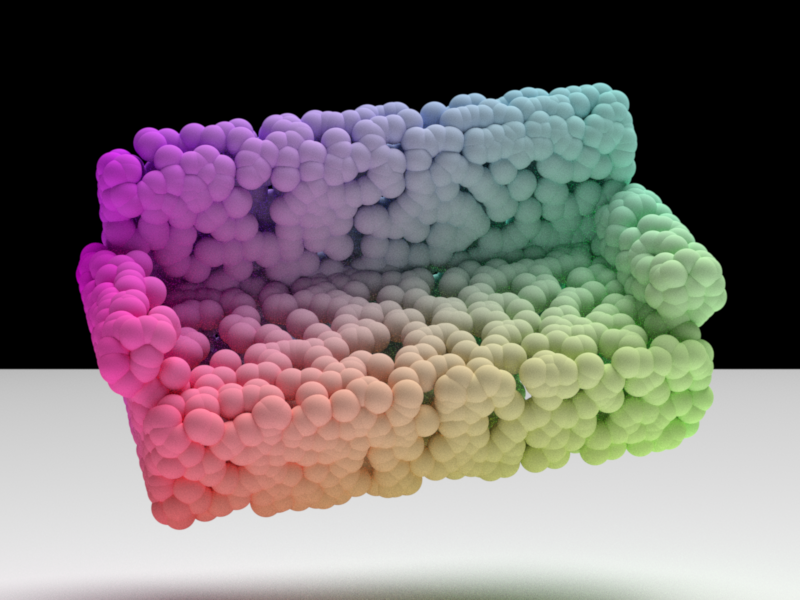}
    \\
    \includegraphics[width=0.34\columnwidth,keepaspectratio]{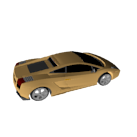} &
    \includegraphics[width=0.32\columnwidth,keepaspectratio]{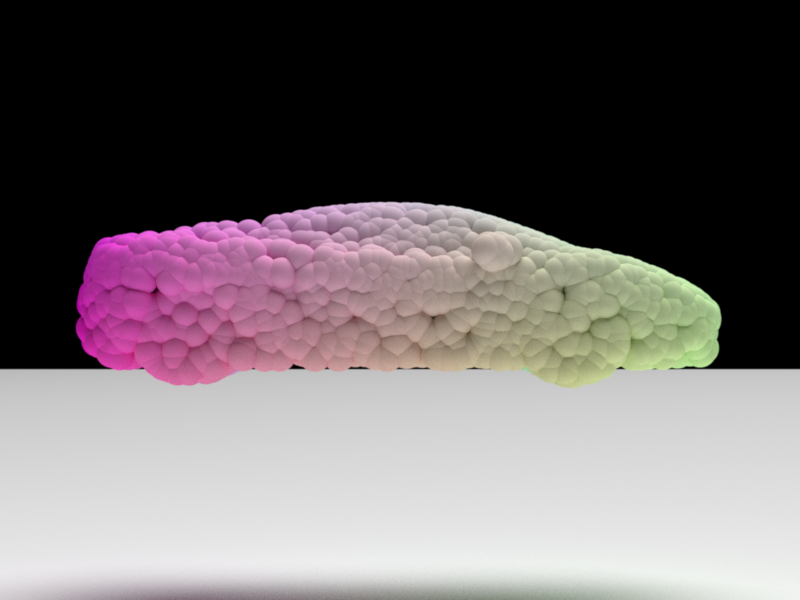} \hspace{1mm} & 
    \includegraphics[width=0.32\columnwidth,keepaspectratio]{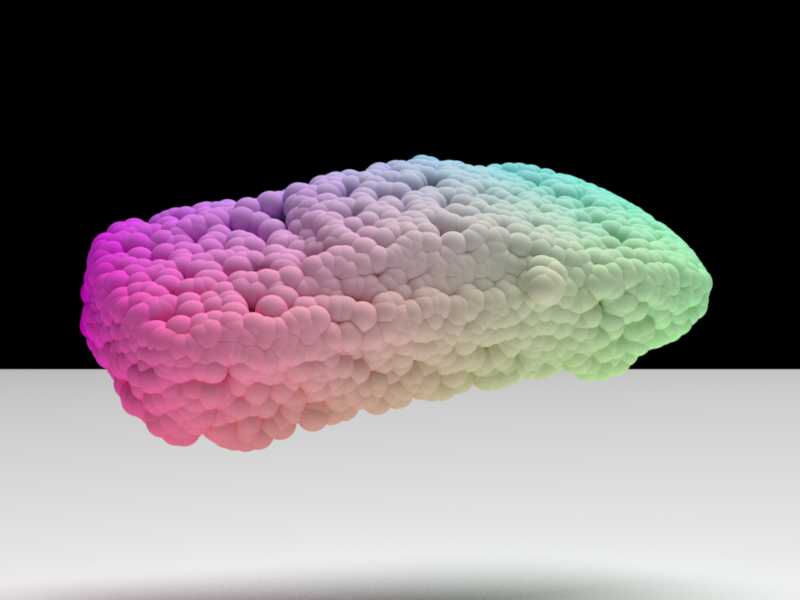}
    \hspace{1mm} &
    \includegraphics[width=0.34\columnwidth,keepaspectratio]{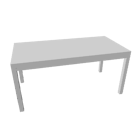} &
    \includegraphics[width=0.32\columnwidth,keepaspectratio]{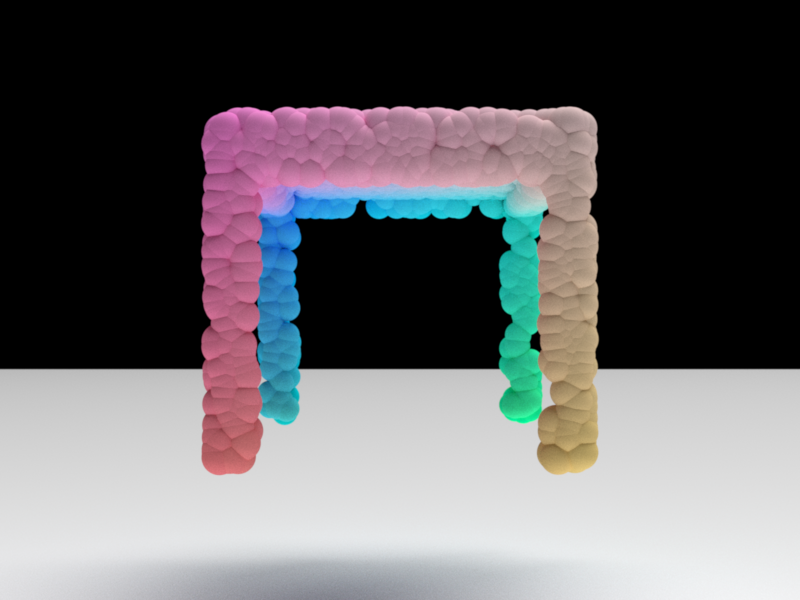} \hspace{1mm} & 
    \includegraphics[width=0.32\columnwidth,keepaspectratio]{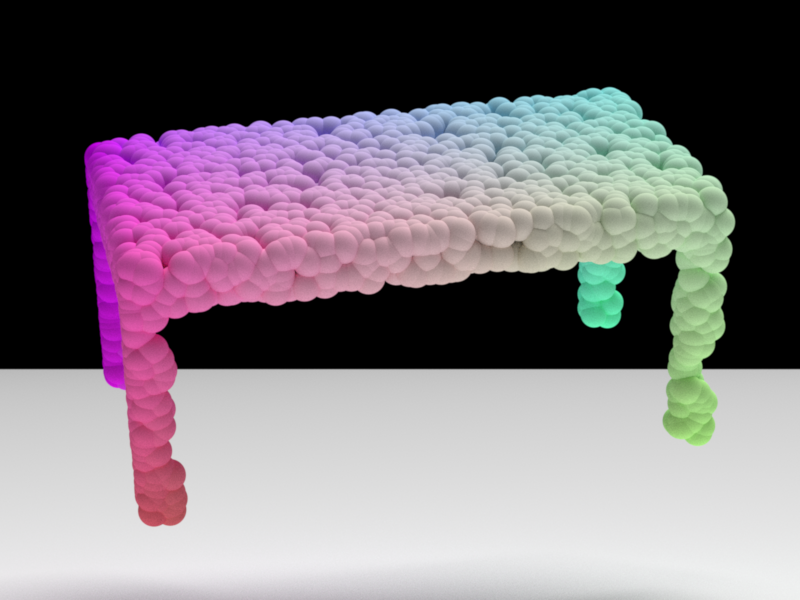}
	\end{tabular}  
	\caption{Visualization of input images and shapes in OC and VC coordinates of ShapeNet~\cite{chang2015shapenet}.}
	\label{fig:oc-vc-vis}
\end{centering}
\end{figure*}

As shown in Figure \ref{fig:oc-vc-vis}, we visualize some shapes in ShapeNet both in OC and VC coordinates. Given a single RGB image as input, we want to predict the 3D shape of the object from which the image is taken. In the OC coordinate, the shapes are predicted in canonical coordinates specified in the training set. For example, in the ShapeNetCore~\cite{chang2015shapenet} dataset, the $\left(\theta_{\mathrm{az}}=0^{\circ}, \theta_{\mathrm{el}}=0^{\circ}\right)$ direction corresponds to the commonly-agreed front of the object, where $\theta_{\mathrm{az}}$ and $\theta_{\mathrm{el}}$ are the azimuth and elevation angle of viewpoint. 
In the VC coordinate, the NN is supervised to predict a pre-aligned 3D shape in the input image’s reference frame. The image-shape pair ensures that $\left(\theta_{\mathrm{az}}=0^{\circ}, \theta_{\mathrm{el}}=0^{\circ}\right)$ in the output coordinate system always corresponds to the input viewpoint. 

We further show the different impacts of the two coordinates on training shapes. The main difference is that shapes in OC are more clustered while shapes in VC are more dispersed. We use all the shapes of the ShapeNetCore~\cite{chang2015shapenet} test set split by \cite{choy20163d}, and represent them both in OC and VC coordinates. The total 8762 shapes covers 13 semantic classes, and each of them is represented as a point cloud with 2500 3D points.
First, we compute distance matrices of shapes using Chamfer Distance as the distance function. 
Then, we visualize the matrices by t-SNE~\cite{10.5555/2968618.2968725} in Figure \ref{fig:tsne-oc} and \ref{fig:tsne-vc}. 
We see that Figure \ref{fig:tsne-oc} shows a clear clustering pattern while Figure \ref{fig:tsne-vc} shows a more dispersed pattern. 
It indicates that shapes in OC are more clustered than that in VC. Besides, we also measure the dispersion score (DS) of shapes. We sweep the cluster-number hyperparameter from 50 to 2000 with step size 50. The results are shown in Figure \ref{fig:ds-ocvc}. The DS of shapes in VC is clearly larger, indicating that the VC coordinate makes shapes more dispersed.

\begin{figure*}[!htb] 
  \captionsetup[subfigure]{labelformat=empty}
  \begin{centering}
  \begin{subfigure}[b]{0.124\linewidth}
       \includegraphics[width=\linewidth]{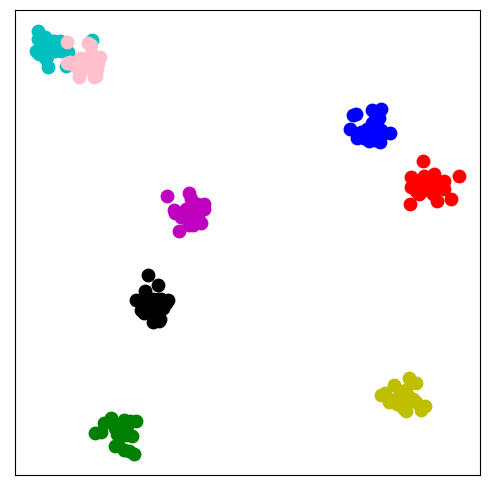}
       \caption{\small cluster std=1}
   \end{subfigure}\hspace*{-0.4em}
  \begin{subfigure}[b]{0.124\linewidth}
       \includegraphics[width=\linewidth]{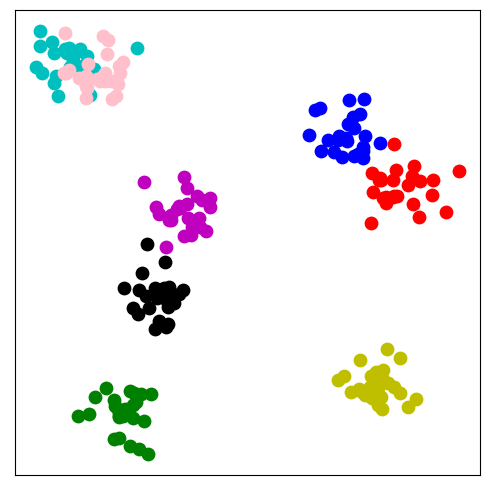}
       \caption{\small cluster std=2}
   \end{subfigure}\hspace*{-0.4em}
  \begin{subfigure}[b]{0.124\linewidth}
       \includegraphics[width=\linewidth]{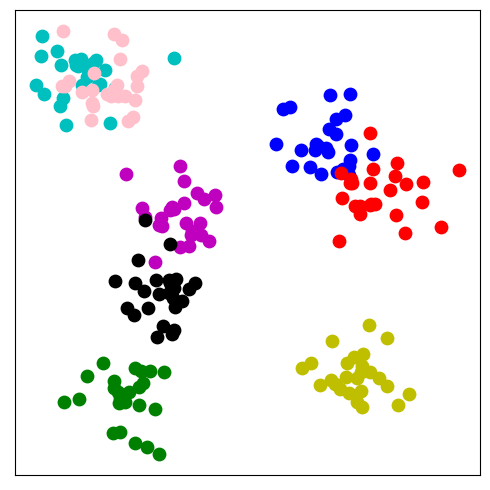}
       \caption{\small cluster std=3}
   \end{subfigure}\hspace*{-0.4em}
  \begin{subfigure}[b]{0.124\linewidth}
       \includegraphics[width=\linewidth]{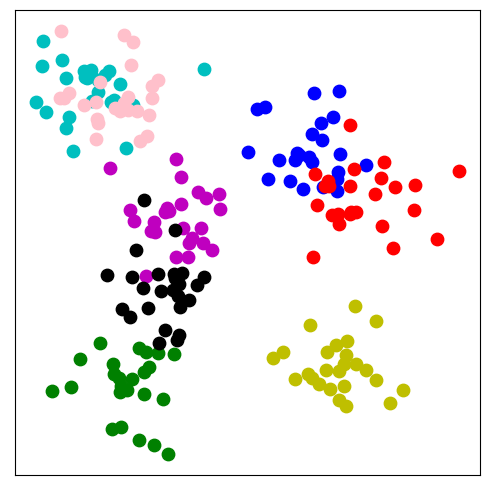}
       \caption{\small cluster std=4}
   \end{subfigure}\hspace*{-0.4em}
  \begin{subfigure}[b]{0.124\linewidth}
       \includegraphics[width=\linewidth]{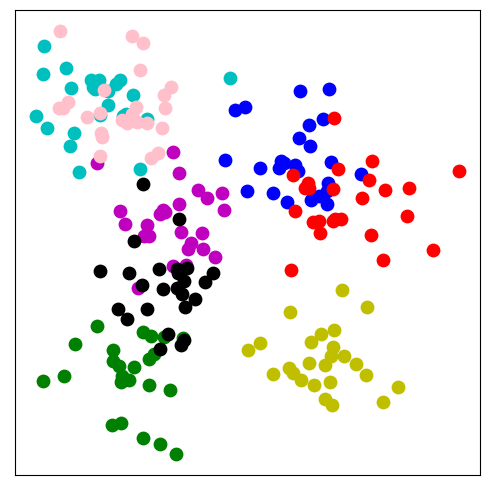}
       \caption{\small cluster std=5}
   \end{subfigure}\hspace*{-0.4em}
  \begin{subfigure}[b]{0.124\linewidth}
       \includegraphics[width=\linewidth]{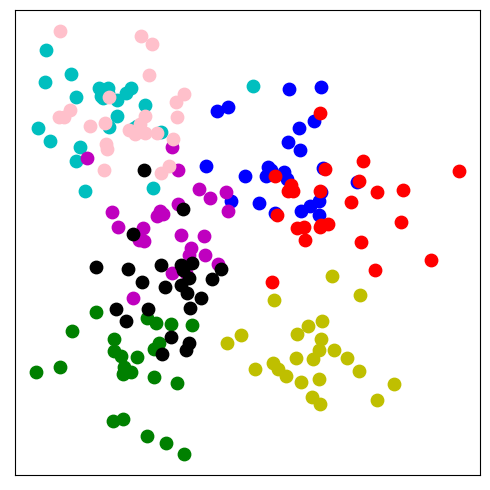}
       \caption{\small cluster std=6}
   \end{subfigure}\hspace*{-0.4em}
  \begin{subfigure}[b]{0.124\linewidth}
       \includegraphics[width=\linewidth]{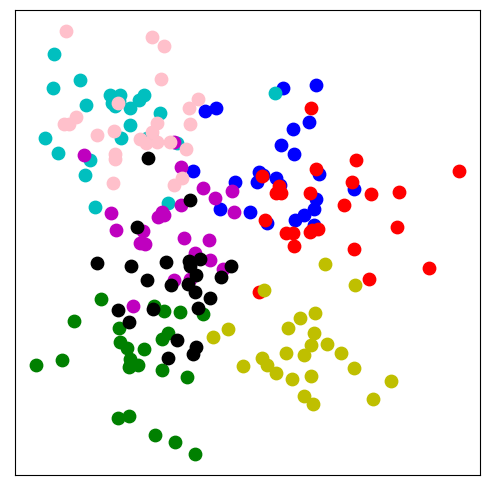}
       \caption{\small cluster std=7}
   \end{subfigure}\hspace*{-0.4em}
  \begin{subfigure}[b]{0.124\linewidth}
       \includegraphics[width=\linewidth]{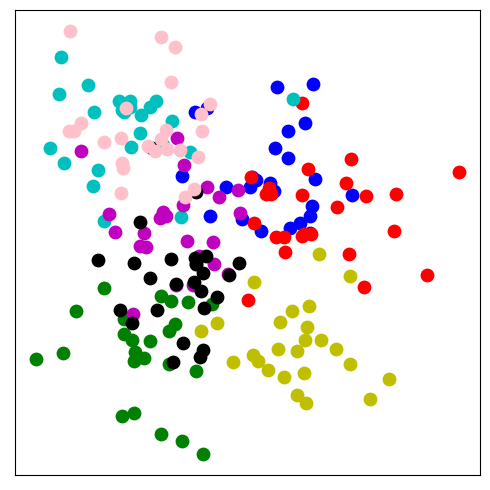}
       \caption{\small cluster std=8}
   \end{subfigure}
   \end{centering}
   \caption{Visualization of toy datasets created for demonstrating dispersion score. Data points with the same color mean the data belong to a common cluster. From left to right, datasets become more dispersed as the standard deviation (std) of all clusters increases by 1.}
   \label{fig:2d-toy-datasets}
\end{figure*}

\section{Hyperparameter of Dispersion Score}\label{sec:metrics}
\vspace{-3mm}
\begin{figure}[!htb]
  \centering
  \includegraphics[width=.95\columnwidth]{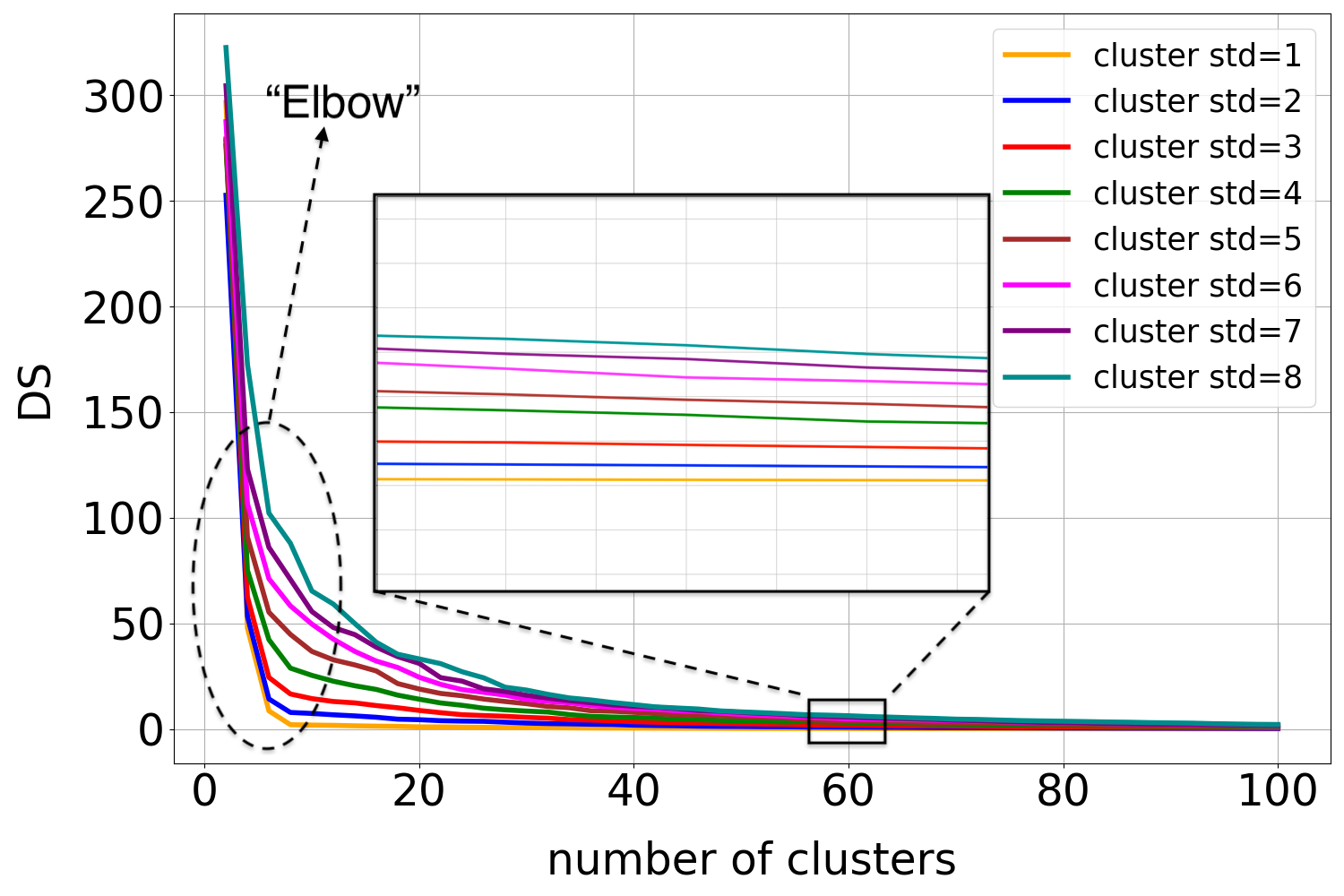} \vspace{-3mm}
  \caption{The metric DS with varying hyperparameter cluster-number in the evaluation of 2D points datasets described in Figure \ref{fig:2d-toy-datasets}. The curves in the black rectangle box are a zoomed-in version of origin curves.}
  \label{fig:num-clusters-search}
\end{figure}

In this section, we explain our method to tune the cluster-number in the proposed metric dispersion score (DS) metric. 
We will call the hyperparameter HP.
To choose the best hyperparameter, we do a parameter sweeping, and choose the value of cluster-number equal to or greater than the ``elbow'' of DS curves. 
``Elbow'' here denotes the value of cluster-number where adding another cluster does not significantly decrease the DS. For instance, see Figure \ref{fig:num-clusters-search}.

Now, we further conduct experiments to explain how to select the HP. We also use these new experiments to demonstrate the effectiveness of the DS metric. 
In Figure \ref{fig:2d-toy-datasets}, we show a group of 2D datasets. For each dataset, we generate 200 2D points $(x, y)$ that forms 8 clusters. We change the extent of how dispersed the data is by increasing the standard deviation (std) of the points in each cluster. 
The eight datasets become increasingly dispersed. Then, we evaluate the DS of eight toy datasets with HP ranging from 2 to 100 with step 2. The distance function is the squared euclidean distance. In Figure \ref{fig:num-clusters-search}, we show the ``elbow'' on the curves. We also observe that, when the cluster-number increases to larger than the value of ``elbow'', the rate at which the DS curves drop decreases, and the DS becomes insensitive to the number of clusters. 
Therefore, we suggest selecting the value of HP equal to or larger than the ``elbow'' of the sweeping curve. Besides, given the observation of Figure \ref{fig:2d-toy-datasets} that datasets with larger cluster std are more dispersed, Figure \ref{fig:num-clusters-search} shows that the DS of the more dispersed datasets are larger. Thus, we further demonstrate that metric DS is useful to measure the extent to which the data is dispersed.

\begin{figure*}[ht!]
\begin{centering}
	\begin{tabular}{c@{}c@{}|c@{}c@{}}
	\toprule
	\multicolumn{2}{c|}{Input DS} & \multicolumn{2}{c}{Output DS} \\
	\toprule
	HP = 2 & HP Sweeping & HP = 2 & HP Sweeping
	\\
	\toprule
    \includegraphics[width=0.23\textwidth]{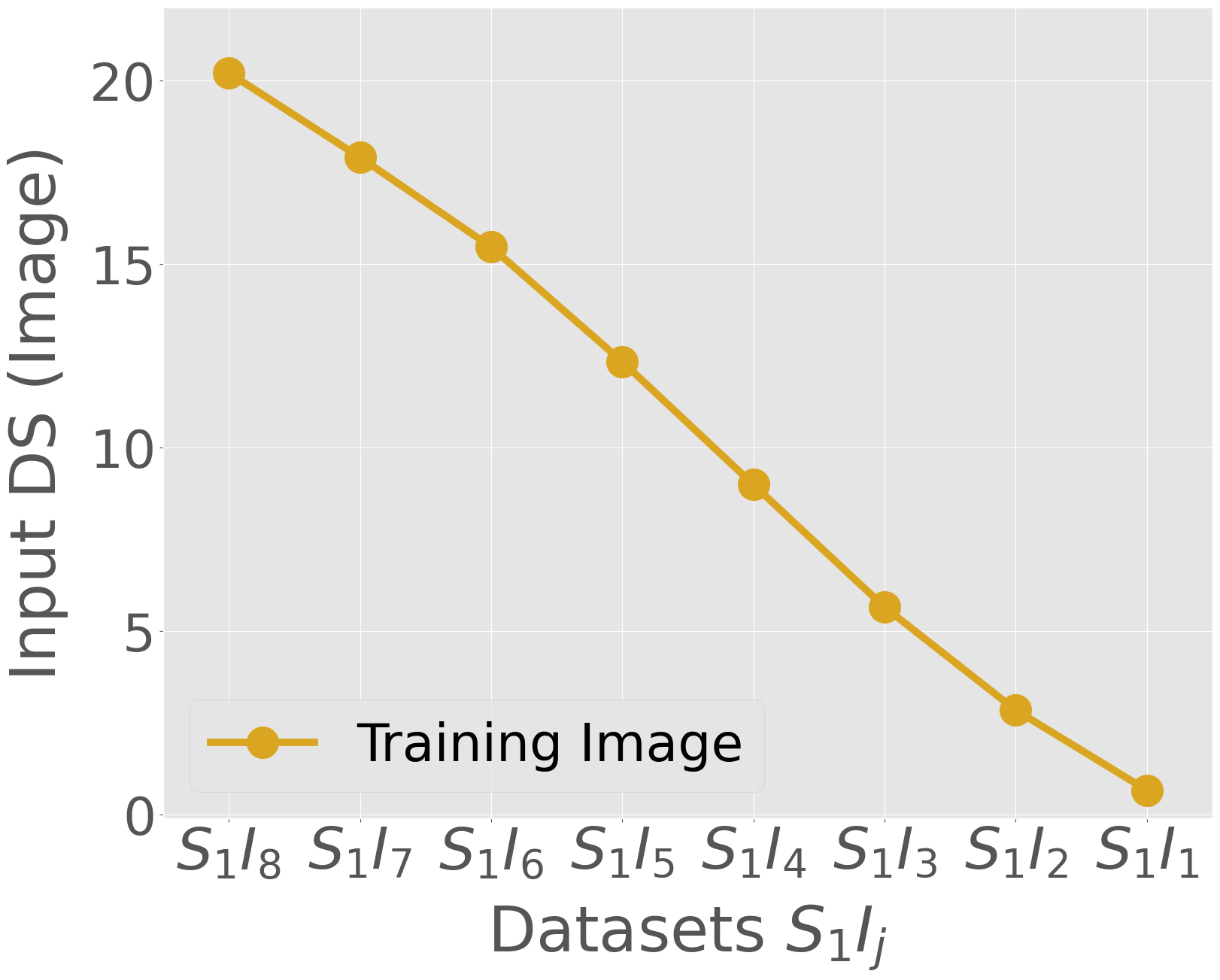} &
    \includegraphics[width=0.24\textwidth]{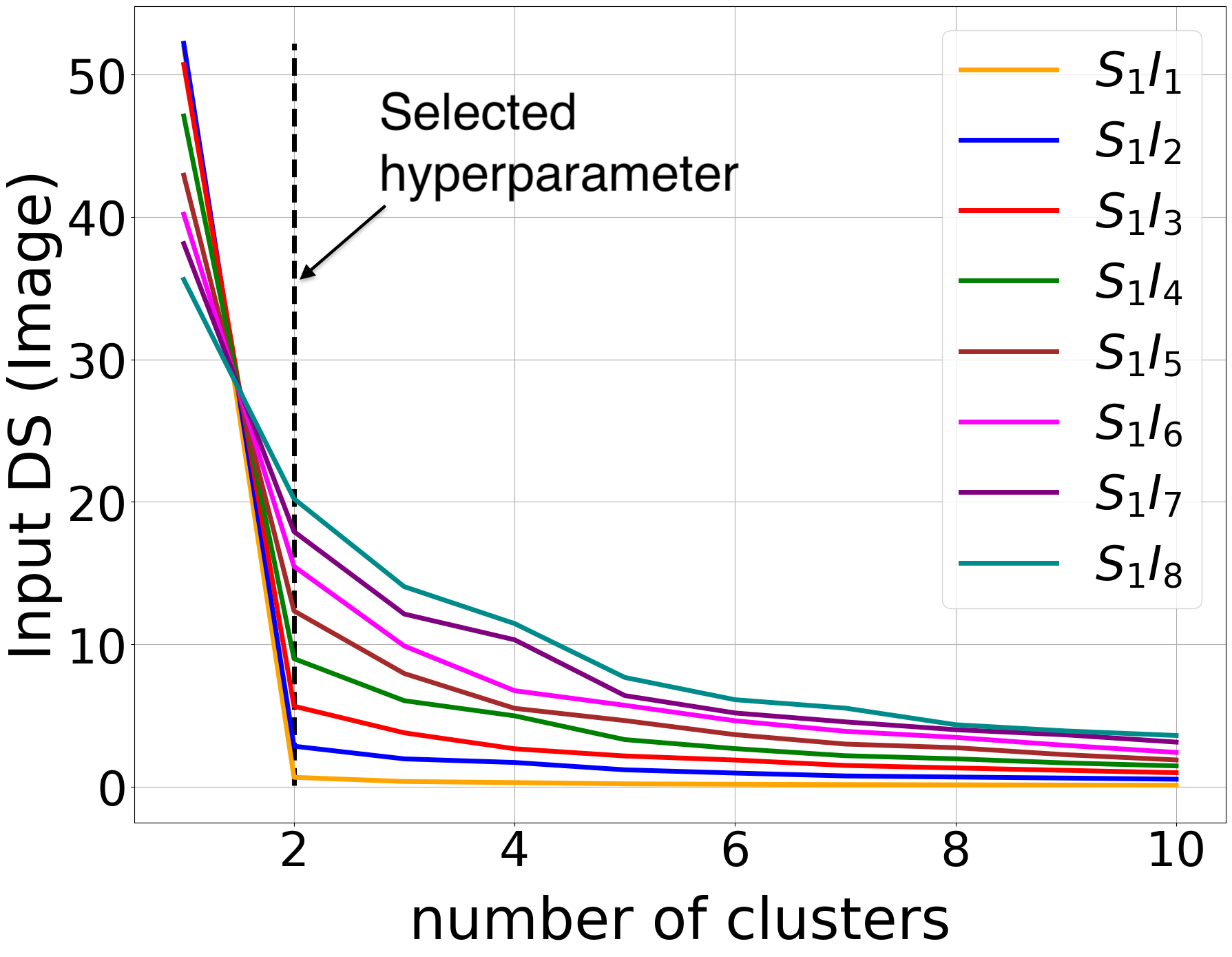} \hspace{2mm}&
    \includegraphics[width=0.24\textwidth]{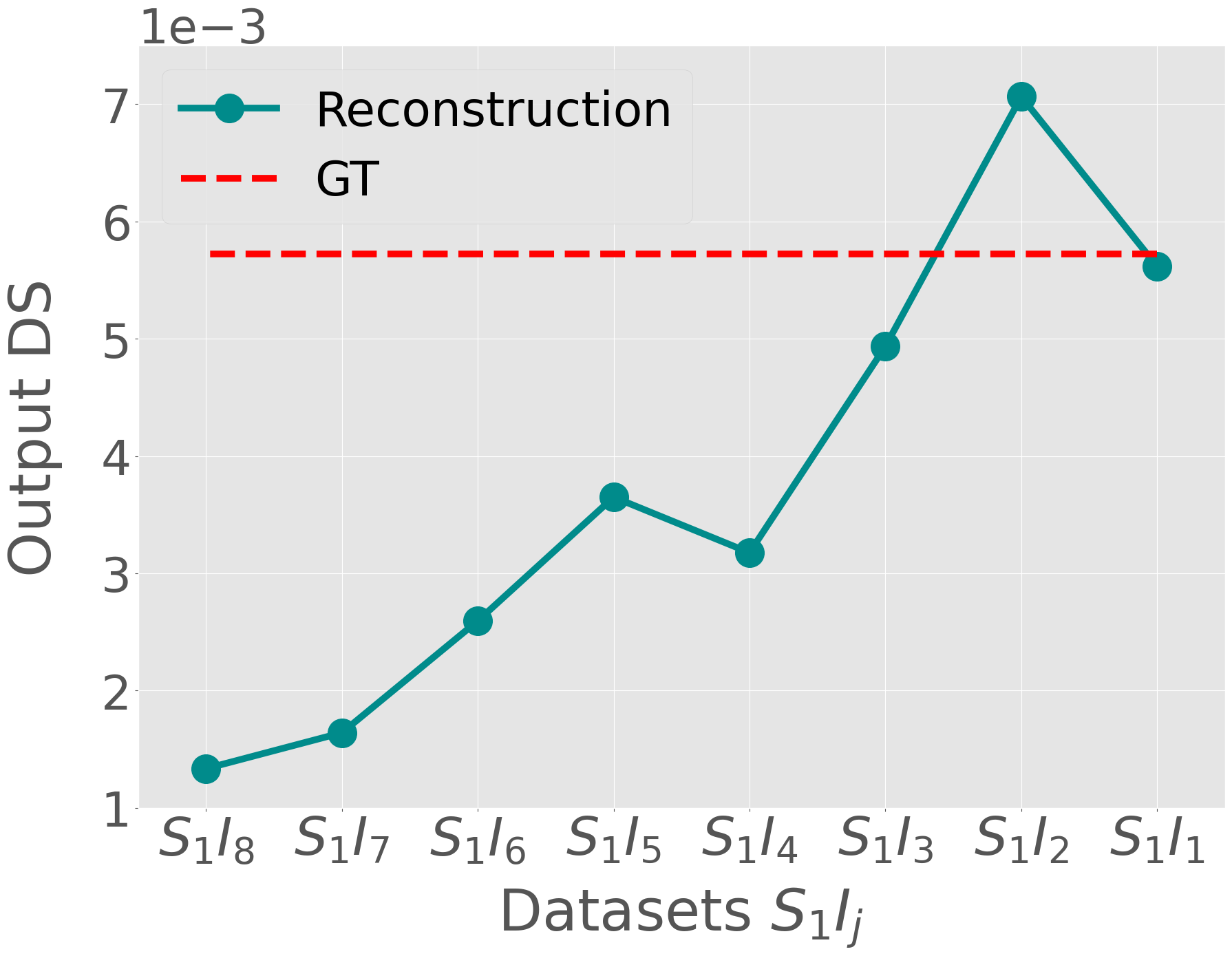} &
    \includegraphics[width=0.24\textwidth]{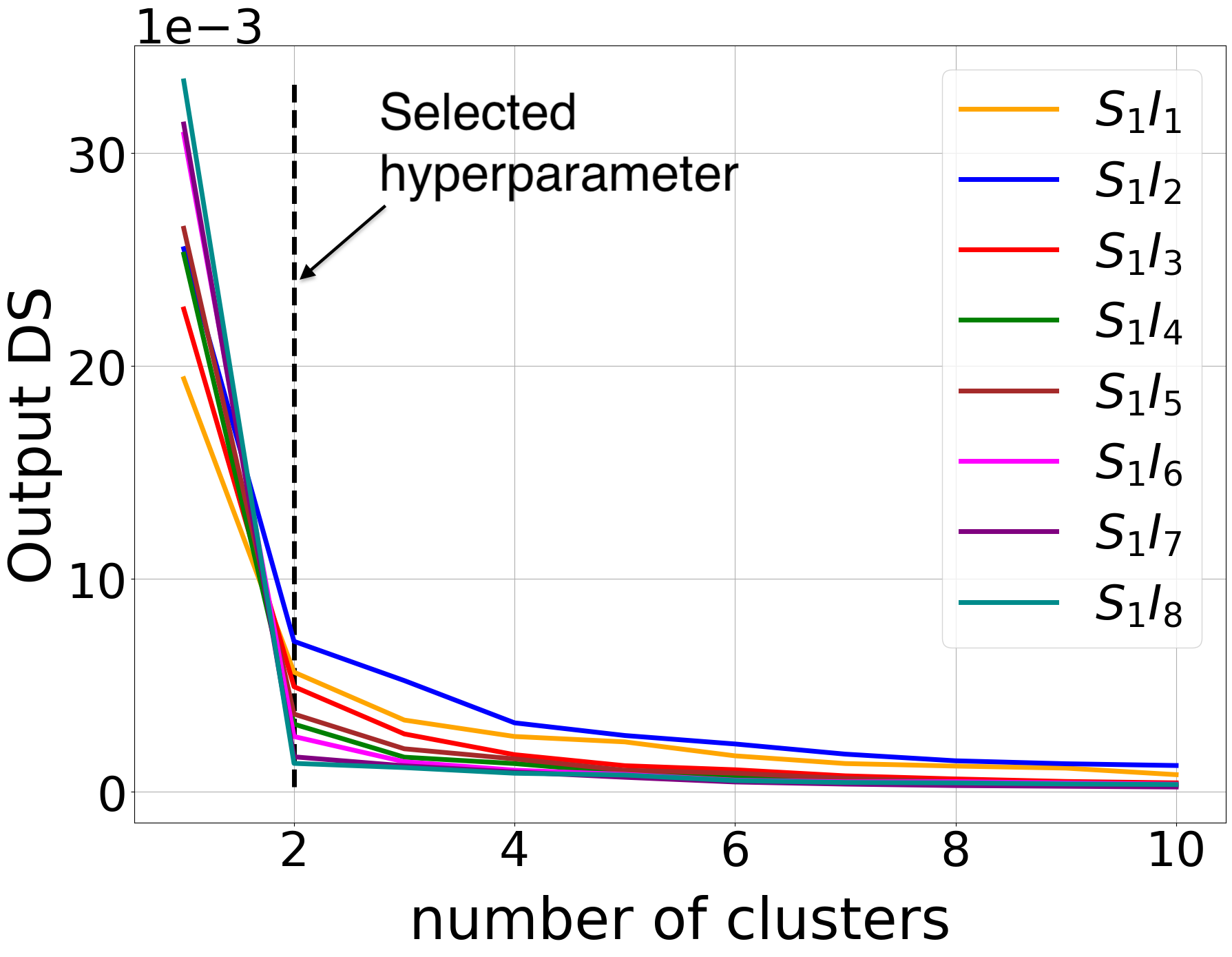} \\
    \includegraphics[width=0.23\textwidth]{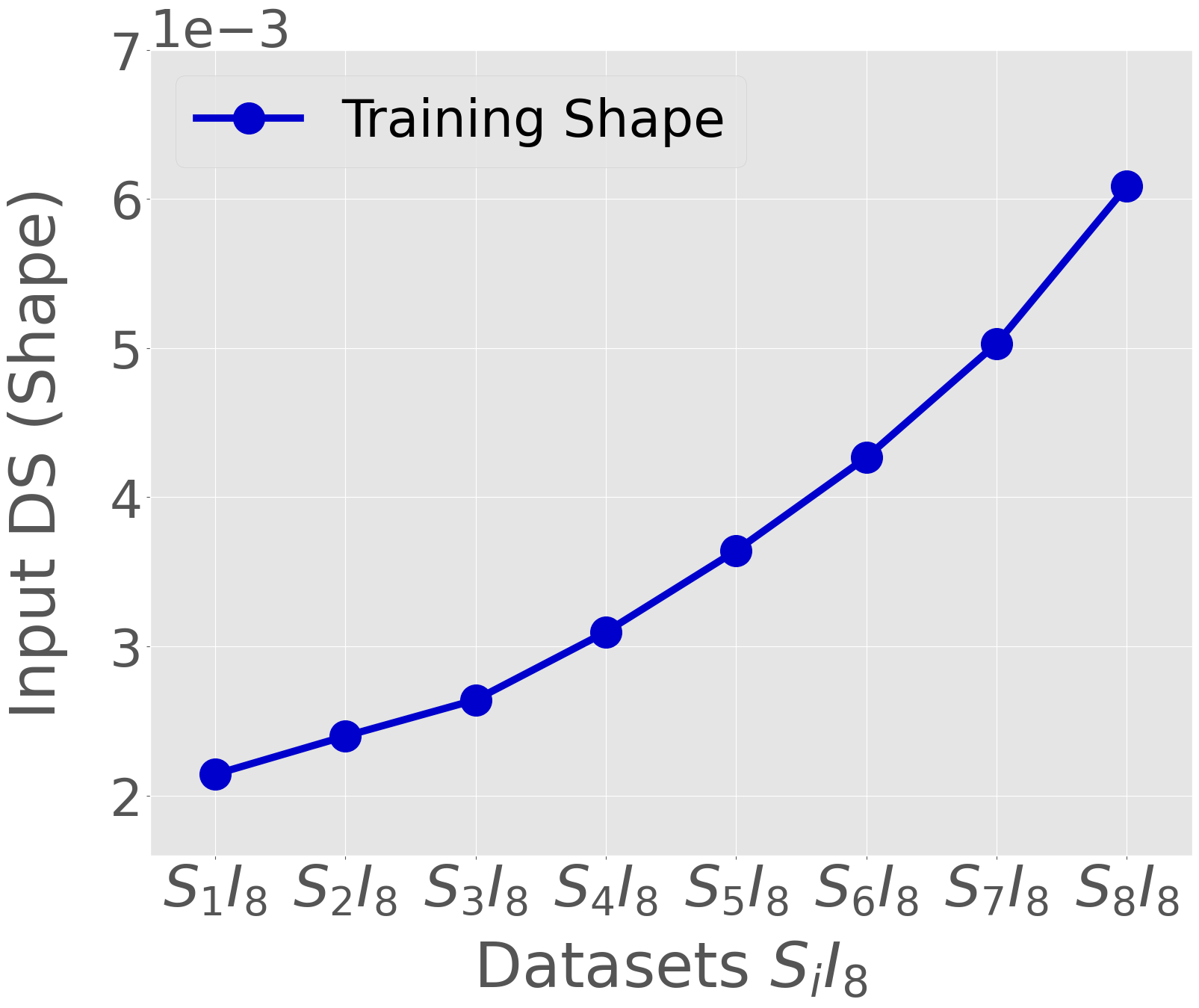} &
    \includegraphics[width=0.24\textwidth]{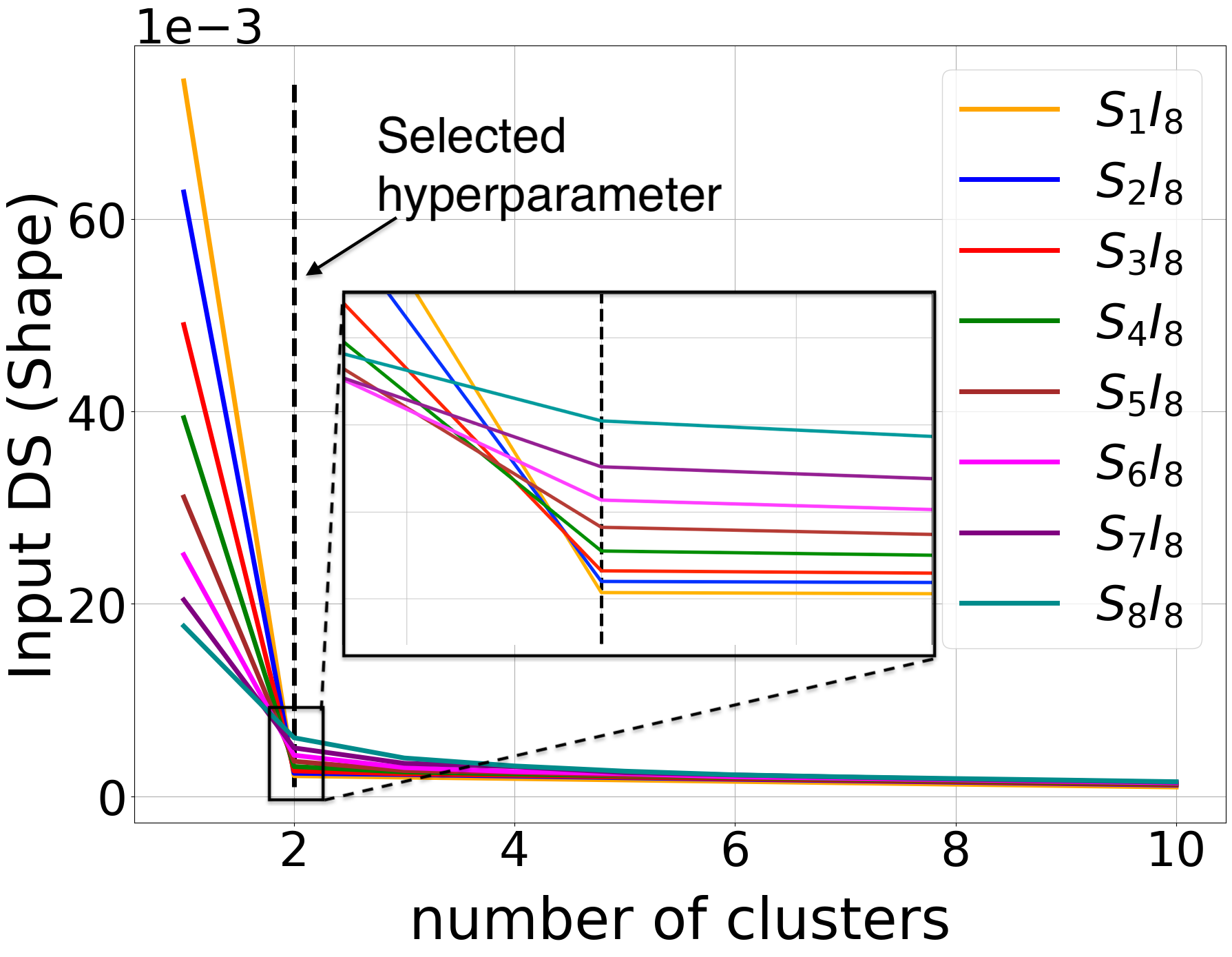} \hspace{2mm}&
    \includegraphics[width=0.24\textwidth]{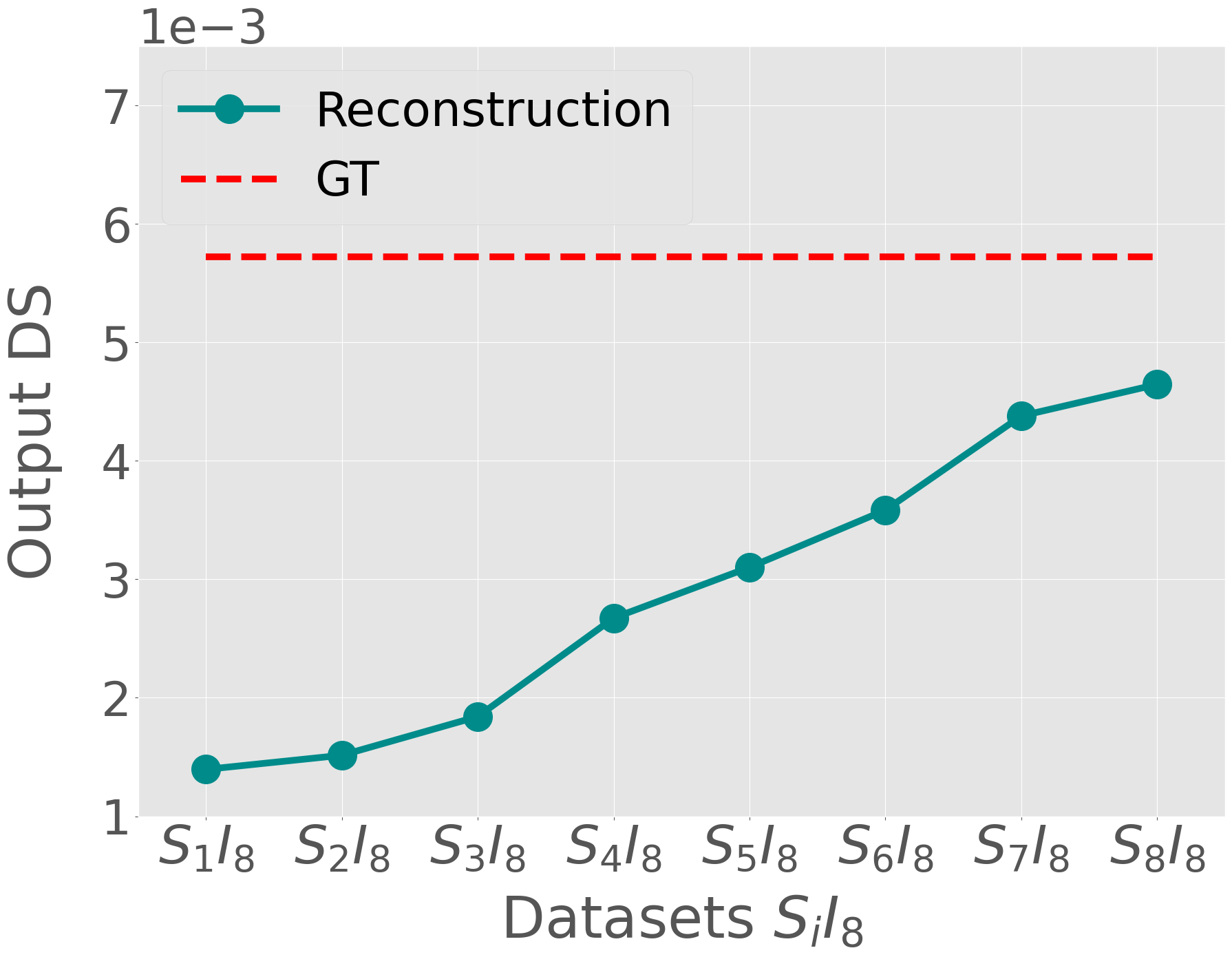} &
    \includegraphics[width=0.24\textwidth]{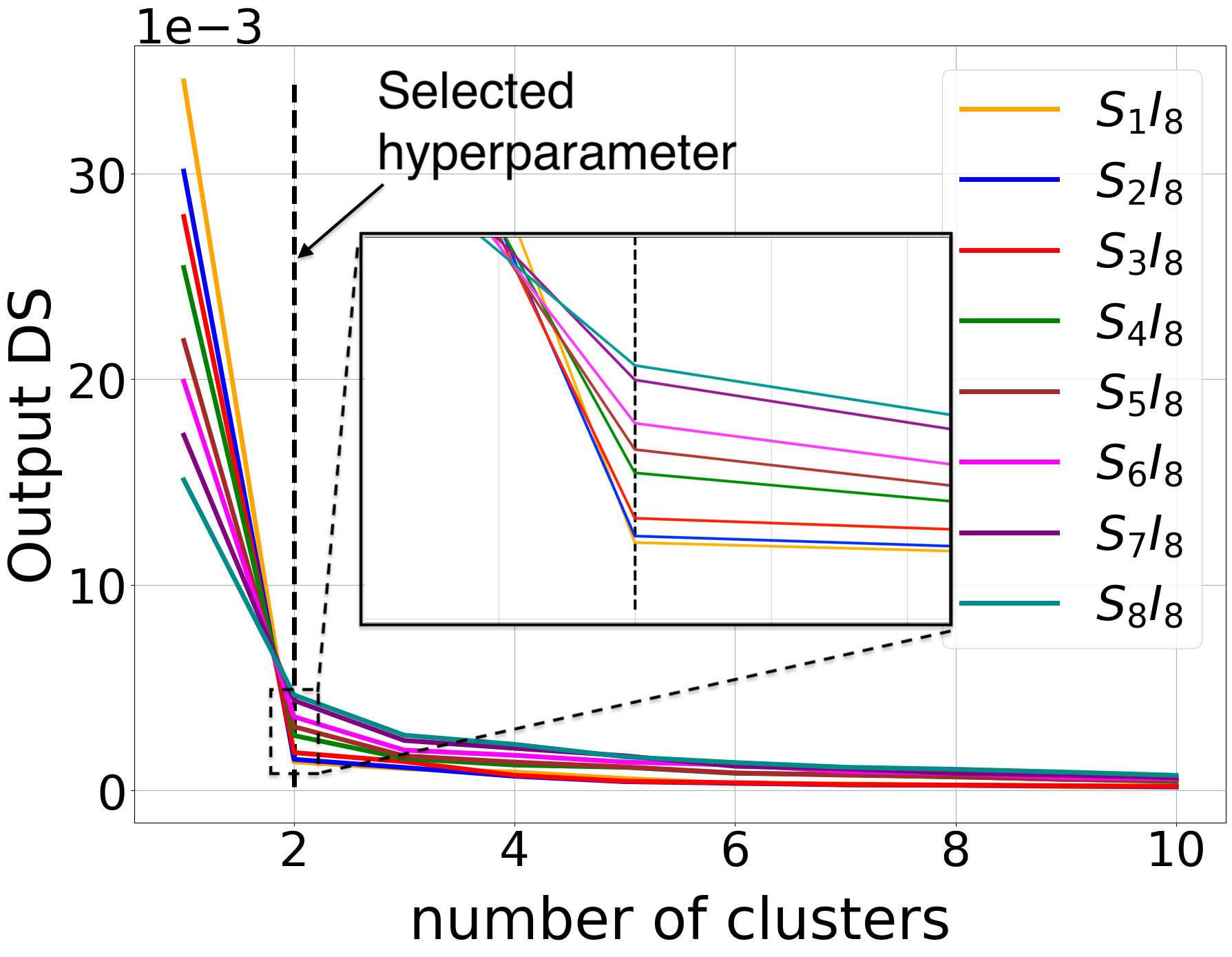} 
\end{tabular}  \vspace{-3mm}
	\caption{Hyperparameter (HP) Sweeping on the input/output DS of the synthetic dataset. The left column titled with ``HP = 2'' shows the DS evaluations with selected HP presented in the main paper, the right column titled with ``HP Sweeping'' shows the DS with sweeping HP. The vertical dash line marks the selected HP in a sweeping curve.}
	\label{fig:synthetic-sweeping}
\end{centering}
\end{figure*}

\begin{figure*}[!ht]
\begin{centering}
	\begin{tabular}{c@{}c@{}|c@{}c@{}}
	\toprule
	\multicolumn{2}{c|}{Input DS} & \multicolumn{2}{c}{Output DS} \\
	\toprule
	HP = 500 & HP Sweeping & HP = 500 & HP Sweeping
	\\
	\toprule 
    \includegraphics[width=0.24\textwidth]{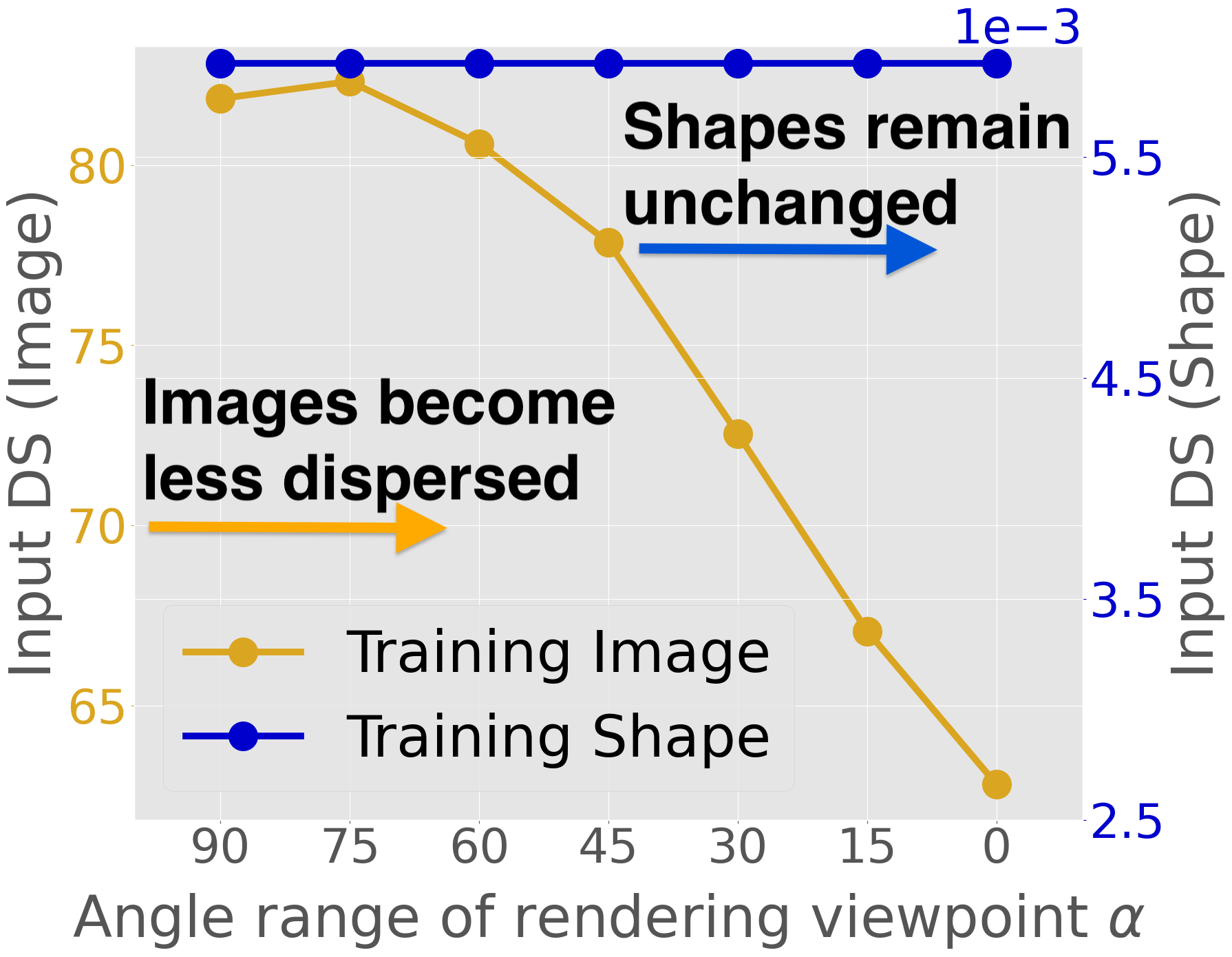} &
    \includegraphics[width=0.24\textwidth]{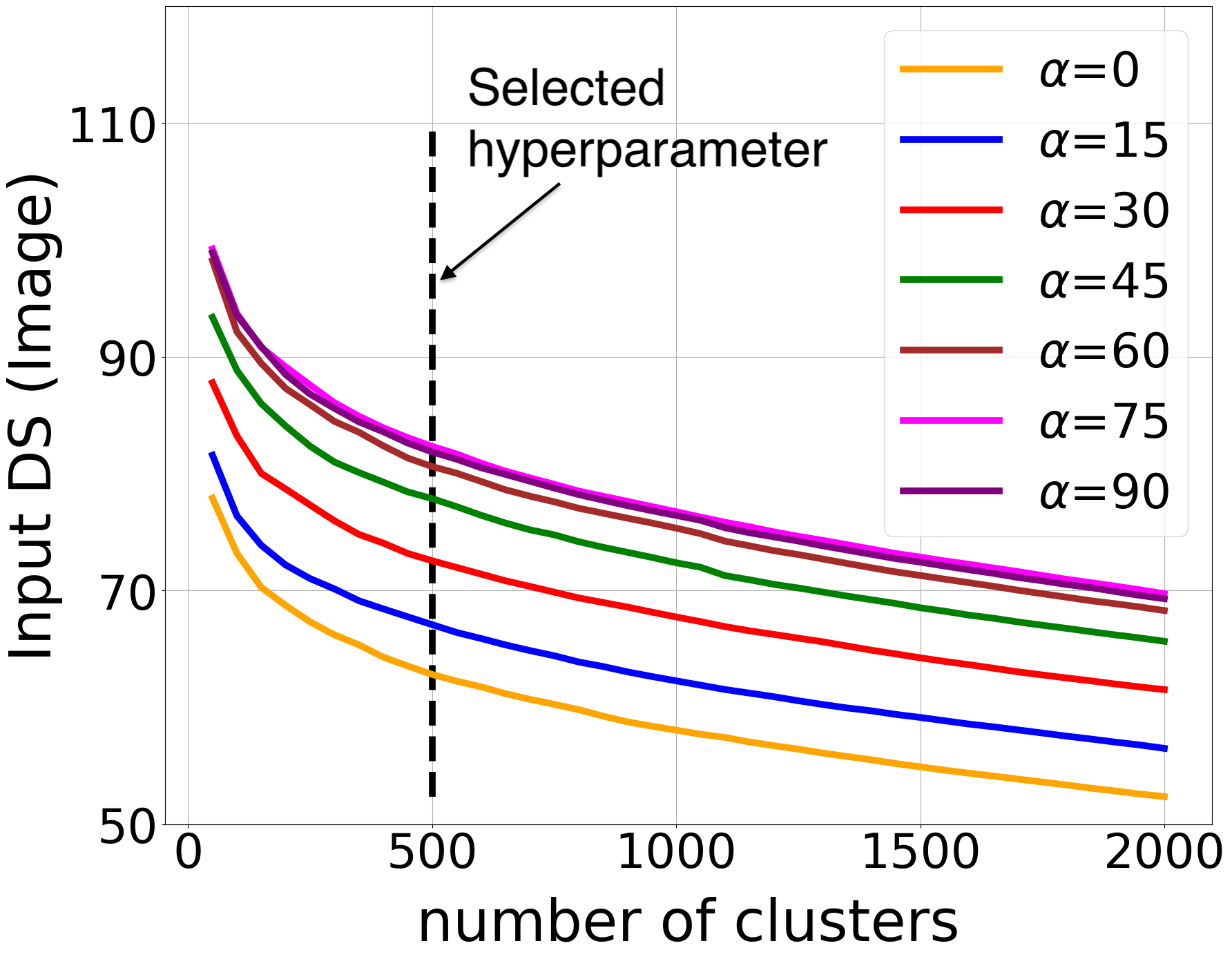} \hspace{2mm}&
    \includegraphics[width=0.24\textwidth]{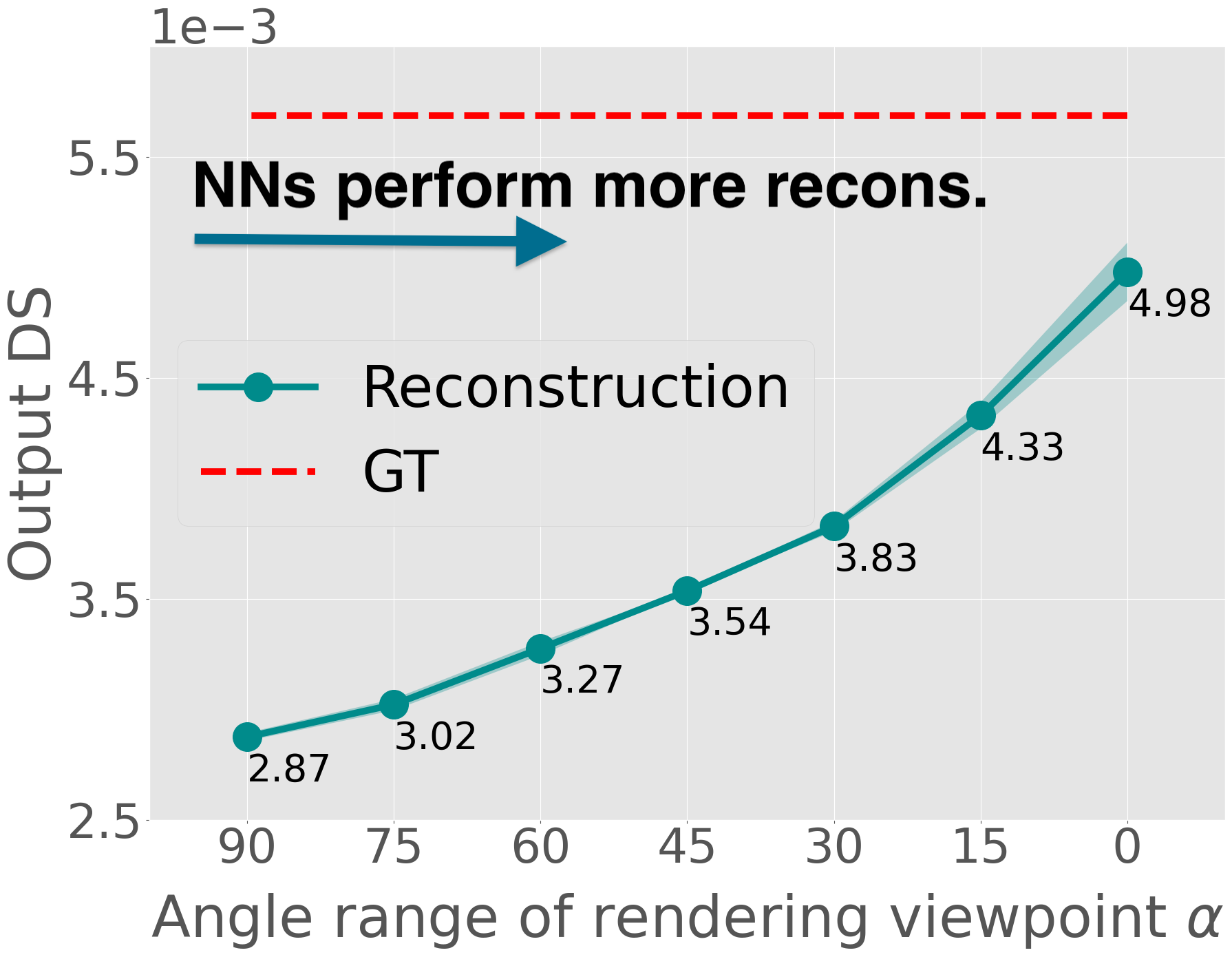} &
    \includegraphics[width=0.24\textwidth]{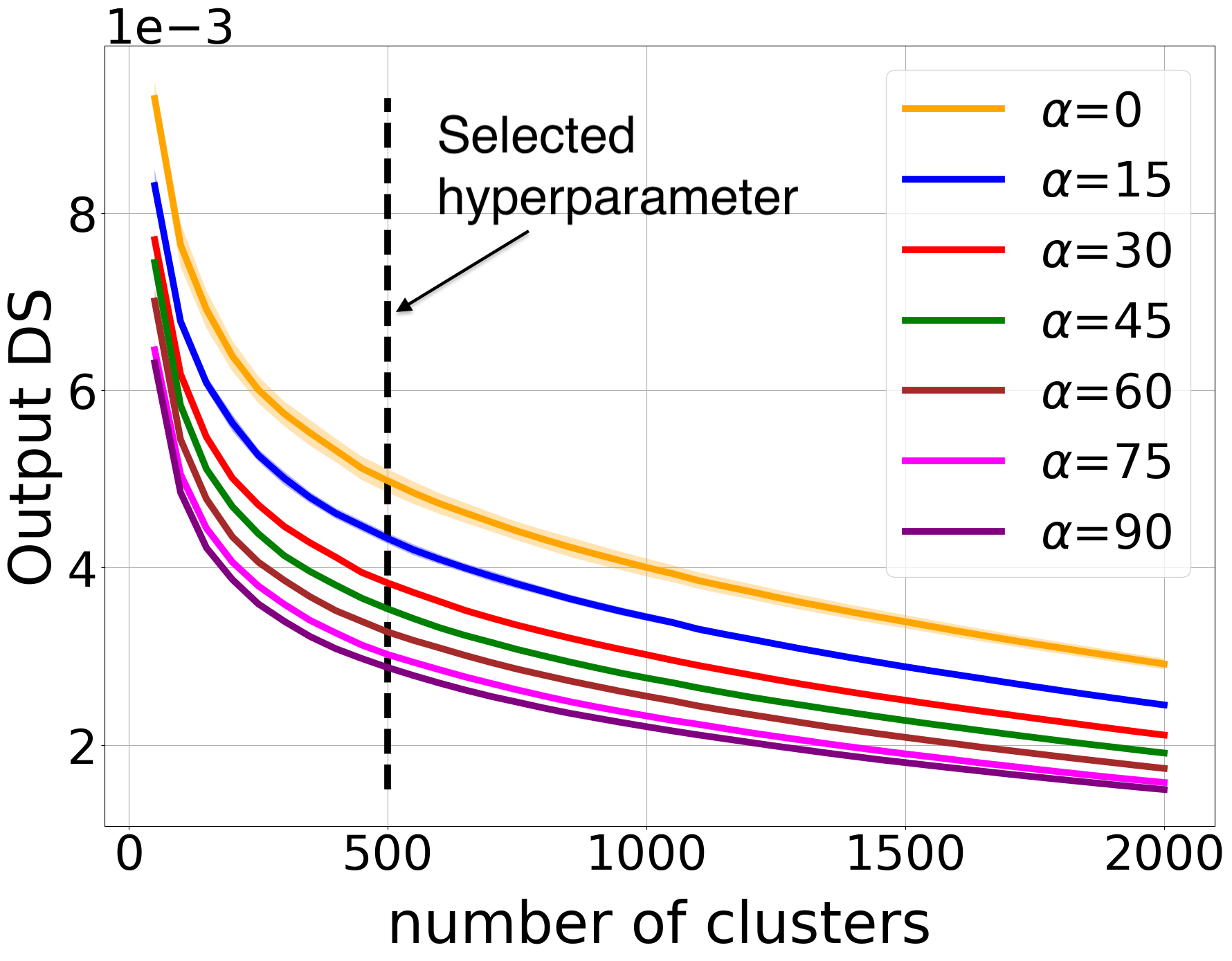} \\
    \includegraphics[width=0.24\textwidth]{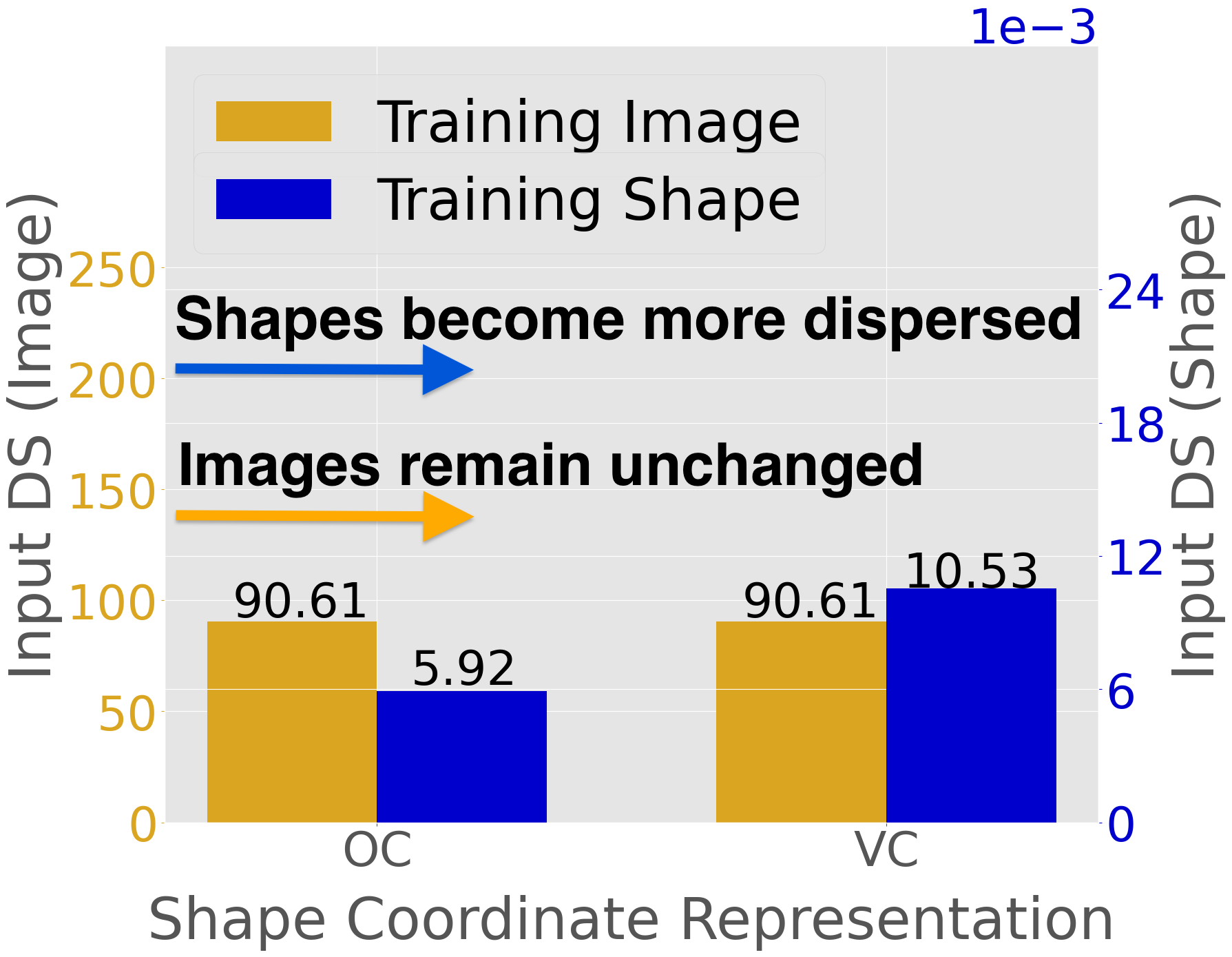} &
    \includegraphics[width=0.24\textwidth]{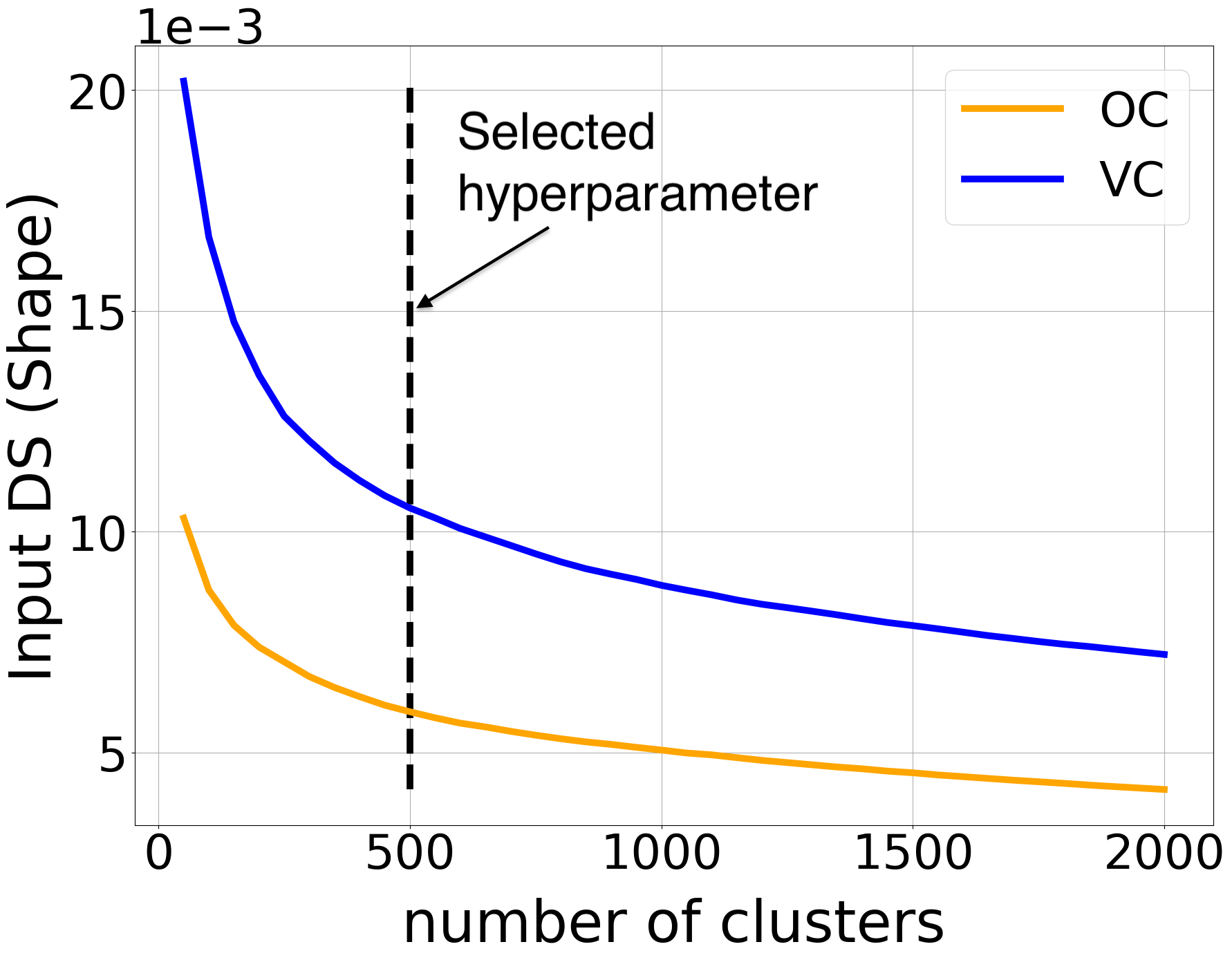} \hspace{2mm}&
    \includegraphics[width=0.24\textwidth]{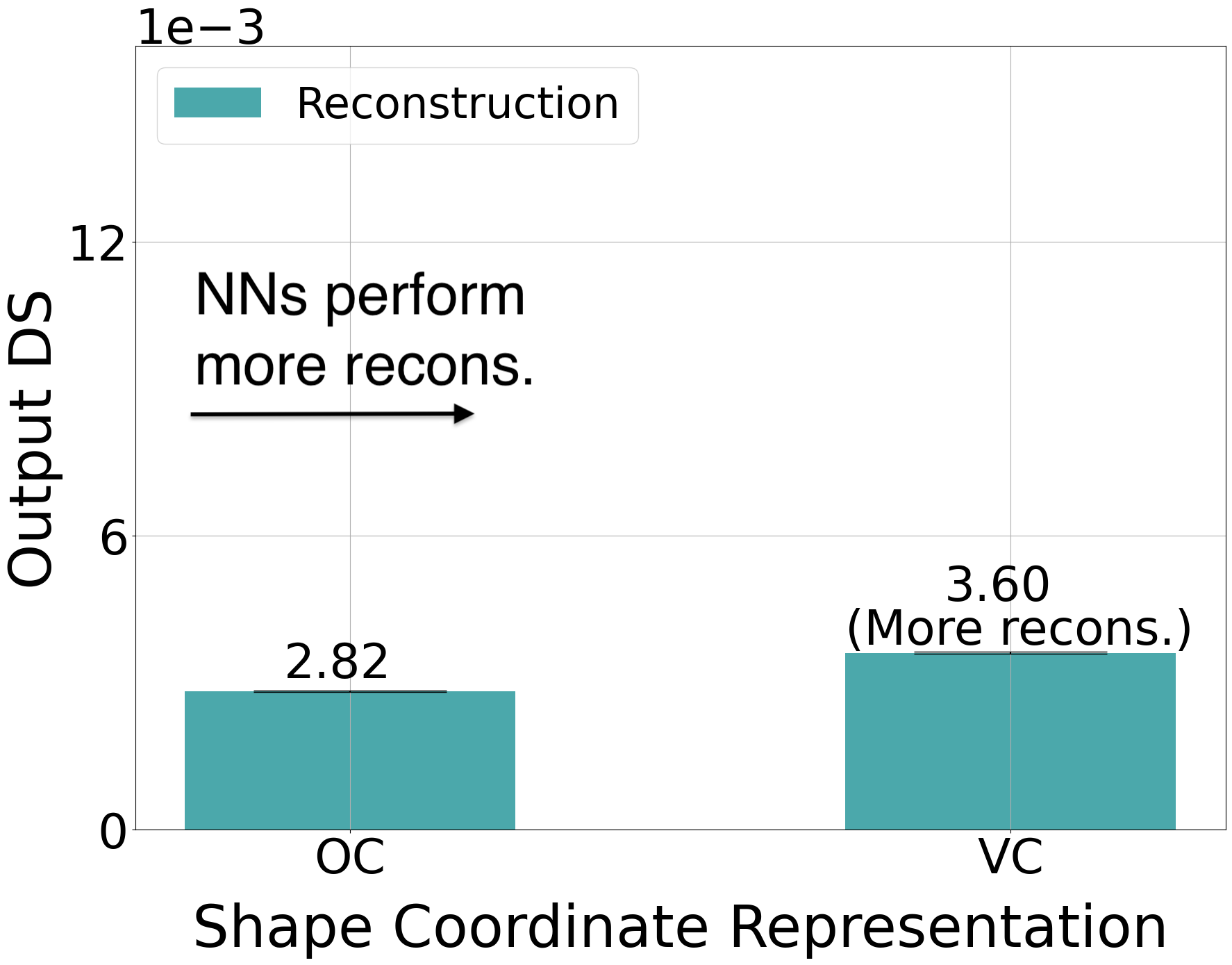} &
    \includegraphics[width=0.24\textwidth]{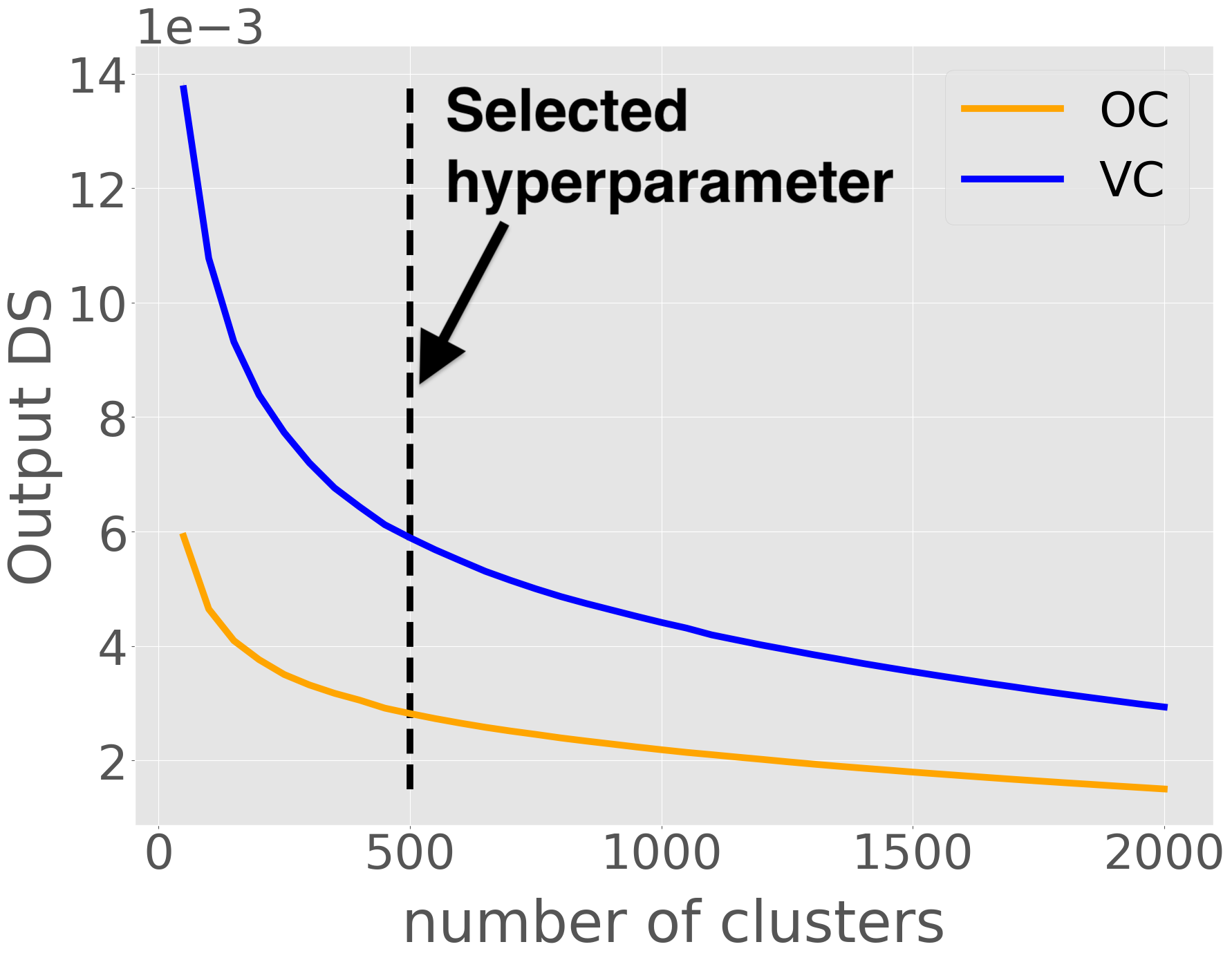} \\ 
    \includegraphics[width=0.24\textwidth]{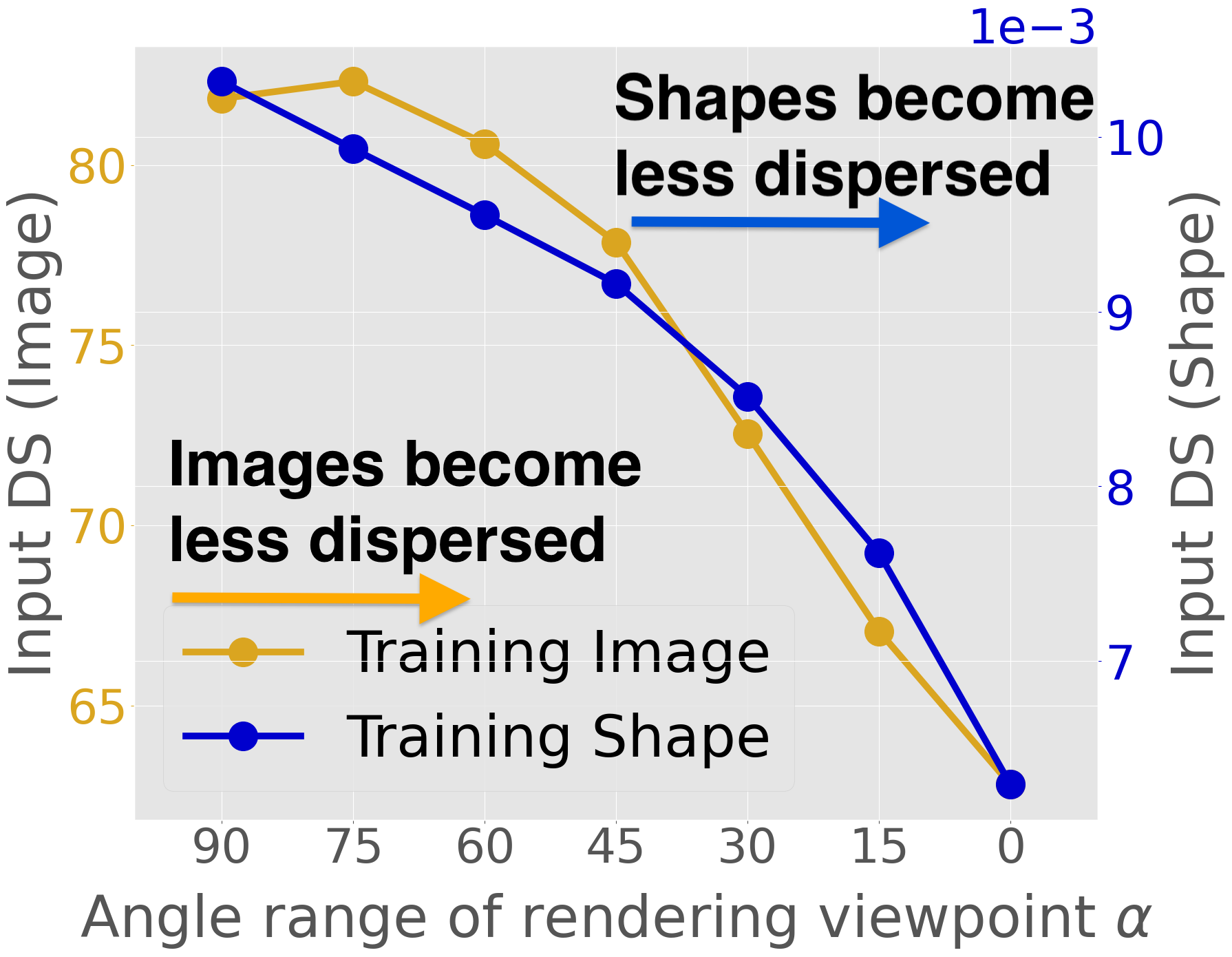} &
    \includegraphics[width=0.24\textwidth]{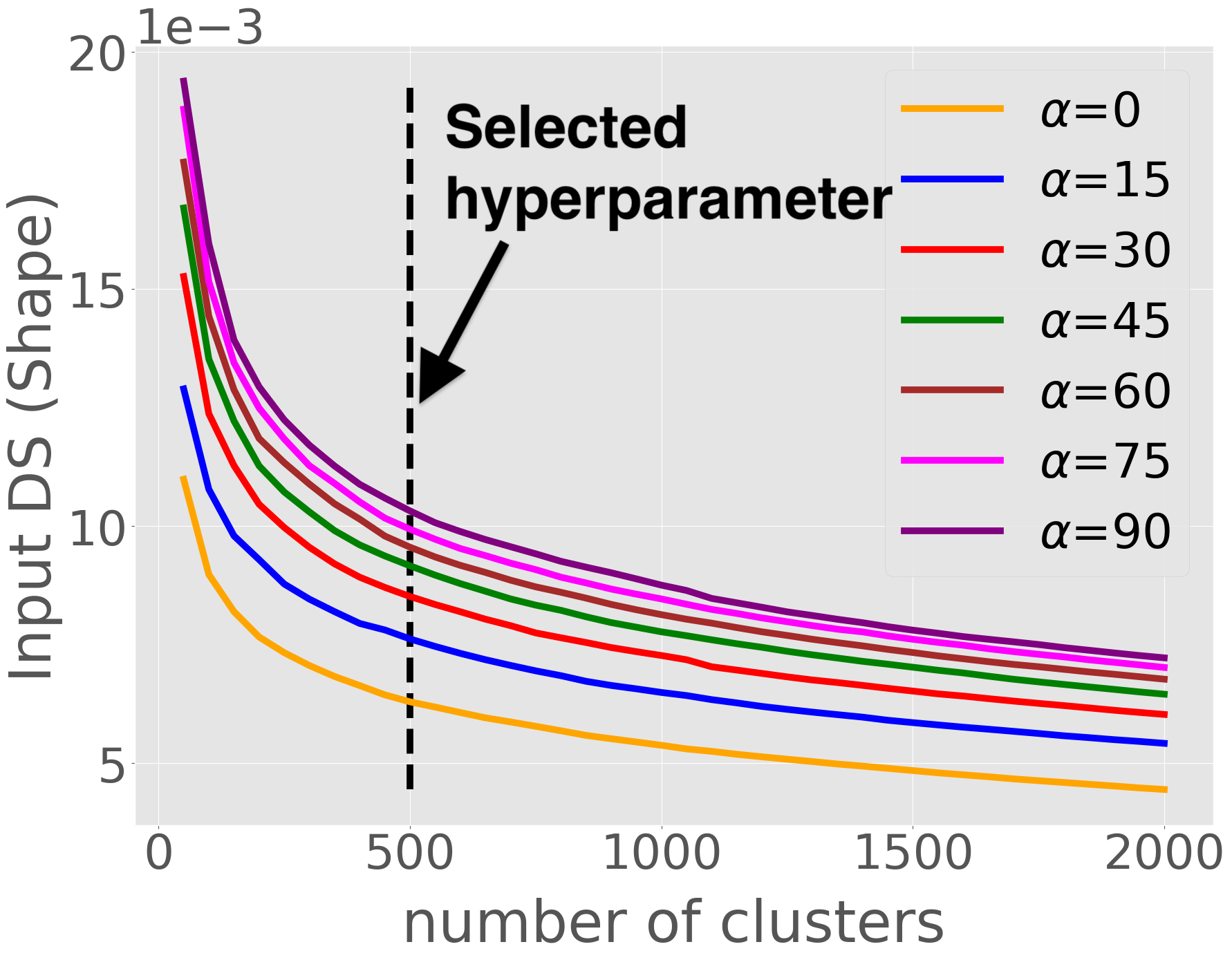} \hspace{2mm}&
    \includegraphics[width=0.24\textwidth]{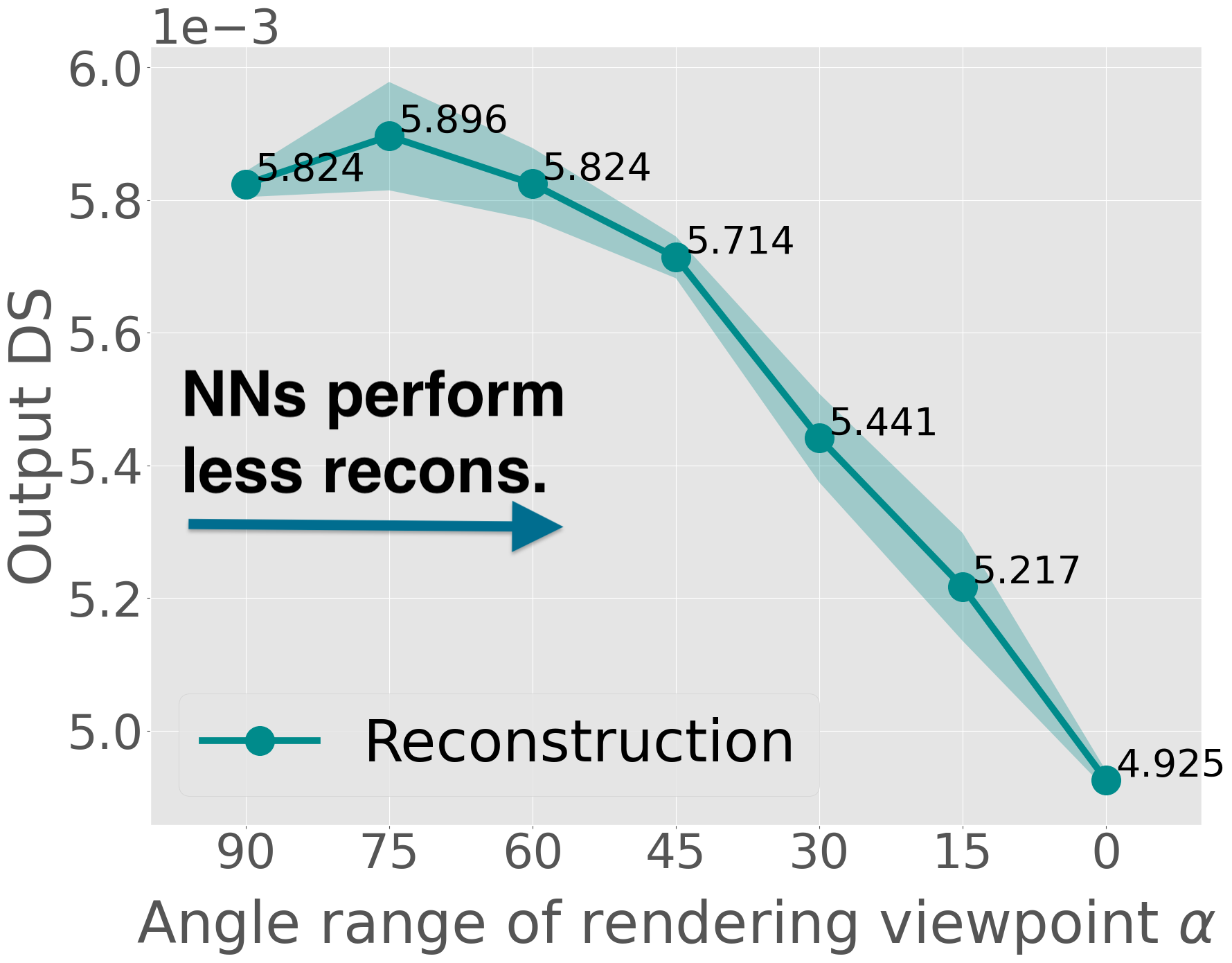} &
    \includegraphics[width=0.24\textwidth]{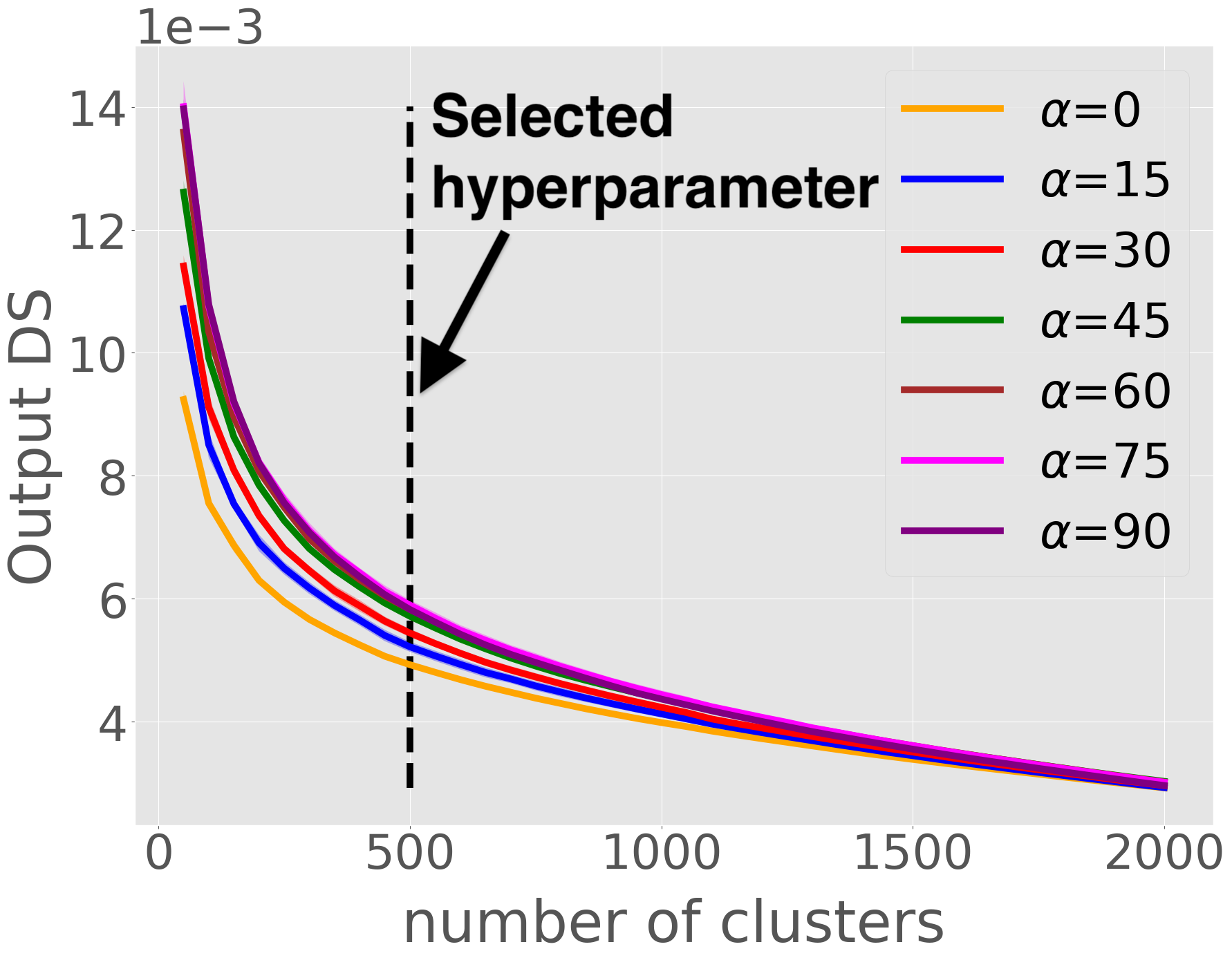}  
\end{tabular}   \vspace{-4mm}
	\caption{Hyperparameter (HP) Sweeping on the input/output DS of ShapeNet. The left column titled with ``HP = 500'' shows the DS evaluations with selected HP presented in the main paper, the right column titled with ``HP Sweeping'' shows the DS with sweeping HP. The vertical dash line marks the selected HP in a sweeping curve.}
	\label{fig:shapenet-sweeping}
\end{centering}
\end{figure*}

We provide the details of selecting the HP of input/output DS in the synthetic dataset and ShapeNet experiments presented in the main paper. Figure \ref{fig:synthetic-sweeping} shows the procedure concerning the synthetic dataset. 
During HP sweeping, we sample the value of HP from 1 to 10 with step 1. The curve of DS stop dropping dramatically when HP equals 2, which indicates the ``elbow'' of sweeping curves. In that, we select the HP cluster-number as 2 in evaluating the DS of synthetic datasets. 
In Figure \ref{fig:shapenet-sweeping}, we repeat the same analysis on ShapeNet. We sweep the HP from 50 to 2000 with step 50 and eventually select HP to be 500. From the two figures, we also observe the consistent relative performance between datasets or reconstructions across varying HP, which indicates the robustness of the metric DS against the change in HP.

\section{Synthetic Data Generation}\label{sec:synthetic-data}

\begin{figure}[!ht]
	\begin{tabular}{c@{}c@{}c@{}c@{}c@{}c@{}c@{}c@{}c@{}c@{}c@{}}
	    sphere & & & & & $\Longleftrightarrow$ & & & &  cube \\
		\includegraphics[width=.10\columnwidth,keepaspectratio]{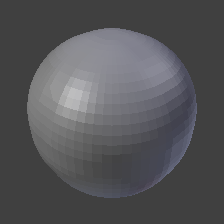} & \hspace{-2mm}
		\includegraphics[width=.10\columnwidth,keepaspectratio]{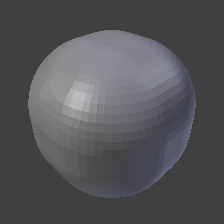} &
		\includegraphics[width=.10\columnwidth,keepaspectratio]{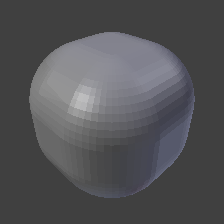} &
		\includegraphics[width=.10\columnwidth,keepaspectratio]{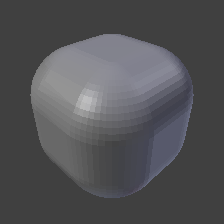} &
		\includegraphics[width=.10\columnwidth,keepaspectratio]{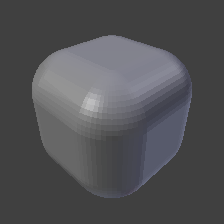} &
		\includegraphics[width=.10\columnwidth,keepaspectratio]{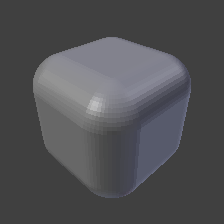} &
		\includegraphics[width=.10\columnwidth,keepaspectratio]{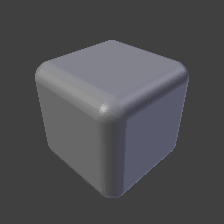} &
		\includegraphics[width=.10\columnwidth,keepaspectratio]{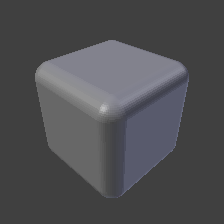} &
		\includegraphics[width=.10\columnwidth,keepaspectratio]{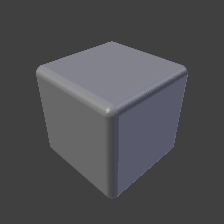} &
		\includegraphics[width=.10\columnwidth,keepaspectratio]{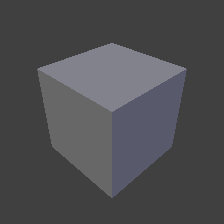} \\  
	    \vspace{2mm}
	    sphere & & & & & $\Longleftrightarrow$ & & & &  cube \\
		\includegraphics[width=.10\columnwidth,keepaspectratio]{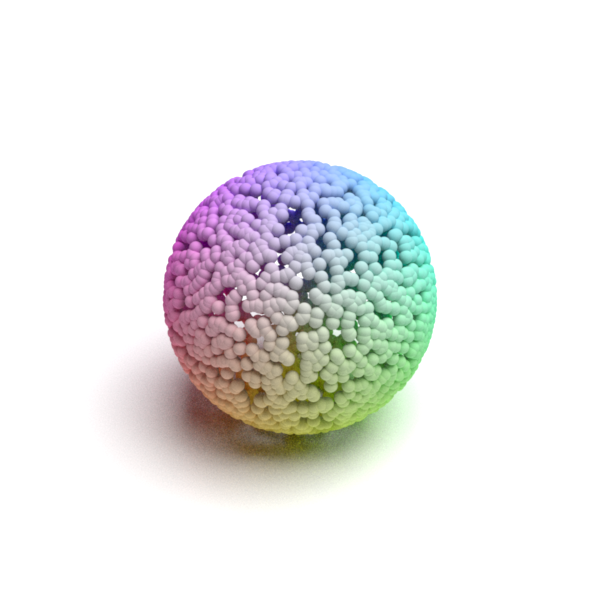} & \hspace{-2mm}
		\includegraphics[width=.10\columnwidth,keepaspectratio]{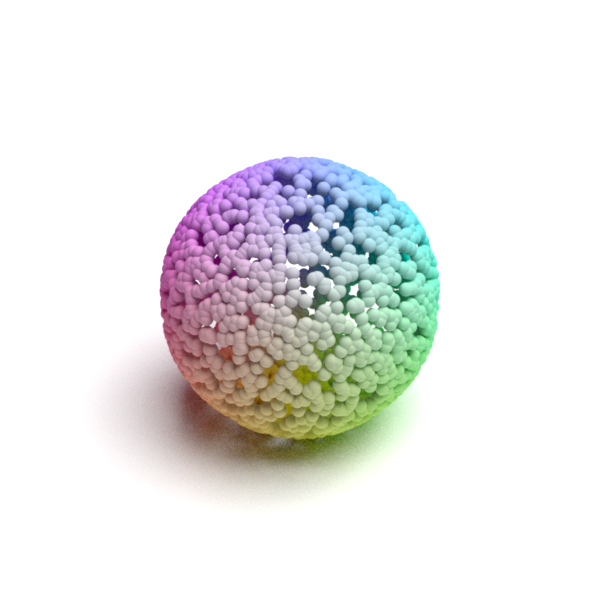} &
		\includegraphics[width=.10\columnwidth,keepaspectratio]{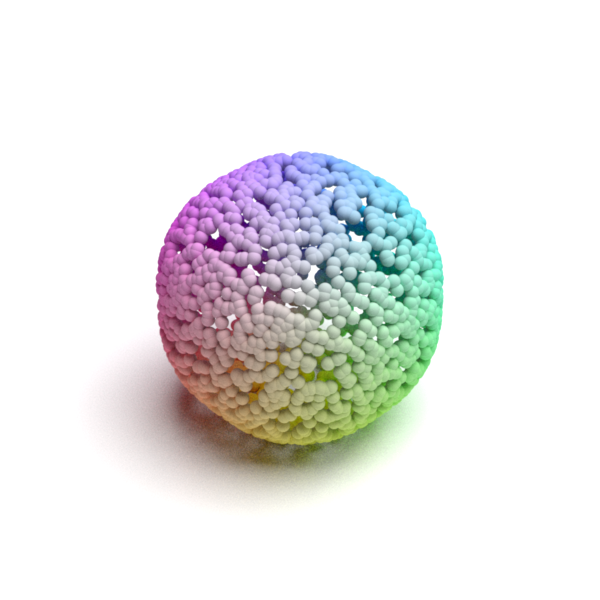} &
		\includegraphics[width=.10\columnwidth,keepaspectratio]{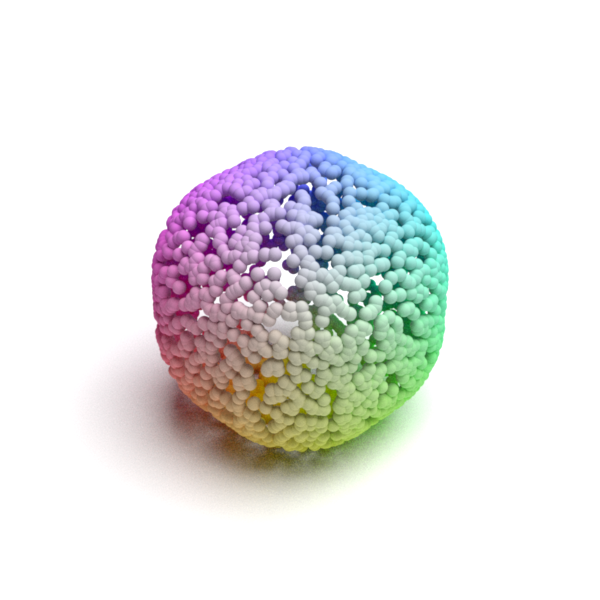} &
		\includegraphics[width=.10\columnwidth,keepaspectratio]{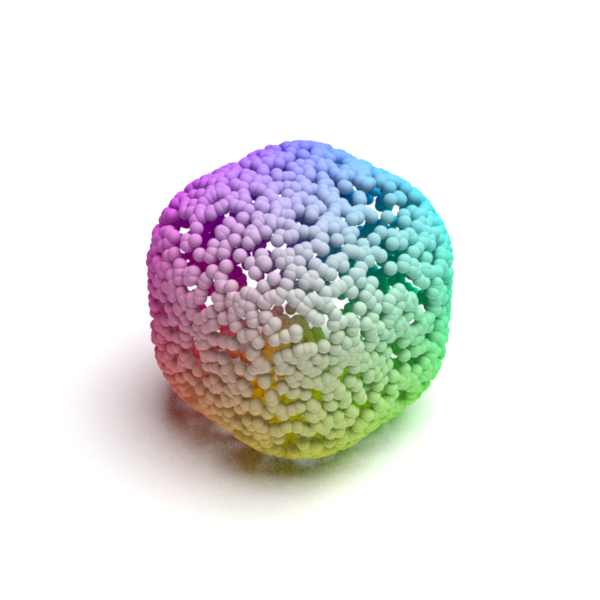} &
		\includegraphics[width=.10\columnwidth,keepaspectratio]{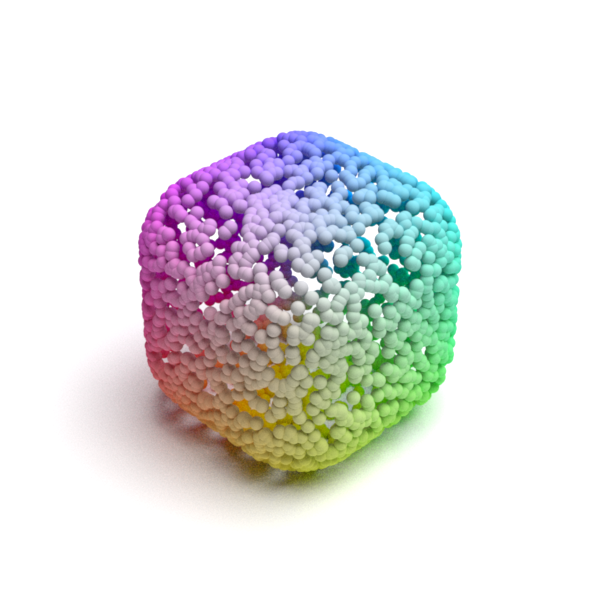} &
		\includegraphics[width=.10\columnwidth,keepaspectratio]{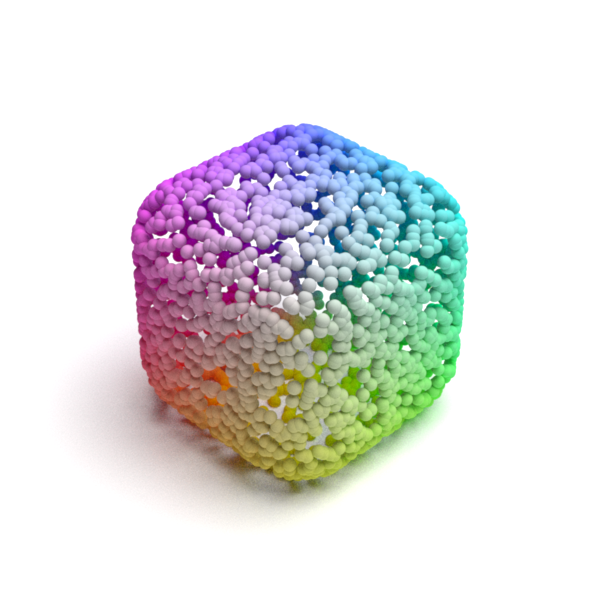} &
		\includegraphics[width=.10\columnwidth,keepaspectratio]{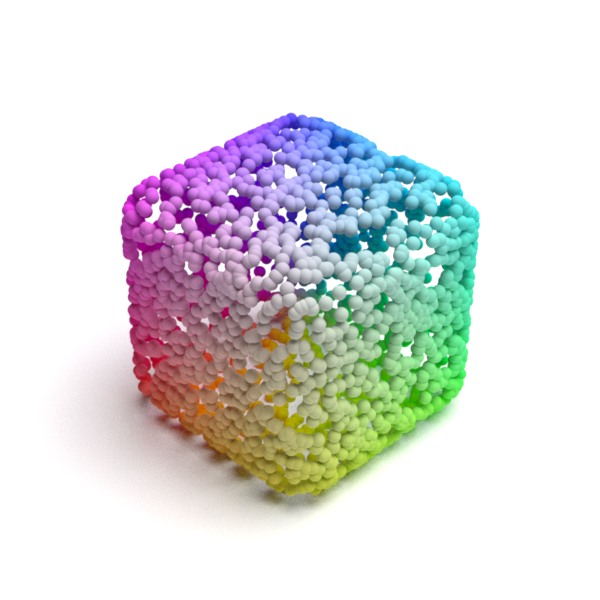} &
		\includegraphics[width=.10\columnwidth,keepaspectratio]{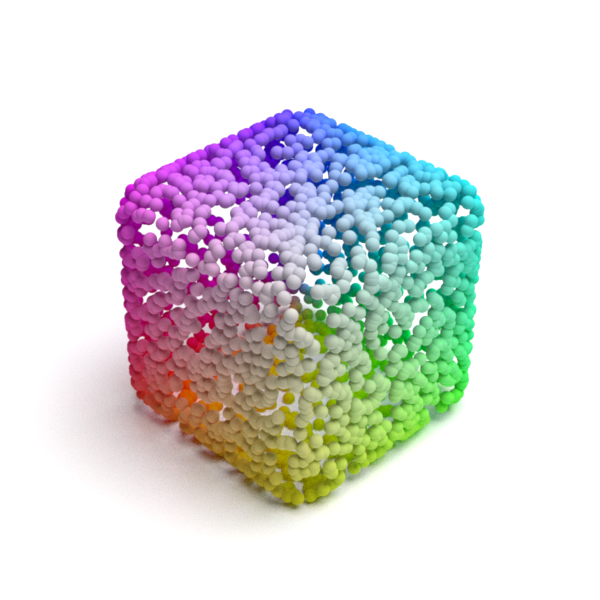} &
		\includegraphics[width=.10\columnwidth,keepaspectratio]{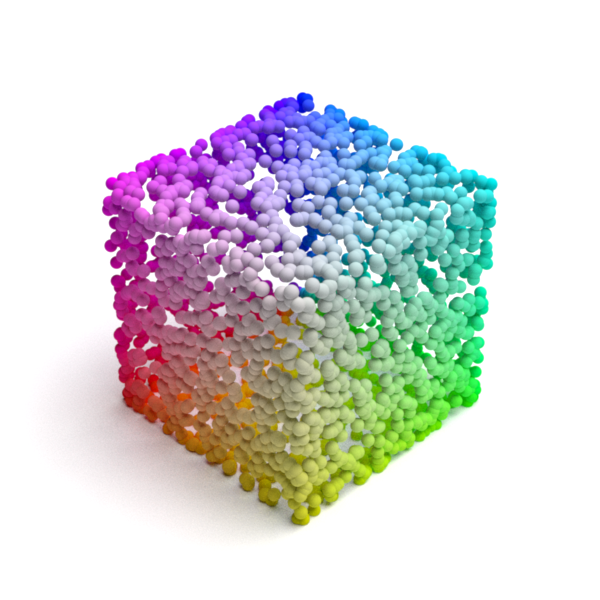} \\  
	\end{tabular}\vspace{-3mm}
	\caption{Image and shapes examples of the synthetic dataset built by interpolating between a sphere and a cube. \emph{Upper}, rendered images. \emph{Lower}, shapes represented by point cloud.}
    \label{fig:cube-sphere-pointcloud} \vspace{-2mm}
\end{figure}

We provide the details of synthetic data generation. The image and shape examples are shown in Figure \ref{fig:cube-sphere-pointcloud}, respectively. We use the software Blender to generate base shapes in a mesh format. Then, we use the Shrinkwrap modifier in ``Nearest Vertex'' mode to define the shape morphing between the two base shapes and control the interpolation progress by the Blender Shape Keys panel. After creating the mesh dataset, we render images and sample point cloud from mesh.

\section{Implementation Details}\label{sec:metrics}
\vspace{-2mm}
In this section, we first provide implementation details of baseline models benchmarked in the main paper. Then, we present more details about our training setup. 

\noindent
\textbf{Baselines}
For NN-based methods, we include PSGN~\cite{fan2017point}, FoldingNet~\cite{yang2018foldingnet}, AtlasNet-Sphere~\cite{groueix2018papier}, AtlasNet-25~\cite{groueix2018papier}. For these models, we all use a ResNet-18 image encoder without any pre-training, the encoder outputs a 1024 dimensional latent vector. We implement the decoder of each model following the original architecture. For recognition-based methods, we include Clustering~\cite{tatarchenko2019single} and Oracle NN~\cite{tatarchenko2019single}. 
For Clustering, we first compute the pair-wise distance matrices of all shapes in the training set, both in OC and VC. The size of the distance matrix is 35021 $\times$ 35021 and the distance function is Chamfer Distance. 
Then we apply K-medoids algorithm~\cite{10.1007/978-3-540-87993-0_19} to the distance matrix and set the cluster-number hyperparameter as 500. Each training shape is assigned to a cluster label by K-medoids. 
After that, we train a classifier that assigns images to cluster labels of corresponding shapes. During testing, we set the training shapes of cluster centroids as the inferred solution. For classification, we use the ResNet-50 architecture~\cite{7780459}, pre-trained on the ImageNet dataset~\cite{5206848}, and fine-tuned for 50 epochs. For Oracle NN, for each shape in the test set, we find the closest shape from the training set in terms of Chamfer Distance. 

\noindent
\textbf{Training Setup}
For experiments on the synthetic dataset, each model is trained for 3600 iterations, using batch size 8. The initial learning rate is 1e-3, and the learning rate decays at 2400, 3000, 3300 iterations by a ratio of 0.1. 
For experiments on ShapeNet, the training epoch is 120, the batch size is 64, the initial learning rate is 1e-3, and it decays at 90, 110, 115 epoch by ratio 0.1. The weight decay is set to be 0. 

\FloatBarrier

{\small
\bibliographystyle{ieee_fullname}
\bibliography{egbib}
}